\documentclass[10pt]{article}
\usepackage[preprint]{tmlr}

\usepackage{amsmath,amsfonts,bm,bbm,mathrsfs,dsfont}
\usepackage{amssymb,mathtools}

\newcommand{\indicator}{\mathds{1}}

\def\vb{{\bm{b}}}
\def\vc{{\bm{c}}}

\def\ve{{\bm{e}}}

\def\vg{{\bm{g}}}
\def\vh{{\bm{h}}}

\def\vk{{\bm{k}}}

\def\vp{{\bm{p}}}
\def\vq{{\bm{q}}}
\def\vr{{\bm{r}}}

\def\vu{{\bm{u}}}

\def\vw{{\bm{w}}}
\def\vx{{\bm{x}}}

\def\vz{{\bm{z}}}

\DeclareMathAlphabet{\mathsfit}{\encodingdefault}{\sfdefault}{m}{sl}
\SetMathAlphabet{\mathsfit}{bold}{\encodingdefault}{\sfdefault}{bx}{n}

\newcommand{\R}{\mathbb{R}}

\def\norm#1{\left\| #1 \right\|}

\def\1{\mathbbm{1}}

\usepackage{amsthm}
\theoremstyle{plain}
\newtheorem{theorem}{Theorem}
\newtheorem{proposition}{Proposition}
\newtheorem{lemma}{Lemma}
\newtheorem{corollary}{Corollary}

\usepackage{booktabs}
\usepackage{array}
\usepackage{ragged2e}
\usepackage{xltabular}
\usepackage{makecell}
\usepackage{pdflscape}
\usepackage{enumitem}
\usepackage[dvipsnames,table]{xcolor}
\usepackage{url}
\usepackage{placeins}
\usepackage{microtype}
\usepackage{algorithm}
\usepackage{algorithmicx}
\usepackage[noend]{algpseudocode}
\usepackage{caption}
\usepackage{dirtytalk}
\usepackage{graphicx}
\usepackage{listings}
\usepackage{subcaption}
\usepackage{tcolorbox}
\tcbuselibrary{breakable}
\usepackage{fancyvrb}
\usepackage{wrapfig}
\usepackage{xspace}
\usepackage{float}
\usepackage{needspace}
\usepackage{longtable}
\usepackage{multirow}
\usepackage{adjustbox}
\usepackage{tabularx}
\usepackage[colorlinks=true]{hyperref}
\usepackage[capitalize,noabbrev]{cleveref}
\usepackage{collcell}
\AtBeginEnvironment{thebibliography}{\hbadness=10000\hfuzz=\maxdimen}
\makeatletter
\g@addto@macro{\hfuzz=\maxdimen}
\let\@expl@@@mark@update@singlecol@structures@@\relax
\makeatother

\newfloat{prompt}{t}{lop}
\floatname{prompt}{Prompt}
\crefname{prompt}{prompt}{prompts}
\Crefname{prompt}{Prompt}{Prompts}

\definecolor{darkblue}{rgb}{0.0,0.0,0.65}
\definecolor{darkred}{rgb}{0.65,0.0,0.0}
\definecolor{darkgreen}{rgb}{0.0,0.5,0.0}
\definecolor{tab:blue}{RGB}{31,119,180}
\definecolor{tab:red}{RGB}{214,39,40}
\definecolor{tab:green}{RGB}{44,160,44}
\definecolor{tab:orange}{RGB}{255,127,14}
\definecolor{highlight}{gray}{0.92}
\definecolor{meState}{RGB}{102,51,153}
\definecolor{meNeighborhood}{RGB}{31,119,180}
\definecolor{meSkip}{RGB}{230,126,34}
\definecolor{meGlobal}{RGB}{44,160,44}
\definecolor{meDomain}{RGB}{214,39,40}
\definecolor{meReadout}{RGB}{0,128,128}

\newcommand{\meneigh}[1]{\textcolor{meNeighborhood}{#1}}

\newcommand{\medomain}[1]{\textcolor{meDomain}{#1}}

\hypersetup{
  linktocpage=true,
  colorlinks=true,
  citecolor=darkblue,
  filecolor=darkblue,
  linkcolor=darkred,
  urlcolor=darkblue
}

\definecolor{thmTheoremFrame}{RGB}{27,54,93}
\definecolor{thmTheoremBack}{RGB}{230,238,248}
\definecolor{thmPropFrame}{RGB}{15,90,87}
\definecolor{thmPropBack}{RGB}{224,242,241}
\definecolor{thmLemmaFrame}{RGB}{153,84,13}
\definecolor{thmLemmaBack}{RGB}{255,246,230}
\definecolor{thmCorFrame}{RGB}{90,40,90}
\definecolor{thmCorBack}{RGB}{243,231,246}

\definecolor{googleblue}{HTML}{4285F4}
\definecolor{googlegreen}{HTML}{34A853}
\definecolor{googlered}{HTML}{EA4335}
\definecolor{googleyellow}{HTML}{F29900}

\definecolor{openQFrame}{HTML}{FF6D00}
\colorlet{openQBack}{openQFrame!10}

\definecolor{repSpatial}{HTML}{1A73E8}
\definecolor{repAttention}{HTML}{F29900}
\definecolor{repSpectral}{HTML}{6A1B9A}
\definecolor{repTransformer}{HTML}{2E7D32}
\definecolor{repHetero}{HTML}{C62828}
\definecolor{repHigher}{HTML}{455A64}
\definecolor{repGeometric}{HTML}{00897B}

\newcolumntype{Y}[1]{>{\raggedright\arraybackslash}p{#1}}
\newcolumntype{Z}[1]{>{\centering\arraybackslash}p{#1}}
\newcolumntype{Q}[1]{>{\raggedright\arraybackslash$}p{#1}<{$}}
\newcolumntype{R}[1]{>{\centering\arraybackslash$}p{#1}<{$}}

\DeclareMathOperator{\Diag}{Diag}
\DeclareMathOperator{\MLP}{MLP}

\definecolor{fsmHeader}{HTML}{D6E4F0}
\definecolor{fsmHeaderText}{RGB}{27,54,93}
\definecolor{fsmSpatial}{HTML}{E3ECFB}
\definecolor{fsmAttention}{HTML}{FDF0DC}
\definecolor{fsmSpectral}{HTML}{F0E6F5}
\definecolor{fsmTransformer}{HTML}{E4F1E5}
\definecolor{fsmHetero}{HTML}{FBE4E4}
\definecolor{fsmHigher}{HTML}{E8ECEE}
\definecolor{fsmGeometric}{HTML}{DFF1EF}

\providecommand{\sm}{\operatorname{softmax}}
\providecommand{\LReLU}{\operatorname{LReLU}}
\providecommand{\LN}{\operatorname{LN}}
\providecommand{\normop}{\operatorname{norm}}
\providecommand{\GNN}{\operatorname{GNN}}
\providecommand{\POOL}{\operatorname{POOL}}
\providecommand{\FC}{\operatorname{FC}}

\newcommand{\mixmsg}{\widetilde{\mathbf{H}}}

\makeatletter
\newcommand{\appendixsection}[1]{%
  \refstepcounter{section}%
  \phantomsection
  \addcontentsline{toc}{subsection}{\protect\numberline{\thesection}#1}%
  \section*{Appendix \thesection\ #1}%
}
\newcommand{\appendixsubsection}[1]{%
  \refstepcounter{subsection}%
  \phantomsection
  \addcontentsline{toc}{subsubsection}{\protect\numberline{\thesubsection}#1}%
  \subsection*{\thesubsection\quad #1}%
}
\makeatother

\newcolumntype{H}{>{\setbox0=\hbox\bgroup}c<{\egroup}@{}}
\newcolumntype{L}[1]{>{\raggedright\arraybackslash}p{#1}}
\newcolumntype{C}[1]{>{\Centering\arraybackslash}p{#1}}

\AtBeginDocument{%
  \setlength{\abovedisplayskip}{6pt plus 2pt minus 2pt}%
  \setlength{\belowdisplayskip}{6pt plus 2pt minus 2pt}%
  \setlength{\abovedisplayshortskip}{3pt plus 2pt minus 1pt}%
  \setlength{\belowdisplayshortskip}{5pt plus 2pt minus 2pt}%
  \setlength{\jot}{3pt}%
  \setlength{\textfloatsep}{10pt plus 2pt minus 2pt}%
  \setlength{\dbltextfloatsep}{10pt plus 2pt minus 2pt}%
  \setlength{\floatsep}{8pt plus 2pt minus 2pt}%
  \setlength{\intextsep}{8pt plus 2pt minus 2pt}%
  \setlength{\abovecaptionskip}{4pt plus 1pt minus 1pt}%
  \setlength{\belowcaptionskip}{2pt plus 1pt minus 1pt}%
  \setlength{\LTpre}{4pt plus 1pt minus 1pt}%
  \setlength{\LTpost}{4pt plus 1pt minus 1pt}%
}

\usepackage{tikz-cd}
\usetikzlibrary{positioning,arrows.meta}

\mathchardef\mhyphen="2D

\title{Unifying Graph Neural Networks Through a Common Layer Equation}

\author{%
\begin{minipage}{0.98\textwidth}
\centering
Sai Karthik Navuluru$^{1}$ \quad
Siddhartha Shankar Das$^{2}$ \quad
Bo Ni$^{3}$ \quad
Hongjie Chen$^{4}$\\
Yu Wang$^{5}$ \quad
Baris Coskunuzer$^{1}$ \quad
Nesreen K. Ahmed$^{6}$\\
Franck Dernoncourt$^{7}$ \quad
Mahantesh Halappanavar$^{2}$ \quad
Tyler Derr$^{3}$ \quad
Ryan A. Rossi$^{7}$ \quad
Lakshman Tamil$^{1}$\\[6pt]
{\normalfont\small
$^{1}$University of Texas at Dallas \quad
$^{2}$Pacific Northwest National Laboratory \quad
$^{3}$Vanderbilt University\\
$^{4}$Dolby Laboratories \quad
$^{5}$University of Georgia \quad
$^{6}$Cisco AI Research \quad
$^{7}$Adobe Research
}\\[4pt]
{\normalfont\footnotesize
\begin{tabular}{@{}r@{\;}l@{}}
$^{1}$ & \texttt{\{saikarthik.navuluru, coskunuz, laxman\}@utdallas.edu}\\
$^{2}$ & \texttt{\{siddhartha.das, Mahantesh.Halappanavar\}@pnnl.gov}\\
$^{3}$ & \texttt{\{bo.ni, tyler.derr\}@vanderbilt.edu}\\
$^{4}$ & \texttt{\{hongjie.chen\}@dolby.com}\\
$^{5}$ & \texttt{\{Yu.Wang6\}@uga.edu}\\
$^{6}$ & \texttt{\{n.kamel\}@gmail.com}\\
$^{7}$ & \texttt{\{franck.dernoncourt\}@gmail.com},
        \texttt{\{ryrossi\}@adobe.com}
\end{tabular}
}
\end{minipage}
}

\begin{document}

\maketitle
\hbadness=10000
\hfuzz=\maxdimen
\vbadness=10000
\makeatletter
\g@addto@macro\@arrayparboxrestore{\hbadness=10000\hfuzz=\maxdimen}
\makeatother

\begin{abstract}
Graph neural networks are commonly presented through family-specific equations whose
notation obscures shared computations and structural differences. We develop a common layer
equation that represents covered architectures through seven components consisting of an
update domain, channel set, propagation bank, per-channel message maps, channel-fusion
operator, ego/residual map, and update map. Its central factorization separates \emph{where}
information moves, encoded by the propagation bank, from \emph{what} moves, encoded by the
message maps. Function-valued component fillings extend the same equation from local message
passing to attention, spectral filtering, global communication, relation-specific channels,
higher-order domains, and geometric messages.

We make the unification checkable through worked reductions of canonical layers and explicit
fillings spanning seven nonexclusive architectural families. A fixed slot discipline assigns
each elementary operation by output role and states the boundary of the covered layer class.
The decomposition yields two component-level conclusions. Under endpoint-local messages and
node-local updates, operator support bounds one-layer dependencies, and a full effective
operator row is necessary for one-layer global mixing under the stated hypotheses. In the
linear additive regime, displayed channel count is not identifiable, whereas the minimum
Kronecker separation rank of the summed layer operator is invariant. These results turn the
decomposition into a unified attribution and identifiability framework. Under the fixed slot
discipline, component-wise comparison identifies where two architectures differ, while
functional equivalence additionally requires agreement among internal mechanisms and component
interactions. The unified system organizes more than 200 architectures, separates
architectural families from reusable operations and problem settings, and connects propagation
choices to oversmoothing, oversquashing, heterophily, and expressivity.

The unified equation provides a representation and component-analysis framework. Selecting the
best architecture for a dataset remains an empirical inverse problem. We use the structured design space to generate
well-formed architectures, translate documented benchmark evidence descriptively into component choices, and
formulate the unresolved inverse problem of mapping measurable graph and task properties to
validated component fillings.
\end{abstract}

\newpage
\begingroup
\small                    
\renewcommand{\baselinestretch}{0.9}\selectfont
\setlength{\parskip}{0pt}
\tableofcontents
\endgroup
\newpage

\section{Introduction}
\label{sec:introduction}
Graph Neural Networks (GNNs) are standard tools for relational data, including social-network
analysis \citep{hamilton2017inductive}, molecular property prediction
\citep{gilmer2017neural}, knowledge-graph reasoning \citep{schlichtkrull2018modeling},
recommendation \citep{ying2018pinsage}, temporal interaction modeling
\citep{rossi2020temporal}, and combinatorial optimization \citep{khalil2017learning}.
Their success has produced hundreds of architectures, from spectral filters
\citep{bruna2014spectral,defferrard2016convolutional} and attention
\citep{velickovic2018gat} to higher-order \citep{morris2019weisfeiler,maron2019provably},
directional \citep{gasteiger2020directional}, and geometric equivariant models
\citep{satorras2021en}. Reusable operations such as rewiring
\citep{Topping2022Curvature} and pooling \citep{ying2018hierarchical} add another source of
variation. Family-specific motivations and notation make these designs difficult to compare.

Existing formalisms explain graph learning through message passing
\citep{gilmer2017neural}, graph networks \citep{battaglia2018relational}, geometric deep
learning \citep{bronstein2021geometric}, the augmented-message-passing view
\citep{velickovic2022message}, and equivariance \citep{maron2019invariant}. Other
organizing schemes emphasize applications, model categories
\citep{wu2020comprehensive,zhou2020graph}, or encoder--decoder structure
\citep{hamilton2020graph,chami2022machine}. These abstractions do not provide one layer-level
component inventory across the seven families considered here. In particular,
message-passing and graph-network formulations do not uniformly factor scalar support and
weighting into a propagation bank $\{\mathbf P_k\}$ and transmitted values into message maps
$\{\Psi_k\}$. Geometric formulations instead foreground symmetry, while encoder--decoder
taxonomies foreground tasks and objectives. DeepGL \citep{rossi2020deep} provides a related
operator-composition view for feature construction rather than a factorized neural layer.
Section~\ref{subsec:framework-scope} compares MPNNs, Graph Network blocks, and augmented
message passing component by component. It states which mappings are exact and where a
multi-stage composition or graded domain is required.

\begin{figure*}[t]
\centering
\includegraphics[width=\textwidth]{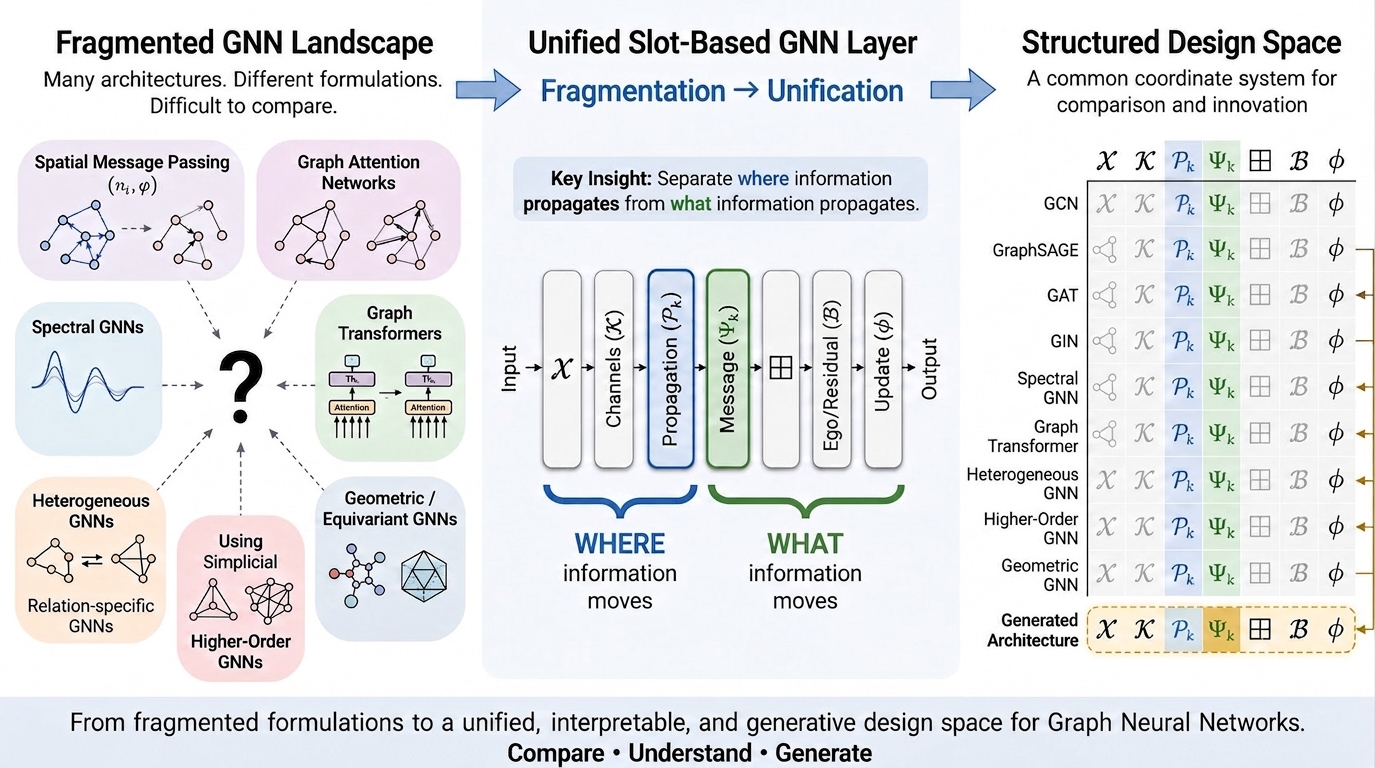}
\caption{\textbf{Our unified GNN design space.} We organize seven previously fragmented architectural families around one common layer. Read the figure from left to right: family-specific formulations are translated into seven shared components: the update domain $\mathcal X$, channel set $\mathcal K$, propagation bank $\{\mathbf P_k\}$, message maps $\{\Psi_k\}$, mixing operator $\boxplus$, ego/residual map $\mathcal B_\ell$, and update map $\phi_\ell$. They then become directly comparable and recombinable. The key split is intuitive: $\mathbf P_k$ specifies \emph{where} information moves, while $\Psi_k$ specifies \emph{what} moves.}
\label{fig:teaser}
\end{figure*}

We address this comparability problem by unifying covered graph neural layers through seven
components. The central split separates \emph{where} information moves
($\mathbf{P}_k$) from \emph{what} moves ($\Psi_k$). Applying one fixed assignment discipline
across families preserves formal differences that a generic message-passing description can
hide. Our contribution is the systematic use of this operator--message factorization as a
unified system across all seven families, together with the attribution and
identifiability results that the separated components make possible. It refines general
message-passing descriptions by exposing computational roles needed for component-level
comparison and analysis. Figure~\ref{fig:teaser} shows how the equation supports unification,
comparison, and architecture generation.

This unification matters because fragmented notation creates more than a presentation problem.
The same mechanism can appear under different names, while methods with similar names can
change different parts of the computation. Comparisons may therefore conflate changes in
support, transmitted values, mixing, and updating, and a result stated for one family may not
visibly transfer to another. The unified system makes these distinctions explicit. It
also exposes two cross-family conclusions. Under endpoint-local messages and node-local
updates, the support of $\{\mathbf P_k\}$ bounds one-layer dependencies, and one-layer global
mixing requires a full effective operator row under the stated hypotheses. In the linear
additive regime, raw channel count is not identifiable, whereas minimum Kronecker separation
rank is an invariant of the summed layer operator. The unified components let an architectural
comparison, theoretical statement, benchmark finding, or proposed remedy be stated through
the components it acts on. They also make underrepresented component combinations visible,
while selecting the best architecture remains an empirical inverse problem.

\raggedbottom

The unified layer has seven components
$(\mathcal X,\mathcal K,\{\mathbf P_k\},\{\Psi_k\},\boxplus,\mathcal B_\ell,\phi_\ell)$.
It can be read as a computational pipeline. The domain $\mathcal X$ specifies what objects are
updated, and $\mathcal K$ lists the available channels. On each channel, $\Psi_k$ constructs
the value to transmit and $\mathbf P_k$ determines where it moves and with what scalar
weight. The mixing operator $\boxplus$ combines channel contributions, $\mathcal B_\ell$
carries the ego or residual state, and $\phi_\ell$ produces the new state.

\[
\bar{\mathbf{H}}^{(\ell+1)}
=
\phi_\ell\!\Bigl(
\underbrace{\mathcal{B}_\ell\!\bigl(\mathbf{H}^{(\ell)},\mathbf{H}^{(0)}\bigr)}_{\text{ego/residual}},\;
\underbrace{\boxplus_{k\in\mathcal K}\,
\mathbf{C}_k^{(\ell)}}_{\mixmsg^{(\ell)}}
\Bigr),
\]
Here $\ell$ indexes layers, $k\in\mathcal K$ indexes channels, and
$\mathbf{H}^{(0)}=\mathbf{X}$. Each contribution $\mathbf C_k^{(\ell)}$ couples
$\mathbf P_k^{(\ell)}$ with $\Psi_k^{(\ell)}$. In the linear case,
$\mathbf C_k^{(\ell)}=\mathbf P_k^{(\ell)}\Psi_k^{(\ell)}$. The symbol
$\mixmsg^{(\ell)}$ denotes only the output of $\boxplus$, not an eighth component. A
single-channel spatial layer sets $\mathcal K=\{1\}$. Equation~\eqref{eq:unified-notation} in
Section~\ref{sec:unified-view} gives the full definition, including the pairwise form needed
for edge-conditioned and geometric messages.

For intuition, GCN~\citep{kipf2017semi} uses one channel with
$\mathbf P_1=\hat{\mathbf A}$, the symmetrically normalized adjacency with self-loops. It
sets $\Psi_1=\mathbf H^{(\ell)}\mathbf W^{(\ell)}$, $\mathcal B_\ell=0$, and
$\phi_\ell(a,b)=\sigma(b)$, recovering
$\bar{\mathbf H}^{(\ell+1)}=\sigma(\hat{\mathbf A}\mathbf H^{(\ell)}\mathbf W^{(\ell)})$.
Section~\ref{sec:unified-view} gives full reductions for GCN, GraphSAGE, GAT, and GIN.

Function-valued fillings allow the same components to express local message passing,
attention, spectral and polynomial filtering, global Transformer attention,
relation-specific channels, higher-order domains, and geometric messages. The tables show the
seven components in six columns because the distinct maps $\mathcal B_\ell$ and $\phi_\ell$
share the final Ego/Update display column. Differing cells then localize the formal choices
recorded by the slot discipline.

The equation provides a unified component system for comparing architectures and formulating
the empirical inverse problem of selecting a model for a dataset. The seven families are
nonexclusive and are defined by their primary axis of variation. Spatial,
attention, spectral, and Graph Transformer families primarily specialize $\mathbf P_k$.
Heterogeneous, higher-order, and geometric families primarily vary $\mathcal K$, $\mathcal X$,
or $\Psi_k$ and $\phi_\ell$ jointly. Hybrid models may therefore occupy several families.

The layer output $\bar{\mathbf H}^{(\ell+1)}$ preserves its domain. A separate pooling
operation may coarsen it between layers (Appendix~\ref{sec:operations}), while higher-order
architectures change $\mathcal X$ itself by lifting the domain to tuples, subgraphs, or cells
(Section~\ref{sec:unified-view}). Table~\ref{tab:representative-slot-sampler} illustrates the
seven-component representation for twenty-one methods, and Table~\ref{tab:family-slot-map}
identifies each family's primary component.

\makeatletter
\@ifundefined{NC@find@Y}{\newcolumntype{Y}[1]{>{\raggedright\arraybackslash}p{#1}}}{}
\@ifundefined{NC@find@B}{\newcolumntype{B}[1]{>{\centering\arraybackslash$}p{#1}<{$}}}{}
\@ifundefined{NC@find@T}{\newcolumntype{T}[1]{>{\centering\arraybackslash}p{#1}}}{}
\makeatother

\newcolumntype{J}[1]{>{\raggedright\arraybackslash}m{#1}}
\newcolumntype{U}[1]{>{\centering\arraybackslash$}m{#1}<{$}}
\newcolumntype{N}[1]{>{\centering\arraybackslash}m{#1}}
\providecommand{\samplermethod}[2]{\makecell[l]{#1\\[-1pt]
{\scriptsize\citeauthor{#2}}\\[-1pt]{\scriptsize(\citeyear{#2})}}}
\makeatletter\@ifundefined{theSSU}{\newlength{\theSSU}}{}\makeatother
\setlength{\theSSU}{0.01\dimexpr\textwidth-42pt\relax}
\begingroup
\footnotesize
\renewcommand{\arraystretch}{1.45}
\arrayrulecolor{black!45}
\setlength{\arrayrulewidth}{0.5pt}
\setlength{\tabcolsep}{3pt}
\setlength{\LTpre}{6pt}
\setlength{\LTpost}{6pt}

\begin{longtable}{@{}
J{15\theSSU}
U{6\theSSU}
N{9\theSSU}
U{19\theSSU}
U{16\theSSU}
U{18\theSSU}
U{17\theSSU}
@{}}
\caption[Our seven-family taxonomy with three anchor methods per family]{
\textbf{Our seven-family taxonomy expressed in one common layer.}
Each colored band is one family, and its three anchor methods show how familiar architectures
fill the same seven components. Read each row from left to right: what objects are updated
($\mathcal X$), which channels are used ($\mathcal K$), where information moves
($\mathbf P_k$), what information moves ($\Psi_k$), how channel contributions are combined
($\boxplus$), and how the result updates the state ($(\mathcal B_\ell,\phi_\ell)$).
The six displayed columns combine the distinct ego/residual and update maps in the final
Ego/Update column. Full source-grounded decompositions and conventions appear in
Table~\ref{tab:family1_spatial_slots}--Table~\ref{tab:family-geometric} and
Sections~\ref{subsec:setup}--\ref{subsec:graded-extension}. Here $\mathrm{cg}$ denotes the
Clebsch--Gordan tensor product.}
\label{tab:representative-slot-sampler}\\[6pt]
\toprule
\textbf{Method} &
\multicolumn{1}{N{6\theSSU}}{\centering$\boldsymbol{\mathcal{X}}$} &
\multicolumn{1}{N{9\theSSU}}{\centering$\boldsymbol{\mathcal{K}}$} &
\multicolumn{1}{N{19\theSSU}}{\centering\makecell[c]{\textbf{Propagation}\\$\boldsymbol{\mathbf{P}_k^{(\ell)}}$}} &
\multicolumn{1}{N{16\theSSU}}{\centering\makecell[c]{\textbf{Message}\\$\boldsymbol{\Psi_k^{(\ell)}}$}} &
\multicolumn{1}{N{18\theSSU}}{\centering\makecell[c]{\textbf{Mixing}\\$\boldsymbol{\boxplus^{(\ell)}}$}} &
  \multicolumn{1}{N{17\theSSU}}{\centering\makecell[c]{\textbf{Ego/Update}\\$\boldsymbol{(\mathcal B_\ell,\phi_\ell)}$}} \\
\midrule
\endfirsthead

\multicolumn{7}{c}{\tablename\ \thetable\ -- continued} \\[6pt]
\toprule
\textbf{Method} &
\multicolumn{1}{N{6\theSSU}}{\centering$\boldsymbol{\mathcal{X}}$} &
\multicolumn{1}{N{9\theSSU}}{\centering$\boldsymbol{\mathcal{K}}$} &
\multicolumn{1}{N{19\theSSU}}{\centering\makecell[c]{\textbf{Propagation}\\$\boldsymbol{\mathbf{P}_k^{(\ell)}}$}} &
\multicolumn{1}{N{16\theSSU}}{\centering\makecell[c]{\textbf{Message}\\$\boldsymbol{\Psi_k^{(\ell)}}$}} &
\multicolumn{1}{N{18\theSSU}}{\centering\makecell[c]{\textbf{Mixing}\\$\boldsymbol{\boxplus^{(\ell)}}$}} &
\multicolumn{1}{N{17\theSSU}}{\centering\makecell[c]{\textbf{Ego/Update}\\$\boldsymbol{(\mathcal B_\ell,\phi_\ell)}$}} \\
\midrule
\endhead

\midrule
\multicolumn{7}{r}{\textit{continued on next page}} \\[4pt]
\endfoot

\bottomrule
\endlastfoot

\rowcolor{repSpatial}
\multicolumn{7}{@{}l}{\textcolor{white}{\textbf{1 Spatial}\quad
\emph{\scriptsize local support in $\mathbf P_k$ with differences in aggregation and ego/update}}} \\
\hline
\rowcolor{repSpatial!8}
\samplermethod{GCN}{kipf2017semi} & V & $\{1\}$ &
\hat{\mathbf A} & \mathbf H^{(\ell)}\mathbf W^{(\ell)} & \mathbf P_1\Psi_1 &
\sigma(\mixmsg^{(\ell)}) \\
\hline
\rowcolor{repSpatial!8}
\samplermethod{GraphSAGE-\\Mean}{hamilton2017inductive} & V & $\{1\}$ &
\mathbf D^{-1}\mathbf A & \mathbf H^{(\ell)} & \mathbf P_1\Psi_1 &
\multicolumn{1}{N{17\theSSU}}{\makecell[c]{$\normop(\sigma([\mathbf H^{(\ell)}\Vert$\\$\mixmsg^{(\ell)}]\mathbf W^{(\ell)}))$}} \\
\hline
\rowcolor{repSpatial!8}
\samplermethod{GIN}{xu2019powerful} & V & $\{1\}$ &
\mathbf A & \mathbf H^{(\ell)} & \mathbf P_1\Psi_1 &
\multicolumn{1}{N{17\theSSU}}{\makecell[c]{$\MLP^{(\ell)}((1{+}\epsilon_\ell)\mathbf H^{(\ell)}{+}$\\$\mixmsg^{(\ell)})$}} \\

\rowcolor{repAttention}
\multicolumn{7}{@{}l}{\textcolor{white}{\textbf{2 Attention}\quad
\emph{\scriptsize state-dependent $\alpha_{ij}(\mathbf H^{(\ell)})$ in $\mathbf P_k$}}} \\
\hline
\rowcolor{repAttention!9}
\samplermethod{GAT}{velickovic2018gat} & V & $\{1,\dots,M\}$ &
\multicolumn{1}{N{19\theSSU}}{\makecell[c]{$\sm_{j\in\widetilde{\mathcal N}(i)}\!\big($\\$\LReLU([\vh_i\mathbf W_m\Vert$\\$\vh_j\mathbf W_m]\mathbf a_m)\big)$}} &
\mathbf H^{(\ell)}\mathbf W_m^{(\ell)} &
\multicolumn{1}{N{18\theSSU}}{\makecell[c]{$\mathbf C_m=\mathbf P_m\Psi_m;$\\$\mixmsg=\Vert_m\mathbf C_m$\\{\scriptsize(avg. at final layer)}}} &
\multicolumn{1}{N{17\theSSU}}{\makecell[c]{$\mathcal B_\ell=0;$\\$\sigma(\mixmsg^{(\ell)})$}} \\
\hline
\rowcolor{repAttention!9}
\samplermethod{GaAN}{zhang2018gaan} & V & $\{1,\dots,M\}$ &
\multicolumn{1}{N{19\theSSU}}{\makecell[c]{$\sm_{j\in\mathcal N(i)}\!\langle$\\$\vh_i\mathbf W_{xa}^{(m)},$\\$\vh_j\mathbf W_{za}^{(m)}\rangle$}} &
\FC_{\theta_v^{(m)}}^h(\vh_j) &
\multicolumn{1}{N{18\theSSU}}{\makecell[c]{$\mathbf C_{m,i}=\mathbf P_m\Psi_m;$\\$\mixmsg_i=\Vert_m g_i^{(m)}\mathbf C_{m,i}$}} &
\multicolumn{1}{N{17\theSSU}}{\makecell[c]{$\mathcal B_{\ell,i}=\vh_i;$\\$\mathbf g_i=\sigma(\FC(\mathbf q_i));$\\$\FC(\mathcal B_{\ell,i}\Vert\mixmsg_i)$}} \\
\hline
\rowcolor{repAttention!9}
\samplermethod{MAGNA}{wang2021magna} & V &
\multicolumn{1}{N{9\theSSU}}{\shortstack[t]{$\{(m,k):$\\$1\le m\le M,$\\$0\le k\le K\}$}} &
\multicolumn{1}{N{19\theSSU}}{\makecell[c]{$c_k(\sm(\mathbf S_m^{(\ell)}))^k;$\\$c_{k<K}=\alpha(1{-}\alpha)^k,$\\$c_K=(1{-}\alpha)^K$}} & \LN(\mathbf H^{(\ell)}) & \multicolumn{1}{N{18\theSSU}}{\makecell[c]{$\mathbf z_m=\sum_{k=0}^K\mathbf P_{m,k}\Psi_{m,k};$\\$\mixmsg=(\Vert_m\mathbf z_m)\mathbf W_o$}} &
\multicolumn{1}{N{17\theSSU}}{\makecell[c]{$\mathbf R=\mathbf H^{(\ell)}+\mixmsg;$\\$\mathbf R+\operatorname{FFN}(\LN\mathbf R)$}} \\

\rowcolor{repSpectral}
\multicolumn{7}{@{}l}{\textcolor{white}{\textbf{3 Spectral}\quad
\emph{\scriptsize polynomial/spectral basis with coefficients folded into $\mathbf P_k$}}} \\
\hline
\rowcolor{repSpectral!8}
\samplermethod{ChebNet}{defferrard2016convolutional} & V &
\multicolumn{1}{N{9\theSSU}}{\makecell[c]{$\{0,\ldots,$\\$K{-}1\}$}} &
\multicolumn{1}{N{19\theSSU}}{\makecell[c]{$T_k(\tilde{\mathbf L}_{\mathrm{ch}})$\\{\scriptsize($[-1,1]$-rescaled $\mathbf L_{\rm sym}$)}}} & \mathbf H^{(\ell)}\boldsymbol\Theta_k & \textstyle\sum_k \mathbf P_k\Psi_k &
\sigma(\mixmsg^{(\ell)}) \\
\hline
\rowcolor{repSpectral!8}
\samplermethod{APPNP}{klicpera2019predict} & V & $\{0,\ldots,K\}$ &
\multicolumn{1}{N{19\theSSU}}{\makecell[c]{$\mathbf P_k=$\\$\begin{cases}\alpha(1{-}\alpha)^k\hat{\mathbf A}^k,&k<K,\\(1{-}\alpha)^K\hat{\mathbf A}^K,&k=K\end{cases}$}} & f_\theta(\mathbf X) & \textstyle\sum_k \mathbf P_k\Psi_k &
\multicolumn{1}{N{17\theSSU}}{\makecell[c]{$\mixmsg^{(\ell)}$\\{\scriptsize final $\sm$}\\{\scriptsize external}}} \\
\hline
\rowcolor{repSpectral!8}
\samplermethod{GPR-GNN}{chien2021adaptive} & V & $\{0,\ldots,K\}$ &
\multicolumn{1}{N{19\theSSU}}{\makecell[c]{$\gamma_k\hat{\mathbf A}^k$\\{\scriptsize(learned, signed)}}} & f_\theta(\mathbf X) & \textstyle\sum_k \mathbf P_k\Psi_k &
\multicolumn{1}{N{17\theSSU}}{\makecell[c]{$\mixmsg^{(\ell)}$\\{\scriptsize final $\sm$}\\{\scriptsize external}}} \\

\rowcolor{repTransformer}
\multicolumn{7}{@{}l}{\textcolor{white}{\textbf{4 Graph Transformer}\quad
\emph{\scriptsize global/sparsified attention in $\mathbf P_k$}}} \\
\hline
\rowcolor{repTransformer!8}
\samplermethod{Graphormer}{ying2021transformers} &
\multicolumn{1}{N{6\theSSU}}{\makecell[c]{$V\cup$\\$\{v_g\}$\\${}+\deg$}} & heads &
\multicolumn{1}{N{19\theSSU}}{\makecell[c]{$\sm(\mathbf Q_m\mathbf K_m^\top/\sqrt d{+}$\\$b_{\phi(i,j)}+c_{ij}^{\rm SP\text{-}edge})$}} & \mathbf H^{(\ell)}\mathbf W_m^V & \Vert_m \mathbf P_m\Psi_m &
\multicolumn{1}{N{17\theSSU}}{\makecell[c]{$H'=$\\$\mathrm{MHA}(\LN H)+H;$\\[2pt]$H^+=$\\$\mathrm{FFN}(\LN H')+H'$}} \\
\hline
\rowcolor{repTransformer!8}
\samplermethod{GPS$^\star$}{rampavsek2022recipe} &
\multicolumn{1}{N{6\theSSU}}{\makecell[c]{$V$\\${}+\mathrm{PE}$\\${}/\mathrm{SE}$}} & lo,gl &
\multicolumn{1}{N{19\theSSU}}{\makecell[c]{$\mathrm{MPNN}_e($\\$\mathbf H,\mathbf E,\mathbf A);$\\[2pt]$\mathrm{GlobalAttn}(\mathbf H)$}} &
\multicolumn{1}{N{16\theSSU}}{\makecell[c]{$\hat{\mathbf X}_M,$\\$\hat{\mathbf X}_T$}} &
\multicolumn{1}{N{18\theSSU}}{\makecell[c]{$\mathbf H_M^{(\ell+1)}$\\${}+\mathbf H_T^{(\ell+1)}$}} &
\multicolumn{1}{N{17\theSSU}}{\makecell[c]{branch res $+$ BN;\\[2pt]$\MLP+$ res $+$ BN}} \\
\hline
\rowcolor{repTransformer!8}
\samplermethod{NodeFormer}{Wu2022NodeFormer} & V & $\scriptstyle\{1{:}K\}_{\rm G}\oplus\{\rm rel\}_{\rm opt}$ &
\multicolumn{1}{N{19\theSSU}}{\makecell[c]{$\tilde\phi_g(\mathbf Q/\sqrt\tau)\tilde\phi_g(\mathbf K/\sqrt\tau)^\top/Z;$\\$\mathbf P_{\rm rel}=\sigma(b^{(\ell)})\mathbf A$}} & \mathbf H^{(\ell)}\mathbf W^V &
\multicolumn{1}{N{18\theSSU}}{\makecell[c]{$K^{-1}\sum_r\mathbf P_r\Psi$\\${}+\mathbf P_{\rm rel}\Psi$}} & \mathbf H^{(\ell+1)}=\mixmsg^{(\ell)} \\

\rowcolor{repHetero}
\multicolumn{7}{@{}l}{\textcolor{white}{\textbf{5 Heterogeneous}\quad
\emph{\scriptsize $\mathcal K$ indexes relations/meta-paths with per-type $\mathbf P_k$}}} \\
\hline
\rowcolor{repHetero!8}
\samplermethod{R-GCN}{schlichtkrull2018modeling} & V & $\mathcal R$ &
\multicolumn{1}{N{19\theSSU}}{\makecell[c]{$\tfrac{1}{c_{i,r}}\mathbf A_r$\\{\scriptsize(per-relation}\\{\scriptsize normalization)}}} & \mathbf H^{(\ell)}\mathbf W_r &
\textstyle\sum_{k\in\mathcal K} \mathbf P_k\Psi_k &
\sigma(\mathbf H^{(\ell)}\mathbf W_0{+}\mixmsg^{(\ell)}) \\
\hline
\rowcolor{repHetero!8}
\samplermethod{HAN}{wang2019heterogeneous} & V &
\multicolumn{1}{N{9\theSSU}}{\makecell[c]{$\{(\Phi,k):$\\$\Phi\in\mathcal P,$\\$1\le k\le K\}$}} &
\multicolumn{1}{N{19\theSSU}}{\makecell[c]{$\alpha_{ij}^{\Phi,k}(\mathbf H)\mathbf A_\Phi$\\{\scriptsize(softmax over $j\in\mathcal N_i^\Phi$)}}} &
\multicolumn{1}{N{16\theSSU}}{\makecell[c]{$\mathbf H^{(\ell)}\mathbf M_{\varphi}$\\{\scriptsize(by node type $\varphi$)}}} &
\multicolumn{1}{N{18\theSSU}}{\makecell[c]{$\textstyle\sum_{\Phi} \beta_\Phi\,$\\$\Vert_k\sigma(\mathbf P_{\Phi,k}\Psi_{\Phi,k})$}} &
\multicolumn{1}{N{17\theSSU}}{\makecell[c]{$\mathcal B_\ell=0;$\\$\phi_\ell(\mixmsg)=\mixmsg$}} \\
\hline
\rowcolor{repHetero!8}
\samplermethod{HGT}{hu2020heterogeneous} & V & $\{m:1\le m\le h\}$ &
\multicolumn{1}{N{19\theSSU}}{\makecell[c]{$\boldsymbol\alpha^{(m)}_{s,e,t}$\\{\scriptsize(meta-relation}\\{\scriptsize conditioned)}}} &
\multicolumn{1}{N{16\theSSU}}{\makecell[c]{$\operatorname{M\text{-}Linear}^{m}_{\tau_s}$\\$(\mathbf H_s^{(\ell)})\mathbf W_r^{\rm MSG,m}$}} &
\multicolumn{1}{N{18\theSSU}}{\makecell[c]{$\mathbf C_{m,t}=\sum_{s\in\mathcal N(t)}$\\$\alpha^{(m)}_{s,e,t}\Psi^{(m)}_{s,e,t};$\\$\mixmsg_t=\Vert_m\mathbf C_{m,t}$}} &
\mathbf W^A_{\tau_t}\sigma(\mixmsg_t){+}\mathbf H_t \\

\rowcolor{repHigher}
\multicolumn{7}{@{}l}{\textcolor{white}{\textbf{6 Higher-order}\quad
\emph{\scriptsize domain $\mathcal X$ beyond nodes: tuples/subgraphs/cells}}} \\
\hline
\rowcolor{repHigher!8}
\samplermethod{PPGN}{maron2019provably} & V^2 & $\{c=1{:}b\}$ &
\multicolumn{1}{N{19\theSSU}}{\makecell[c]{$\MLP_1(\mathbf H^{(\ell)})$\\{\scriptsize(state-dependent}\\{\scriptsize operator)}}} & \MLP_2(\mathbf H^{(\ell)}) &
\mathbf P_\times\Psi_\times & \multicolumn{1}{N{17\theSSU}}{\makecell[c]{$[\MLP_3(\mathbf H^{(\ell)})\Vert$\\$\mixmsg^{(\ell)}]$\\{\scriptsize($\MLP_3$ gives $\mathcal B_\ell$)}}} \\
\hline
\rowcolor{repHigher!8}
\samplermethod{NGNN$^\ddagger$}{zhang2021nested} & \mathcal S & $\{1\}$ &
\mathbf A_{\mathcal S_i} &
\multicolumn{1}{N{16\theSSU}}{\makecell[c]{$M_t(\vh_{v,G_w^h}^t,$\\$\vh_{u,G_w^h}^t,e_{vu})$}} &
\multicolumn{1}{N{18\theSSU}}{\makecell[c]{$\displaystyle\sum_{u\in N(v\mid G_w^h)}$\\$\mathbf P_{vu}\Psi_{vu}$}} &
\multicolumn{1}{N{17\theSSU}}{\makecell[c]{$U_t(\vh_{v,G_w^h}^t,$\\$\mixmsg_v^{t+1});$\\$R_0\text{ then }R_1\text{ pooling}$}} \\
\hline
\rowcolor{repHigher!8}
\samplermethod{MPSN}{bodnar2021simplicial} & \bigsqcup_p X_p & $\scriptstyle\{\mathrm B,\mathrm C,\downarrow,\uparrow\}$ &
\multicolumn{1}{N{19\theSSU}}{\makecell[c]{$B(\sigma),C(\sigma),$\\$N_\downarrow(\sigma),N_\uparrow(\sigma)$}} & M_{\mathrm B},M_{\mathrm C},M_\downarrow,M_\uparrow &
\multicolumn{1}{N{18\theSSU}}{\makecell[c]{$[m_{\mathrm B}\Vert m_{\mathrm C}$\\${}\Vert m_\downarrow\Vert m_\uparrow]$}} &
U(\vh_\sigma^{(\ell)},\mixmsg_\sigma^{(\ell)}) \\

\rowcolor{repGeometric}
\multicolumn{7}{@{}l}{\textcolor{white}{\textbf{7 Geometric}\quad
\emph{\scriptsize $\Psi_k,\phi$ respect a geometric symmetry group}}} \\
\hline
\rowcolor{repGeometric!8}
\samplermethod{SchNet}{schutt2017schnet} & {\scriptstyle V;r_{ij}} & $\{1\}$ &
1 {\scriptstyle(\text{all pairs})} & (\vh_j\mathbf W_{\rm in})\odot W_\theta(\mathrm{RBF}(r_{ij})) & \mathrm{id}\ {\scriptstyle(C_{1,i}=\sum_jP_{ij}\Psi_{ij})} &
\mathbf H^{(\ell)}{+}\MLP(\mixmsg^{(\ell)}) \\
\hline
\rowcolor{repGeometric!8}
\samplermethod{EGNN}{satorras2021en} &
\multicolumn{1}{N{6\theSSU}}{\makecell[c]{$V;$\\$(\vh_i,x_i)$}} & $\{h,x\}$ &
\multicolumn{1}{N{19\theSSU}}{\makecell[c]{$P_{h,ij}=$\\$\indicator[j\in\mathcal N(i)];$\\[2pt]$P_{x,ij}=$\\$C\indicator[j\ne i]$\\${}\cdot\phi_x(m_{ij})$}} &
\multicolumn{1}{N{16\theSSU}}{\makecell[c]{$\Psi_{h,ij}=m_{ij}=$\\$\phi_e(\vh_i,\vh_j,$\\$\|x_i{-}x_j\|^2,\ve_{ij});$\\[2pt]$\Psi_{x,ij}=$\\$x_i-x_j$}} &
\mathrm{id}\ {\scriptstyle(C_h,C_x\text{ typed})} & \multicolumn{1}{N{17\theSSU}}{\makecell[c]{$x_i'=x_i+\Delta x_i;$\\$\vh_i'=\phi_h(\vh_i,m_i)$}} \\
\hline
\rowcolor{repGeometric!8}
\samplermethod{SE(3)-Tr.}{fuchs2020se3} &
\multicolumn{1}{N{6\theSSU}}{\makecell[c]{$V;$\\$Y_{\ell m}(\hat r_{ij})$}} & $\{1,\dots,M\}$ &
\multicolumn{1}{N{19\theSSU}}{\makecell[c]{$\alpha^m_{ij}=$\\$\mathrm{softmax}_{j\in N_i\setminus i}$\\$\bigl(q_i^{m\top}k_{ij}^m\bigr)$}} &
\multicolumn{1}{N{16\theSSU}}{\makecell[c]{$\Psi^m_{ij}=$\\$\displaystyle\bigoplus_{\ell,k}$\\$W^{\ell k}_{V,m}(x_j-x_i)f_j^k$}} &
\displaystyle\big\Vert_m C_m &
\multicolumn{1}{N{17\theSSU}}{\makecell[c]{self-interaction $+$\\norm nonlinearity}} \\

\end{longtable}
\endgroup
\FloatBarrier

\begin{table*}[t!]
\centering
\footnotesize
\renewcommand{\arraystretch}{1.35}
\setlength{\tabcolsep}{4pt}
\caption[How our seven families occupy the unified layer]{\textbf{How our seven families occupy the unified layer.}
This table is the compact map behind our taxonomy: the bold \textbf{Defining axis} identifies
the component that most clearly separates each family, while the remaining cells summarize
its common fillings. Families~1--4 primarily change where information moves through
$\mathbf P_k$; Family~5 changes the relation or meta-path channels $\mathcal K$; Family~6
changes the updated domain $\mathcal X$; and Family~7 changes the geometric message and update
maps $\Psi_k$ and $\phi$. Thus the table can be read as a guide from a family name to the
part of Eq.~\eqref{eq:unified-notation} that it principally redesigns. Here
$V^k$, $\mathcal S$, and $\mathcal X_p$ denote tuples, subgraphs, and cells, respectively
(Section~\ref{subsec:graded-extension}).}
\label{tab:family-slot-map}
\makebox[\textwidth][c]{\resizebox{0.98\textwidth}{!}{
\begin{tabular}{@{}l L{2.5cm} L{1.6cm} L{2.0cm} L{2.5cm} L{2.4cm} L{2.5cm} L{1.8cm}@{}}
\toprule
\textbf{Family} & \textbf{Defining axis} & \textbf{Domain $\mathcal{X}$} & \textbf{Channels $\mathcal{K}$}
& \textbf{Propagation $\mathbf{P}_k$} & \textbf{Message $\Psi_k$}
& \textbf{Mixing $\boxplus$} & \textbf{Ego/Update $(\mathcal B_\ell,\phi_\ell)$} \\
\midrule
\rowcolor{googleblue!12}
\textbf{1 Spatial} & \textbf{$\mathbf{P}_k$: local support} & $V$ & one or more local supports & fixed sparse local operators &
linear or pairwise local messages & id. / sum / concat &
$\sigma$ / MLP / gate / residual \\
\rowcolor{googlered!10}
\textbf{2 Attention} & \textbf{$\mathbf{P}_k$: state-dependent} & $V$ & heads & $\boldsymbol{\alpha}(\mathbf{H}^{(\ell)})$
on local support & linear / pairwise values & concat / mean / gated mixing & $\sigma$ / MLP / residual \\
\rowcolor{googleyellow!18}
\textbf{3 Spectral} & \textbf{$\mathbf{P}_k$: polynomial basis} & $V$ & spectral order $k$ &
$g_k(\mathbf{S})$ / $\hat{\mathbf{A}}^k$ & $\mathbf{H}^{(\ell)}\boldsymbol{\Theta}_k$ / $f_\theta(\mathbf{X})$ &
$\sum_k\mathbf{P}_k\Psi_k$ (coefficients folded into $\mathbf P_k$) & id. / $\sigma$ / residual \\
\rowcolor{googlegreen!12}
\textbf{4 Graph Transformer} & \textbf{$\mathbf{P}_k$: global support} & $V$ (or tokens) & heads / tokens &
global or sparsified attention (optionally local--global) & projected values / branch outputs &
concat / sum / local--global mixing & residual + norm + MLP \\
\rowcolor{googleblue!12}
\textbf{5 Heterogeneous} & \textbf{$\mathcal{K}$: relations / meta-paths} & $V$ & relations / meta-paths &
typed adjacency or relation-conditioned attention & type-/relation-specific messages &
sum / concat / semantic attention & type-specific $\sigma$ / MLP / residual \\
\rowcolor{googlered!10}
\textbf{6 Higher-order} & \textbf{$\mathcal{X}$: lifted domain} & $V^k$ / $\mathcal{S}$ / $\mathcal{X}_p$ &
replacement / incidence / support types & lifted or inter-rank operators & inner-GNN / pairwise / equivariant messages &
sum / concat / typed mixing & lifted update / MLP / base GNN \\
\rowcolor{googlegreen!12}
\textbf{7 Geometric} & \textbf{$\Psi_k,\phi$: geometric equivariance} & $V$ / directed edges ($+$ geometry) & scalar / vector / tensor channels &
invariant scalar support or attention weights & $G$-equivariant tensor / geometric messages &
typed sum / direct sum / concat & $G$-equivariant update \\
\bottomrule
\end{tabular}}}
\end{table*}

We make the coverage claim checkable through step-by-step reductions of GCN
\citep{kipf2017semi}, GraphSAGE \citep{hamilton2017inductive}, GAT
\citep{velickovic2018gat}, and GIN \citep{xu2019powerful}, supplemented by representative
fillings across all seven families. Section~\ref{sec:unified-view} extends the same components
to state-dependent attention, multi-hop and spectral bases, global communication, relation
and meta-path channels, lifted domains, and geometric equivariant messages.

\paragraph{What the unification establishes.}
A flexible equation risks making coverage vacuous because the same computation can be
assigned to different components. We therefore fix a slot discipline that assigns each
elementary operation by output role. Scalar edge and channel weights enter $\mathbf{P}_k$,
vector- or matrix-valued messages enter $\Psi_k$, $\boxplus$ combines channel contributions,
$\mathcal{B}_\ell$ forms the ego or residual input, and $\phi_\ell$ performs the update. The
fixed discipline makes component-wise comparisons meaningful even when equivalent
factorizations exist. It determines entries up to stated redundancies, including channel-rank
non-identifiability in the linear regime (Proposition~\ref{prop:channel-rank}) and equivalent
iterative or polynomial forms.

The coverage boundary is equally explicit. Channels use either a matrix linear-value form
$\mathbf{P}_k\Psi_k$ or a pairwise message with permutation-invariant aggregation, followed
by permutation-equivariant mixing. Section~\ref{sec:unified-view} identifies methods outside
these primitives and the structural property responsible. These cases bound the framework
rather than establish impossibility.

Within this discipline, differing components locate recorded computational differences.
GraphSAGE and GIN differ in operator normalization and ego handling, while GCN and GAT differ
in fixed versus state-dependent scalar weights. Functional equivalence additionally depends on
internal mechanisms and component interactions beyond the displayed granularity. The fixed
coordinates nevertheless expose relationships obscured by paper-specific notation and reveal
combinations absent from the catalogued inventory. Section~\ref{sec:generation} uses those
combinations to generate structurally consistent candidates for empirical evaluation.

The available evidence also rules out a context-free ranking of component fillings. Rankings
depend on graph structure under heterophily \citep{zhu2020beyond,pei2020geom}, tuning
\citep{Tonshoff2023GapGo}, and evaluation protocol
\citep{luo2024classic,luo2025graphlevel,RubioMadrigal2026FAF}. These finite comparisons neither
establish a no-free-lunch theorem nor preclude dominance on a specified distribution class.
They motivate the unresolved problem of mapping measurable graph and task properties to
validated component choices (Section~\ref{sec:outlook}).

The paper proceeds from notation (Section~\ref{sec:preliminaries}) to the unified equation and
families (Section~\ref{sec:unified-view}), generated architectures (Section~\ref{sec:generation}),
propagation analysis (Section~\ref{sec:propagation}), benchmark interpretation
(Section~\ref{sec:benchmarks}), and open problems (Section~\ref{sec:outlook}). The family map
and detailed component tables provide the primary reference path through the paper.

\flushbottom
\subsection{Contributions}

The unified equation supports four contributions.

\begin{itemize}
    \item We define a \textbf{seven-component unification} of covered graph neural layers,
    together with a fixed slot discipline, a named coverage boundary, worked reductions of
    four canonical layers, and representative fillings across all seven families
    (Section~\ref{sec:unified-view}).

    \item We organize more than 200 catalogued architectures from over 400 cited works into
    \textbf{seven nonexclusive, component-anchored families}. The comparison tables record
    assignments for all seven components at a declared reporting granularity and distinguish
    architectural families from reusable operations and problem settings
    (Table~\ref{tab:family1_spatial_slots}--Table~\ref{tab:family-geometric}).

    \item We derive \textbf{component-level attribution and identifiability results}. Under
    node-local updates, operator support localizes one-layer dependencies and yields a necessary
    full-row condition for one-layer global mixing. In the linear additive regime, raw channel
    count is non-identifiable and minimum Kronecker channel rank provides the corresponding
    invariant. These are the central consequences of exposing the propagation bank and channel
    decomposition. Seventeen results connect these and related component interactions to
    oversmoothing, oversquashing, heterophily, expressivity, and permutation symmetry, with
    sixteen proofs in Appendix~\ref{sec:analysis}.
    Section~\ref{sec:benchmarks} records documented benchmark properties and evaluation findings,
    then translates them descriptively into the component vocabulary while reserving causal
    attribution for controlled evidence.

    \item We treat the component inventory as a \textbf{structured design space}. It supports
    the generation of well-formed architectures (Section~\ref{sec:generation}) and exposes
    the inverse problem of selecting fillings from measurable graph, task, and deployment
    properties (Section~\ref{sec:outlook}).
\end{itemize}

\subsection{Scope and Terminology}
\label{subsec:scope}

We use \emph{component} and \emph{slot} interchangeably for one coordinate of the unified
equation. A \emph{filling} is a concrete choice for that coordinate. A compatible assignment
to all seven components specifies a layer at the declared reporting granularity. Within the
broader graph-learning stack, an
\emph{architectural family} groups layers by a primary pattern of component variation. A
family-agnostic \emph{operation}, such as a positional encoding, rewiring, skip connection,
normalization, or pooling, is a reusable module with a primary component or external location
and may affect other components (Appendix~\ref{sec:operations}). A \emph{problem setting} or
\emph{training paradigm} instead specifies data, task, or loss without fixing the layer.
Architectural families organize the unification, operations are treated separately, and
problem settings and training paradigms remain outside its primary scope. Temporal graphs and
pre-training may add architectural machinery, but are treated primarily as a problem setting
and training paradigm, respectively.

\paragraph{Architectural families.}
Table~\ref{tab:family-slot-map} groups the seven families by their primary component. Spatial message-passing, attention, spectral/polynomial, and Graph Transformer families specialize $\mathbf{P}_k$, whereas heterogeneous/multi-relational, higher-order/subgraph, and geometric equivariant families vary $\mathcal K$, $\mathcal X$, and jointly $\Psi_k,\phi$, respectively. Section~\ref{sec:unified-view} introduces each family, Appendix~\ref{app:family-comparison-tables} provides the tables, and Section~\ref{sec:propagation} further studies the propagation families.

\paragraph{Operations and diagnostics.}
Appendix~\ref{sec:operations} maps positional/structural encodings, rewiring, virtual nodes, skip connections, normalization, pooling, continuous-depth/ODE views, and mini-batch sampling to their primary components or external locations. Appendix~\ref{sec:analysis} gives sixteen proofs concerning expressivity, locality/globality of $\mathbf{P}_k$, spectral filtering, oversmoothing, heterophilic realizability, and permutation equivariance. The immediate concatenation corollary is proved with its statement in Section~\ref{sec:unified-view}. Section~\ref{sec:propagation} uses these results.

\paragraph{Scope beyond the base-layer equation.}
The seven-component equation describes single-layer graph computation rather than complete
training or deployment pipelines. Recurrent and fixed-point GNNs
\citep{scarselli2009graph,li2016gated} largely overlap the MPNN framework
\citep{gilmer2017neural} and are treated elsewhere. Graph autoencoders and generative models
center on decoders and generation processes, so we cover only their encoders when these belong
to the seven families. Temporal and dynamic graph networks are primarily problem settings,
despite often adding memory, message, and update machinery. Pre-training objectives, graph
generative decoders, language-model coupling, retrieval systems, and multi-agent control
therefore remain outside the formal scope of the equation. Section~\ref{sec:outlook}
nevertheless uses the component vocabulary to formulate open questions about temporal memory,
cross-domain transfer, retrieval-grounded generation, personalization, reasoning graphs, and
graph-structured agent systems, as well as weight-space model editing. These discussions
identify possible extensions and do not
claim that the seven-component equation represents the complete surrounding system. Their
constituent GNN layers remain representable when they satisfy our primitives. We cover what a
layer computes, not its training loss, enclosing task pipeline, or deployment system.
Adaptation to data structure within the heterogeneous and geometric families remains in scope
because it is a layer property.

\section{Preliminaries}
\label{sec:preliminaries}

This section fixes the notation and definitions used throughout the paper.

\subsection{Basic Graph-Theoretic Objects}
\label{subsec:graph-theory-prelim}

A graph $\mathcal{G}=(\mathcal{V},\mathcal{E})$ consists of a node set $\mathcal{V}$ and an edge set $\mathcal{E}$. Nodes represent entities, and edges represent relations between them. We write $n=|\mathcal{V}|$ for the number of nodes and $m=|\mathcal{E}|$ for the number of edges.

Graphs can be \emph{undirected} (edges are symmetric, i.e., $(u,v)$ is the same as $(v,u)$) or \emph{directed} (order matters). They can be \emph{unweighted} (edges are present or absent) or \emph{weighted} (each edge $(u,v)$ has a weight $w_{uv}$).

The adjacency matrix $\mathbf{A}\in\mathbb{R}^{n\times n}$ encodes connectivity. Specifically, $A_{uv}=1$ if $(u,v)\in\mathcal{E}$ (unweighted), or $A_{uv}=w_{uv}$ (weighted). For undirected graphs, $\mathbf{A}$ is symmetric. The neighborhood of node $v$ is $\mathcal{N}(v)$, with degree $d_v=|\mathcal{N}(v)|$ in the unweighted case and $d_v=\sum_u A_{vu}$ for weighted graphs. Directed graphs use separate in- and out-degree conventions. The degree matrix $\mathbf{D}$ is diagonal with $D_{vv}=d_v$.

For normalized \emph{propagation} operators, inverse normalizations $\mathbf{D}^{-1}$ and $\mathbf{D}^{-1/2}$ adopt the convention that isolated nodes ($d_v=0$) contribute $0$. The spectral statements below assume no isolated nodes, so that $\mathbf{D}^{-1/2}$ is invertible and the normalized Laplacian's null space tracks connected components. Isolates, if present, are dropped from the spectral discussion.

Self-loops connect each node to itself. With self-loops, the adjacency becomes $\tilde{\mathbf{A}}=\mathbf{A}+\mathbf{I}$ with corresponding degree matrix $\tilde{\mathbf{D}}$. A normalized adjacency is $\hat{\mathbf{A}}=\tilde{\mathbf{D}}^{-1/2}\tilde{\mathbf{A}}\tilde{\mathbf{D}}^{-1/2}$. Normalization prevents high-degree nodes from dominating aggregation. The self-loop-free normalized adjacency is $\bar{\mathbf{A}}=\mathbf{D}^{-1/2}\mathbf{A}\mathbf{D}^{-1/2}$, used below in spectral analysis.

The combinatorial (unnormalized) graph Laplacian is $\mathbf{L}=\mathbf{D}-\mathbf{A}$, and the symmetric normalized Laplacian is $\mathbf{L}_{\mathrm{sym}}=\mathbf{I}-\bar{\mathbf{A}}=\mathbf{I}-\mathbf{D}^{-1/2}\mathbf{A}\mathbf{D}^{-1/2}$. Unless stated otherwise, all spectral statements refer to $\mathbf{L}_{\mathrm{sym}}$; its spectrum, eigenbasis, and the resulting high-pass/low-pass filtering language are developed in Section~\ref{subsec:spectral-spatial-prelim}.

When a model uses the self-loop-normalized support $\hat{\mathbf{A}}$ instead, we write the reference propagation Laplacian as $\mathbf{L}_{\mathrm{ref}}=\mathbf{I}-\hat{\mathbf{A}}$, and distinguish it explicitly. For an undirected graph with nonnegative weights, $\mathbf{L}_{\mathrm{ref}}=\mathbf{I}-\hat{\mathbf{A}}$ is symmetric positive semidefinite with spectrum in $[0,2)$; the self-loop on every node rules out a bipartite component, and hence rules out the eigenvalue $2$ itself. $\mathbf{L}_{\mathrm{ref}}$ therefore admits its own exact orthonormal eigendecomposition and Fourier picture. Adding self-loops gives it a generally \emph{different} eigenbasis from $\mathbf{L}_{\mathrm{sym}}$. The two can still share eigenvectors in special cases, notably regular graphs, where the normalizations differ only by a constant. The clean Fourier picture developed in Section~\ref{subsec:spectral-spatial-prelim} is stated for $\mathbf{L}_{\mathrm{sym}}$, and the same construction applies verbatim to $\mathbf{L}_{\mathrm{ref}}$ in its own eigenbasis.

Real-world graphs are typically sparse ($m\ll n^2$). Sparsity makes message passing tractable, since naive dense propagation over all $n^2$ pairs is usually infeasible. This is also why graph Transformers, which attend over all pairs \citep{ying2021transformers,rampavsek2022recipe}, often require scaling mechanisms such as sparsification or kernelization, as discussed under the Graph Transformer family (Section~\ref{sec:unified-view}).

\subsection{Graph Data and Notation}
\label{subsec:graph-data-notation}

Node features are stored in $\mathbf{X}\in\mathbb{R}^{n\times d}$, where row $\vx_v$ is the feature vector of node $v$. Edge features are denoted by $\ve_{uv}$ or a tensor $\mathbf{E}$. Some tasks also use graph-level features $\vx_{\mathcal{G}}$.

Labels depend on the task; for node-level tasks, node labels are $\mathbf{Y}\in\mathbb{R}^{n\times c}$.

Hidden representations at layer $\ell$ are $\mathbf{H}^{(\ell)}$, with $\mathbf{H}^{(0)}=\mathbf{X}$ by default (some models instead take $\mathbf{H}^{(0)}$ to be a learned embedding of $\mathbf{X}$). A GNN layer transforms $\mathbf{H}^{(\ell)}$ to $\mathbf{H}^{(\ell+1)}$ by combining feature transformations with graph-structured propagation. The final output before prediction is $\mathbf{Z}$; predicted labels are $\hat{\mathbf{Y}}$.

In most node-level models the update domain is the node set, written $\mathcal X=\mathcal V$, so rows of $\mathbf{H}^{(\ell)}$ are indexed by nodes. In the unified notation, $\mathcal X$ denotes the update domain more generally. Higher-order models may index rows by tuples, rooted subgraphs, simplices, cells, or motifs. Geometric models may attach scalar, vector, or tensor features and coordinates to each object. We keep $\mathbf{X}$ for input features and $\mathcal X$ for the domain to avoid ambiguity. The per-family comparison tables render this same node-set domain value as plain $V$ instead of $\mathcal V$, a typographic compression for column width, not a separate object.

\subsection{Learning Tasks on Graphs}
\label{subsec:graph-learning-tasks}

Graph learning spans several task types that differ in their prediction target. This paper's
layer designs primarily target the first five below; the last two (generation and
temporal/dynamic tasks) are problem settings we note only for completeness
(Section~\ref{subsec:scope}).

\paragraph{Node classification and regression.}
Predict a label or continuous value for each node, e.g., semi-supervised document or user
classification~\citep{kipf2017semi,velickovic2018gat}.

\paragraph{Edge and link prediction.}
Predict whether an edge exists, its type, or an associated value, e.g., knowledge-graph
completion~\citep{schlichtkrull2018modeling,rossi2021knowledge}.

\paragraph{Graph classification and regression.}
Predict a label or value for an entire graph, e.g., molecular property
prediction~\citep{gilmer2017neural,hu2020ogb}.

\paragraph{Community detection and clustering.}
Identify groups of nodes with similar roles or dense internal
connectivity~\citep{wang2019daegc}.

\paragraph{Combinatorial optimization on graphs.}
Predict or construct a combinatorial structure that optimizes a graph-defined objective, e.g.,
the traveling salesman problem or minimum vertex cover~\citep{khalil2017learning}, or influence
maximization.

\paragraph{Graph generation.}
Generate new graphs or components satisfying desired properties, e.g., molecule or
layout synthesis~\citep{yan2024swingnn}.

\paragraph{Temporal and dynamic graph tasks.}
When nodes, edges, features, or labels change over time, tasks include forecasting future
links, predicting event times, and learning adaptive
representations~\citep{rossi2020temporal,huang2023tgb}.

As Section~\ref{subsec:scope} makes explicit, graph generation is itself a task and
temporal/dynamic graphs are a problem setting, not a layer design.

\subsection{Homophily and Heterophily}
\label{subsec:homophily-prelim}

Homophily means connected nodes tend to be similar. Heterophily means they tend to differ. In label settings, homophily means adjacent nodes often share the same class. The distinction strongly affects which propagation operators perform well, since standard GNN propagation implicitly assumes homophily and degrades when it is absent \citep{pei2020geom,zhu2020beyond}.

A common measure \citep{zhu2020beyond}, defined when $|\mathcal{E}|>0$, is the fraction of edges connecting same-label nodes. Formally, $h_{\mathrm{edge}} = \frac{1}{|\mathcal{E}|}\sum_{(u,v)\in\mathcal{E}}\mathbb{1}\{y_u=y_v\}$. Values near one indicate strong homophily. Values near zero indicate few same-label edges, though this does not automatically mean strong heterophily without considering class balance.

Node-level homophily \citep{pei2020geom} averages this locally over the non-isolated nodes $\mathcal{V}_{+}=\{v:|\mathcal{N}(v)|>0\}$. This gives $h_{\mathrm{node}}=\frac{1}{|\mathcal{V}_{+}|}\sum_{v\in\mathcal{V}_{+}}\frac{1}{|\mathcal{N}(v)|}\sum_{u\in\mathcal{N}(v)}\mathbb{1}\{y_u=y_v\}$. Feature homophily can be measured using similarity functions, e.g., cosine similarity, $h_{\mathrm{feat}}=\frac{1}{|\mathcal{E}|}\sum_{(u,v)\in\mathcal{E}}\mathrm{sim}(\vx_u,\vx_v)$. Edge homophily is also sensitive to the number and balance of classes, which makes raw comparisons across datasets unreliable; adjusted homophily corrects for this and is comparable across datasets. Label informativeness instead tracks how much structure helps GNN performance \citep{platonov2023characterizing} (Section~\ref{sec:outlook}).

These are descriptive statistics, not fixed properties of a graph. A graph can be label-homophilic but feature-heterophilic, and homophily can vary across classes and neighborhoods. They are also global averages: a single graph routinely contains both homophilic and heterophilic neighborhoods at once, so a single graph-level ratio can obscure node-level mixture. A layer's propagation should accommodate that mixture rather than treat the graph as uniformly homophilic or heterophilic.

\subsection{Message Passing Intuition}
\label{subsec:message-passing-prelim}

The core of many GNNs is message passing \citep{gilmer2017neural}. Each node collects information from its neighbors, transforms it, aggregates it, and updates its representation. A single layer has the form
\begin{equation}
\vh_v^{(\ell+1)}=\mathrm{Update}^{(\ell)}\!\left(\vh_v^{(\ell)},\mathrm{Aggregate}^{(\ell)}\{\mathrm{Message}^{(\ell)}(\vh_v^{(\ell)},\vh_u^{(\ell)},\ve_{uv}):u\in\mathcal{N}(v)\}\right).
\label{eq:message-passing-prelim}
\end{equation}
After one layer, a node has information from its immediate neighbors. After $L$ layers, its receptive field can include nodes within $L$ hops. This changes for architectures with multi-hop propagation, graph rewiring, or global tokens, which let information travel farther than one hop per layer (Section~\ref{sec:unified-view} and Appendix~\ref{sec:operations} develop these mechanisms).

Repeated propagation can cause several problems. \emph{Oversmoothing} describes representations that become too similar after many steps \citep{li2018deeper,oono2020expressive}. \emph{Oversquashing} describes information compressed through graph bottlenecks \citep{Topping2022Curvature,DiGiovanni2023OverSquashing}. \emph{Long-range dependencies} arise when predictions depend on distant nodes. Many later architectures respond by modifying one specific part of the layer (propagation, aggregation, or the graph structure itself) to control this flow. Section~\ref{sec:propagation} analyzes these phenomena in detail; the next subsection names each such part of the layer precisely.

\subsection{Layer Slots in the Unified Equation}
\label{subsec:unified-slots-prelim}

To compare architectures that change only one part of the same underlying computation, we
give each modifiable part of the layer a name. A \emph{component} (or slot) is such a named
position in the layer equation. The domain $\mathcal X$ and channel index set $\mathcal K$ are
sets; the remaining components are operators or maps. Different architectures fill these
positions differently, and a family is defined by the component it primarily specializes.

Eq.~\eqref{eq:message-passing-prelim} above is the node-local case of the unified
equation, Eq.~\eqref{eq:unified-notation}, whose compact top-level form is
$\bar{\mathbf{H}}^{(\ell+1)}=\phi_\ell\big(\mathcal{B}_\ell(\mathbf{H}^{(\ell)},\mathbf{H}^{(0)}),\,\mixmsg^{(\ell)}\big)$,
an ego/residual term combined with the mixed message across channels. Its seven components
$(\mathcal X,\mathcal K,\mathbf{P}_k,\Psi_k,\boxplus,\mathcal B_\ell,\phi_\ell)$ are defined in
full in Section~\ref{sec:unified-view}. We record here only the
detail left implicit there. The channel set $\mathcal K$ indexes hops, heads, relations,
meta-paths, spectral terms, lifted ranks, or tensor degrees, and the message map carries its explicit
arguments $\Psi_k^{(\ell)}(\mathbf{H}^{(\ell)},\mathbf{H}^{(0)},\mathbf{E})$. The mixed message is the output of the mixing component
\[
\mixmsg^{(\ell)}
=
\boxplus_{k\in\mathcal K}
\mathbf{C}_k^{(\ell)},
\qquad
[\mathbf{C}_k^{(\ell)}]_{i:}=
\begin{cases}
[\mathbf{P}_k^{(\ell)}\Psi_k^{(\ell)}]_{i:} & \text{(linear-value channel)},\\
\operatorname*{Agg}_{j}\!\big([\mathbf{P}_k^{(\ell)}]_{ij}\Psi_k^{(\ell)}(\vh_i,\vh_j,\ve_{ij})\big) & \text{(pairwise channel)},
\end{cases}
\]
where $\boxplus$ denotes the channel-mixing operator (sum, concatenation, or gated/attention combination across
the channel index $k$, equivariant under node relabeling), and the channel contribution
$\mathbf{C}_k$ is the matrix product for a linear value or, row by row, the aggregated per-edge
message for an edge-conditioned/geometric one. The neighbor
aggregation $\operatorname{Agg}_j$ ranges over $j$ with $[\mathbf{P}_k^{(\ell)}]_{ij}\neq 0$ and is part of the channel contribution, not of $\boxplus$.

A scalar per-channel coefficient (fixed, learned, signed, or constrained) is
absorbed into the operator,
$\mathbf{P}_k^{(\ell)}\leftarrow\gamma_k^{(\ell)}\mathbf{P}_k^{(\ell)}$, so the mix $\mixmsg^{(\ell)}$ carries
no separate coefficient slot and no additive residual term. The update map then combines
$\mixmsg^{(\ell)}$ with an ego or residual term
$\mathcal{B}_\ell(\mathbf{H}^{(\ell)},\mathbf{H}^{(0)})$ to produce the node-preserving intermediate
state $\bar{\mathbf{H}}^{(\ell+1)}$, which an optional pooling step then maps to the next domain.

After $L$ layers, a readout $\rho$ maps the stack of layer states to the task
representation, $\mathbf{Z}=\rho(\mathbf{H}^{(0)},\dots,\mathbf{H}^{(L)})\mathbf{W}_{\mathrm{out}}$.
For most node-level models $\rho$ simply selects the last state $\mathbf{H}^{(L)}$, while
jumping-knowledge designs instead combine states from multiple layers.

This slot vocabulary is the backbone of the seven families, each varying a particular
slot or sub-property as its primary lever. Four families specialize the propagation operator
$\mathbf{P}_k$. Spatial models vary the local operator together with the message and
update. Attention models make $\mathbf{P}_k$ state-dependent. Spectral models vary the
basis and the coefficient rule folded into $\mathbf{P}_k$. Graph Transformers use global
or tokenized attention operators. The remaining three vary a different slot instead.
Heterogeneous models index channels by relation or meta-path, higher-order models
change the domain $\mathcal X$, and geometric models constrain $\Psi_k$ and $\phi$ to
respect an additional symmetry group. Figure~\ref{fig:unified-framework} and the slot
legend in Section~\ref{sec:unified-view} (Table~\ref{tab:notation} in
Appendix~\ref{sec:notation}) give the visual and tabular counterparts of this decomposition.

\subsection{Spectral and Spatial Perspectives}
\label{subsec:spectral-spatial-prelim}

Two views complement each other. The \emph{spatial} perspective starts from neighborhoods. A node receives messages from adjacent nodes and updates its representation. The \emph{spectral} perspective starts from graph signals.

A graph signal is a vector $\mathbf{x}\in\mathbb{R}^{n}$ where $x_v$ is the value at node $v$ \citep{shuman2013emerging}. Node features are signals on graph vertices. Each column of the feature matrix $\mathbf{X}\in\mathbb{R}^{n\times d}$ is one signal, representing one feature channel. This view lets GNN layers be interpreted as operations that transform, smooth, sharpen, or mix signals over graph structure.

The graph Laplacian~\citep{chung1997spectral} makes the frequency view concrete. For a signal $\mathbf{x}$ on an undirected graph (where $\mathbf{A}$ is symmetric), the Laplacian quadratic form is $\mathbf{x}^{\top}\mathbf{L}\mathbf{x} = \frac{1}{2}\sum_{u,v}A_{uv}(x_u-x_v)^2$. This is small when adjacent nodes have similar values and large when they differ, making it the graph's discrete Dirichlet energy; the normalized $\mathbf{L}_{\mathrm{sym}}$ used from here on carries the same reading with an added degree weighting. This is why Laplacians formalize smoothness and frequency.

For undirected graphs with nonnegative edge weights, $\mathbf{L}_{\mathrm{sym}}$ is symmetric positive semidefinite. Its eigenvalues are real and nonnegative, $0=\lambda_1\leq \lambda_2\leq \cdots \leq \lambda_n \leq 2$. The multiplicity of the zero eigenvalue equals the number of connected components (assuming no
isolated nodes, as fixed in Section~\ref{subsec:graph-theory-prelim}; under the
$\mathbf{D}^{-1/2}$ convention above, an isolated node instead contributes an eigenvalue of
$1$, which is exactly why it must be excluded for the correspondence to hold).

The eigendecomposition is $\mathbf{L}_{\mathrm{sym}}=\mathbf{U}\boldsymbol{\Lambda}\mathbf{U}^{\top}$, where $\mathbf{U}$ contains orthonormal eigenvectors and $\boldsymbol{\Lambda}=\mathrm{diag}(\lambda_1,\ldots,\lambda_n)$ contains eigenvalues. The eigenvectors are graph Fourier basis functions. Small eigenvalues correspond to smooth, low-frequency eigenvectors; large eigenvalues correspond to high-frequency eigenvectors that change rapidly across edges. This frequency interpretation explains how propagation preserves, suppresses, or amplifies modes of variation.

This construction assumes an undirected graph with symmetric Laplacian, so the eigenvalues are real and the Fourier interpretation applies directly. Directed, signed, temporal, or heterogeneous graphs may instead call for non-symmetric or complex operators, for which this picture does not transfer automatically. The relevant families adopt symmetrized or magnetic Laplacians, and we state the spectral caveats where those operators are introduced, rather than assuming the undirected filtering intuition carries over unchanged.

The graph Fourier transform of a signal $\mathbf{x}$ is $\widehat{\mathbf{x}}=\mathbf{U}^{\top}\mathbf{x}$, representing $\mathbf{x}$ in the frequency domain. The inverse is $\mathbf{x}=\mathbf{U}\widehat{\mathbf{x}}$.

A spectral graph filter modifies frequency components \citep{bruna2014spectral}. Formally, $g_{\theta}(\mathbf{L}_{\mathrm{sym}})\mathbf{x} = \mathbf{U}g_{\theta}(\boldsymbol{\Lambda})\mathbf{U}^{\top}\mathbf{x}$, where $g_{\theta}(\boldsymbol{\Lambda})$ is diagonal with $i$th entry $g_{\theta}(\lambda_i)$. A low-pass filter preserves smooth components. A high-pass filter emphasizes variation. A band-pass filter preserves an intermediate range. Because $\mathbf{L}_{\mathrm{sym}}=\mathbf{I}-\bar{\mathbf{A}}$ measures disagreement between neighbors while $\bar{\mathbf{A}}$ measures agreement with them, a propagation operator built on the Laplacian tends to act as a high-pass filter and one built on the normalized adjacency tends to act as a low-pass filter. For multi-dimensional features, $g_{\theta}(\mathbf{L}_{\mathrm{sym}})\mathbf{X} = \mathbf{U}g_{\theta}(\boldsymbol{\Lambda})\mathbf{U}^{\top}\mathbf{X}$.

Explicit spectral filtering through the eigendecomposition is expensive on large graphs. The standard alternative is a degree-$K$ polynomial $p_\theta(\lambda)=\sum_{k=0}^{K}\theta_k\lambda^k$ applied to the Laplacian \citep{defferrard2016convolutional}. This gives $p_{\theta}(\mathbf{L}_{\mathrm{sym}})\mathbf{x} = \sum_{k=0}^{K}\theta_k\mathbf{L}_{\mathrm{sym}}^{k}\mathbf{x}$. On a fixed graph where the operator in use has $q$ distinct eigenvalues, polynomials realize spectral responses \emph{exactly}. Any prescribed response is interpolated by a polynomial of degree at most $q-1$ (Appendix~\ref{sec:analysis}; the count $q$ is operator-dependent, since $\mathbf{L}_{\mathrm{sym}}$ and $\hat{\mathbf{A}}$ generally have different numbers of distinct eigenvalues). A truncation $K<q-1$ may still be exact for responses that happen to admit a lower-degree realization; it only approximates responses that require higher degree, trading expressivity for locality and cost. Polynomial filters avoid eigendecomposition and use sparse matrix multiplication. They are localized. $\mathbf{L}_{\mathrm{sym}}^{k}\mathbf{x}$ depends on nodes at most $k$ hops away. This connects spectral filtering to message passing and multi-hop propagation.

Chebyshev polynomial bases are common in practice \citep{defferrard2016convolutional,he2022chebnetii}, and GCN \citep{kipf2017semi} follows from this construction after a first-order truncation, parameter tying across the two terms, and the renormalization trick $\mathbf{I}+\mathbf{D}^{-1/2}\mathbf{A}\mathbf{D}^{-1/2}\to\hat{\mathbf{A}}$, which rescales the $[0,2]$ spectrum of $\mathbf{I}+\mathbf{D}^{-1/2}\mathbf{A}\mathbf{D}^{-1/2}$ to mitigate the numerical instability of repeated application.

The two views are compatible. Consider a propagation operator $\mathbf{P}$ that acts as a
\emph{scalar on each eigenspace} of $\mathbf{L}_{\mathrm{sym}}$, equivalently a spectral
function $g(\mathbf{L}_{\mathrm{sym}})$, using the same spectral-function form as $g_\theta$
above but without asserting it is learned or parameterized. Such an operator has both a
spectral form $\mathbf{P} = \mathbf{U}g(\boldsymbol{\Lambda})\mathbf{U}^{\top}$
and, by the polynomial realizability above, a spatial form as a weighted combination of Laplacian
powers, $\mathbf{P} = \sum_k\theta_k\mathbf{S}_k$ with $\mathbf{S}_k=\mathbf{L}_{\mathrm{sym}}^{k}$.
This equivalence is what links polynomial filters to multi-hop propagation. It is not universal.
A general propagation operator (a data-dependent attention matrix, a per-relation operator)
typically does \emph{not} act as a scalar on each $\mathbf{L}_{\mathrm{sym}}$ eigenspace. It may
fail to commute with $\mathbf{L}_{\mathrm{sym}}$ at all, or split a repeated eigenspace, and so
has no such $g(\boldsymbol{\Lambda})$ form.

The slot framework therefore takes structural supports
$\{\mathbf{S}_k\}$ (adjacency powers, Laplacian powers, attention operators, relation matrices)
as primary, with the spectral form available for the operators that admit it. In
Eq.~\eqref{eq:unified-notation}, each weighted support $\theta_k\mathbf{S}_k$ becomes a
propagation channel $\mathbf{P}_k^{(\ell)}$ with its scalar coefficient $\theta_k$ folded
in (the spectral instance of the generic per-channel coefficient $\gamma_k$ introduced
above), so the coefficient rule lives on the operator rather than in a separate slot. We
reserve $\Psi_k^{(\ell)}$ for the channel message/value, so spectral basis supports
and messages are not overloaded. The first form emphasizes graph frequencies; the second
emphasizes the operators through which information flows.
These spectral and spatial objects and their roles are summarized alongside the rest of the
paper's notation in Table~\ref{tab:notation}, given in Appendix~\ref{sec:notation}.

\subsection{Symmetry, Invariance, and Equivariance}
\label{subsec:symmetry-prelim}

Graphs have no canonical node ordering. Reordering nodes does not change the underlying graph, only its matrix representation. Graph learning methods should respect this symmetry.

A function is \emph{permutation invariant} if reordering input nodes does not change the output. Graph-level predictions should be invariant. A molecule's predicted property should not depend on node indexing. A function is \emph{permutation equivariant} if reordering input nodes reorders output nodes the same way. Node-level predictions should be equivariant. Permuting nodes permutes predictions consistently. These symmetries are the defining design constraint of GNNs \citep{maron2019invariant}.

Message passing satisfies this when the neighbor aggregation is order-invariant (sum, mean, max) and the attention scores transform equivariantly with the node relabeling. This is why GNNs differ from sequence models. The architecture respects relational structure rather than an arbitrary node ordering. For lifted higher-order domains, equivariance is with respect to the induced permutation action on tuples, subgraphs, or cells. For geometric families, permutation equivariance is combined with an additional group equivariance or invariance, such as translation, rotation, or Euclidean symmetry.

This symmetry constraint has an exact combinatorial counterpart. Sum-aggregation message passing distinguishes two nodes, or two graphs, only as well as the Weisfeiler--Leman (WL) test does. Equivariant message passing is therefore no more powerful than 1-WL at telling non-isomorphic graphs apart \citep{xu2019powerful,morris2019weisfeiler}. Section~\ref{sec:propagation} develops this expressivity ceiling in full, alongside the substructure-counting and higher-order refinements that exceed it.

Section~\ref{sec:unified-view} develops the unified equation in full and organizes the literature into the seven families. The notation introduced above, together with the unified layer-slot notation, is collected for reference in Appendix~\ref{sec:notation} (Table~\ref{tab:notation}).

\section{The Unified Layer Equation for Graph Neural Networks}
\label{sec:unified-view}

\subsection{Coverage, Claims, and the Seven Families}
\label{subsec:framework-scope}
Graph neural networks are often presented as distinct families, yet many differ only in local choices within a shared layer. Our unified equation forms a message on each channel, propagates it with a channel operator, combines the channels, and updates the state. Its seven components are the update domain $\mathcal X$, channel set $\mathcal K$, propagation bank $\{\mathbf P_k\}$, message maps $\{\Psi_k\}$, mixing operator $\boxplus$, ego/residual map $\mathcal B_\ell$, and update map $\phi_\ell$. This unified system records \emph{which} object an architecture changes and \emph{how}, enabling conditional attribution and identifiability results across families. Selecting the best filling for a task remains the separate empirical inverse problem. Appendix~\ref{app:family-comparison-tables} uses the components to organize seven primary-axis families. These are not disjoint sets, so a hybrid model may occupy several categories depending on the component being compared.
For the appendix tables, each method is assigned to one primary family according to the dominant mechanism of its layer update. Secondary mechanisms are reported as cross-references or boundary notes rather than duplicate rows. Appendix~\ref{app-family-assignment-rules} states the assignment rules in family order. When more than one rule applies, the primary family is the one that best describes the contribution claimed for the cited layer.
The coverage claim is about the core propagation/update mechanism of a layer, not about every procedure surrounding a cited model. When the cited layer is itself one synchronous propagation/update rule, the displayed slots give a full-layer decomposition. When a method additionally contains a solver, inner loop, routing procedure, search procedure, readout, pooling step, or typed edge node global schedule, the table retains the covered core equation and marks the remaining machinery as an exact slice or named boundary in the row caption.

\paragraph{A first pass, concretely.} Before the claims and equivariance theorem below reuse this notation, one instance fixes what each name denotes. GCN's layer $\mathbf{H}^{(\ell+1)}=\sigma(\hat{\mathbf{A}}\mathbf{H}^{(\ell)}\mathbf{W}^{(\ell)})$ updates nodes ($\mathcal{X}=V$) through a single channel ($\mathcal{K}=\{1\}$, so $\boxplus$ is a no-op). Within that channel, $\Psi(\mathbf{H}^{(\ell)})=\mathbf{H}^{(\ell)}\mathbf{W}^{(\ell)}$ is \emph{what} moves and the normalized adjacency $\hat{\mathbf{A}}$, playing the role of $\mathbf{P}_1$, is \emph{where} it moves. There is no separate ego term ($\mathcal{B}_\ell=0$), and $\phi_\ell$ is the elementwise nonlinearity $\sigma$ applied to the mixed message. Each row is read with the same slot questions: name the domain being updated, the channels available, what moves on each channel and where, how the channels combine, how the prior state carries forward, and how the result becomes the new state. Whether that reading covers the full layer or only the core propagation/update equation is part of the row-level claim, not something inferred from the presence of slot entries alone. Section~\ref{subsec:setup} makes each slot precise and Eq.~\eqref{eq:unified-notation} states them jointly; this walkthrough only aims to make the claims and theorem below legible on a first pass.

\paragraph{Relation to message passing and graph networks.}
The MPNN formulation~\citep{gilmer2017neural} computes
$m_i^{(\ell+1)}=\sum_{j\in\mathcal N(i)}M_\ell(h_i^{(\ell)},h_j^{(\ell)},e_{ij})$
and then $h_i^{(\ell+1)}=U_\ell(h_i^{(\ell)},m_i^{(\ell+1)})$. Equation~\eqref{eq:unified-notation}
recovers it exactly by taking $\mathcal X=\mathcal V$, $\mathcal K=\{1\}$,
$[\mathbf P_1]_{ij}=\mathbf 1_{\{j\in\mathcal N(i)\}}$, $\Psi_1=M_\ell$,
$\operatorname*{Agg}=\sum$, $\boxplus=\mathrm{id}$,
$\mathcal B_\ell(\mathbf H^{(\ell)},\mathbf H^{(0)})=\mathbf H^{(\ell)}$, and
$\phi_\ell=U_\ell$. Its graph readout is the external map $R$, downstream of the seven
layer components.

The Graph Network block~\citep{battaglia2018relational} first updates each edge with
$\phi^e$, aggregates updated incident edges with $\rho^{e\to v}$, and updates each node
with $\phi^v$. It also permits global aggregation and a global update. Its node-update slice
is recovered by using edge support in $\mathbf P_1$, taking $\Psi_1=\phi^e$,
$\operatorname*{Agg}=\rho^{e\to v}$, $\boxplus=\mathrm{id}$, and
$\mathcal B_\ell(\vh_i)=(\vh_i,u)$ with $\phi_\ell=\phi^v$, after broadcasting the global
attribute as context when it is present.
The full edge node global block is not one node-only instance of
Eq.~\eqref{eq:unified-notation}. Its typed states use the graded domains of
Section~\ref{subsec:graded-extension}, while its prescribed edge node global update order
is a composition of such updates. Section~\ref{sec:generation} makes this distinction
explicit by adding an execution schedule in its Level~3 schema demonstration. The
augmented-message-passing view~\citep{velickovic2022message} argues that computations that
appear to go beyond message passing can often be expressed through pairwise messages on a
suitably modified graph. It does not prescribe the support--value factorization or the
channel and mixing inventory used here. Table~\ref{tab:framework-comparison} records these
relations without imposing an expressive ordering.

\begin{table*}[!t]
\centering
\begingroup
\scriptsize
\setlength{\tabcolsep}{4pt}
\renewcommand{\arraystretch}{1.32}
\arrayrulecolor{black!30}
\setlength{\arrayrulewidth}{0.5pt}
\caption{Comparison with prior message-passing abstractions.}
\label{tab:framework-comparison}
\begin{tabularx}{\textwidth}{
|>{\raggedright\arraybackslash}p{0.15\textwidth}
|>{\raggedright\arraybackslash}p{0.24\textwidth}
|>{\raggedright\arraybackslash}p{0.25\textwidth}
|>{\raggedright\arraybackslash}X|}
\hline
\rowcolor{googleblue!18}
\textcolor{googleblue!45!black}{\textbf{Formalism}} &
\textcolor{googleblue!45!black}{\textbf{Canonical computational objects}} &
\textcolor{googleblue!45!black}{\textbf{Not separately exposed}} &
\textcolor{googleblue!45!black}{\textbf{Exact relation or boundary}} \\
\hline

\rowcolor{googleblue!5}
\cellcolor{googleblue!14}
\textcolor{googleblue!45!black}{\textbf{MPNN}~\citep{gilmer2017neural}} &
Node states, a pairwise message
$M_\ell(h_i,h_j,e_{ij})$, summation over neighbors, a node update
$U_\ell$, and a graph-level readout $R$ &
The canonical equations expose a single neighbor-aggregated message
$m_i=\sum_{j\in\mathcal N(i)}M_\ell(\cdot)$.
They do not separately expose a channel index set, a decomposition of
propagation support from message values, a distinct cross-channel mixing
operator, or an independently named ego/residual construction &
Exact node-domain specialization of the pairwise branch with
$\mathcal K=\{1\}$,
$[\mathbf P_1]_{ij}=\mathbf 1_{\{j\in\mathcal N(i)\}}$,
$\Psi_1=M_\ell$,
$\operatorname*{Agg}=\sum$,
$\boxplus=\mathrm{id}$,
$\mathcal B_\ell(\mathbf H^{(\ell)},\mathbf H^{(0)})
=\mathbf H^{(\ell)}$, and
$\phi_\ell=U_\ell$.
The graph-level readout $R$ lies outside the base-layer equation \\
\hline

\rowcolor{googlegreen!5}
\cellcolor{googlegreen!14}
\textcolor{googlegreen!45!black}{\textbf{Graph Network block}~\citep{battaglia2018relational}} &
A graph $(\mathbf u,V,E)$; edge update $\phi^e$;
edge-to-node aggregation $\rho^{e\to v}$; node update $\phi^v$;
edge-to-global and node-to-global aggregations
$\rho^{e\to u}$ and $\rho^{v\to u}$; and global update $\phi^u$ &
The GN formalism explicitly types edge , node, and global domain
updates and aggregations, but it does not separately parameterize them
by a channel inventory $\mathcal K$ with distinct support operators
$\mathbf P_k$, value maps $\Psi_k$, and a cross-channel mixing operator
$\boxplus$ &
Its node-update slice is represented exactly by taking edge incidence
as the support of $\mathbf P_1$, $\Psi_1=\phi^e$,
$\operatorname*{Agg}=\rho^{e\to v}$,
$\boxplus=\mathrm{id}$,
$\mathcal B_\ell(\vh_i)=(\vh_i,\mathbf u)$, and
$\phi_\ell=\phi^v$, with $\mathbf u$ supplied as shared context.
A full GN block additionally updates edge and global states:
in the canonical ordering, $\phi^v$ consumes aggregated updated edges
and $\phi^u$ consumes aggregated updated edges and updated nodes.
Representing the full block therefore requires multiple typed domains
and staged updates rather than a single node-domain base-layer
application \\
\hline

\rowcolor{googleyellow!7}
\cellcolor{googleyellow!18}
\textcolor{googleyellow!45!black}{\textbf{Augmented message-passing view}~\citep{velickovic2022message}} &
Pairwise message passing applied to a suitably augmented computational
graph, where features, connectivity, or represented entities may be
modified so that a richer computation is realized through pairwise
communication &
This is a broad representational viewpoint rather than a prescribed
layer tuple. It does not require an explicit channel inventory,
support value factorization, or separately named cross channel mixing
component &
Broader in representational scope. A computation is captured by one
seven-component base-layer application only when the relevant
synchronous step admits the stated channel-wise normalization.
Computations requiring graph augmentation, replicated entities,
multiple dependent stages, or other transformations may still fall
within augmented message passing while requiring more than one
base-layer application or an augmented computational graph \\
\hline

\rowcolor{googlered!4}
\cellcolor{googlered!12}
\textcolor{googlered!45!black}{\textbf{Seven-component equation (this work)}} &
$\mathcal X$, $\mathcal K$, $\{\mathbf P_k\}$, $\{\Psi_k\}$,
$\boxplus$, $\mathcal B_\ell$, and $\phi_\ell$ &
The training objective, post-stack readout, external domain transitions
such as pooling, and multi-stage control flow are outside the base layer &
A typed reporting form for the covered layer class.
It refines the canonical MPNN description by separately exposing
support, channel-wise value computation, mixing, and base-state
construction. It captures the node-update slice of a Graph Network
block but not an entire staged edge node global block in one
node-domain application, and it is intentionally narrower than the
broader augmented-message-passing viewpoint \\
\hline
\end{tabularx}
\endgroup
\end{table*}

\paragraph{Component-level consequences of the decomposition.}
The decomposition yields two conclusions that are not available from an undivided message
function without first introducing equivalent structure. First, under the endpoint-local
message and node-local update assumptions of Proposition~\ref{prop:receptive}, the union of
the supports of $\{\mathbf P_k\}$ fixes the support-level envelope of the one-layer dependency
graph. Changing only
$\Psi_k$, $\boxplus$, $\mathcal B_\ell$, or $\phi_\ell$ cannot create dependence on a node
outside that support. Proposition~\ref{prop:global} consequently localizes one-layer global
mixing to a full effective operator row under its stated assumptions. Second, in the linear
additive regime, Proposition~\ref{prop:channel-rank} shows that the displayed channel count
$|\mathcal K|$ is not an invariant of the layer function because channels can be split or
cancelled. The minimum Kronecker separation rank of
$\sum_k\mathbf W_k^\top\otimes\mathbf P_k$ is invariant. Thus two models cannot be declared
functionally different merely because their tables use different numbers of heads, relations,
or polynomial terms. Their contribution is component-level attribution and identifiability.
Expressive comparison with unrestricted message passing is a separate question.

\FloatBarrier
We make three claims, all checkable against the comparison tables rather than asserted.

\begin{enumerate}
\item \label{claim:coverage} \textbf{Coverage with a named boundary.} The framework covers layers built from the two channel types formalized in Theorem~\ref{thm:equivariance}. Matrix linear-value channels use $\mathbf{P}_k\Psi_k$, while pairwise-message channels invariantly aggregate $\Psi_k(\vh_i,\vh_j,\ve_{ij})$. The former permits fixed or state-dependent node-level and inter-rank matrices, and the latter covers edge-conditioned and geometric messages that no single matrix product expresses. An equivariant $\boxplus$ combines either type before $\phi_\ell$. With permutation-commuting slots and an invariant readout, the graph function is permutation-invariant. Section~\ref{subsec:worked-reductions} verifies representative reductions.

This condition is independently checkable from the channel form and mix symmetry. Tables distinguish full-layer decompositions from exact covered slices, flag layers that fail the condition, and exclude methods without a single-layer graph-propagation description. Section~\ref{subsec:setup} details five cases outside the channel primitives, and Section~\ref{subsec:graded-extension} gives a sixth case outside the single-layer scope. Alternative representations of these methods require primitives beyond the stated reduction rules.
\item \label{claim:determinacy} \textbf{Determinacy under a fixed slot discipline.} Section~\ref{subsec:slot-discipline} assigns each ingredient to a slot, determining the normalized placements in Section~\ref{subsec:worked-reductions} relative to the convention rather than absolutely. Determinacy remains subject to the linear-regime channel-rank redundancy of Proposition~\ref{prop:channel-rank} and computationally equivalent forms. For example, APPNP places its teleport in $\mathcal{B}_\ell=\alpha f_\theta(\mathbf{X})$ iteratively but absorbs it into $\mathbf{P}_k$ with $\mathcal{B}_\ell=0$ when unrolled (Section~\ref{subsec:setup}).
\item \label{claim:discrimination} \textbf{Discrimination relative to that discipline.} Under the stated rule, differing table cells localize architectural differences. This claim is conditional because another discipline could relocate differences and the slots can interact (Section~\ref{subsec:worked-reductions}).
\end{enumerate}

We do \emph{not} claim that the slots are orthogonal or that every instantiation is trainable, stable, scalable, or expressive. These remain properties of specific models.

The theorem below states the equivariance guarantee using notation defined in Sections~\ref{subsec:setup} and~\ref{subsec:graded-extension}. Its proof is in \hyperref[proof:thm-equivariance]{Appendix~\ref*{sec:analysis}}.

\begin{theorem}[Permutation equivariance]
\label{thm:equivariance}
Suppose each operator is an equivariant function of the graph, in the sense made precise by the following hypotheses on the node domain, the graded domain, and pooling.

\textbf{Hypotheses (node domain).} A \emph{linear-value channel} has a row-wise message $\Psi_k$ and a matrix product $\mathbf{P}_k\Psi_k$, with $\mathbf{P}_k(\boldsymbol{\Pi}\mathbf{A}\boldsymbol{\Pi}^\top,\boldsymbol{\Pi}\mathbf{H})=\boldsymbol{\Pi}\mathbf{P}_k(\mathbf{A},\mathbf{H})\boldsymbol{\Pi}^\top$. When edge features modulate the operator, it takes the three-argument form $\mathbf{P}_k(\boldsymbol{\Pi}\mathbf{A}\boldsymbol{\Pi}^\top,\boldsymbol{\Pi}\mathbf{H},\boldsymbol{\Pi}\!\cdot\!\mathbf{E})=\boldsymbol{\Pi}\mathbf{P}_k(\mathbf{A},\mathbf{H},\mathbf{E})\boldsymbol{\Pi}^\top$, and the channel remains a matrix product. In edge-aware GAT, the score reads $\ve_{ij}$ while the value remains $\Psi_k=\mathbf{H}\mathbf{W}_k$.

A \emph{pairwise-message channel} has a message $\Psi_k(\vh_i,\vh_j,\ve_{ij})$ combined by a permutation-invariant per-node aggregation $\operatorname*{Agg}_{j\in N(i)}$, with no matrix product. Here the edge-feature tensor $\mathbf{E}\in\mathbb{R}^{n\times n\times p}$ is relabeled jointly with the nodes by the action $(\boldsymbol{\Pi}\!\cdot\!\mathbf{E})_{ij:}=\mathbf{E}_{\pi^{-1}(i)\,\pi^{-1}(j)\,:}$, which reduces to $\boldsymbol{\Pi}\mathbf{E}\boldsymbol{\Pi}^\top$ only when $p=1$. If edge features influence the operator, the equivariance hypothesis includes $\mathbf{E}$ as an argument and requires $\mathbf{P}_k(\boldsymbol{\Pi}\mathbf{A}\boldsymbol{\Pi}^\top,\boldsymbol{\Pi}\mathbf{H},\boldsymbol{\Pi}\!\cdot\!\mathbf{E})=\boldsymbol{\Pi}\mathbf{P}_k(\mathbf{A},\mathbf{H},\mathbf{E})\boldsymbol{\Pi}^\top$.

Throughout, ``row-wise'' means the \emph{same} node function applied at every row, a shared row map; this is what makes the layer commute with relabeling, whereas a node-indexed family of distinct row maps need not. In both channel types $\boxplus$ commutes with row permutation, and the node-local specialization takes $\mathcal{B}_\ell$ and $\phi_\ell$ to be shared row maps.

\textbf{Hypotheses (graded domain and pooling).} On a graded domain, let $\boldsymbol{\Pi}_r$ and $\boldsymbol{\Pi}_s$ relabel the source and target objects at ranks $r$ and $s$. Replace conjugation by the typed condition
\[
\mathbf{P}_{k,r\to s}(\boldsymbol{\Pi}\!\cdot\!\mathcal{G},\boldsymbol{\Pi}\!\cdot\!\mathbf{H})
=\boldsymbol{\Pi}_s\mathbf{P}_{k,r\to s}(\mathcal{G},\mathbf{H})\boldsymbol{\Pi}_r^\top,
\]
and require the source message to transform by $\boldsymbol{\Pi}_r$, pairwise relation features to be relabeled jointly by $(\boldsymbol{\Pi}_s,\boldsymbol{\Pi}_r)$, and the target-rank mix, ego map, and update to commute with $\boldsymbol{\Pi}_s$. Finally, an external pooling map preserves equivariance when
$\mathbf{Q}_\ell(\boldsymbol{\Pi}\mathbf{A}\boldsymbol{\Pi}^\top,\boldsymbol{\Pi}\mathbf{H})=\boldsymbol{\Pi}_{\mathrm{out}}\mathbf{Q}_\ell(\mathbf{A},\mathbf{H})\boldsymbol{\Pi}^\top$ and the coarsened graph/operator bank is constructed equivariantly.

\textbf{Conclusion.} Under these hypotheses, the node-domain layer, the graded layer, and the optional pooling step are permutation equivariant. Composing them with a permutation-invariant readout yields a permutation-invariant graph function, the defining symmetry of message-passing and invariant graph networks~\citep{gilmer2017neural,maron2019invariant}.
\end{theorem}

The row-wise hypothesis on $\phi_\ell$ used above is sufficient but stronger than
necessary. It excludes the cross-node normalizations (PairNorm~\citep{zhao2020pairnorm},
GraphNorm~\citep{Cai2021GraphNorm}) that the framework folds into $\phi_\ell$, which are
equivariant but couple rows. Replacing ``row-wise'' by ``equivariant'' for $\Psi_k$,
$\mathcal{B}_\ell$, and $\phi_\ell$ extends the conclusion to those updates.

\paragraph{Representation, analysis, and prescription.} The equation represents architectures,
localizes their component differences, and supports the attribution and identifiability results
above. It also states where documented dataset, task, or deployment properties enter the
computation. Selecting a component filling for a new task is a distinct inverse problem that requires a
reliable mapping from measurable properties to component choices, supported by controlled
evidence that isolates each choice. Sections~\ref{sec:benchmarks} and~\ref{sec:outlook}
formulate this inverse-design problem.

\subsection{Setup and Notation}
\label{subsec:setup}

To make the claims of Section~\ref{subsec:framework-scope} checkable slot by slot, let $\mathcal{G}^{(\ell)}=(\mathcal{V}^{(\ell)},\mathcal{E}^{(\ell)})$ denote the graph at layer $\ell$, with $n_\ell=|\mathcal{V}^{(\ell)}|$, and let $\mathbf{H}^{(\ell)}\in\mathbb{R}^{n_\ell\times d_\ell}$ have row $\vh_i^{(\ell)}\in\mathbb{R}^{1\times d_\ell}$. We write $n_\ell$ because coarsening operations can change the node count between layers (Appendix~\ref{subsec:op-pooling}), whereas node-preserving stacks have $n_\ell=n$. Row-vector feature transformations act on the right, as in $\vh_i^{(\ell)}\mathbf{W}^{(\ell)}$ and $\mathbf{H}^{(\ell)}\mathbf{W}^{(\ell)}$.

We develop the predominant node-level case first and give the higher-order extension to tuples, subgraphs, and cells in Section~\ref{subsec:graded-extension}.

\paragraph{Propagation operators $\{\mathbf{P}_k^{(\ell)}\}$.} Each possibly state-dependent operator $\mathbf{P}_k^{(\ell)}\in\mathbb{R}^{n_\ell\times n_\ell}$ specifies communication support and scalar weights on channel $k$. Within-layer propagation preserves the domain, so the operator is square, $\mathcal{B}_\ell$ has the same row count, and $\bar{\mathbf{H}}^{(\ell+1)}\in\mathbb{R}^{n_\ell\times d_{\ell+1}}$. Only the external pooling operation of Appendix~\ref{sec:operations} changes $n_\ell$. The channel index may denote hop powers, attention heads, relations, meta-paths, spectral terms, lifted ranks, or tensor degrees. Examples include $\hat{\mathbf{A}}$, $\hat{\mathbf{A}}^k$, $T_k(\tilde{\mathbf{L}}_{\mathrm{ch}})$ with $\tilde{\mathbf{L}}_{\mathrm{ch}}=2\mathbf{L}_{\mathrm{sym}}/\lambda_{\max}-\mathbf{I}$, and a data-dependent attention matrix $\mathbf{P}_k(\mathbf{H}^{(\ell)})$. Feature transformations remain in $\mathbf{W}_k$ inside the message, preventing duplicate edge weighting. The induced neighborhood is $\mathcal{N}_k(i):=\{j:[\mathbf{P}_k]_{ij}\neq 0\}$.

\paragraph{Message map $\Psi_k^{(\ell)}$.} Each $\Psi_k^{(\ell)}$ forms the message propagated by $\mathbf{P}_k^{(\ell)}$. The common linear form is $\Psi_k^{(\ell)}(\mathbf{H}^{(\ell)})=\mathbf{H}^{(\ell)}\mathbf{W}_k^{(\ell)}$. Edge-conditioned or geometric messages instead contribute $\operatorname*{Agg}_{j}\bigl([\mathbf{P}_k^{(\ell)}]_{ij}\Psi_k^{(\ell)}(\vh_i^{(\ell)},\vh_j^{(\ell)},\ve_{ij}^{(\ell)})\bigr)$. For a linear value $\Psi_k=\mathbf{H}\mathbf{W}_k$, a linear sum or mean folds into $\mathbf{P}_k$ as bare support or $\mathbf{D}^{-1}\mathbf{A}$, yielding $\mathbf{P}_k\Psi_k$. A pairwise message cannot be folded this way, even under summation, and remains an explicit neighbor contraction in $\mathbf{C}_k$ of \eqref{eq:unified-notation}. A nonlinear aggregator such as max or power mean is also recorded separately because no matrix $\mathbf{P}_k$ realizes it. This convention places the GIN~\citep{xu2019powerful} MLP in $\phi_\ell$, distinguishes GraphSAGE mean~\citep{hamilton2017inductive} through $\mathbf{D}^{-1}\mathbf{A}$ and ego handling, and places GAT's~\citep{velickovic2018gat} state dependence in $\mathbf{P}_k$.

\paragraph{Channel coefficients fold into the operator.} A scalar channel weight is absorbed as $\mathbf{P}_k^{(\ell)}\leftarrow\gamma_k^{(\ell)}\mathbf{P}_k^{(\ell)}$, leaving $\boxplus$ as a pure combiner and $\mixmsg^{(\ell)}$ without ego or residual terms. Per-node update gates remain in $\phi_\ell$. Coefficient constraints remain discriminating properties of $\mathbf{P}_k$. GPR-GNN~\citep{chien2021adaptive} uses signed $\gamma_k$, BernNet~\citep{he2021bernnet} uses nonnegative $\theta_k$, and APPNP has profile $\alpha(1-\alpha)^k$ only in the infinite-depth personalized-PageRank limit, with finite-$K$ top-order mass $(1-\alpha)^K$. ChebNet has no scalar operator coefficient, as Lemma~\ref{lem:spectral-subsumption} explains.

\paragraph{Mixing $\boxplus$ and its output $\mixmsg$.} The component $\boxplus$ produces $\mixmsg^{(\ell)}=\boxplus_k\mathbf{C}_k^{(\ell)}$. Table headers therefore name $\boxplus$, while cells give its rule and $\mixmsg$ denotes only the resulting tensor. Common rules are summation, concatenation, and learned per-node or per-channel gating. Mixing commutes with node relabeling as required by Theorem~\ref{thm:equivariance}, but need not be invariant to channel order. Concatenation changes when heads exchange output blocks, as Proposition~\ref{prop:concat} and Corollary~\ref{cor:concat-fusion} formalize. It contains no ego or residual term.

Neighbor aggregation and channel mixing are distinct operations. In a pairwise channel,
$\operatorname*{Agg}_{j\in\mathcal N_k(i)}$ constructs $[\mathbf C_k]_{i:}$ before
$\boxplus_k$ acts. The family tables therefore write a pairwise cell as a channel definition
$\mathbf C_k=\operatorname*{Agg}_j(\cdot)$ followed by
$\mixmsg=\boxplus_k\mathbf C_k$. When $\mathcal K=\{1\}$, the table may display the
single contribution $\mathbf C_1$ explicitly, but the mixing filling is
$\boxplus=\mathrm{id}$. No neighbor aggregator is treated as the filling of $\boxplus$.

\begin{proposition}[Concatenation is additive mixing]
\label{prop:concat}
Multi-head concatenation satisfies
\[
\big\Vert_{h=1}^{M} \mathbf{P}_h \mathbf{H} \mathbf{W}_h \;=\; \sum_{h=1}^{M} \mathbf{P}_h \mathbf{H}\,(\mathbf{W}_h \mathbf{J}_h),
\]
where $\mathbf{J}_h\in\mathbb{R}^{d_h\times \sum_{h'} d_{h'}}$ is the $0/1$ block-placement matrix (the identity on block $h$ and zero elsewhere) that places the width-$d_h$ output of head $h$ into block $h$, so $\mathbf{W}_h\mathbf{J}_h$ maps into the full concatenated width. Hence concatenation is additive mixing into disjoint output blocks and introduces no separate mixing mechanism beyond summation.
\end{proposition}

\begin{corollary}[Concatenation expressivity]
\label{cor:concat-fusion}
In the linear specialization, multi-head concatenation and additive mixing realize the same function class only \emph{after} projection to a common target width. Writing $\mathbf{G}_h=\mathbf{P}_h\mathbf{H}\mathbf{W}_h\in\mathbb{R}^{n\times d_h}$ (using $\mathbf{G}$, not the framework's reserved final-representation $\mathbf{Z}$) for the head outputs and partitioning a projection $\mathbf{W}_o\in\mathbb{R}^{(\sum_h d_h)\times d'}$ into block rows $\mathbf{V}_h\in\mathbb{R}^{d_h\times d'}$,
\[
[\mathbf{G}_1\Vert\cdots\Vert\mathbf{G}_M]\,\mathbf{W}_o=\sum_{h=1}^{M}\mathbf{G}_h\mathbf{V}_h,
\]
and conversely every projected sum $\sum_h\mathbf{G}_h\mathbf{V}_h$ arises this way. So the projected concatenation and the (width-$d'$) additive mixing realize the same class. The heads may have distinct widths $d_h$ (raw summation $\sum_h\mathbf{G}_h$ is then not even defined), which is exactly why the equivalence is asserted only at the common width $d'$. Concatenation itself \emph{does} create explicit head separation, placing each head in disjoint output coordinates. The downstream $\mathbf{W}_o$ then decides whether to exploit that separation (block-distinct $\mathbf{V}_h$) or erase it (a projection summing the blocks). A per-head (blockwise) elementwise nonlinearity, whether applied before concatenation or blockwise afterward in $\phi_\ell$, further preserves that separation, since it commutes with the block arrangement.
\end{corollary}

The proof of Proposition~\ref{prop:concat} is in \hyperref[proof:prop-concat]{Appendix~\ref*{sec:analysis}}, while the corollary follows directly from the block partition. Additive mixing depends on $\{(\mathbf{P}_k,\mathbf{W}_k)\}$ only through the summed contributions, so $K$ is not recoverable. Proposition~\ref{prop:channel-rank} identifies the remaining rank-like invariant, with proof in \hyperref[proof:prop-channel-rank]{Appendix~\ref*{sec:analysis}}.

\begin{proposition}[Channel count is not identifiable in the linear regime]
\label{prop:channel-rank}
The slots have an identifiability caveat. The linear pre-activation is the sum-of-Kronecker-products operator $\mathbf{T}=\sum_k \mathbf{W}_k^\top\otimes\mathbf{P}_k$ acting on $\operatorname{vec}(\mathbf{H})$ \citep{roth2024rank}, using $\operatorname{vec}(\mathbf{P}_k\mathbf{H}\mathbf{W}_k)=(\mathbf{W}_k^\top\otimes\mathbf{P}_k)\operatorname{vec}(\mathbf{H})$. Splitting any channel $(\mathbf{P}_k,\mathbf{W}_k)$ into $(\mathbf{P}_k,\mathbf{W}_k')$ and $(\mathbf{P}_k,\mathbf{W}_k'')$ with $\mathbf{W}_k'+\mathbf{W}_k''=\mathbf{W}_k$ (or appending a pair of channels that cancel) leaves $\mathbf{T}$, and hence the layer function, unchanged while raising $K$, so the channel count is not recoverable. The well-defined invariant is the \emph{minimum} number of Kronecker terms representing $\mathbf{T}$, the Kronecker (separation) rank of the rearranged operator, which we read here as the channel rank. Channels cannot generally be \emph{merged} to lower $K$ because a sum of Kronecker products need not be a single Kronecker product. If the admissible operators are constrained (e.g.\ a fixed support pattern or an equivariance requirement on each $\mathbf{P}_k$), the relevant invariant is the minimum over decompositions \emph{respecting that constraint}, which may exceed the unconstrained rank.
\end{proposition}

\paragraph{Ego/residual map $\mathcal{B}_\ell$ and update $\phi_\ell$.} The map $\mathcal{B}_\ell$ carries the node's own state, commonly as $\mathcal{B}_\ell(\mathbf{H}^{(\ell)},\mathbf{H}^{(0)})=\mathbf{H}^{(\ell)}\mathbf{B}^{(\ell)}$, and $\phi_\ell$ combines it with the mixed signal, commonly under an elementwise nonlinearity $\sigma$. Ego handling may use support folding, a scaled sum, concatenation, or an initial-state teleport $\alpha\mathbf{H}^{(0)}$. This axis distinguishes GCN, GIN, and GraphSAGE in Section~\ref{subsec:worked-reductions}, and APPNP in Lemma~\ref{lem:spectral-subsumption}.

The node-preserving output is $\bar{\mathbf{H}}^{(\ell+1)}\in\mathbb{R}^{n_\ell\times d_{\ell+1}}$, and the next input is $\mathbf{H}^{(\ell+1)}=\mathbf{Q}_\ell\bar{\mathbf{H}}^{(\ell+1)}$. The node-to-cluster map $\mathbf{Q}_\ell\in\mathbb{R}^{n_{\ell+1}\times n_\ell}$ is the identity unless external pooling coarsens the domain (Appendix~\ref{sec:operations}). Pooling must also construct the coarsened graph and operator bank, which $\mathbf{Q}_\ell\bar{\mathbf{H}}^{(\ell+1)}$ alone does not specify. Appendix~\ref{subsec:op-pooling} covers pooling variants, equivariance of $\mathbf{Q}_\ell$, and realignment of initial residuals.

\paragraph{The linear specialization.} Many spectral, polynomial-filter, and multi-hop layers use linear message and ego maps with an additive update under an elementwise nonlinearity. Taking
$\Psi_k^{(\ell)}(\mathbf{H}^{(\ell)})=\mathbf{H}^{(\ell)}\mathbf{W}_k^{(\ell)}$, additive
mixing, and the additive update $\phi_\ell(a,b)=\sigma(a+b)$ with $\sigma$ an elementwise
nonlinearity, \eqref{eq:unified-notation} reduces to the closed form
\begin{equation}
\bar{\mathbf{H}}^{(\ell+1)}
=
\sigma\!\Bigl(
\textstyle
\sum_{k=1}^{K_\ell}
\mathbf{P}_k^{(\ell)}\,
\mathbf{H}^{(\ell)}\,
\mathbf{W}_k^{(\ell)}
\;+\;
\mathcal{B}_\ell\!\bigl(\mathbf{H}^{(\ell)},\mathbf{H}^{(0)}\bigr)
\Bigr),
\qquad
\mathbf{H}^{(\ell+1)}=\mathbf{Q}_\ell\bar{\mathbf{H}}^{(\ell+1)},
\label{eq:unified-notation-linear}
\end{equation}
with $\mathbf{Q}_\ell=\mathbf{I}$ absent pooling.
GCN~\citep{kipf2017semi}, SGC~\citep{wu2019simplifying}, ChebNet~\citep{defferrard2016convolutional}, and the spectral and multi-hop families take
$\mathcal{B}_\ell=\mathbf{H}^{(\ell)}\mathbf{B}^{(\ell)}$ or $\mathcal{B}_\ell=0$.

\colorlet{eqboxframe}{googlered}
\colorlet{eqboxback}{googlered!6}

\begin{tcolorbox}[
  colback=eqboxback,
  colframe=eqboxframe,
  boxrule=0.6pt,
  arc=1mm,
  fonttitle=\bfseries\small,
  coltitle=white,
  title={Unified Equation},
  width=\linewidth,
  left=2mm, right=2mm, top=1.5mm, bottom=1.5mm,
  boxsep=1mm,
  before skip=4pt, after skip=4pt,
  breakable
]
\small
\begin{equation}
\begin{aligned}
\Psi_k^{(\ell)}
  &= \mathbf{H}^{(\ell)}\mathbf{W}_k^{(\ell)}
  \ \ (\text{or an edge-conditioned / geometric map}),
  && \quad\text{\footnotesize per-channel message} \\[3pt]
\mathbf{C}_k^{(\ell)}
  &= \begin{cases}
       \mathbf{P}_k^{(\ell)}\Psi_k^{(\ell)}, & \text{linear-value channel},\\[2pt]
       \bigl[\operatorname*{Agg}\limits_{j\in\mathcal{N}_k(i)}\!\bigl([\mathbf{P}_k^{(\ell)}]_{ij}\,\Psi_k^{(\ell)}(\vh_i^{(\ell)},\vh_j^{(\ell)},\ve_{ij}^{(\ell)})\bigr)\bigr]_i, & \text{pairwise channel},
     \end{cases}
  && \quad\text{\footnotesize channel contribution} \\[3pt]
\mixmsg^{(\ell)}
  &= \mathop{\boxplus}\limits_{k=1}^{K_\ell}\,\mathbf{C}_k^{(\ell)},
  && \quad\text{\footnotesize mix channels} \\[3pt]
\bar{\mathbf{H}}^{(\ell+1)}
  &= \phi_\ell\!\Bigl(\mathcal{B}_\ell\!\bigl(\mathbf{H}^{(\ell)},\mathbf{H}^{(0)}\bigr),\,\mixmsg^{(\ell)}\Bigr).
  && \quad\text{\footnotesize update}
\end{aligned}
\label{eq:unified-notation}
\end{equation}

\vspace{0.4em}
\scriptsize
\setlength{\tabcolsep}{4pt}
\renewcommand{\arraystretch}{1.25}
\begin{tabularx}{\linewidth}{
  @{}
  >{\centering\arraybackslash}p{0.07\linewidth}
  >{\raggedright\arraybackslash}p{0.30\linewidth}
  >{\raggedright\arraybackslash}X
  @{}
}
\toprule
\textbf{Slot} & \textbf{Question answered} & \textbf{Typical choices} \\
\midrule
$\mathcal{X}$ & What object is updated? & nodes, directed edges, tuples, subgraphs, simplices, cells \\
$\mathcal{K}$ & What channels are available? & hops, heads, relations, meta-paths, spectral terms, lifted ranks, tensor degrees \\
$\mathbf{P}_k$ & Where and with what weight does it move? & adjacency, hop powers, spectral filters, attention, relation operators \\
$\Psi_k$ & What value is moved? & linear value, edge-conditioned message, geometric or tensor-valued equivariant message \\
$\boxplus$ & How are channels combined? & sum, concatenation, semantic attention, gating \\
$\mathcal{B}$ & How is the prior state carried? & identity, learned self term, residual, initial-state teleport \emph{(the family comparison tables apply $\mathcal{B}$ within the $\phi$ column, following the nesting $\phi_\ell(\mathcal{B}_\ell(\cdot),\mixmsg^{(\ell)})$ of Eq.~\eqref{eq:unified-notation})} \\
$\phi$ & How is the new state produced? & nonlinearity, MLP, gate, equivariant update \\
\bottomrule
\end{tabularx}

\vspace{0.4em}
\footnotesize
Each $\Psi_k^{(\ell)}$ forms a message, and $\mathbf{C}_k^{(\ell)}$ propagates it through a matrix product or a permutation-invariant pairwise aggregation. The component $\boxplus$ mixes channels, $\mathcal{B}_\ell$ carries the prior state, and $\phi_\ell$ updates it. Matrix products cover linear-value channels, while edge-conditioned and geometric pairwise messages require the second form.
\end{tcolorbox}

This is comparison notation, not a new model. Its only modeling commitment is to separate support and scalar weights ($\mathbf{P}_k$), values ($\Psi_k$), channel mixing ($\boxplus$), and updating ($\phi_\ell$), making spectral, spatial, attention, and multi-hop layers directly comparable.

The lemma below gives exact spectral-calculus fillings for ChebNet, SGC, APPNP, GPR-GNN, and BernNet. Its proof is in \hyperref[proof:lem-spectral-subsumption]{Appendix~\ref*{sec:analysis}}.

\begin{lemma}[Spectral subsumption]
\label{lem:spectral-subsumption}
Any layer $\mathbf{H}^{(\ell+1)}=\sigma\big(\sum_k g_k(\mathbf{S})\,\Psi_k+\mathcal{B}_\ell\big)$, with each $g_k$ a fixed or learned scalar function applied through the spectral calculus of a fixed graph shift operator $\mathbf{S}$ and $\Psi_k$ a linear message, is an instance of the unified equation with $\mathbf{P}_k^{(\ell)}=g_k(\mathbf{S})$, additive mixing, $\mathcal{B}_\ell$ as the separate ego/residual input, and $\phi_\ell(a,b)=\sigma(a+b)$ combining it with the mixed channels. In particular, ChebNet~\citep{defferrard2016convolutional}, SGC~\citep{wu2019simplifying}, APPNP~\citep{klicpera2019predict}, GPR-GNN~\citep{chien2021adaptive}, and BernNet~\citep{he2021bernnet} are each \emph{exactly} a single point of this specialization. The operator entry alone does not distinguish them. They differ along three recorded axes.
\begin{enumerate}\setlength{\itemsep}{1pt}
\item[(i)] The first axis is the shift operator and basis inside $\mathbf{P}_k$. ChebNet uses $g_k=T_k(\tilde{\mathbf{L}}_{\mathrm{ch}})$ on the $[-1,1]$-rescaled $\mathbf{L}_{\mathrm{sym}}$, and BernNet uses a Bernstein basis in $\mathbf{L}_{\mathrm{sym}}/2\in[0,1]$, which presupposes this scaling. APPNP and GPR-GNN instead use powers of $\hat{\mathbf{A}}=\mathbf{I}-\mathbf{L}_{\mathrm{ref}}$ (Section~\ref{subsec:spectral-spatial-prelim}) as a multi-order bank, while SGC uses the single power $\hat{\mathbf{A}}^{K}$.
\item[(ii)] The second axis is the coefficient constraint folded into $\mathbf{P}_k$, including GPR-GNN's signed learned $\gamma_k$ and BernNet's nonnegative Bernstein weights. ChebNet carries \emph{no} scalar operator coefficient here. Its fixed basis is $\mathbf{P}_k=T_k(\tilde{\mathbf{L}}_{\mathrm{ch}})$, and its learnable freedom lies in the per-order feature matrices $\boldsymbol{\Theta}_k$ in the message (axis~(iii)), not in a scalar $\theta_k$ in $\mathbf{P}_k$. APPNP propagates a pre-transformed signal. With $\mathbf{G}^{(0)}=f_\theta(\mathbf{X})$ (an MLP of the input, distinct from the framework's state initialization $\mathbf{H}^{(0)}=\mathbf{X}$, and written $\mathbf{G}$ rather than the framework's reserved final-representation $\mathbf{Z}$), it is recovered in either of two ways. As a $K$-step iteration $\mathbf{G}^{(\ell+1)}=(1-\alpha)\hat{\mathbf{A}}\mathbf{G}^{(\ell)}+\alpha\mathbf{G}^{(0)}$ it is the single-channel filling $\mathbf{P}=(1-\alpha)\hat{\mathbf{A}}$ with ego term $\mathcal{B}_\ell=\alpha\mathbf{G}^{(0)}=\alpha f_\theta(\mathbf{X})$ (the teleport is to the transformed signal, not to $\alpha\mathbf{H}^{(0)}=\alpha\mathbf{X}$). Unrolled into a fixed polynomial in $\hat{\mathbf{A}}$, it has coefficients $\alpha(1-\alpha)^k$ for $k<K$ but top-order coefficient $(1-\alpha)^K$, since
\[
\mathbf{G}^{(K)}=(1-\alpha)^K\hat{\mathbf{A}}^K\mathbf{G}^{(0)}+\alpha\sum_{k=0}^{K-1}(1-\alpha)^k\hat{\mathbf{A}}^k\mathbf{G}^{(0)}.
\]
The uniform profile $\alpha(1-\alpha)^k$ across \emph{all} orders is exact only in the infinite-depth personalized-PageRank limit ($K\to\infty$), not for finite APPNP.
\item[(iii)] The third axis is the message slot $\Psi_k$. ChebNet propagates the current state with per-order feature matrices, $\Psi_k=\mathbf{H}^{(\ell)}\boldsymbol{\Theta}_k$, which are its learnable parameters rather than scalars. APPNP, GPR-GNN, and BernNet are \emph{decoupled} and propagate the layer-independent signal $\Psi_k=f_\theta(\mathbf{X})$, shared across channels without per-channel $\mathbf{W}_k$. SGC is instead the single channel $\mathbf{P}=\hat{\mathbf{A}}^{K}$ with one weight, $\hat{\mathbf{A}}^{K}\mathbf{X}\mathbf{W}$.
\end{enumerate}
Thus the relationship is exact special-case recovery, not equivalence between the framework and any one model. The phrase ``spectral versus multi-hop spatial'' describes differences along axes (i)--(iii) within the framework rather than a separate mechanism.
\end{lemma}

In APPNP's one-shot polynomial view, folding the teleport into $\mathbf{P}_k$ leaves $\mathcal{B}_\ell=0$, matching Table~\ref{tab:family3_spectral_slots}. The lemma below interprets each diagonal self-weight in this multi-order bank as a return probability. Its proof is in \hyperref[proof:lem-return]{Appendix~\ref*{sec:analysis}}.

\begin{lemma}[Return probabilities as structural self-weights]
\label{lem:return}
For $\mathbf{P}_k=\hat{\mathbf{A}}^{k}$, the diagonal entry $[\hat{\mathbf{A}}^{k}]_{ii}$ multiplying node $i$'s transformed feature $(\mathbf{H}\mathbf{W}_k)_i$ is the $k$-step return probability. Multi-hop channels therefore induce structural self-weighting through closed walks at $i$, including self-loops and immediate backtracking rather than only simple cycles.
\end{lemma}

Section~\ref{subsec:worked-reductions} recovers the two canonical departures from this specialization, namely GIN's nonlinear $\phi_\ell$ and GAT's state-dependent $\mathbf{P}_k$.

\paragraph{Non-core ingredients are folded into core slots.} Architectures may use objects beyond node features and fixed adjacency, but these enter the existing slots of \eqref{eq:unified-notation}.
\begin{itemize}\setlength{\itemsep}{2pt}
\item \emph{Cross-layer skips} use the ego/update pathway. $\mathcal{B}_\ell$ forms the skip and $\phi_\ell$ combines it, so tables record both in the joint $(\mathcal{B}_\ell,\phi_\ell)$ column. A dense skip reading every earlier layer exceeds the two-argument signature $\mathcal{B}_\ell(\mathbf{H}^{(\ell)},\mathbf{H}^{(0)})$ and must extend it or move to the readout $\rho$ (Appendix~\ref{subsec:op-residual}).
\item \emph{Global registers} such as virtual nodes or \textsc{[cls]} tokens enlarge $\mathcal X$ or add a channel in $\mathcal K$ and $\mathbf{P}_k$.
\item \emph{Geometric and positional information} enters $\mathbf{P}_k$, $\Psi_k$, or $\phi_\ell$.
\item \emph{Pooling, coarsening, sparsification, and rewiring} change the domain
$\mathcal X$ and/or the propagation operators $\mathbf{P}_k$ across layers.
\item \emph{Higher-order ranks} are handled by choosing $\mathcal X$ to be tuples,
subgraphs, cells, or graded objects and by using the corresponding inter-rank
propagation operators $\mathbf{P}_k$ (Section~\ref{subsec:graded-extension}).
\end{itemize}

\paragraph{Five excluded coverage cases.} Five boundary cases named in Section~\ref{subsec:framework-scope} fall outside the channel primitives. NeuralWalker~\citep{Chen2025NeuralWalker} (Family~4) uses a non-commutative, order-dependent walk mixer rather than invariant aggregation. NLGNN~\citep{liu2021non} (Family~2) hard-sorts nodes by learned scores before sequence convolution, creating multi-stage feature-dependent reindexing rather than either channel type. Sorting with distinct scores is equivariant, but ties require an equivariant rule. GPNN~\citep{yang2022graph} (Family~2) processes a sampled, possibly truncated BFS sequence with an LSTM pointer before sequence convolution and pooling, so enumeration can change under relabeling. MR-GNAS~\citep{zheng2022meta} is a search space rather than a layer. TransE/DistMult-style knowledge-graph embeddings (surveyed in~\citet{rossi2021knowledge}, Family~5) statically score triples without a layer-wise neighborhood update and are therefore excluded from Table~\ref{tab:family-heterogeneous}. These facts show that our reductions fail, not that no equivalent representation exists. Section~\ref{subsec:graded-extension} discusses the sixth case, NGNN.

\begin{figure*}[t]
    \centering
    \includegraphics[
        width=0.92\textwidth,
        keepaspectratio
    ]{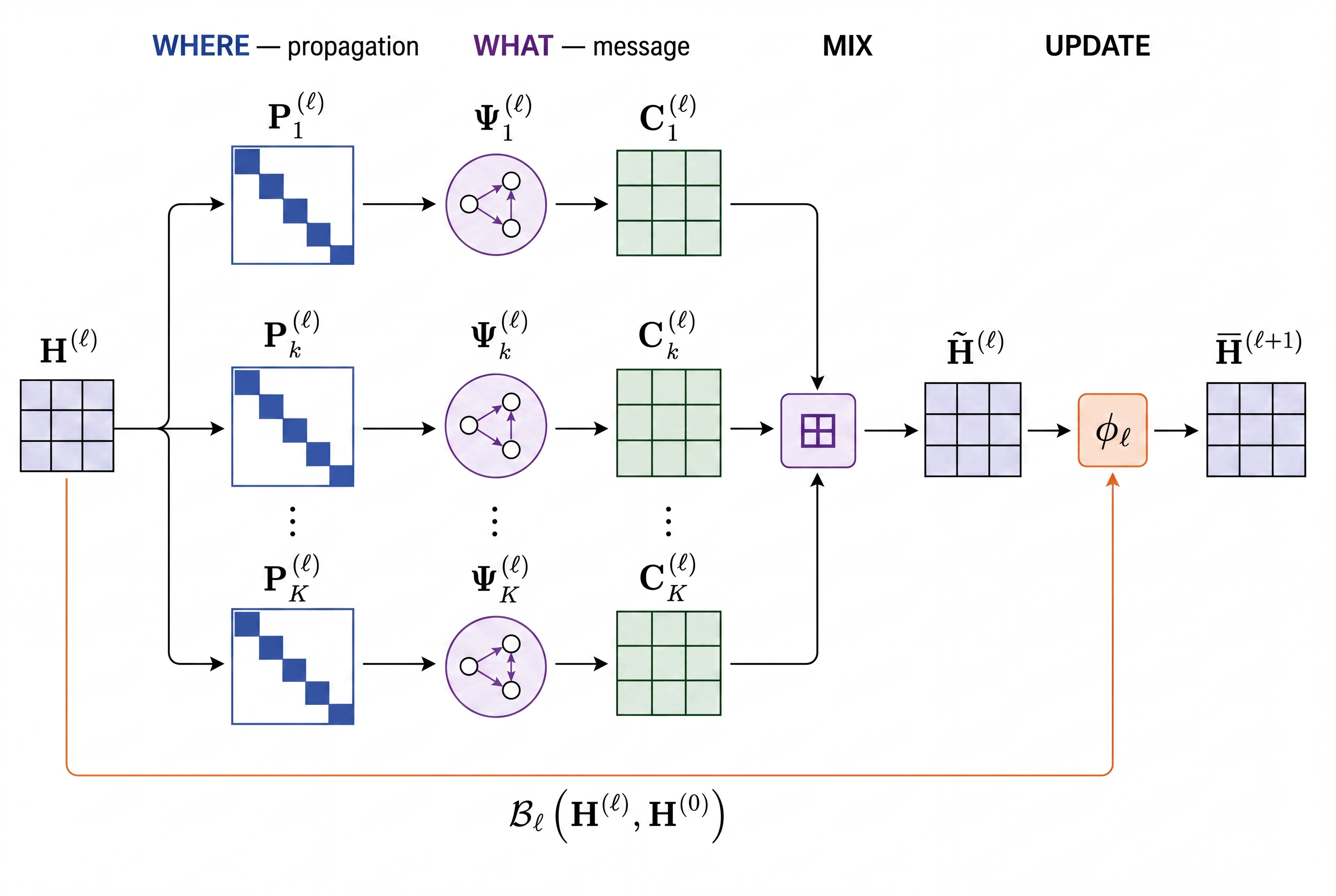}
    \caption{\textbf{The unified equation, slot by slot.}
    Each channel separates \emph{where} information propagates,
    set by the propagation operator $\mathbf{P}_k^{(\ell)}$,
    from \emph{what} information is transmitted, set by the
    message map $\Psi_k^{(\ell)}$. Combining the two gives the channel contribution
    $\mathbf{C}_k^{(\ell)}$. The channel-mixing operator
    $\boxplus$ mixes the contributions from every channel into the mixed message
    $\widetilde{\mathbf{H}}^{(\ell)}$ (written $\mixmsg^{(\ell)}$ in the text). The
    update map $\phi_\ell$ then combines $\widetilde{\mathbf{H}}^{(\ell)}$ with the
    residual/ego pathway $\mathcal{B}_\ell(\mathbf{H}^{(\ell)},\mathbf{H}^{(0)})$,
    which carries the node's own prior state, to produce
    the next-layer intermediate state $\overline{\mathbf{H}}^{(\ell+1)}$
    (Eq.~\eqref{eq:unified-notation}). Architectures differ only in how each of
    these seven components is filled. The domain $\mathcal X$ and channel set $\mathcal K$
    determine the objects and indexed branches shown in the diagram.}
    \label{fig:unified-framework}
\end{figure*}

The task representation is produced by a readout
\[
\mathbf{Z}
=
\rho\!\big(\mathbf{H}^{(0)},\mathbf{H}^{(1)},\dots,\mathbf{H}^{(L)}\big)\,
\mathbf{W}_{\mathrm{out}},
\]
which is usually the last-layer state. Jumping-knowledge designs (JKNet~\citep{xu2018representation}, Family~1) instead use the full stack. The readout lies outside a layer. Figure~\ref{fig:unified-framework} summarizes the within-layer decomposition, whose remaining placement ambiguity is resolved next.

\subsection{Slot Discipline}
\label{subsec:slot-discipline}
Because the equation admits multiple representations of one architecture, we apply the following assignment rule uniformly across families.
\begin{itemize}\setlength{\itemsep}{3pt}
\item[(i)] \emph{Operator $\mathbf{P}_k$ carries scalars only.} Its entries include structural normalizations ($\hat a_{ij}$), attention weights ($\alpha_{ij}$), and folded signed or spectral coefficients. Its sparsity pattern defines support.
\item[(ii)] \emph{Message $\Psi_k$ carries vector- and matrix-valued maps.} Edge-conditioned maps $\Theta(\ve_{ij})$, geometric or equivariant messages, and relation-composition maps belong here rather than in $\mathbf{P}_k$.
\item[(iii)] \emph{Ego map $\mathcal{B}_\ell$ forms the prior-state input, and update $\phi_\ell$ combines it with the mixed channels.} The map $\mathcal{B}_\ell$ assembles self, residual, initial-state teleport, and layer-wise skip terms. A self-loop instead belongs to $\mathbf{P}_k$ when it uses the same shared message and scalar-weighting rules as neighboring edges, even if its value differs. Thus GCN and GAT have $\mathcal{B}_\ell=0$, and GAT obtains both $\alpha_{ii}$ and $\alpha_{ij}$ from the same rule. GIN's independently parameterized $(1+\epsilon)$ is an ego term. Gates, normalization, and the combination of $\mathcal{B}_\ell$ with mixed channels belong to $\phi_\ell$. Unless noted otherwise, both maps are node-local and row-wise, with precise assumptions given in Appendix~\ref{sec:analysis}.
\item[(iv)] \emph{Mixing $\boxplus$ purely combines channel contributions, and $\mixmsg=\boxplus_k\mathbf{C}_k$ is its output.} Residual and ego terms instead enter through $\mathcal{B}_\ell$ and $\phi_\ell$.
\end{itemize}
Rule (iv) places GCNII's~\citep{chen2020simple} initial residual and APPNP's iterative teleport in $\mathcal{B}_\ell$ and $\phi_\ell$, while APPNP's equivalent polynomial form folds the teleport into $\mathbf{P}_k$ with $\mathcal{B}_\ell=0$ (Section~\ref{subsec:setup}). It also makes GraphSAGE's self-state concatenation an update rather than a channel. This discipline is a convention whose consistent application makes table cells comparable.

\subsection{Discriminating Axes}
\label{subsec:axes}

Because the equation alone cannot distinguish layers satisfying the two-channel-type condition of Section~\ref{subsec:framework-scope}, the tables record the axes exposed by the canonical reductions of Section~\ref{subsec:worked-reductions}.
\begin{itemize}\setlength{\itemsep}{1pt}
\item The operator bank and scalar weighting rule are $\{\mathbf{P}_k\}$.
\item The channel count is $K_\ell$, with mixing $\boxplus$.
\item The coefficient constraint is folded into $\mathbf{P}_k$.
\item The message map $\Psi_k$ may be a linear $\mathbf{W}_k$, an MLP, or an attention-value map,
together with its aggregator $\mathrm{Agg}\in\{\mathrm{max},\text{power-mean},\ldots\}$ when
it is nonlinear. Linear sum and mean are instead folded into $\mathbf{P}_k$.
\item Ego handling is recorded in $\mathcal B_\ell$, with its combination rule in $\phi_\ell$.
\item Any non-core ingredient is folded into $\mathcal X$, $\mathbf{P}_k$, $\Psi_k$, or $\phi_\ell$.
\end{itemize}

These axes are empirically discriminating rather than theoretically minimal. Removing the weight-source distinction collapses fixed and state-dependent edge weights, while removing ego handling collapses distinct ways of inserting the self state into $\phi$. Section~\ref{subsec:worked-reductions} demonstrates both cases, and Table~\ref{tab:family-slot-map} marks each family's defining axis.

The Family~1/Family~3 split uses the checkable operational distinction between fixed support and a polynomial basis, independent of deeper equivalence in realizable functions. \citet{jiang2026spectralneither} argue that spectral $\mathbf{P}_k$ may not be theoretically distinct from spatial message passing for node classification. If confirmed, this would motivate merging the families in that setting. We retain the operational split pending resolution of this narrower question (Section~\ref{subsec:op-evaluation}).

Tables~\ref{tab:family-slot-map} and~\ref{tab:representative-slot-sampler} in Section~\ref{sec:introduction} display seven components in six columns, $(\mathcal X,\mathcal K,\mathbf P_k,\Psi_k,\boxplus,(\mathcal B_\ell,\phi_\ell))$. The shared Ego/Update column reports distinct $\mathcal B_\ell$ and $\phi_\ell$ components, including support-folded self states in GCN/GAT, GraphSAGE concatenation, GIN's scaled sum, and APPNP's teleport. The Mixing column reports the action of $\boxplus$, not an additional $\mixmsg$ component. The sampler gives three abbreviated methods per family, while Appendix~\ref{app:family-comparison-tables} gives full expressions. Its bands mark each family's defining axis. We now derive a representative subset.

\subsection{Worked Reductions for the Canonical Models}
\label{subsec:worked-reductions}

We verify coverage on GCN, GraphSAGE, GAT, and GIN. To express GraphSAGE concatenation and GIN's MLP, write the row-vector state as $h_i^{(\ell)}\equiv\vh_i^{(\ell)}\in\mathbb{R}^{1\times d_\ell}$ and expand the message and aggregation per node. With $\mathbf{Q}_\ell=\mathbf{I}$, one channel, and no skip or global terms, the layer is
\begin{equation}
h_i^{(\ell+1)}
=
\phi_\ell\!\Bigl(
\mathcal{B}_\ell(h_i^{(\ell)}),\;
\underbrace{
\textstyle
\operatorname*{Agg}_{j\in\mathcal{N}_1(i)}
\left(
[\mathbf{P}_1^{(\ell)}]_{ij}\,
\Psi^{(\ell)}(h_i^{(\ell)},h_j^{(\ell)},e_{ij}^{(\ell)})
\right)
}_{[\mathbf{C}_1^{(\ell)}]_{i:}}
\Bigr),
\label{eq:unified-notation-pernode}
\end{equation}
The underbrace is $[\mathbf{C}_1^{(\ell)}]_{i:}=\mixmsg_i^{(\ell)}$. With multiple channels, $\mixmsg_i^{(\ell)}=\boxplus_k[\mathbf{C}_k^{(\ell)}]_{i:}$. Here $\mathcal{N}_1(i)=\{j:[\mathbf{P}_1]_{ij}\neq 0\}$ and $[\mathbf{P}_1]_{ij}$ is the scalar propagation weight. GCN and GAT include the self state in $\mathcal{N}_1(i)$ and set $\mathcal{B}_\ell=0$. GraphSAGE and GIN exclude it and use $\mathcal{B}_\ell=\mathrm{id}$, after which $\phi_\ell$ respectively concatenates the ego state or applies an MLP to $(1+\epsilon)h_i^{(\ell)}+\mathrm{agg}$. The scaling $(1+\epsilon)$ belongs to $\phi_\ell$.

\paragraph{GCN \citep{kipf2017semi}.}
\emph{Slot choices.} $\mathcal{N}_1(i)=\mathcal{N}(i)\cup\{i\}$, with the self-loop support
defined by $\tilde{\mathbf{A}}=\mathbf{A}+\mathbf{I}$.
\[
[\mathbf{P}_1]_{ij}
=
\hat a_{ij}
=
\frac{\tilde{\mathbf{A}}_{ij}}{\sqrt{\tilde d_i\tilde d_j}}
\quad\text{for }j\in\mathcal{N}_1(i),
\]
The entry is zero otherwise. The message is $\Psi^{(\ell)}(h_i,h_j,e_{ij})=h_j\mathbf{W}^{(\ell)}$, $\mathrm{Agg}=\mathrm{sum}$, and $\phi_\ell(\cdot,\mixmsg_i)=\sigma(\mixmsg_i)$.
\emph{Substitution into \eqref{eq:unified-notation-pernode}:}
\[
h_i^{(\ell+1)}
=
\sigma\!\Bigl(
\textstyle
\sum_{j\in\mathcal{N}(i)\cup\{i\}}
\hat a_{ij}\,
h_j^{(\ell)}\mathbf{W}^{(\ell)}
\Bigr)
=
\sigma\!\bigl([\hat{\mathbf{A}}\mathbf{H}^{(\ell)}\mathbf{W}^{(\ell)}]_i\bigr),
\]
which is the defining update $\mathbf{H}^{(\ell+1)}=\sigma(\hat{\mathbf{A}}\mathbf{H}^{(\ell)}\mathbf{W}^{(\ell)})$. The scalar normalization $\hat a_{ij}$ appears only in $\mathbf{P}_1$.

\paragraph{GraphSAGE mean \citep{hamilton2017inductive}.}
\emph{Slot choices.} Self is excluded from $\mathcal{N}_1(i)=\mathcal{N}(i)$. The mean enters $[\mathbf{P}_1]_{ij}=1/d_i=[\mathbf{D}^{-1}\mathbf{A}]_{ij}$, where $d_i=|\mathcal{N}(i)|$. With $\Psi(h_i,h_j,e_{ij})=h_j$ and $\mathrm{Agg}=\mathrm{sum}$, $[\mathbf{C}_1^{(\ell)}]_{i:}=\mathrm{mean}_{j\in\mathcal{N}(i)}h_j^{(\ell)}$, and $\phi$ concatenates the ego input,
\[
\phi_\ell(h_i,[\mathbf{C}_1^{(\ell)}]_{i:})
=
\sigma\!\bigl([\,h_i\Vert [\mathbf{C}_1^{(\ell)}]_{i:}\,]\mathbf{W}^{(\ell)}\bigr).
\]
\emph{Substitution:}
\[
h_i^{(\ell+1)}
=
\sigma\!\Bigl(
\bigl[
h_i^{(\ell)}
\Vert
\mathrm{mean}_{j\in\mathcal{N}(i)}h_j^{(\ell)}
\bigr]
\mathbf{W}^{(\ell)}
\Bigr),
\]
which is the core mean GraphSAGE update under the row-vector convention, excluding sampling and optional post-update normalization. Unlike GCN, it supplies the self state separately to $\phi$.

\paragraph{GAT \citep{velickovic2018gat}.}
\emph{Slot choices.} The data-dependent operator uses $\mathcal{N}_1(i)=\mathcal{N}(i)\cup\{i\}$ and attention scores
\[
s_{ij}^{(\ell)}
=
\mathrm{LeakyReLU}
\!\left(
[
h_i^{(\ell)}\mathbf{W}^{(\ell)}
\Vert
h_j^{(\ell)}\mathbf{W}^{(\ell)}
]
\mathbf{a}^{(\ell)}
\right),
\qquad
\alpha_{ij}^{(\ell)}
=
\frac{\exp(s_{ij}^{(\ell)})}
{\sum_{r\in\mathcal{N}_1(i)}\exp(s_{ir}^{(\ell)})}.
\]
Thus $[\mathbf{P}_1(\mathbf{H}^{(\ell)})]_{ij}=\alpha_{ij}^{(\ell)}$ for $j\in\mathcal{N}_1(i)$ and zero otherwise. The message is $\Psi^{(\ell)}(h_i,h_j,e_{ij})=h_j\mathbf{W}^{(\ell)}$, with edge features unused in the original GAT layer. Also, $\mathrm{Agg}=\mathrm{sum}$ and $\phi_\ell(\cdot,\mixmsg_i)=\sigma(\mixmsg_i)$.
\emph{Substitution:}
\[
h_i^{(\ell+1)}
=
\sigma\!\Bigl(
\textstyle
\sum_{j\in\mathcal{N}(i)\cup\{i\}}
\alpha_{ij}^{(\ell)}\,
h_j^{(\ell)}\mathbf{W}^{(\ell)}
\Bigr),
\]
which is the defining single-head GAT update. Unlike GCN's fixed normalization, $[\mathbf{P}_1]_{ij}$ is learned and state-dependent. Although $\mathbf{W}^{(\ell)}$ appears in both the score and message, $\mathbf{P}_1$ stores only the scalar $\alpha_{ij}$ and $\Psi$ the value $h_j\mathbf{W}^{(\ell)}$. The lemma below extends this reduction to multi-head mixing and, in the single-head edge-feature-free case, GATv2. Its proof is in \hyperref[proof:lem-attention-subsumption]{Appendix~\ref*{sec:analysis}}.

\paragraph{GIN \citep{xu2019powerful}.}
\emph{Slot choices.} Self is excluded from $\mathcal{N}_1(i)=\mathcal{N}(i)$, $[\mathbf{P}_1]_{ij}=1$ for $j\in\mathcal{N}(i)$, $\Psi(h_i,h_j,e_{ij})=h_j$, and $\mathrm{Agg}=\mathrm{sum}$. The ego input enters $\phi$ as
\[
\phi_\ell(h_i,[\mathbf{C}_1]_{i:})
=
\mathrm{MLP}^{(\ell)}
\!\left(
(1+\epsilon^{(\ell)})h_i+[\mathbf{C}_1]_{i:}
\right).
\]
\emph{Substitution:}
\[
h_i^{(\ell+1)}
=
\mathrm{MLP}^{(\ell)}
\!\Bigl(
(1+\epsilon^{(\ell)})h_i^{(\ell)}
+
\textstyle
\sum_{j\in\mathcal{N}(i)}
h_j^{(\ell)}
\Bigr),
\]
which is the defining GIN update. Unlike GraphSAGE, the self term enters as a scaled sum
inside the MLP rather than as a separate concatenated block.

\begin{lemma}[Attention subsumption]
\label{lem:attention-subsumption}
Any single-head layer whose neighbor weight is a scalar function
$\alpha_{ij}(\mathbf{H})$ of node states multiplying a linear value
$h_j\mathbf{W}$ is an instance with state-dependent operator
$\mathbf{P}_k=[\alpha_{ij}(\mathbf{H})]$ and message
$\Psi_k=\mathbf{H}\mathbf{W}$. Multiple heads use per-head mixing
$\boxplus$ over the head index. This is concatenation $\Vert_m$ at intermediate
layers (Prop.~\ref{prop:concat}) and, at a final fixed-width layer, the
head average $\tfrac1M\sum_m(\cdot)$, which is the additive mixing of
Prop.~\ref{prop:concat} rescaled. Edge features, when present, enter
$\mathbf{P}_k$ if they modulate the scalar attention score, $\Psi_k$ if
they are added on the value side, or both. Restricted to the single-head,
edge-feature-free case, GAT~\citep{velickovic2018gat} and
GATv2~\citep{brody2022attentive} differ only in the functional form of
$\alpha_{ij}$ inside $\mathbf{P}_k$.
\end{lemma}

\paragraph{Reading the reductions.} The tables assign all seven components under the fixed slot discipline and are therefore slot-complete at the declared reporting granularity. Each component cell records its assigned computational role and may abstract internal parameterizations that preserve that role. Two attention layers may, for example, compute $\boldsymbol\alpha(\mathbf H^{(\ell)})$ with different scoring functions while sharing every displayed component assignment. Differing slots localize the distinctions represented at this granularity. GraphSAGE and GIN differ in operator normalization ($\mathbf{D}^{-1}\mathbf{A}$ versus $\mathbf{A}$) and ego handling (concatenation versus scaled addition), while GCN and GAT differ in fixed versus state-dependent scalar weights. This role-based localization underlies Appendix~\ref{app:family-comparison-tables}. Higher-order models additionally require a domain extension.

\subsection{Graded Extension for Higher-Order Models}
\label{subsec:graded-extension}

Higher-order models update tuples, subgraphs, or cells, so we replace the node domain $\mathcal{X}=\mathcal{V}$ by the graded support
\[
\mathcal{X}^{(\ell)}
=
\bigsqcup_{r=0}^{R}\mathcal{X}_r^{(\ell)},
\qquad
\mathbf{H}^{(\ell)}
=
(\mathbf{H}_r^{(\ell)})_{r=0}^{R}.
\]
where rank $0$ contains nodes and higher ranks contain higher-order objects. Propagation uses inter-rank operators
\[
\mathbf{P}_{k,r\to s}^{(\ell)}
\in
\mathbb{R}^{|\mathcal{X}_s^{(\ell)}|\times |\mathcal{X}_r^{(\ell)}|},
\]
where $k$ indexes channels and $r\to s$ identifies source and target ranks. Unlike the square operators of Section~\ref{subsec:setup}, these matrices are generally rectangular for $r\neq s$. The coverage condition of Section~\ref{subsec:framework-scope} applies to this \emph{inter-rank linear} branch. Pairwise $\mathbf{C}_k$ similarly extends to geometric and edge-conditioned messages $\Psi_k(\vh_i,\vh_j,\ve_{ij})$ that read one target--source pair before aggregation.

Substitution into \eqref{eq:unified-notation} gives one equation per target rank $s$, mixing all source ranks mapped into it.
\begin{equation}
\mathbf{C}_{k,r\to s}^{(\ell)} = \mathbf{P}_{k,r\to s}^{(\ell)}\Psi_k^{(\ell)}(\mathbf{H}_r^{(\ell)}),
\qquad
\mixmsg_s^{(\ell)} = \mathop{\boxplus}_{k,r}\mathbf{C}_{k,r\to s}^{(\ell)},
\qquad
\bar{\mathbf{H}}_s^{(\ell+1)} = \phi_\ell\big(\mathcal{B}_\ell(\mathbf{H}_s^{(\ell)},\mathbf{H}_s^{(0)}),\,\mixmsg_s^{(\ell)}\big),
\label{eq:graded-notation}
\end{equation}
The pairwise branch generalizes one target source pair at a time. Taking $R=0$ and $\mathbf{P}_{k}^{(\ell)}=\mathbf{P}_{k,0\to0}^{(\ell)}$ recovers \eqref{eq:unified-notation} exactly.

PPGN's bilinear tuple product~\citep{maron2019provably}, $\mathbf{Y}_{ijc}=\sum_t \mathbf{M}_{1,itc}\mathbf{M}_{2,tjc}$ with $\mathbf{M}_1=\mathrm{MLP}_1(\mathbf{H})$ and $\mathbf{M}_2=\mathrm{MLP}_2(\mathbf{H})$, fits the \emph{inter-rank linear} branch after expanding output channels $c$. On $\mathcal{X}=\mathcal{V}^2$, define
\[
[\mathbf{P}_c(\mathbf{H})]_{(i,j),(t,j')}=\mathbf 1_{\{j=j'\}}\,\mathbf{M}_{1,itc},
\qquad
[\Psi_c(\mathbf{H})]_{(t,j')}=\mathbf{M}_{2,tj'c},
\]
which gives $[\mathbf{C}_c]_{(i,j)}=\sum_{t,j'}[\mathbf{P}_c]_{(i,j),(t,j')}[\Psi_c]_{(t,j')}=\mathbf{Y}_{ijc}$. Concatenating over $c$ recovers the product exactly. Because each $[\mathbf{P}_c]$ is a state-dependent scalar, this is the propagation bank of Section~\ref{subsec:setup} indexed by output channel rather than a new primitive, at the cost of one operator per output channel. Let $\mathbf{C}_\times:=\boxplus_c\mathbf{C}_c$ denote the concatenated bank, shown as $\{\times\}$ in the slot sampler. The full PPGN block separately applies $\mathrm{MLP}_3$ to the retained pair state and concatenates it with $\mathbf{C}_\times$. Applying one $\mathrm{MLP}_3$ after their joint concatenation would misstate the block. Under Section~\ref{subsec:slot-discipline}, the retained state is $\mathcal{B}_\ell$ and enters $\phi_\ell$, not $\mixmsg$.

NGNN~\citep{zhang2021nested}, the sixth boundary case of Section~\ref{subsec:framework-scope}, instead runs an inner GNN and pooling inside each subgraph. This multi-stage composition does not admit a one-layer channel filling.

The extension covers dense tuple supports $\mathcal{X}_r=\mathcal{V}^{r+1}$ used by invariant/equivariant and Weisfeiler Leman-style models, as well as sparse cell complexes whose boundary or incidence maps define propagation, as in CW Networks~\citep{Bodnar2021CWNetworks}. Restricting this notation to the higher-order family keeps the remaining tables node-level.

\section{Generated Architectures from the Unified Equation} \label{sec:generation}

Once covered architectures are expressed through the unified system, the component inventory exposes a
structured design space. We illustrate this consequence through three increasingly open
forms of architecture generation. Level~1 holds Eq.~\eqref{eq:unified-notation} and the catalogued
filler inventory fixed while recombining choices across families. Level~2 holds the seven
components fixed while proposing a compatible filler absent from that inventory. Level~3
extends the equation with a typed component that expresses computation outside the current
boundary. The levels differ only in what the unification holds fixed. They generate
structurally checkable architectures whose empirical performance remains untested.
Appendix~\ref{sec:operations} catalogs a complementary source of moves at this granularity:
reusable operations, such as positional encodings, rewiring, residuals, normalization,
pooling, continuous-depth views, and sampling, that already carry a primary-slot assignment,
so transplanting one across families or introducing a new filler for the slot it targets reads
directly as a Level~1 or Level~2 move without a separate proposal step.

\subsection{Procedure for Generated Architectures}
\label{subsec:generation-method}
\begin{figure}[!t]
    \centering
    \includegraphics[
        width=\linewidth,
        trim=0 5 0 0,
        clip
    ]{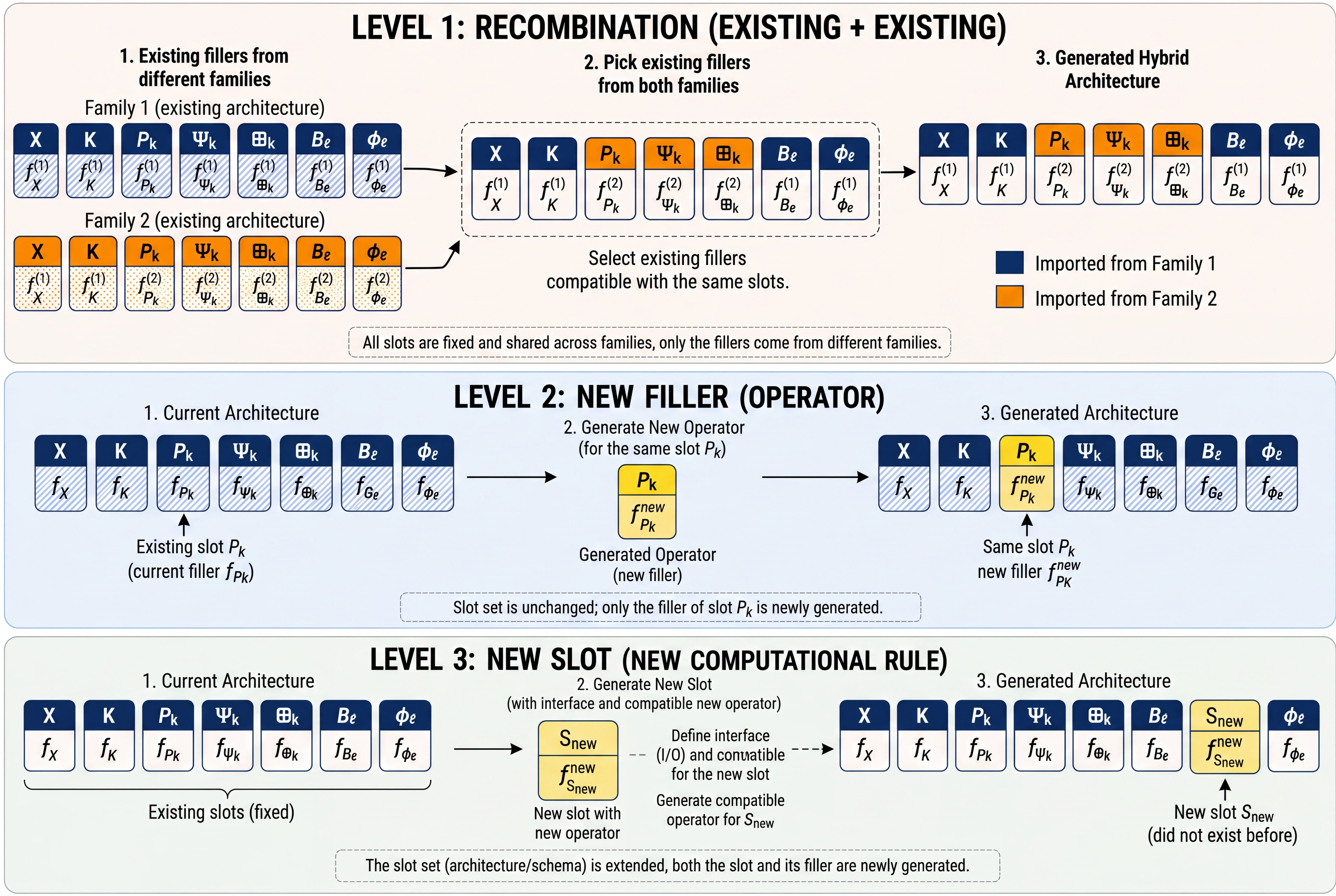}
    \caption{\textbf{Generated architectures at three levels.}
    The fixed inventory in Levels~1--2 contains the seven components
    $(\mathcal X,\mathcal K,\mathbf P_k,\Psi_k,\boxplus,\mathcal B_\ell,\phi_\ell)$.
    Level~1 recombines compatible existing fillings across families. Level~2 replaces the
    filling of an existing component (illustrated for $\mathbf P_k$) without changing that
    inventory. Level~3 extends the schema with a compatible new computational component.
    The generic labels $f_{\bullet}$ denote component fillings and do not imply that the
    set-valued components $\mathcal X$ and $\mathcal K$ are functions.
    Removing the added slot recovers Eq.~\eqref{eq:unified-notation} on the node domain, or,
    for a graded-domain Level~3 layer, first the synchronous graded form
    Eq.~\eqref{eq:graded-notation} and then Eq.~\eqref{eq:unified-notation} at its rank-zero
    case. The sequential node--edge example instantiates Level~3 with a schedule that makes
    the edge update precede the node update.}
    \label{fig:unified-gnn-generation}
\end{figure}

Only Level~1 is a search in the neural architecture search sense
\citep{wang2022automated} because it selects from a finite inventory. Levels~2 and~3 require
a proposal mechanism for a new filler or component. We use a language-model controller of
the kind applied to graph NAS \citep{gao2025llm4gnas}, supplying the unified equation, slot
definitions, admissible inventory, and generation level. The controller proposes functional
forms in the shared vocabulary. We audit each proposal against the type, symmetry, dimensional,
inventory, and reduction conditions defined for its level, and retain only those that satisfy
these conditions. Each retained Level~1--2 architecture specifies all seven components, while a
Level~3 architecture also specifies its added component and extended layer equation.
Figure~\ref{fig:unified-gnn-generation} summarizes which part of this specification is held
fixed and which part changes at each of the three levels.

We constrain architecture generation by level. At \textcolor{googleblue}{\bfseries level~1}, a generated architecture must be permutation
equivariant over the node index, be dimensionally consistent across channels,
absorb scalar per-channel weights into the operator, not reduce to a named
method in the per-family tables, and draw its fillers from at least two
families. At \textcolor{googlegreen}{\bf level~2}, the new filler must respect the type of the slot it fills
and the same conventions, and the resulting layer must not reduce to an existing
method. 
At \textcolor{googlered}{\bfseries level~3}, the added slot must be typed and must preserve permutation equivariance over nodes, and removing it must recover Eq.~\eqref{eq:unified-notation}.
This last check keeps an expansion anchored to the base equation rather than free
of it.

We retain generated architectures that pass the checks for their level and record the level, source
families or fillers, and closest existing method together with the component that differs.
The full protocol is given in Prompt~\ref{prompt:gen-prompt}. Six generated architectures
survive this audit, five at Levels~1--2 and the Sequential Co-Lifted Node--Edge Layer at
Level~3. The Level~3 architecture adds an
execution schedule to distinguish sequential writeback from synchronous graded coupling.
Together, the generated architectures instantiate all three generation levels.

\definecolor{pLvlOne}{HTML}{D97706}
\definecolor{pLvlTwo}{HTML}{0891B2}
\definecolor{pLvlThree}{HTML}{7C3AED}

\begin{SaveVerbatim}[commandchars=\\\{\}]{GenPrompt}
You generate graph neural network layers from a fixed unified equation.
You may operate at one of three levels.

\textcolor{pLvlOne}{\textbf{Level 1}} (recombination): keep the equation and the filler
  inventory fixed, and choose a new combination of existing fillers across families.
\textcolor{pLvlTwo}{\textbf{Level 2}} (new filler): keep the equation and slot structure
  fixed, and introduce a new admissible filler for one slot the inventory lacks.
\textcolor{pLvlThree}{\textbf{Level 3}} (expansion): add or generalize a slot of the
  equation to express layers the current equation cannot.

\textbf{Base layer:}
  H^\{l+1\} = phi( B(H^l, H^0),  [+]_\{k in K\} C_k )
  where each C_k is either P_k Psi_k (linear-value channel)
  or Agg_\{j in N_k(i)\} [P_k]_\{ij\} Psi_k(h_i,h_j,e_ij) (pairwise channel).

\textbf{Slots and admissible fillers:}
  X   update domain: nodes | edges | node tuples | rooted subgraphs | cells
  K   channel set:   single | hop or polynomial order | relation or type
                     | attention heads | geometric components
  P_k propagation:   fixed normalized adjacency | polynomial of the Laplacian
                     | per-relation adjacency | state-dependent attention
                     | global dense attention
  Psi_k message:     linear map of state | gated or MLP message
                     | relation-specific map | equivariant scalar-vector message
  [+] mix:           sum | mean | max | concat-then-project
  B   ego/residual:  none | self term | initial-state residual | gated residual
  phi update:        identity | activation | MLP | norm-then-activation
                     | equivariant update

\textbf{Conventions:}
  - Scalar per-channel coefficients are absorbed into P_k  (P_k <- gamma_k P_k).
  - The layer must be permutation equivariant over nodes.
  - Messages across channels must be dimensionally consistent.
  - B and phi are distinct components reported jointly in the Ego/Update column.

\textbf{Validity by level:}
  - \textcolor{pLvlOne}{\textbf{Level 1:}} must not reduce to a method in the reference
    list, and must draw fillers from at least two families.
  - \textcolor{pLvlTwo}{\textbf{Level 2:}} the new filler must respect the slot type and
    conventions, and the layer must not reduce to an existing method.
  - \textcolor{pLvlThree}{\textbf{Level 3:}} the added slot must be typed and permutation
    equivariant, and removing it must recover the base equation.

\textbf{For each generated layer return:}
  1. name and level (1, 2, or 3)
  2. the six-column display of seven components (X, K, P_k, Psi_k, mix, (B, phi))
     for level 3, also display the added component
  3. the fully instantiated layer equation
  4. for level 2, the new filler and the slot it fills
     for level 3, the added slot and the extended equation
  5. the closest existing method and the single slot that differs

\textbf{Reference list:}
  GCN, GraphSAGE, GAT, GIN, ChebNet, GPR-GNN, APPNP, RGCN, HAN,
  Graphormer, GPS, EGNN, ESAN
\end{SaveVerbatim}

\begin{prompt}[t]
\centering
\begin{tcolorbox}[
  colback=eqboxback,
  colframe=eqboxframe,
  boxrule=0.6pt,
  arc=1mm,
  fonttitle=\bfseries\small,
  coltitle=white,
  title={Generation protocol for the unified equation, slot inventory, and per-level validity constraints},
  width=\linewidth,
  left=2mm, right=2mm, top=1mm, bottom=1mm,
  boxsep=0.5mm,
]
\BUseVerbatim[fontsize=\scriptsize,baselinestretch=1.15]{GenPrompt}
\end{tcolorbox}
\caption{The generation protocol supplied to a language-model controller. It comprises the unified
equation~\eqref{eq:unified-notation}, the slot inventory and admissible fillers, and the
per-level validity conditions a generated layer must satisfy, for recombining fillers, introducing
a new filler, or expanding the equation.}
\label{prompt:gen-prompt}
\end{prompt}

\subsection{Generated Architectures}
\label{subsec:generated-architectures}
We describe six generated architectures. The first five make controlled changes to existing
fillings at Levels~1--2. The Sequential Co-Lifted Node--Edge Layer adds an execution
schedule that orders two graded updates. Table~\ref{tab:generated} gives their component
views.

\begin{table*}[t]
\centering
\scriptsize
\renewcommand{\arraystretch}{1.25}
\setlength{\tabcolsep}{2pt}
\renewcommand{\tabularxcolumn}[1]{m{#1}}
\newcommand{\genann}[1]{{\fontsize{6.5}{7}\selectfont #1}}
\caption{
\textbf{Generated architectures from the unified equation.}
Architectures retained by the procedure of Section~\ref{sec:generation}.
Levels~1--2 use the seven components in six base columns, while Level~3 adds the new slot
$\mathcal S_\ell$. The Mixing and Ego/Update columns follow the conventions of
Section~\ref{subsec:setup}. Here $\mathbf B_{\mathrm{spec}}=\sum_{k=0}^{K}\beta_k
T_k(\tilde{\mathbf L}_{\mathrm{ch}})$ is the learned spectral bias. The symbol
$^{\star}$ marks the component changed
relative to the closest method, and $^{\dagger}$ identifies the two rows using the
linear-specialization update
$\sigma(\mathbf H^{(\ell)}\mathbf W_0+\mixmsg^{(\ell)})$ of
Eq.~\eqref{eq:unified-notation-linear}.}
\label{tab:generated}
\begin{tabularx}{\textwidth}{@{}
>{\raggedright\arraybackslash}m{2.00cm}
>{\centering\arraybackslash}m{0.85cm}
>{\centering\arraybackslash}m{1.15cm}
>{\centering\arraybackslash}m{3.05cm}
>{\centering\arraybackslash}m{2.45cm}
>{\centering\arraybackslash}m{2.15cm}
>{\centering\arraybackslash}m{2.45cm}
>{\centering\arraybackslash}m{1.30cm}
@{}}
\toprule
\textbf{Method} &
\makecell[c]{$\boldsymbol{\mathcal X}$} &
\makecell[c]{$\boldsymbol{\mathcal K}$} &
\makecell[c]{\textbf{Propagation}\\$\boldsymbol{\mathbf P_k^{(\ell)}}$} &
\makecell[c]{\textbf{Message}\\$\boldsymbol{\Psi_k^{(\ell)}}$} &
\makecell[c]{\textbf{Mixing}\\$\boldsymbol{\boxplus^{(\ell)}}$} &
\makecell[c]{\textbf{Ego/Update}\\$\boldsymbol{(\mathcal B_\ell,\phi_\ell)}$} &
\makecell[c]{\textbf{New Slot}\\$\boldsymbol{\mathcal S_\ell}$}
\\
\midrule

\rowcolor{googleblue}
\multicolumn{8}{@{}>{\raggedright\arraybackslash}m{16.38cm}@{}}{
\textcolor{white}{\textbf{Level 1\ \ Recombination}\quad
\emph{equation and inventory fixed, with fillers combined across families}}}
\\[2pt]

\rowcolor{googleblue!25}
\makecell[l]{Spectrally-Gated\\Propagation} &
$V$ &
\makecell[c]{$\{1,\ldots,$\\[-1pt]$K\}$} &
\makecell[c]{$\Diag\!\bigl(\boldsymbol\gamma_k($\\[-1pt]$\mathbf H^{(\ell)})\bigr)^{\!\star}$\\[-1pt]${}\cdot T_k(\tilde{\mathbf L}_{\mathrm{ch}})$\\[1pt]\genann{(node-adaptive)}} &
$\mathbf H^{(\ell)}\mathbf W_k$ &
$\displaystyle\sum_{k=1}^{K}\mathbf P_k\Psi_k$ &
\makecell[c]{$\sigma\!\bigl(\mathbf H^{(\ell)}\mathbf W_0$\\[-1pt]${}+\mixmsg^{(\ell)}\bigr)^{\dagger}$} &
--
\\[3pt]

\rowcolor{googleblue!25}
\makecell[l]{Relation-\\Equivariant\\Network} &
\makecell[c]{$V$\\\genann{$(s,v)$}} &
\makecell[c]{$\mathcal R$\\\genann{(relations)}} &
$\hat{\mathbf A}_r$ &
\makecell[c]{$\Psi_r^{\star}\!\bigl(\mathbf h_i,\mathbf h_j,$\\[-1pt]$\lVert\mathbf x_i-\mathbf x_j\rVert\bigr)$\\[1pt]\genann{(equiv.\ $(s,v)$)}} &
\makecell[c]{$\mathbf C_{r,i}=\operatorname*{Agg}$\\[-1pt]$_{j\in\mathcal N_r(i)}[\mathbf P_r]_{ij}\Psi_r,$\\[2pt]$\mixmsg_i=\displaystyle\sum_{r\in\mathcal R}\mathbf C_{r,i}$} &
\makecell[c]{$\phi^{s},\phi^{v}$\\\genann{(scalar/vector)}} &
--
\\[3pt]

\rowcolor{googleblue!25}
\makecell[l]{Subgraph-\\Attentive Layer} &
\makecell[c]{$\mathcal S$\\\genann{(subgr.)}} &
\makecell[c]{$\{1,\ldots,$\\[-1pt]$H\}$\\\genann{(heads)}} &
\makecell[c]{$\mathrm{softmax}\!\bigl($\\[-1pt]$\mathbf Q_h\mathbf K_h^{\top}\bigr)^{\!\star}$\\[1pt]\genann{(state-dependent)}} &
$\mathbf H_{\mathcal S}^{(\ell)}\mathbf W_{V,h}$ &
$\displaystyle\big\Vert_{h}\mathbf P_h\Psi_h$ &
\makecell[c]{$\MLP(\mixmsg^{(\ell)})$\\[-1pt]${}+\mathbf H_{\mathcal S}^{(\ell)}$\\[1pt]\genann{(external $\mathbf Q_\ell$ fold}\\[-1pt]\genann{$\mathcal S\!\to\!V$)}} &
--
\\[3pt]

\rowcolor{googleblue!25}
\makecell[l]{Spectral-Biased\\Transformer} &
$V$ &
\makecell[c]{$\{1\}$\\\genann{(global)}} &
\makecell[c]{$\mathrm{softmax}\!\bigl($\\[-1pt]$\mathbf Q\mathbf K^{\top}/\sqrt d$\\[-1pt]${}+\mathbf B_{\mathrm{spec}}^{\star}\bigr)$} &
$\mathbf H^{(\ell)}\mathbf W_V$ &
$\mathbf P_1\Psi_1$ &
\makecell[c]{$\MLP(\mixmsg^{(\ell)})$\\${}+\mathbf H^{(\ell)}$} &
--
\\[3pt]

\rowcolor{googlegreen}
\multicolumn{8}{@{}>{\raggedright\arraybackslash}m{16.38cm}@{}}{
\textcolor{white}{\textbf{Level 2\ \ New fillers}\quad
\emph{equation fixed, with a filler introduced that the inventory lacks}}}
\\[2pt]

\rowcolor{googlegreen!25}
\makecell[l]{Curvature-Gated\\Propagation} &
$V$ &
$\{1\}$ &
\makecell[c]{$g(\kappa_{ij})^{\star}\,\hat{\mathbf A}$\\[1pt]\genann{(curvature-gated)}} &
$\mathbf H^{(\ell)}\mathbf W$ &
$\mathbf P_1\Psi_1$ &
\makecell[c]{$\sigma\!\bigl(\mathbf H^{(\ell)}\mathbf W_0$\\[-1pt]${}+\mixmsg^{(\ell)}\bigr)^{\dagger}$} &
--
\\[3pt]

\rowcolor{googlered}
\multicolumn{8}{@{}>{\raggedright\arraybackslash}m{16.38cm}@{}}{
\textcolor{white}{\textbf{Level 3\ \ Schema expansion}\quad
\emph{add a typed execution schedule to the graded equation}}}
\\[2pt]

\rowcolor{googlered!20}
\makecell[l]{Sequential\\Co-Lifted\\Node--Edge Layer} &
$V\sqcup E$ &
\makecell[c]{$\{V{\to}E,$\\[-1pt]$E{\to}V\}$} &
\makecell[c]{$\mathbf P_{V\to E}=\mathbf M$\\[2pt]$\mathbf P_{E\to V}=\mathbf M^\top$} &
\makecell[c]{$\Psi_{V\to E}(\mathbf H_V^{(\ell)})$\\[2pt]$\Psi_{E\to V}(\mathbf H_E^{(\ell+1)})$} &
\makecell[c]{$\boxplus=\mathrm{id}$\\[1pt]\genann{per stage}} &
\makecell[c]{$(\mathcal B_E,\phi_E)$\\[2pt]$(\mathcal B_V,\phi_V)$} &
\makecell[c]{$\mathcal S_\ell^{\star}=$\\[-1pt]$\bigl((V{\to}E),$\\[-1pt]$(E{\to}V)\bigr)$}
\\[5pt]

\bottomrule
\end{tabularx}
\end{table*}

The first four generated architectures hold the equation and its inventory fixed and recombine
fillers across families. Spectrally-Gated Propagation places a polynomial basis
in the propagation slot and makes the per-channel coefficient a node-wise function
of the state,
\begin{equation}
\mathbf{H}^{(\ell+1)}
=\sigma\!\Bigl(\mathbf{H}^{(\ell)}\mathbf{W}_0
+\sum_{k=1}^{K}\operatorname{diag}\!\bigl(\boldsymbol{\gamma}_k(\mathbf{H}^{(\ell)})\bigr)\,
T_k(\tilde{\mathbf{L}}_{\mathrm{ch}})\,\mathbf{H}^{(\ell)}\mathbf{W}_k\Bigr).
\end{equation}
The closest method is GPR-GNN \citep{chien2021adaptive}, which fixes one global coefficient per order. The
generated layer replaces that scalar with a node-adaptive coefficient, giving a
spatially modulated polynomial propagation operator. Because the node-wise diagonal gate
need not commute with $T_k(\tilde{\mathbf L}_{\mathrm{ch}})$, the resulting operator is not in
general a scalar spectral multiplier with a globally defined passband. This extends rule~(i) of the slot discipline
(Section~\ref{subsec:slot-discipline}). The absorbed coefficient is a per-node diagonal matrix
$\operatorname{diag}(\boldsymbol{\gamma}_k(\mathbf{H}^{(\ell)}))$ rather than a scalar, so the propagation operator
is a state-dependent, node-wise weighted polynomial basis rather than a pure scalar-times-polynomial. Relation-Equivariant
Network indexes the message by relation and constrains it to be equivariant to
rotations and translations. It differs from RGCN \citep{schlichtkrull2018modeling} in the message slot, where the
relation-specific linear map becomes an equivariant scalar-vector message, and from EGNN
\citep{satorras2021en} in the channel slot. Because the message
$\Psi_r(\mathbf{h}_i,\mathbf{h}_j,\lVert\mathbf{x}_i-\mathbf{x}_j\rVert)$ is geometric
rather than a linear value, the channel takes the pairwise branch of
Eq.~\eqref{eq:unified-notation}. It forms
$\mathbf C_{r,i}=\operatorname*{Agg}_{j\in\mathcal{N}_r(i)}\bigl([\mathbf{P}_r]_{ij}\,\Psi_r(\mathbf{h}_i,\mathbf{h}_j,\lVert\mathbf{x}_i-\mathbf{x}_j\rVert)\bigr)$
rather than a matrix product $\mathbf{P}_r\Psi_r$, and additive mixing gives
$\mixmsg_i=\sum_{r\in\mathcal R}\mathbf C_{r,i}$. The scalar message reads only
$\mathbf{h}_i,\mathbf{h}_j$ and the invariant distance $\lVert\mathbf{x}_i-\mathbf{x}_j\rVert$,
while coordinates use the EGNN equivariant update \citep{satorras2021en}. Under that update,
rotating or translating the inputs leaves scalar channels invariant and transforms vector
channels equivariantly. Subgraph-Attentive Layer lifts the update domain to rooted
subgraphs and propagates among them with state-dependent attention, then folds
back to nodes through the external equivariant pooling map $\mathbf{Q}_\ell$
associated with the unified system, the same map whose hypotheses Theorem~\ref{thm:equivariance}
verifies. If rooted-subgraph construction and $\mathbf Q_\ell$ commute with node relabeling,
the shared softmax attention is permutation equivariant over the subgraph domain and folds
back to nodes without breaking that symmetry.
It differs from
ESAN \citep{bevilacqua2022equivariant} in the propagation slot and from GAT \citep{velickovic2018gat} in the domain slot. Spectral-Biased
Transformer keeps global attention and replaces the structural bias term with a
learned polynomial of the Laplacian,
\begin{equation}
\mathbf{H}^{(\ell+1)}
=\operatorname{MLP}\!\Bigl(\operatorname{softmax}\!\Bigl(
\tfrac{(\mathbf{H}\mathbf{W}_Q)(\mathbf{H}\mathbf{W}_K)^{\top}}{\sqrt{d}}
+\sum_{k=0}^{K}\beta_k\,T_k(\tilde{\mathbf{L}}_{\mathrm{ch}})\Bigr)\mathbf{H}\mathbf{W}_V\Bigr)
+\mathbf{H}^{(\ell)},
\end{equation}
differing from Graphormer \citep{ying2021transformers} in the propagation slot, where the additive bias is
spectral rather than encoded from degree or shortest-path distance.

The fifth generated architecture keeps the slot structure fixed and introduces a propagation
operator the inventory did not contain. Curvature-Gated Propagation reweights the
one-hop operator by a learned function of discrete edge curvature, a quantity
introduced for graph rewiring \citep{Topping2022Curvature,Nguyen2023BORF} but used
here as a gate inside the propagation slot,
\begin{equation}
\vh_i^{(\ell+1)}
=\sigma\!\Bigl(\vh_i^{(\ell)}\mathbf{W}_0
+\textstyle\sum_{j\in\mathcal N(i)}g(\kappa_{ij})\,\hat{\mathbf{A}}_{ij}\,
\vh_j^{(\ell)}\mathbf{W}\Bigr),
\end{equation}
where $\kappa_{ij}$ is the discrete curvature of edge $(i,j)$. It differs from GCN
in the propagation slot, where the fixed normalized adjacency is replaced by a
curvature-gated operator.

\paragraph{Level~3 generated architecture.}
The Sequential Co-Lifted Node--Edge Layer distinguishes typed state from execution order.
Let $\mathbf H_V^{(\ell)}$ and $\mathbf H_E^{(\ell)}$ denote persistent node and edge states,
and let $\mathbf M\in\{0,1\}^{|E|\times |V|}$ be the unsigned edge--node incidence matrix.
The schedule $\mathcal S_\ell=((V\!\to E),(E\!\to V))$ gives
\begin{align}
\mathbf H_E^{(\ell+1)}
&=
\phi_E\!\left(
\mathcal B_E(\mathbf H_E^{(\ell)},\mathbf H_E^{(0)}),
\mathbf M\Psi_{V\to E}(\mathbf H_V^{(\ell)})
\right),
\label{eq:sequential-colifted-edge}\\
\mathbf H_V^{(\ell+1)}
&=
\phi_V\!\left(
\mathcal B_V(\mathbf H_V^{(\ell)},\mathbf H_V^{(0)}),
\mathbf M^\top\Psi_{E\to V}(\mathbf H_E^{(\ell+1)})
\right).
\label{eq:sequential-colifted-node}
\end{align}
The node update therefore reads the newly written edge state rather than
$\mathbf H_E^{(\ell)}$. Without this writeback, simultaneous node and edge updates reduce
to Eq.~\eqref{eq:graded-notation} with $\mathcal X_0=V$, $\mathcal X_1=E$, and typed
inter-rank channels. The layer instead adds the execution-schedule slot $\mathcal S_\ell$. Taking
one stage with all channels restores the synchronous graded equation, whose rank-zero case
is Eq.~\eqref{eq:unified-notation}. Because the two scheduled stages and their composition
are permutation equivariant whenever their incidence operators, messages, and updates obey
Theorem~\ref{thm:equivariance}, the layer satisfies the Level~3 symmetry condition.

The new slot makes within-layer execution order an explicit, removable component of the schema.
Its closest established construction is the sequential edge--node update of a Graph Network
block~\citep{battaglia2018relational}, whose dependency the added schedule records in the
unified vocabulary.

Each generated Level~1--2 architecture is a single structural move from a catalogued method. Level~1
changes a combination of existing fillings, Level~2 changes one filling, and Level~3 adds
the schedule component.

Architecture generation only proposes a filling. It does not certify that the filling behaves well
under repeated application. Whichever slot a generated architecture edits, its viability as a deep
architecture still rests on the propagation slot $\mathbf{P}_k$. Generation does not
constrain that operator's spectral or topological behavior by construction. Section~\ref{sec:propagation}
turns to that slot directly, treating it as an object of theory rather than a further
source of fillers.

\section{GNN Propagation: Limits, Expressivity, and Remedies}
\label{sec:propagation}

The architectures in Appendix~\ref{app:family-comparison-tables} and generated architectures from Section~\ref{sec:generation} differ chiefly in the propagation slot $\mathbf{P}_k^{(\ell)}$ of Eq.~\eqref{eq:unified-notation}. An operator can propagate signal yet also cause oversmoothing, oversquashing, failure under heterophily, or limited expressivity. We study what its repeated application can and cannot do. Under stated assumptions on the message, update, and input maps, oversmoothing, oversquashing, and heterophily depend largely on the operator's \emph{spectrum} and \emph{topology}. Expressivity instead depends on the structural information accessible to the components. Localizing these limits and remedies within the unified equation is this section's contribution.

Sections~\ref{subsec:prop-oversmoothing}--\ref{subsec:prop-tradeoff} analyze forward and backward smoothing, topological bottlenecks, pass-band mismatch under heterophily, and the spectral-gap trade-off, including recent critiques. Section~\ref{subsec:prop-expressivity} turns to what propagation can distinguish through the WL hierarchy, substructure counting, and associated generalization costs. Section~\ref{subsec:prop-transformer} treats global attention as another propagation filling, and Section~\ref{subsec:prop-synthesis} connects the diagnostics to the generated architectures of Section~\ref{sec:generation}. Tables~\ref{tab:prop-diagnostics}, \ref{tab:prop-expressivity}, and \ref{tab:prop-remedies} summarize governing quantities, expressivity relations, and component-specific remedies, respectively.

Unless stated otherwise, $\ell$ indexes GNN layers and $t$ indexes repeated propagation
or diffusion steps. $n=|\mathcal V|$ denotes the node count and $|\mathcal E|$ denotes
the edge count. Complexity terms below are written in $n$ and $|\mathcal E|$. We write
$\hat{\mathbf A}=\tilde{\mathbf D}^{-1/2}(\mathbf A+\mathbf I)\tilde{\mathbf D}^{-1/2}$, as defined
in Section~\ref{sec:preliminaries}, for the self-loop normalized adjacency used throughout this
section, with its eigenvalues $\mu_i$ ordered $1=\mu_1>\mu_2\ge\cdots\ge\mu_n>-1$ on a connected
graph. We write $\widehat{\mathbf S}$ for the
normalized sensitivity or support operator used in oversquashing bounds. We reserve
$\mathbf{P}_k^{(\ell)}$ for the framework's propagation slot.

\subsection{Oversmoothing: collapse under repeated propagation}
\label{subsec:prop-oversmoothing}

\begin{figure*}[t]
    \centering
    \includegraphics[
        width=\textwidth,
        keepaspectratio
    ]{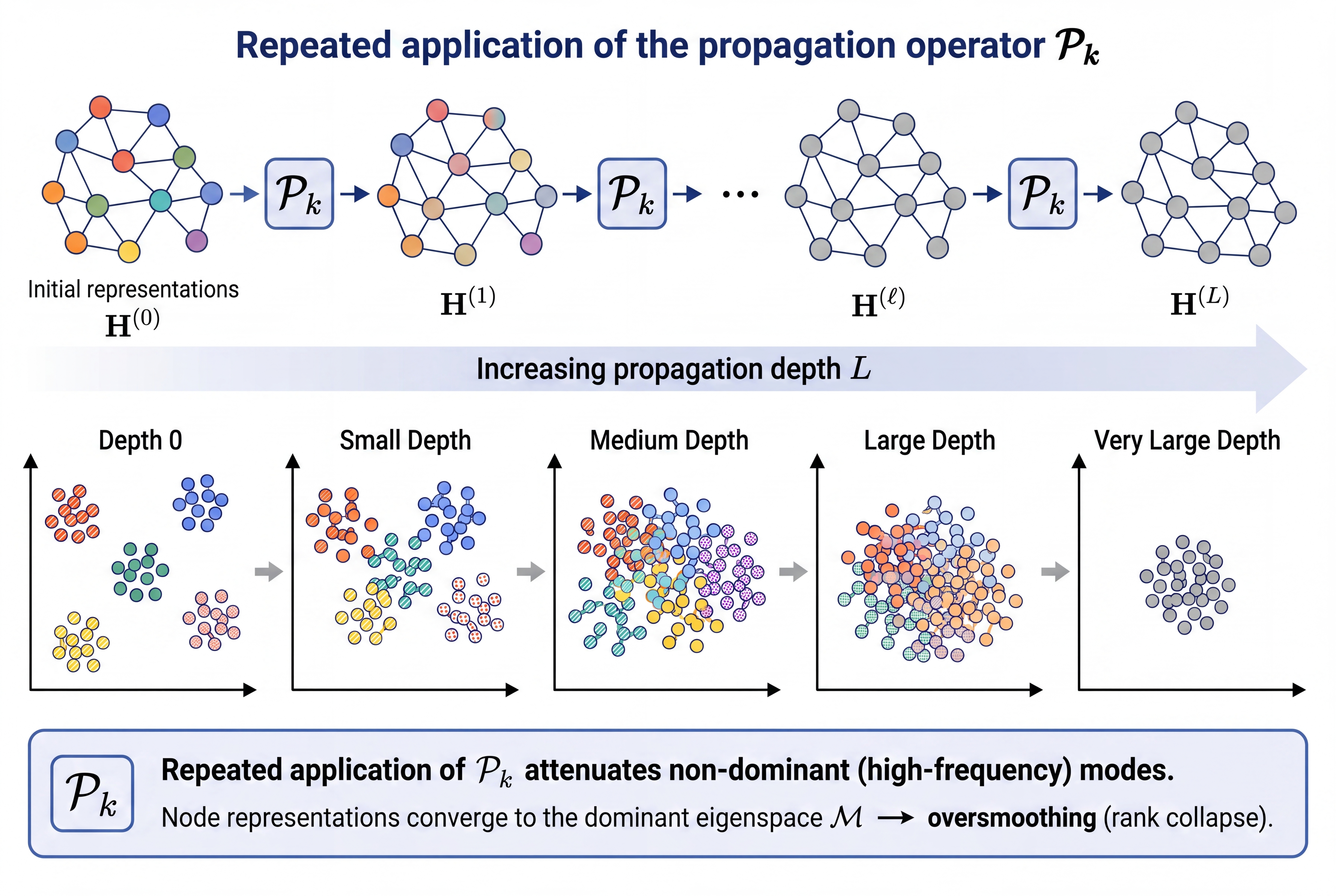}

    \caption{\textbf{Repeated propagation leads to oversmoothing.}
    Repeated application of the propagation operator $\mathbf{P}_k$ progressively
    suppresses non-dominant (high-frequency) modes. As propagation depth
    $L$ increases, initially distinct node representations become increasingly
    mixed and eventually converge toward the dominant invariant subspace,
    producing oversmoothing (rank collapse).}
    \label{fig:prop-oversmoothing}
\end{figure*}

Oversmoothing describes the spectral collapse caused by repeated low-pass propagation, as Figure~\ref{fig:prop-oversmoothing} illustrates.

Under the linear specialization~\eqref{eq:unified-notation-linear}, repeated low-pass propagation contracts states toward the operator's dominant invariant subspace. This subspace is mapped into itself and corresponds here to the largest-eigenvalue eigenspace. For self-loop normalized adjacency, it is the one-dimensional degree-dominated direction on a connected graph and one component-indicator direction per connected component otherwise. \citet{li2018deeper} identified the collapse empirically, and \citet{oono2020expressive} established contraction under specified activation and weight assumptions.

Let $\mathcal M$ be this subspace, $s_\ell=\Vert\mathbf{W}^{(\ell)}\Vert_2$, $\bar s=\max_\ell s_\ell$, and $\mu_\star=\max_{i\ge2}|\mu_i|<1$, the largest eigenvalue magnitude of $\hat{\mathbf A}$ on $\mathcal M^\perp$ and the specialization of \citet{oono2020expressive}'s general-operator radius. Then
\begin{equation}
d_{\mathcal M}\!\left(\mathbf{H}^{(\ell)}\right)\;\le\;(\bar s\,\mu_\star)^{\ell}\,
d_{\mathcal M}\!\left(\mathbf{H}^{(0)}\right),
\label{eq:oono}
\end{equation}
so when $\bar s\mu_\star<1$ node features collapse to $\mathcal M$ exponentially,
independently of the labels.

The same $\mu_\star$ governs the one-step contraction below, proved in \hyperref[proof:lem-contraction]{Appendix~\ref*{sec:analysis}}.

\begin{lemma}[One-step contraction]
\label{lem:contraction}
On a connected graph, $E(\hat{\mathbf{A}} \mathbf{H})\le \mu_\star^{2}\,E(\mathbf{H})$. This is the elementary spectral step underlying the oversmoothing analyses of \citet{nt2019revisiting,oono2020expressive,cai2020note}. We restate it here in the framework's notation.
\end{lemma}

The Dirichlet-energy form, following \citet{oono2020expressive,cai2020note} and restated for the linear specialization, makes its activation assumptions explicit. The proof appears in \hyperref[proof:thm-oversmoothing]{Appendix~\ref*{sec:analysis}}.

\begin{theorem}[One-layer energy bound]
\label{thm:oversmoothing}
Let $\mathbf{H}'=\sigma(\hat{\mathbf{A}} \mathbf{H} \mathbf{W})$ with $\sigma$ entrywise, $1$-Lipschitz, positively homogeneous, and $\sigma(0)=0$. Then
\[
E(\mathbf{H}')\;\le\;\big(s\,\mu_\star\big)^2\,E(\mathbf{H}),\qquad s=\Vert \mathbf{W}\Vert_2 .
\]
Iterating over an $L$-layer stack with $s_\ell=\Vert\mathbf{W}^{(\ell)}\Vert_2$ and $\bar s=\max_\ell s_\ell$ gives the geometric bound $E(\mathbf{H}^{(L)})\le\big(\bar s\,\mu_\star\big)^{2L}E(\mathbf{H}^{(0)})$, geometric Dirichlet-energy decay when $\bar s\mu_\star<1$. Energy controls the distance to the degree-dominated direction $\mathbf{v}_1$ (the $\mu=1$ eigenvector). Write $\mathbf{c}_i^\top=\mathbf{v}_i^\top\mathbf{H}$ for $\mathbf{H}$'s coefficient in eigendirection $\mathbf{v}_i$. Since $E(\mathbf{H})=\sum_{i\ge 2}(1-\mu_i)\Vert\mathbf{c}_i\Vert^2\ge(1-\mu_2)\sum_{i\ge 2}\Vert\mathbf{c}_i\Vert^2$,
\[
\big\Vert(\mathbf{I}-\mathbf{v}_1\mathbf{v}_1^\top)\mathbf{H}\big\Vert_F^2\;\le\;\frac{E(\mathbf{H})}{1-\mu_2}.
\]
Hence if $\bar s\,\mu_\star<1$ the energy decays geometrically with depth, and by this inequality the states collapse toward the degree-dominated subspace $\operatorname{span}\{\mathbf{v}_1\}$.
\end{theorem}

Positive homogeneity makes $E(\sigma(\mathbf{M}))\le E(\mathbf{M})$ exact for the (Leaky)ReLU family. Because GELU, tanh, and sigmoid lack this property, the per-node $1/\sqrt{\tilde D_{ii}}$ scaling cannot pass through $\sigma$, leaving only a weaker Lipschitz bound.

Because $\mu_\star$ is the spectral radius on $\mathcal M^\perp$, Eq.~\eqref{eq:oono} includes negative eigenvalues. Replacing it by a quantity based only on the second-smallest normalized-Laplacian eigenvalue requires additional assumptions. Decay therefore depends jointly on the nontrivial spectrum, feature-map norms, and activation.

Within the broader theory of \citet{cai2020note,rusch2023survey}, let $\mathrm{Osm}(\mathbf H)\ge0$ vanish exactly when all rows coincide (avoiding the source's $\mu$, already used for eigenvalues). Oversmoothing means $\mathrm{Osm}(\mathbf H^{(\ell)})\le C_1e^{-C_2\ell}$ for $C_1,C_2>0$. Dirichlet energy satisfies these axioms, whereas the mean-average distance of \citet{chen2020measuring} does not. Preventing this decay is necessary but insufficient for a useful deep network.

Three refinements qualify this baseline without changing its mechanism. First, state dependence alone does not prevent contraction. Under connectivity, non-bipartiteness, positivity, and boundedness, \citet{Wu2023DemystifyingOversmoothing} show that an input-dependent row-stochastic $\mathbf{P}^{(t)}$ still satisfies $\mathrm{Osm}(\mathbf H^{(t)})\le C_1\kappa^t$ for some $\kappa<1$, where $\mathrm{Osm}(\mathbf H)=\lVert(\mathbf I-\tfrac1n\mathbf 1\mathbf 1^\top)\mathbf H\rVert_F$.

Here the joint spectral radius $\kappa$ is the long-run contraction factor of the changing operator sequence, unrelated to the eigenvalue count $q$ in Lemma~\ref{lem:realizability}. It is lower-bounded by the second eigenvalue of the fixed, self-loop-free normalized adjacency $\bar{\mathbf A}$. Thus GAT-style attention can oversmooth through joint-product contraction, conditional on the four assumptions rather than universally.

Second, \citet{digiovanni2023understanding} interpret the linear update as gradient flow of a channel-weighted energy. Positive \emph{channel-mixing} eigenvalues produce attraction and low-frequency smoothing (Section~\ref{subsec:spectral-spatial-prelim}), whereas negative ones produce repulsion and high-frequency \emph{oversharpening}. This spectrum is distinct from the propagation spectrum, making low-pass bias a design choice rather than a necessity of message passing.

Third, \citet{roth2024rank} locate the mechanism in rank collapse, whereby repeated use of one operator drives representations into its dominant eigenspace regardless of per-layer weights, which rescale outputs without changing the surviving spectral component. This yields both row-wise oversmoothing and column-wise overcorrelation. A sum-of-Kronecker-products update instead lets distinct weights act on distinct operators, allowing feature directions to retain different dominant components. Across these refinements, the operative quantities remain the joint-product radius, channel-mixing eigenvalue signs, and operator eigenspaces.

Do architectural fixes slow collapse, and is energy a reliable diagnostic? For stacks without residual or skip connections, \citet{chen2025residual} show that the decay rate of the squared energy-type similarity $\mathrm{Osm}$ converges to the spectral radius of $\mathbf P$ on its non-dominant eigenspace, regardless of layer weights. Residual updates provably raise this rate for broad weight distributions, including Ginibre ensembles (i.i.d.\ Gaussian matrices), but the benefit is distribution-dependent rather than universal.

An initial-residual teleport does more than slow decay because it keeps energy above a positive floor. Personalized-PageRank convergence is standard \citep{klicpera2019predict}. The floor below is a mild extension locating the escape in $\mathcal B_\ell$, proved in \hyperref[proof:thm-floor]{Appendix~\ref*{sec:analysis}}.

\begin{theorem}[Initial-residual energy floor]
\label{thm:floor}
Let $\mathbf{G}^{(0)}$ be a fixed anchor and initialize the propagation state at that anchor. (We write $\mathbf{G}$ rather than the framework's final-representation $\mathbf{Z}$ to keep this internal PPR-propagation state distinct from that reserved symbol.) The initial-residual iteration $\mathbf{G}^{(\ell+1)}=(1-\eta)\hat{\mathbf{A}} \mathbf{G}^{(\ell)}+\eta \mathbf{G}^{(0)}$ uses $\eta\in(0,1)$, the personalized-PageRank teleport probability of \citet{klicpera2019predict} (for APPNP, $\mathbf{G}^{(0)}=f_\theta(\mathbf{X})$). It converges to the fixed point
\[
\mathbf{G}^\star=\eta\big(\mathbf{I}-(1-\eta)\hat{\mathbf{A}}\big)^{-1}\mathbf{G}^{(0)}
=\eta\sum_{k\ge 0}(1-\eta)^k\hat{\mathbf{A}}^k \mathbf{G}^{(0)}.
\]
If $\mathbf{G}^{(0)}$ carries any nonzero non-DC content, that is $E(\mathbf{G}^{(0)})>0$ (equivalently, a nonzero component outside the $\mu=1$ degree-dominated eigenspace, not necessarily high frequency), its fixed-point energy is bounded strictly from below by a fixed fraction of the anchor energy:
\[
E(\mathbf{G}^\star)\;>\;\Big(\tfrac{\eta}{2-\eta}\Big)^2 E(\mathbf{G}^{(0)}).
\]
\end{theorem}

Energy may itself be the wrong diagnostic. \citet{zhang2026measuring} show numerical rank collapsing to one across broad GNN families without weight-norm assumptions, whereas Dirichlet-energy guarantees require restrictive norms and empirically miss degradation around ten layers. Rank is therefore the more faithful signature in their analysis.

Depth also has a non-monotone effect. \citet{keriven2022nottoo} show that a finite, nonzero number of aggregation steps can improve learning by shrinking within-community and non-principal variation before eventual collapse, yielding an optimum rather than monotone harm.

\paragraph{A backward view.}
Oversmoothing also acts backward. \citet{keriven2025backward} show that backpropagated error undergoes \emph{linear} smoothing even with a nonlinear forward pass, creating spurious near-stationary, flat high-loss regions once the last layer is fit. An equally deep graph-agnostic MLP does not exhibit this failure.

\citet{park2024taming} distinguish gradient oversmoothing from the gradient \emph{expansion} induced by residual connections that repair forward contraction. Frobenius-norm weight normalization constrains each layer's Lipschitz constant, enabling stable training at hundreds of layers. Together with \citet{chen2025residual}, this shows that residuals relocate the difficulty to a backward/Lipschitz regime rather than eliminate it.

\subsection{Oversquashing: sensitivity, curvature, and effective resistance}
\label{subsec:prop-oversquashing}

\begin{figure}[!t]
    \centering
    \includegraphics[
        width=\linewidth,
        keepaspectratio
    ]{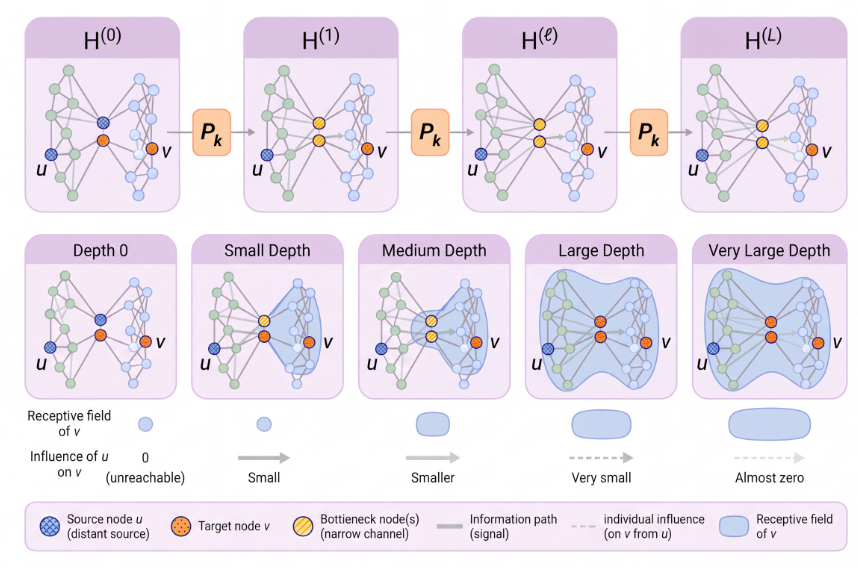}
    \caption{\textbf{Local propagation through a bottleneck leads to oversquashing.}
    Repeated application of $\mathbf{P}_k$ expands the receptive field of target
    node $v$ with depth $L$. Once distant nodes become reachable, increasingly
    many signals must traverse the same narrow topological bottleneck, compressing
    their individual influence on $v$.}
    \label{fig:prop-oversquashing}
\end{figure}

Oversquashing is insufficient rather than excessive mixing because task-relevant distant signals traverse narrow topological channels as the receptive field expands~\citep{Alon2021Bottleneck}. Figure~\ref{fig:prop-oversquashing} illustrates this compression. (Node labels $u,v$ below are unrelated to eigenvectors $\mathbf v_i$.) The proposition formalizes the $L$-hop receptive-field bound and its locality assumptions. The proof appears in \hyperref[proof:prop-receptive]{Appendix~\ref*{sec:analysis}}.

\begin{proposition}[Receptive field and underreaching]
\label{prop:receptive}
The underreaching bound~\citep{Alon2021Bottleneck} is recast here on the slots, so the hypothesis falls on the support of $\mathbf{P}_k$ and the locality of its entries. Suppose every channel is \emph{locally computed}, a condition stated directly on the current-layer states with no reference to the receptive field. Concretely, the support lies in the closed one-hop neighborhood, $\operatorname{supp}(\mathbf{P}_k^{(\ell)})\subseteq\{(i,j): d(i,j)\le 1\}$, and each entry $[\mathbf{P}_k^{(\ell)}]_{ij}$ and each message on edge $(i,j)$ is a function only of the two endpoint states $\vh_i^{(\ell)},\vh_j^{(\ell)}$ and the local edge feature $\ve_{ij}$. Here $\ve_{ij}$ must itself be \emph{local}, i.e.\ not a function of input features at nodes outside $\{i,j\}$. Suppose further that $\mathcal{B}_\ell$ and $\phi_\ell$ act per node, so that row $i$ of the update depends only on row $i$ of its arguments (the initial-state teleport $\eta\mathbf{H}^{(0)}$ contributes only $\vh_i^{(0)}$ to row $i$). Then $\vh_i^{(L)}$ is independent of $\vh_j^{(0)}$ whenever $d(i,j)>L$, and the depth-$L$ receptive field of node $i$ is contained in the $L$-hop ball $\bar{B}_L(i):=\{j: d(i,j)\le L\}$ (the bar keeps this closed hop-ball visually distinct from the ego/residual map $\mathcal{B}_\ell$ used just above).
\end{proposition}

Both locality conditions are necessary because an attention score on $(i,j)$ may read only its endpoints, whereas an edge attribute computed from a global statistic such as $\sum_z\vh_z^{(0)}$ defeats the bound despite being attached to $(i,j)$.

\citet{Topping2022Curvature} split the sensitivity of $v$ to a source $u$ at distance $r$ into a model-dependent Lipschitz amplification and a topological transition weight:
\begin{equation}
\left\lVert \partial \vh_v^{(r+1)}/\partial \vh_u^{(0)}\right\rVert
\;\le\;(L_\Psi L_\phi)^{\,r+1}\,(\hat{\mathbf A}^{\,r+1})_{vu},
\label{eq:oversquash}
\end{equation}
where $L_\Psi,L_\phi$ bound the message and update maps, and $(\hat{\mathbf A}^{r+1})_{vu}$ is the $(r{+}1)$-step transition weight. Strongly negative \emph{Balanced-Forman curvature}, a discrete Ricci-curvature measure, is associated with bottlenecks having small transition weight, few common neighbors, or few short connecting cycles.

At depth $t=r+1$, \citet{DiGiovanni2023OverSquashing} refine the prefactor into activation Lipschitz constant $c_\sigma$, maximum weight magnitude $\omega$, and width $p$. They replace $\hat{\mathbf A}$ by a family of generalized shifts $\widehat{\mathbf S}$ built from the self-loop-free normalized adjacency $\bar{\mathbf A}$ with two tunable weights. The operator $\hat{\mathbf A}$ is one member of this family. In the resulting bound $(c_\sigma\omega p)^t(\widehat{\mathbf S}^t)_{vu}$, the prefactor scales sensitivity, while the topological factor determines whether $u$ and $v$ can interact.

Because depth enters only as an exponent, it cannot repair a bottleneck, and vanishing gradients dominate beyond the graph's topological scale.

\citet{black2023resistance} relate $(\widehat{\mathbf S}^t)_{vu}$ to \emph{effective resistance} $R_{uv}=\tau(u,v)/(2|\mathcal E|)$, where $\tau(u,v)$ is round-trip commute time. The total $R_{\mathrm{tot}}=\sum_{u<v}R_{uv}$ defines a graph-level oversquashing budget targeted by resistance-minimizing rewiring. \citet{digiovanni2024oversquashing} convert the sensitivity bound into a capacity requirement under which guaranteeing fixed interaction between $u$ and $w$ forces depth and weight norms to grow with $\tau(u,w)$. Thus bounded-capacity MPNNs face a quantitative impossibility, not merely a patchable design flaw.

When $t<d(u,v)$, $(\widehat{\mathbf S}^t)_{vu}=0$ and the nodes cannot interact (Proposition~\ref{prop:receptive}). This shallow-depth \emph{underreaching} precedes oversquashing. A virtual node shortens paths but remains topology-dependent because the shared hub assigns uniform importance, requiring a topology-aware heterogeneous variant for node-dependent sensitivity at $O(n)$ cost \citep{southern2025virtualnodes}.

\subsection{Heterophily: a spectral mismatch}
\label{subsec:prop-heterophily}

Unlike the preceding depth pathologies, heterophily can be read---usefully but not exhaustively---as a mismatch between $\mathbf P_k$'s response and the frequency band carrying class signal.

\citet{nt2019revisiting} show that $\hat{\mathbf A}=\mathbf I-\mathbf L_{\mathrm{ref}}$ scales Laplacian frequency $i$ by $(1-\lambda_i)=\mu_i$. Low frequencies pass nearly unchanged, whereas higher frequencies may be attenuated or sign-reversed. Negative adjacency eigenvalues can make their magnitude non-monotone across steps.

On the connected self-loop graph used here, repeated $\hat{\mathbf A}$ propagation nevertheless suppresses every non-dominant component because $|\mu_i|<1$ for $i\ge2$. This favors low-frequency discriminative signal, but heterophily alone does not locate that signal. Fixed low-pass propagation may instead remove it.

The response can instead be shaped. The lemma realizes any discriminative band with a polynomial channel, with the proof given in \hyperref[proof:lem-realizability]{Appendix~\ref*{sec:analysis}}. The corollary gives a heterophilic high-pass configuration by zeroing the smoothing frequency $\nu=1$, with the proof given in \hyperref[proof:cor-heterophily]{Appendix~\ref*{sec:analysis}}.

\begin{lemma}[Filter realizability]
\label{lem:realizability}
Let $\hat{\mathbf{A}}$ have $q$ distinct eigenvalues $\nu_1,\dots,\nu_q$ with spectral projectors $\boldsymbol{\Gamma}_1,\dots,\boldsymbol{\Gamma}_q$. These are the same eigenvalues denoted $\mu_i$ elsewhere in this section, here relabeled by their $q$ distinct values rather than by the $n$ possibly-repeated indices $i=1,\dots,n$. The projectors $\boldsymbol{\Gamma}_j$ are the eigenprojectors of $\hat{\mathbf{A}}$, not of $\mathbf{L}_{\mathrm{sym}}$. For any target responses $r_1,\dots,r_q\in\mathbb{R}$ there exist coefficients $\gamma_0,\dots,\gamma_{q-1}$ with
\[
\sum_{k=0}^{q-1}\gamma_k\hat{\mathbf{A}}^k=\sum_{j=1}^{q} r_j\,\boldsymbol{\Gamma}_j .
\]
A polynomial channel of degree at most $q-1$ realizes any frequency response, low-pass, band-pass, or high-pass. This is the standard interpolation argument for polynomial spectral filters~\citep{defferrard2016convolutional,wang2022how,balcilar2021analyzing}, stated here for $\hat{\mathbf{A}}$.
\end{lemma}

\begin{corollary}[Heterophilic configuration]
\label{cor:heterophily}
A high-pass response that places weight on the high-frequency band and zero on the smoothing direction $\nu=1$ is realizable inside $\mathbf{P}_k$ at degree at most $q-1$. Suppose class-discriminative signal is carried by high-frequency eigenvectors. Heterophily can produce this, though heterophily does not by itself imply that all such signal lies in any designated high-frequency band. Such a filling \emph{retains} those components because their response is nonzero, though not necessarily at unchanged amplitude. Repeated propagation by $\hat{\mathbf{A}}$, by contrast, annihilates them, since its powers send every non-DC response to zero. A generic low-pass filter need not. More generally, the realizability of Lemma~\ref{lem:realizability} lets $\mathbf{P}_k$ match whatever band the discriminative signal occupies, rather than presuming it is high-frequency.
\end{corollary}

\begin{figure*}[t]
\centering
\includegraphics[width=\textwidth]{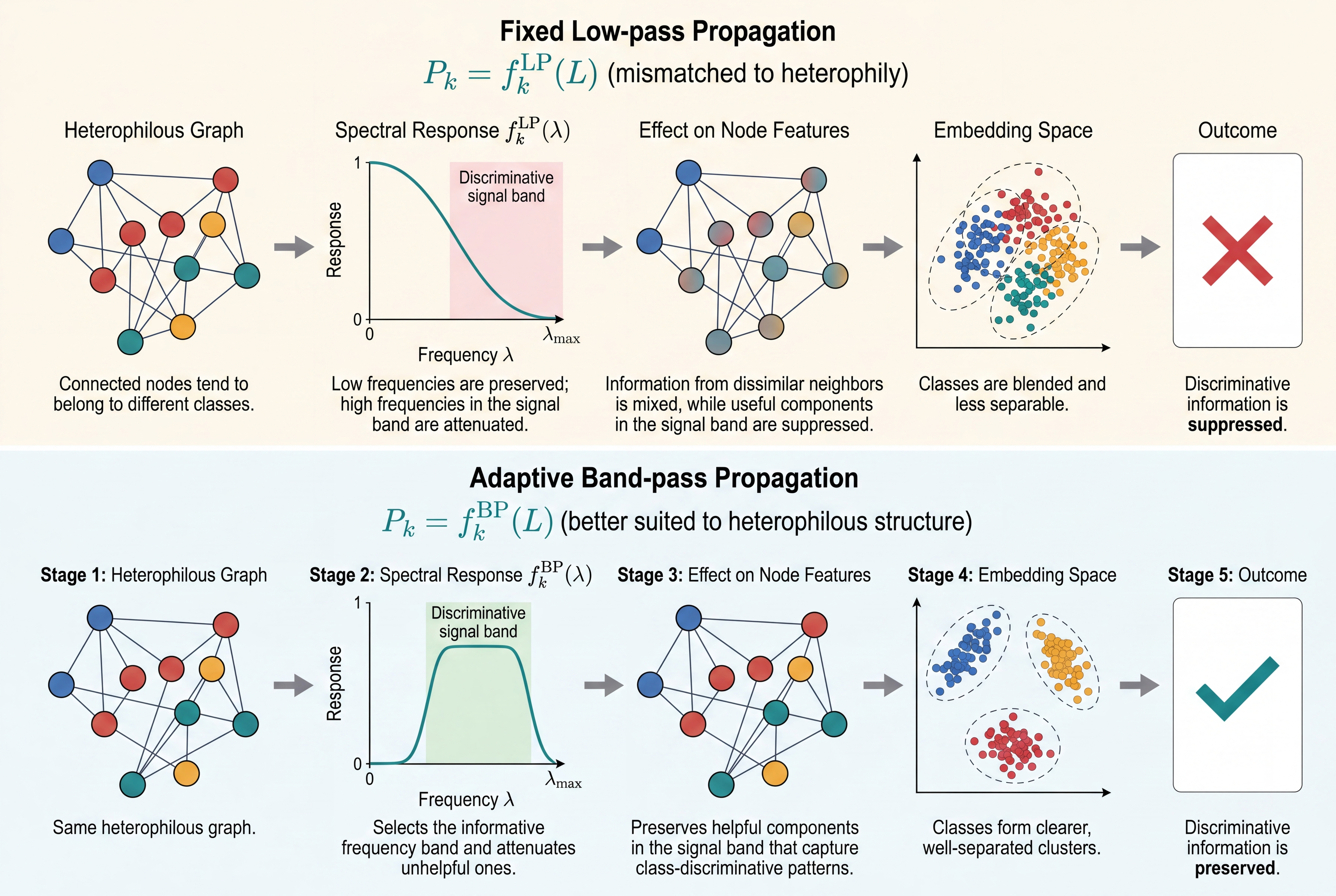}
\caption{\textbf{Matching the operator's pass-band to the signal.}
In the illustrated heterophilic setting, the class-discriminative signal
occupies frequencies attenuated by standard low-pass propagation.
Consequently, repeated propagation suppresses these informative components,
causing the class representations to become less distinguishable, as
characterized in Corollary~\ref{cor:heterophily}.
An adaptive band-selective $\mathbf{P}_k$ instead shapes its spectral response
to retain the informative frequency band and thereby preserve discriminative
information.}
\label{fig:prop-heterophily}
\end{figure*}

Figure~\ref{fig:prop-heterophily} contrasts fixed low-pass propagation with adaptive operator banks. FAGCN interpolates low- and high-pass filters per edge using $\alpha_{ij}=\tanh(\cdot)\in[-1,1]$, with high-pass propagation increasing distances between connected nodes \citep{bo2021beyond}. GPR-GNN learns signed coefficients in $\sum_k\gamma_k\hat{\mathbf A}^k$ \citep{chien2021adaptive}. BernNet and ChebNetII use nonnegative Bernstein coefficients and Gauss-node Chebyshev interpolation \citep{he2021bernnet,he2022chebnetii}. MidGCN targets a mid-frequency pass-band, and RFA-GNN makes the response relation-dependent \citep{Huang2023MidGCN,Wu2023RFAGNN}.

Sheaf methods replace $\mathbf P_k$ by a learned cellular-sheaf Laplacian $\mathbf L_{\mathcal F}$, assigning vector-space \emph{stalks} to nodes and edges with linear maps between them. The ordinary Laplacian is the trivial-sheaf case. Suitable non-trivial kernels preserve nonconstant signals, so the $t\to\infty$ limit need not be constant. In the analyzed constructions, the number of linearly separable limiting classes grows with stalk dimension, addressing heterophily and oversmoothing jointly \citep{Bodnar2022DiagNSD}. \citet{wang2025signed} reinterpret such remedies as structured negative edges with non-collapsing equilibria. These designs reshape the operator's pass-band or kernel to match the signal.

These remedies require diagnosing the mismatch, yet plain edge homophily (Section~\ref{subsec:homophily-prelim}) poorly predicts GNN difficulty \citep{Zheng2024DisentanglingHomophily}. \citet{zhu2020beyond} identify ego/neighbor separation, higher-order channels, and intermediate-layer combination. \citet{luan2022revisiting} propose aggregation homophily from post-aggregation similarity and an adaptive low/high/identity mix. GCN can succeed when same-class nodes share neighborhood distributions \citep{Ma2022HomophilyNecessary}. Structured heterophily is comparatively benign, while high intra-class neighborhood variance causes harm rather than low homophily itself~\citep{Lee2024FeatureDistribution}.

To predict when the mismatch costs accuracy relative to a graph-agnostic baseline, \citet{luan2023whenhelp} use Contextual Stochastic Block Model (CSBM) node-distinguishability/Bayes error, yielding a hardest mid-homophily regime and a threshold at which a GNN beats an MLP.

Table~\ref{tab:prop-remedies} maps these remedies, including oversquashing interventions revisited in Section~\ref{subsec:prop-tradeoff}, to the components they modify.

\colorlet{premBar}{googlered}
\colorlet{premRow}{googlered!9}
\begin{table}[t]
\centering
\footnotesize
\setlength{\tabcolsep}{4pt}
\renewcommand{\arraystretch}{1.3}
\caption{\textbf{Remedies are interventions on a named slot.} Each remedy class, the
pathology it targets, the unified-equation slot~\eqref{eq:unified-notation} it modifies, and the
operator/topology quantity it moves (with the trade-off it incurs). Architectures are
instances.}
\label{tab:prop-remedies}
\resizebox{\linewidth}{!}{
\begin{tabular}{@{}L{3.5cm} L{2.6cm} L{2.3cm} L{4.9cm} L{2.7cm}@{}}
\toprule
\rowcolor{premBar}
{\color{white}\textbf{Remedy class (instance)}} & {\color{white}\textbf{Targets}} & {\color{white}\textbf{Slot}} &
{\color{white}\textbf{Quantity moved (trade-off)}} & {\color{white}\textbf{Key refs}} \\
\midrule
Curvature rewiring (SDRF, BORF) & oversquashing & $\mathbf{P}_k$ support/weights & raises
$(\widehat{\mathbf S}^t)_{vu}$ on negative-curvature edges; pruning can slow smoothing &
\citealp{Topping2022Curvature,Nguyen2023BORF,Liu2023CurvDrop} \\
\rowcolor{premRow}
Spectral-gap rewiring (FoSR, GTR) & oversquashing & $\mathbf{P}_k$ support; relation channels &
raises $\lambda_2$ / lowers $R_{\mathrm{tot}}$; relational split avoids induced smoothing &
\citealp{Karhadkar2023FoSR,black2023resistance} \\
Spectral pruning (Braess) & squashing \emph{and} smoothing & $\mathbf{P}_k$ support/weights & selected edge
deletions can raise $\lambda_2$ and co-mitigate both under the cited criterion & \citealp{jamadandi2024spectralpruning,Liang2025SpectrumSparsification} \\
\rowcolor{premRow}
Learned rewiring (PR-MPNN) & oversquashing, underreaching & $\mathbf{P}_k$ sampled support &
differentiable $k$-subset edges; conditions for ${>}$ random rewiring & \citealp{qian2024prmpnn} \\
Virtual / global channel (MPNN${+}$VN) & oversquashing, long-range & $\mathcal{X}$ and
$\mathbf{P}_k$ for VN; $\mathbf{P}_k$ for dense attention & VN adds a two-edge path
(two synchronous propagation steps); classical VN forces uniform importance &
\citealp{southern2025virtualnodes,Cai2023MPNNGTConnection} \\
\rowcolor{premRow}
Signed / high-pass bank (FAGCN, GPR-GNN, BernNet, ChebNetII) & heterophily, oversmoothing &
$\mathbf{P}_k$ bank ($\gamma_k$ folded in) & shifts the response high or uses signed
$\gamma_k$; can suppress the dominant smoothing response & \citealp{bo2021beyond,chien2021adaptive,he2021bernnet,he2022chebnetii} \\
Sheaf / signed propagation (NSD, SBP) & heterophily \emph{and} oversmoothing &
$\mathbf{P}_k$ built from $\mathbf{L}_{\mathcal F}$ & suitable learned kernels preserve
nonconstant signals; structured negative edges & \citealp{Bodnar2022DiagNSD,wang2025signed} \\
\rowcolor{premRow}
Residual / skip & oversmoothing (forward) & $\mathcal{B}_\ell$ (combined by $\phi_\ell$) & raises decay exponent
above $\max_{\nu\neq 1}|\nu|$; induces gradient expansion &
\citealp{chen2025residual,park2024taming} \\
Ego/neighbor split ${+}$ multi-hop (H2GCN, ACM-GCN) & heterophily & $\phi_\ell$
(ego/neighbor split), $\mathbf{P}_k$ bank & separates identity from aggregate; adds high-pass/identity
channels & \citealp{zhu2020beyond,luan2022revisiting} \\
\rowcolor{premRow}
Learnable depth/filtering (AMP) & smoothing, squashing, underreaching & per-layer
$\mathbf{P}_k$ weights (latent) & infers effective depth/filter by ELBO; no global rewiring &
\citealp{Errica2025AMP} \\
Storage capacity (gLSTM) & oversquashing (short-range) & $\Psi_k$ / $\phi_\ell$ &
enlarges per-node bandwidth at fixed topology & \citealp{blayney2026glstm} \\
\rowcolor{premRow}
Kronecker-product update & rank collapse & $\mathbf{P}_k$ bank / $\Psi_k$ & prevents collapse to
dominant eigenspace; preserves feature rank & \citealp{roth2024rank} \\
Normalization (PairNorm) & oversmoothing (forward) & $\phi_\ell$ / state & holds total pairwise
feature distance roughly constant across layers; can distort useful smoothing & \citealp{zhao2020pairnorm} \\
\rowcolor{premRow}
Stochastic edge dropping (DropEdge) & oversmoothing (forward) & $\mathbf{P}_k$ support & random
per-epoch support sparsification slows the smoothing convergence & \citealp{rong2020dropedge} \\
\bottomrule
\end{tabular}}
\end{table}

\subsection{The oversmoothing-oversquashing trade-off}
\label{subsec:prop-tradeoff}

\begin{figure*}[t]
\centering
\includegraphics[width=\textwidth]{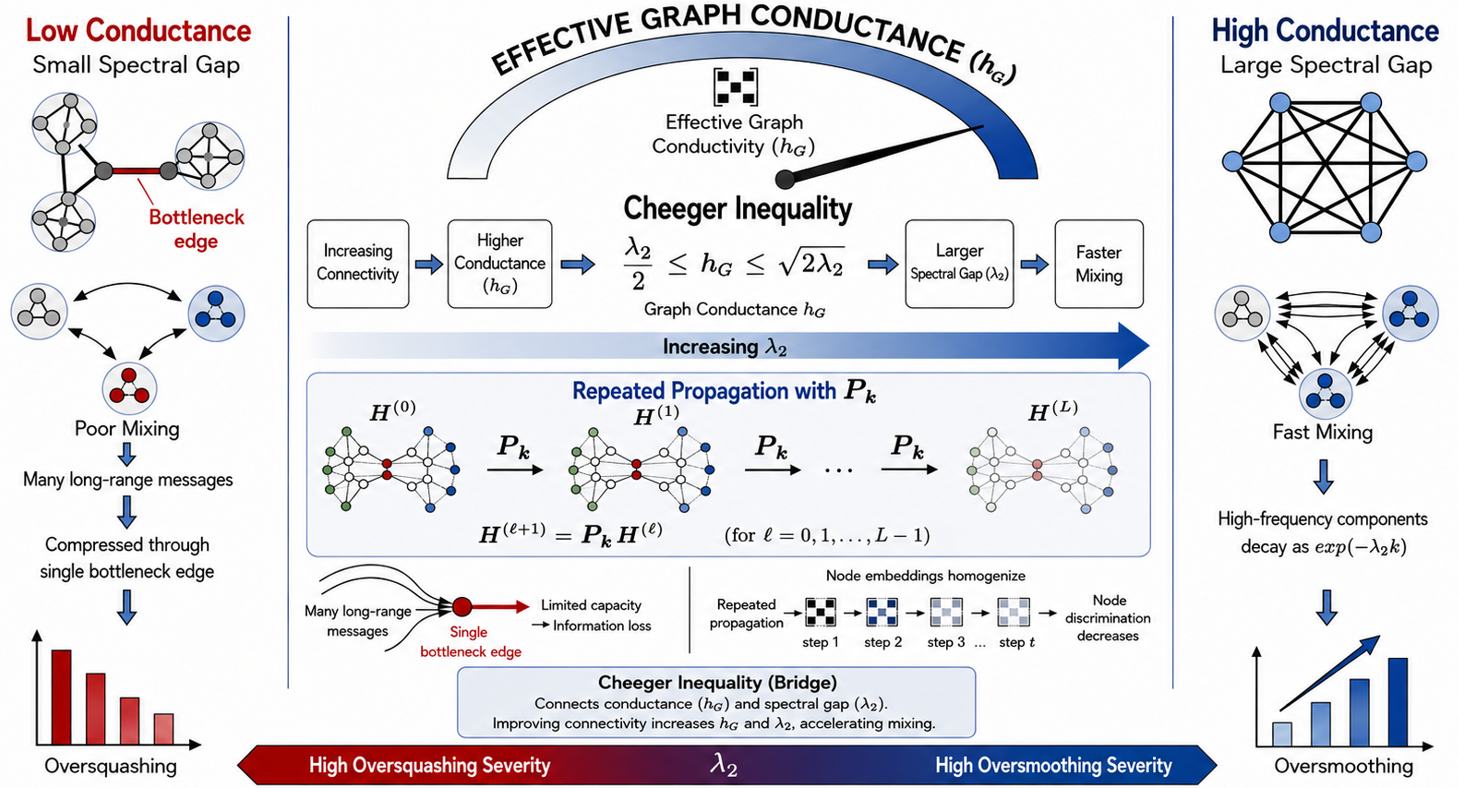}
\caption{\textbf{The spectral-gap trade-off.} Oversquashing and oversmoothing are governed
by the same spectral gap $\lambda_2$ in opposite directions. A low-conductance graph with a
bottleneck edge (left) has a small $\lambda_2$. Long-range messages compress through the
bottleneck, so oversquashing is severe while repeated propagation barely mixes the graph.
A high-conductance, densely connected graph (right) has a large $\lambda_2$. Mixing is
fast, relieving oversquashing, but the same speed drives high-frequency components to decay
as $\exp(-\lambda_2 k)$, so node embeddings homogenize just as quickly. The Cheeger bound
$\lambda_2/2\le h_G\le\sqrt{2\lambda_2}$ ties both effects to the same graph conductance
$h_G$. Moving $\lambda_2$ trades one pathology for the other rather than fixing both
independently.}
\label{fig:prop-tradeoff}
\end{figure*}

\citet{giraldo2023tradeoff} couple oversmoothing and oversquashing through the normalized-Laplacian spectral gap $\lambda_2$, in opposite directions. Here $\lambda_2$ is the second-smallest eigenvalue of $\mathbf L_{\mathrm{ref}}=\mathbf I-\hat{\mathbf A}$, not of $\mathbf L_{\mathrm{sym}}$ (Section~\ref{sec:preliminaries}), so $\lambda_2=1-\mu_2$. The sharper energy rate $\mu_\star=\max_{i\ge2}|\mu_i|$ equals $1-\lambda_2$ only when $\mu_2\ge0$ and $|\mu_2|\ge|\mu_n|$. The $\lambda_2$ interpretation applies only in this regime. Increasing the gap accelerates mixing and relieves oversquashing but also speeds the energy decay in Eq.~\eqref{eq:oono}. The Cheeger bound $\lambda_2/2\le h_G\le\sqrt{2\lambda_2}$ links both effects to conductance $h_G$, the worst-cut bottleneck targeted by curvature rewiring and illustrated in Figure~\ref{fig:prop-tradeoff}. Because the classical bound omits self-loops, $h_G$ here refers to the self-loop-augmented graph underlying $\mathbf L_{\mathrm{ref}}$. The fixed self-loop shift does not change the qualitative reading. Under the result's operator family and assumptions, changing the gap selects a trade-off point rather than independently improving both objectives, but this is not an impossibility theorem for all fixed-topology architectures or interventions.

\citet{Nguyen2023BORF} add edges at the most negative and prune at the most positive Ollivier--Ricci curvature. \citet{Karhadkar2023FoSR} increase $\lambda_2$ by first-order edge perturbation and use a relational split to avoid the induced smoothing fixed point. \citet{qian2024prmpnn} learn a differentiable $k$-subset edge distribution, specifying when it outperforms random rewiring and how it affects expressivity. Table~\ref{tab:prop-remedies} places these and earlier remedies in the component map.

Three results escape the trade-off. Edge \emph{deletions} can raise the spectral gap, letting spectral pruning mitigate both pathologies and linking the problem to graph lottery tickets \citep{jamadandi2024spectralpruning}. \citet{Errica2025AMP} learn effective depth and per-layer filters as ELBO latent variables, jointly curbing smoothing, squashing, and underreaching without global rewiring. \citet{blayney2026glstm} instead treat oversquashing as node-vector storage saturation relieved by greater effective storage at fixed topology.

Other work questions the vocabulary itself. Through counterexamples, \citet{arnaiz2026demystifying} separate oversmoothing (an operator property) from accuracy-related node non-separability, split oversquashing into topological and computational bottlenecks, and reject a general homophily-good/heterophily-bad rule. \citet{mishayev2025shortrange} further identify a short-range bottleneck missed by resistance and curvature and unresolved by virtual nodes but handled by graph transformers, separating bottleneck from range.

Our account preserves these distinctions. Table~\ref{tab:prop-diagnostics} separates the spectrum of $\mathbf P_k$ from non-separability and does not claim that an operator property alone predicts accuracy. Section~\ref{subsec:prop-oversquashing} separates the model factor $(c_\sigma\omega p)^t$ from topology $(\widehat{\mathbf S}^t)_{vu}$, while the short-range result separates bottleneck and range. Section~\ref{subsec:prop-heterophily} likewise ties harm to within-class structural variance and pass-band mismatch rather than heterophily alone. The spectral-gap trade-off is therefore a first-order organizing principle whose universality remains under investigation.

Table~\ref{tab:prop-diagnostics} summarizes oversmoothing and its backward/rank-collapse readings, oversquashing, underreaching, heterophily, their spectral-gap trade-off, and the structurally governed expressivity limits discussed next.

\colorlet{pdiagBar}{googleyellow}
\colorlet{pdiagRow}{googleyellow!9}
\begin{table}[t]
\centering
\footnotesize
\setlength{\tabcolsep}{4pt}
\renewcommand{\arraystretch}{1.3}
\caption{\textbf{Propagation diagnostics and their assumptions.} Each pathology, the measure
that detects it, and the operator, topology, capacity, or optimization quantity that governs
it (with the inequality direction where available). $\mathbf{L}_{\mathrm{ref}}$ is the self-loop-normalized reference Laplacian of Section~\ref{sec:preliminaries},
$\lambda_2$ the relevant normalized-Laplacian spectral gap in the cited result,
$\widehat{\mathbf S}$ the normalized sensitivity/support operator,
$R_{uv}$ effective resistance, and $\tau$ commute time.}
\label{tab:prop-diagnostics}
\resizebox{\linewidth}{!}{
\begin{tabular}{@{}L{3.1cm} L{4.0cm} L{6.4cm} L{3.4cm}@{}}
\toprule
\rowcolor{pdiagBar}
{\color{white}\textbf{Phenomenon}} & {\color{white}\textbf{Diagnostic measure}} &
{\color{white}\textbf{Governing quantity (and bound)}} & {\color{white}\textbf{Key refs}} \\
\midrule
Oversmoothing (forward) & Dirichlet energy $E$; MAD (Mean Average Distance); numerical rank &
largest nontrivial eigenvalue magnitude:
$d_{\mathcal M}(\mathbf H^{(\ell)}){\le}(\bar s\mu_\star)^{\ell}$;
$E(\mathbf H'){\le}(s\mu_\star)^2E(\mathbf H)$ under Theorem~\ref{thm:oversmoothing}; rank${\to}1$ under the cited conditions &
\citealp{oono2020expressive,cai2020note,roth2024rank,zhang2026measuring} \\
\rowcolor{pdiagRow}
Oversharpening (dual) & high-frequency energy growth & sign of channel-mixing eigenvalues:
most-negative dominant ${\Rightarrow}$ repulsion & \citealp{digiovanni2023understanding} \\
Oversmoothing (backward) & smoothness of backprop.\ error; flat high-loss regions &
linear backward smoothing; per-layer Lipschitz bound & \citealp{keriven2025backward,park2024taming} \\
\rowcolor{pdiagRow}
Oversquashing & Jacobian $\lVert\partial\vh_v^{(t)}/\partial\vh_u^{(0)}\rVert$;
mixing & $(\widehat{\mathbf S}^t)_{vu}$; $R_{uv}{=}\tau(u,v)/(2|\mathcal E|)$; capacity ${\gtrsim}\tau(u,v)$ &
\citealp{Topping2022Curvature,DiGiovanni2023OverSquashing,black2023resistance,digiovanni2024oversquashing} \\
Underreaching & receptive radius ${<}$ task distance & depth $t{<}d(u,v)$:
$(\widehat{\mathbf S}^t)_{vu}{=}0$ & \citealp{DiGiovanni2023OverSquashing,Errica2025AMP} \\
\rowcolor{pdiagRow}
Heterophily / band mismatch & aggregation homophily; Contextual Stochastic Block Model
(CSBM) Bayes error & pass-band of
$\mathbf{P}_k$ vs.\ signal band; low-pass response of $\hat{\mathbf A}$ is $(1{-}\lambda_i)$ &
\citealp{nt2019revisiting,luan2022revisiting,Ma2022HomophilyNecessary,luan2023whenhelp} \\
Smoothing$\leftrightarrow$squashing trade-off & mixing time; Cheeger $h_G$ &
spectral gap $\lambda_2$: $\lambda_2/2{\le}h_G{\le}\sqrt{2\lambda_2}$; Ollivier--Ricci sign &
\citealp{giraldo2023tradeoff,Nguyen2023BORF,jamadandi2024spectralpruning} \\
\rowcolor{pdiagRow}
Limited expressivity & WL level; substructure count; VC dim & ${\le}k$-WL refinement;
VC governed by $\#\{1\text{-WL-distinguishable}\}$ & \citealp{xu2019powerful,morris2023wlmeetvc} \\
\bottomrule
\end{tabular}}
\end{table}

\subsection{Expressivity: the WL ladder and substructure counting}
\label{subsec:prop-expressivity}

The commute-time capacity bound of Section~\ref{subsec:prop-oversquashing} already limits what a model can realize. More generally, expressivity asks what propagation can distinguish independently of optimization. The Weisfeiler--Leman (WL) test iteratively hashes each node's color with its neighbor-color multiset. Ordinary message passing can be simulated round by round and therefore cannot distinguish graphs that WL confuses. The proposition restates this ceiling \citep{xu2019powerful,morris2019weisfeiler,balcilar2021breaking} under explicit node-domain component assumptions and identifies three ways to break them. The proof appears in \hyperref[proof:prop-wl]{Appendix~\ref*{sec:analysis}}.

\begin{proposition}[$1$-WL ceiling and the escape mechanisms]
\label{prop:wl}
Consider a stack of node-domain layers with $\mathcal{X}=\mathcal{V}$ in which every map is \emph{shared and node-wise, with no independent access to $\mathbf{A}$ or global graph statistics}:
\begin{itemize}\setlength{\itemsep}{1pt}
\item one-hop operators whose entries $[\mathbf{P}_k]_{ij}$ are functions of the $1$-WL colors of the incident nodes $i$ and $j$ through a color-to-weight rule shared across graphs (the \emph{refinability hypothesis});
\item shared node-wise messages $\Psi_k$;
\item a shared node-wise channel mixing $\boxplus$;
\item shared node-wise ego and update maps $\mathcal{B}_\ell,\phi_\ell$;
\item no node identifiers in $\mathbf{H}^{(0)}$; and
\item a readout that is a shared function of the node-state multiset $\{\!\{\vh_i^{(L)}\}\!\}$ alone.
\end{itemize}
Such a stack is bounded by $1$-WL. Graphs that $1$-WL fails to distinguish receive equal outputs. We single out \emph{three structural escape mechanisms} that each negate one hypothesis above:
\begin{enumerate}\setlength{\itemsep}{1pt}
\item[(i)] the domain $\mathcal{X}$, through the graded extension to tuples, subgraphs, or cells~\citep{Zhou2023RelationalPooling};
\item[(ii)] the input message $\mathbf{H}^{(0)}$, through positional and structural encodings or node identifiers~\citep{murphy2019relational};
\item[(iii)] the operator $\mathbf{P}_k$ itself, when its entries depend on graph structure beyond the incident colors (for instance substructure counts such as per-edge triangle counts), which violates the refinability hypothesis and is the route taken by substructure-aware operators~\citep{bouritsas2023improving}.
\end{enumerate}
\end{proposition}

The color-to-weight rule suffices for one-round color refinability, while the structure-blind restrictions are essential. A readout that re-reads $\mathbf A$, or a $\mathcal B_\ell$, $\phi_\ell$, or $\boxplus$ that broadcasts a global invariant, can count substructures regardless of the operator. Permutation invariance alone does not exclude this. Thus the three listed escapes are not exhaustive because structure-aware $\Psi_k$, $\boxplus$, $\phi_\ell$, or readout can also exceed $1$-WL. For example, broadcasting $\operatorname{tr}(\mathbf A^3)$ separates the prism from $K_{3,3}$ with ordinary $\mathcal X$, $\mathbf H^{(0)}$, and $\mathbf P_k$. The proposition lists the escapes compatible with keeping all other components $1$-WL-bounded.

Changing a component need not yield a strict gain because a $1$-WL-refinable encoding such as degree, or an uninformative lift, preserves the ceiling. \citet{Pellizzoni2025SubstructureCounting} study which substructures ordinary message passing can count under graph-structural conditions.

Being bounded by $1$-WL does not imply attaining it because aggregation and update must preserve the relevant multiset information. The lemma gives the injectivity condition and explains why mean and max fail. The proof appears in \hyperref[proof:lem-nbr-agg-inj]{Appendix~\ref*{sec:analysis}}.

\begin{lemma}[Neighborhood-aggregation injectivity]
\label{lem:nbr-agg-inj}
The aggregation is over the neighbor index, realized by the operator (the bare support $\mathbf{P}_k=\mathbf{A}$ sums neighbor messages, $\mathbf{P}_k=\mathbf{D}^{-1}\mathbf{A}$ averages them), not by the channel mixing $\boxplus$. The constructions below use a single channel, so $\boxplus$ plays no role. For multisets of bounded size drawn from a countable set of node states, there exists an encoding $f$ such that the neighbor sum of the encoded states, $S\mapsto\sum_{x\in S}f(x)$, is injective on such multisets. Injective neighbor aggregation alone does \emph{not} attain the $1$-WL ceiling. It must be paired with an injective combination of the aggregate with the root state, i.e.\ an update of the form
\[
\vh_i^{(\ell+1)}=g\!\Big(\vh_i^{(\ell)},\,\textstyle\sum_{j\in \mathcal{N}(i)}f(\vh_j^{(\ell)})\Big)
\]
with $g$ injective in its pair of arguments. Placing $f$ in the message map $\Psi_k$, summing with $\mathbf{P}_k=\mathbf{A}$, and routing this $g$ through the ego map $\mathcal{B}_\ell$ and update $\phi_\ell$, then attains the ceiling. The raw neighbor sum without such an encoding is \emph{not} injective on multisets. No encoding makes the neighbor aggregators $\mathrm{mean}$ or $\max$ injective either. Mean is invariant to multiplicity duplication and max retains only extremes.
\end{lemma}

GIN realizes this construction, whereas mean and max lose multiplicity information \citep{xu2019powerful,morris2019weisfeiler}.

\colorlet{pexprBar}{googlegreen}
\colorlet{pexprRow}{googlegreen!9}
\begin{table}[t]
\centering
\footnotesize
\setlength{\tabcolsep}{4pt}
\renewcommand{\arraystretch}{1.25}
\caption{\textbf{Slot-aware expressivity hierarchy.} Each row places a model
\emph{class} (architectures in parentheses are instances) on the relevant
Weisfeiler--Leman or orthogonal expressivity axis and states the
substructure/homomorphism count it can or cannot achieve. Rows on incomparable axes
should not be read as a total bottom-to-top ordering.}
\label{tab:prop-expressivity}
\resizebox{\linewidth}{!}{
\begin{tabular}{@{}L{4.3cm} L{3.1cm} L{6.0cm} L{1.6cm} L{2.6cm}@{}}
\toprule
\rowcolor{pexprBar}
{\color{white}\textbf{Model class (instance)}} & {\color{white}\textbf{WL power}} &
{\color{white}\textbf{Counting power}} & {\color{white}\textbf{Memory}} & {\color{white}\textbf{Key refs}} \\
\midrule
Mean/max/attention-aggregation MPNN (GCN, GraphSAGE-mean, GAT) & ${\le}\,1$-WL (strict for mean/max) & loses
multiset multiplicities (mean/attention) or repeats (max) & $O(n{+}|\mathcal E|)$ & \citealp{xu2019powerful} \\
\rowcolor{pexprRow}
Injective sum-aggregation MPNN (GIN) & ${=}\,1$-WL (the ceiling, attained iff aggregate and
readout are injective) & stars only; no induced connected pattern ${\ge}3$ nodes; no
girth/diameter/$k$-clique/biconnectivity &
$O(n{+}|\mathcal E|)$ & \citealp{xu2019powerful,chen2020substructures,garg2020generalization,Zhang2023Biconnectivity} \\
Polynomial spectral GNN (GPR-GNN, BernNet, ChebNetII) & ${\le}\,1$-WL under the
structure-blind node-domain hypotheses of Proposition~\ref{prop:wl}; fixed-graph reachability is a separate axis & filter-band
dependent; reaches any $1$-D target on one fixed graph if the spectrum is simple and no frequency is missing &
$O(K|\mathcal E|)$ & \citealp{wang2022how} \\
\rowcolor{pexprRow}
Spectral-invariant GNN (eigenbasis invariants) & strictly between $1$-WL and the
subgraph-GNN band above it &
homomorphism-counts (hom-counts) exactly ``parallel trees''; class grows with depth & $O(n^2)$ &
\citealp{gai2025spectralhom} \\
Hom-count-augmented MPNN ($F$-MPNN; spasm basis) & ${=}\,F$-WL; gain over $1$-WL
depends on $F$; for every fixed $k$ some $F$ escapes $k$-WL
& counts patterns via $\mathrm{sub}(Q,G){=}\sum_{F\in\mathrm{spasm}(Q)}\alpha_F\hom(F,G)$ & $O(n)$ &
\citealp{Barcelo2021LocalGraphParams,jin2024hombasis} \\
\rowcolor{pexprRow}
Node-based subgraph GNN (ESAN, GNN-AK, SUN, SSWL) & ${>}\,1$-WL,
${<}\,2$-FWL ($=3$-WL) for the six ordered classes & the analyzed subgraph-MPNN
class node-counts cycles only through length $4$ &
$O(n(n{+}|\mathcal E|))$ & \citealp{Frasca2022SubgraphSymmetries,Zhang2023SubgraphWLHierarchy} \\
I$^2$-GNN & partially (not totally) stronger than $3$-WL & counts all
$3$--$6$-cycles in the cited construction & sparse; model-dependent & \citealp{huang2023i2gnn} \\
\rowcolor{pexprRow}
Generalized-distance WL (SPD-/RD-WL; Graphormer-GD) & orthogonal; gets biconnectivity
standard MPNNs cannot & SPD${\to}$vertex-, RD${\to}$edge-biconnectivity & $O(n^2)$ &
\citealp{Zhang2023Biconnectivity} \\
$k$-WL-aligned transformer (higher-order) & ${\ge}\,k$-WL (strictly more for $k{>}1$; $\delta$-$k$-WL) & inherits
$k$-WL; PE choice sets recoverable properties & $O(n^k)$ &
\citealp{Muller2024AligningTransformers} \\
\rowcolor{pexprRow}
$k$-GNN / set-$k$-WL & ${=}$ set-$k$-WL, strict in $k$ & bounded patterns scaling with $k$
& $O(n^k)$ & \citealp{morris2019weisfeiler} \\
PPGN / $2$-IGN${+}$matmul ($k$-FWL) & ${=}\,2$-FWL${=}3$-WL; $k$-FWL${=}(k{+}1)$-WL &
patterns up to bounded size & $O(n^2)$--$O(n^k)$ & \citealp{maron2019provably,chen2020substructures} \\
Recursive neighborhood pooling (RNP-GNN) & beyond fixed $k$-WL for counting & any connected
size-$k$ substructure; near-tight lower bound on \#subgraphs & adaptive & \citealp{tahmasebi2023recursion} \\
\bottomrule
\end{tabular}}
\end{table}

Table~\ref{tab:prop-expressivity} places model \emph{classes}, with architectures as instances, on three kinds of expressivity axis. These are exact classical WL rungs, subgraph-model bands between rungs, and orthogonal properties. On the classical ladder, $k$-GNNs match the strict set-$k$-WL hierarchy at $O(n^k)$ cost \citep{morris2019weisfeiler}. PPGN realizes $2$-FWL${}={}$$3$-WL with $O(n^2)$ memory, and generally $k$-FWL${}={}$$(k{+}1)$-WL \citep{maron2019provably}.

Between rungs, node-based subgraph classes lie strictly above $1$-WL and below $3$-WL. SUN subsumes its unified subgraph GNNs \citep{Frasca2022SubgraphSymmetries}. \citet{Zhang2023SubgraphWLHierarchy} divide this band into six totally ordered classes below $2$-FWL, including ESAN \citep{bevilacqua2022equivariant} and GNN-AK \citep{zhao2022from}.

Orthogonally, standard MPNNs cannot compute biconnectivity at any depth, whereas generalized-distance tests recover vertex biconnectivity with SPD-WL and edge biconnectivity with RD-WL \citep{Zhang2023Biconnectivity}. The classical ladder does not order this property.

Counting power measures pattern multiplicities rather than graph distinguishability. MPNNs, $2$-WL, and $2$-IGNs cannot count induced patterns on at least three nodes, although they can count stars, while $k$-WL counting power grows with $k$ \citep{chen2020substructures}. Recursive neighborhood pooling counts any connected size-$k$ pattern, induced or not, with a near-tight lower bound on the number of encoded subgraphs \citep{tahmasebi2023recursion}. The analyzed subgraph message-passing class cannot node-count cycles longer than four, whereas I$^2$-GNN counts cycles of length $3$--$6$ in linear time under bounded degree and is only partially stronger than $3$-WL---incomparable on some inputs and a strict superset on others \citep{huang2023i2gnn}. The $r$-loopy WL hierarchy instead indexes visible cycle length, strictly exceeds $1$-WL, and remains incomparable to every fixed $k$-WL \citep{paolino2024loopy}.

Homomorphism counts unify these results. A homomorphism is an edge-preserving map from pattern $F$ to $G$, counted by $\hom(F,G)$. Injecting $\hom(F,\cdot)$ gives exactly $F$-WL power, strictly beyond $1$-WL when the chosen $F$ adds unavailable information. Such $F$-MPNNs use $O(n)$ memory and are bounded by $k$-WL when $F$ has treewidth $k$ (a measure of how tree-like $F$ is), yet for every fixed $k$ some pattern family distinguishes graphs that $k$-WL cannot \citep{Barcelo2021LocalGraphParams}. \citet{zhou2024finegrained} characterize exactly which homomorphism counts each WL variant computes. Because subgraph counts decompose over the \emph{spasm} basis of vertex-merged versions of $Q$, basis hom-count features strictly subsume single-pattern counts at no extra asymptotic cost \citep{jin2024hombasis}.

On a fixed graph, a polynomial filter reaches every one-dimensional target under the simple-spectrum, no-missing-frequency conditions below. The proof appears in \hyperref[proof:thm-universality]{Appendix~\ref*{sec:analysis}}.

\begin{theorem}[Spectral universality]
\label{thm:universality}
If $\hat{\mathbf{A}}$ has distinct eigenvalues and the one-dimensional signal $\mathbf{x}$ has nonzero projection on every eigenvector, then
\[
\big\{g(\hat{\mathbf{A}})\,\mathbf{x} : g \text{ a polynomial of degree at most } n-1\big\}=\mathbb{R}^n .
\]
This is \emph{cyclic-vector reachability}. Because $\mathbf{x}$ has a nonzero component along every eigendirection of $\hat{\mathbf{A}}$, repeated application of $\hat{\mathbf{A}}$ to $\mathbf{x}$ generates the whole space. It is the same simple-spectrum, no-missing-frequency condition under which spectral GNNs are universal in the sense of \citet{wang2022how}.
\end{theorem}

This is fixed-graph filter reachability, not graph-distinguishing universality or a WL classification. Separately, spectral-invariant networks count exactly the ``parallel tree'' homomorphisms, with the class growing in depth \citep{gai2025spectralhom}.

Spatial and spectral $\mathbf P_k$ fillings are therefore complementary. Parallel spatial message passing and a parametrically bounded spectral filter widen the receptive field and mitigate oversquashing without extra message-passing depth \citep{geisler2024spatiospectral}. Spatially adaptive per-node aggregation can likewise complement a global spectral filter under heterophily \citep{guo2024spatially}. Properly conditioned Chebyshev filtering \citep{defferrard2016convolutional} is competitive on long-range tasks that motivated graph transformers \citep{hariri2025chebnet}, showing that the limits concern particular fillings, not the component itself.

\paragraph{Expressivity's generalization cost.}
Worst-case analyses often associate greater distinguishing power with looser generalization bounds, but not unconditionally. The Rademacher bound of \citet{garg2020generalization} grows exponentially with depth through branching and multiplicatively with maximum degree. The same class cannot decide girth, circumference, diameter, or $k$-clique. The tighter PAC-Bayes bound of \citet{liao2021pacbayesian} grows, up to logarithmic factors, with depth, degree, $\sqrt p$ (using the width $p$ of Section~\ref{subsec:prop-oversquashing}), and $\prod_\ell\lVert\mathbf W^{(\ell)}\rVert_2$ (whose factors are $s_\ell$ in Section~\ref{subsec:prop-oversmoothing}), and shrinks with margin and the square root of sample size. Non-uniformly, the VC dimension of $L$-layer GNNs is governed by the number of graphs distinguished by $1$-WL after $L$ rounds. Uniformly, it is infinite for $d,L\ge2$ \citep{morris2023wlmeetvc}.

Data-dependent bounds replace worst-case $d^L$ by instance-specific quantities. \citet{li2024homgen} use homomorphism-count entropy. The tension is not fundamental under every metric. \citet{rauchwerger2025generalization} define a hierarchical-optimal-transport pseudometric on attributed computation trees under which message passers are Lipschitz, fully separating at $1$-WL resolution, and defined on a relatively compact space, yielding both Stone--Weierstrass approximation and distribution-free generalization. This construction does not settle the general question. \citet{morris2024future} identify alignment among WL expressivity, generalization, and optimization as a central open direction.

\subsection{Global attention as a propagation channel}
\label{subsec:prop-transformer}

Global attention is the densest graph-transformer filling of $\mathbf P_k$ in Eq.~\eqref{eq:unified-notation}. Under its fixed-operator and node-local hypotheses, the proposition states when one-layer global mixing requires a dense row, why a virtual node needs two propagation steps, and why $\Psi_k$, $\boxplus$, or $\phi_\ell$ alone cannot supply globality. The proof appears in \hyperref[proof:prop-global]{Appendix~\ref*{sec:analysis}}.

\begin{proposition}[One-layer global mixing]
\label{prop:global}
Suppose the operators are \emph{fixed} (state-independent), the messages $\Psi_k$ are node-wise and linear in $\mathbf{H}$, the mixing is additive, the ego map is node-local (so $\mathcal{B}_\ell(\mathbf{H},\mathbf{H}^{(0)})_i$ depends only on $\vh_i$ and $\vh_i^{(0)}$), and the update $\phi_\ell$ is node-local and \emph{message-faithful}. We call the update message-faithful if, for every fixed ego input $a$, the partial map $b\mapsto\phi_\ell(a,b)$ from the mixed message to the output is injective (the common $\phi_\ell(a,b)=\sigma(a+b)$ with injective $\sigma$ qualifies). Then having every off-diagonal entry of row $i$ present in the combined support $\bigcup_k\operatorname{supp}(\mathbf{P}_k)$ (i.e.\ some $k$ with $[\mathbf{P}_k]_{ij}\neq 0$ for each $j\neq i$) is \emph{necessary} for $\vh_i^{(\ell+1)}$ to depend on every \emph{other} node $\vh_j^{(\ell)}$ ($j\neq i$). For generic channel weights $\{\mathbf{W}_k\}$ it is also sufficient. Consequently, under these hypotheses one-layer global mixing requires the effective off-diagonal row $i$ to be genuinely full (a dense operator, or a support rewiring that actually fills that row) and cannot be produced by $\Psi_k$, $\boxplus$, or $\phi_\ell$ alone.
\end{proposition}

A node-local ego map is essential because $\mathcal B_\ell(\mathbf H)_i=\sum_j\vh_j$ with $\phi_\ell(a,b)=a+b$ would create global dependence without operator support. Self-dependence is separate because $\mathcal B_\ell$ can carry $\vh_i$ even when all $[\mathbf P_k]_{ii}=0$. Including the ego path in the support recovers a full-row statement.

\begin{figure}[t]
    \centering
    \includegraphics[
        width=\linewidth,
        keepaspectratio
    ]{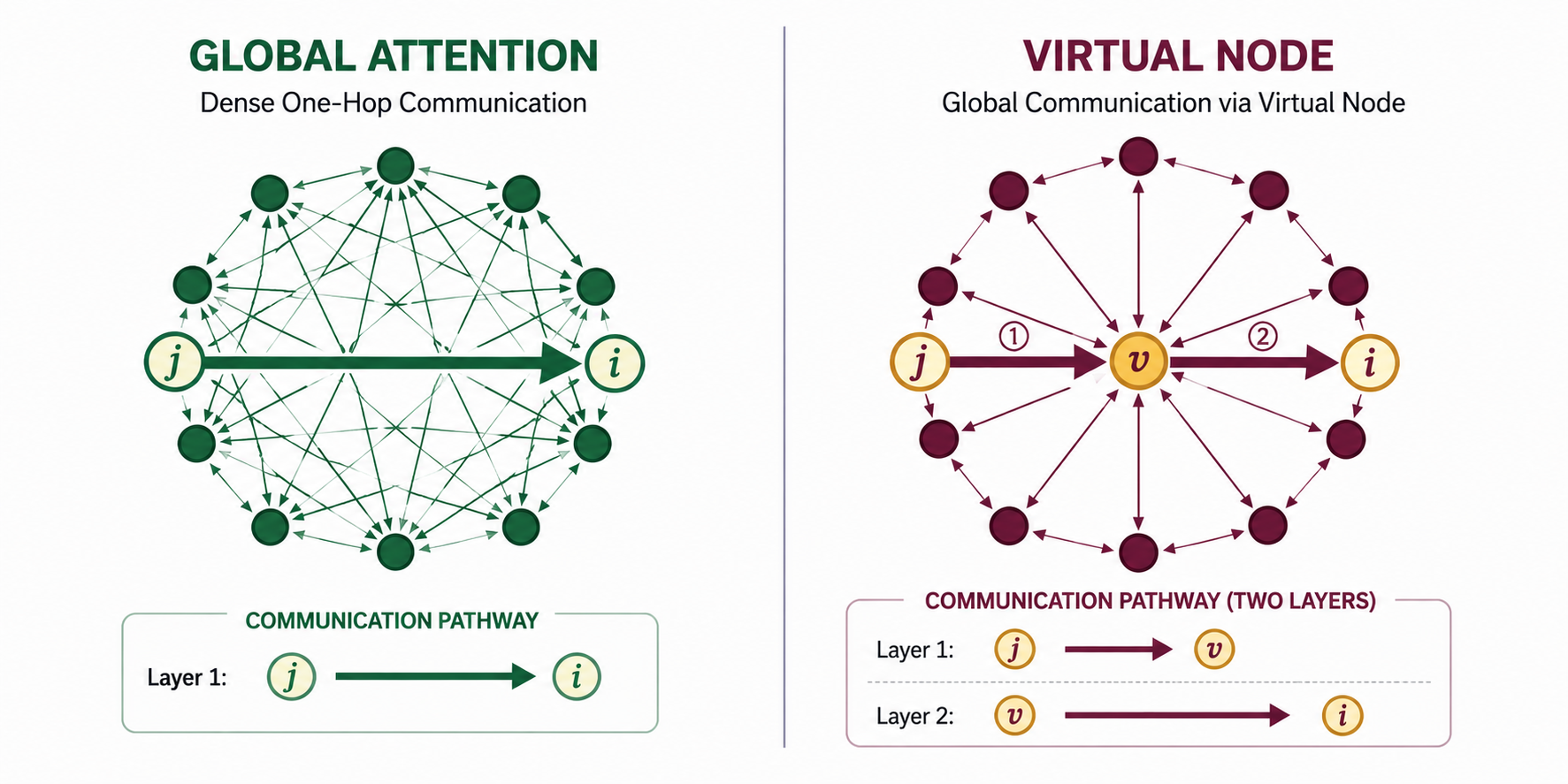}
    \caption{\textbf{Global mixing in one layer versus two.} A dense propagation operator
    (row $i$ full) lets node $i$ depend on every other node $\vh_j^{(\ell)}$ after a single
    application of $\mathbf{P}_k$ (Proposition~\ref{prop:global}). A virtual node instead
    adds only a one-hop path to a shared hub. Node $i$ reads the hub's pre-update state, but
    the hub itself must first collect from every real node, so a signal from $j$ reaches $i$
    only after two propagation steps, $j\to v\to i$. The same one-layer/two-layer gap holds
    for any support that leaves row $i$ short of full.}
    \label{fig:prop-global}
\end{figure}

A virtual-node channel is not global in one synchronous layer because $i$ reads the hub's pre-update state, so $j\to v\to i$ requires two steps, as Figure~\ref{fig:prop-global} shows. One-step globality requires a pre-aggregated hub, and rewiring provides it only if it fills row $i$.

The fixed-operator hypothesis is also essential. The globally state-dependent zero-hop operator $\mathbf P(\mathbf H)=\mathrm{sum}(\mathbf H)\mathbf I$, with $\mathrm{sum}(\mathbf H)=\sum_{j,c}H_{jc}$ (Prop.~\ref{prop:receptive}), creates all-node dependence through diagonal support alone.

An MPNN with a virtual node approximates linear attention at $O(1)$ depth and width \citep{Cai2023MPNNGTConnection}. Full softmax attention instead requires either $O(n^{d_{\mathrm{in}}})$ width at $O(1)$ depth or $O(1)$ width at $O(n)$ depth, where $d_{\mathrm{in}}$ is the input feature dimension, narrowing the expressivity gap under explicit resource trade-offs.

Positional and structural encodings also control expressivity. Higher-order $k$-tuple transformers are at least $k$-WL-powerful and, for $k>1$, strictly stronger by simulating $\delta$-$k$-WL \citep[Thms.~4--5]{Muller2024AligningTransformers}. Shortest-path or resistance-distance encodings can exceed $1$-WL, with the encoding determining recoverable properties \citep{Zhang2023Biconnectivity}. This power trades against stability. Laplacian-eigenvector encodings are unstable under eigenspace degeneracy, whereas the soft-partition encoding of \citet{huang2024spe}, obtained by applying a smooth function $g$ of the Laplacian eigenvalues in the eigenbasis, is Lipschitz-stable and universal over basis-invariant functions. The smoothness of $g$ mediates the trade-off. These Graph Transformer architectures (Appendix~\ref{app:family-comparison-tables}) instantiate the same propagation and input components and inherit their limits.

\subsection{Constraining Generated Architectures by Propagation Limits}
\label{subsec:prop-synthesis}

Oversmoothing (Section~\ref{subsec:prop-oversmoothing}), oversquashing (Section~\ref{subsec:prop-oversquashing}), and heterophily (Section~\ref{subsec:prop-heterophily}) depend on the spectrum and topology of $\mathbf P_k$, with the first two coupled through $\lambda_2$ (Section~\ref{subsec:prop-tradeoff}). Expressivity (Section~\ref{subsec:prop-expressivity}) depends instead on accessible structural information. Remedies therefore localize to named components (Table~\ref{tab:prop-remedies}), and changing a filling, support, or schedule changes which limit binds rather than eliminating all limits. This operator-level account remains partial because node separability and feature distributions are independent axes not determined by the spectrum alone.

A generative process over Eq.~\eqref{eq:unified-notation} (Section~\ref{sec:generation}) can propose $\mathbf P_k^{(\ell)}$, $\Psi_k$, $\phi_\ell$ (including $\mathcal B_\ell$), and $\boxplus$ jointly. Before pre-training, Tables~\ref{tab:prop-diagnostics} and \ref{tab:prop-remedies} support criteria that penalize a spectral-gap operating point that worsens both smoothing and squashing (Section~\ref{subsec:prop-tradeoff}) or favor a pass-band conditioned on expected domain homophily over a fixed low-pass response. The diagnosed limits thus constrain architecture generation, and known component-specific remedies provide generation moves.

Which limit binds is an empirical question. Section~\ref{sec:benchmarks} relates documented benchmark evidence to the same component vocabulary without causal attribution.

\section{Benchmark Properties in the Component Vocabulary}
\label{sec:benchmarks}

Section~\ref{sec:propagation} characterized what a propagation operator's spectrum and
topology can and cannot do. This section records what widely used benchmark papers establish
about their datasets and protocols, then translates those properties into the component
vocabulary. The translation identifies where a model can respond to a documented property. It
does not identify a causal bottleneck, attribute accuracy to one component, or predict the best
filling.

\paragraph{Node-classification benchmarks.}
Cora, Citeseer, and Pubmed are homophilous citation networks under edge and adjusted
homophily~\citep{yang2016planetoid,platonov2023characterizing}. This makes the topology encoded
by $\mathbf P_k$ relevant, but it does not isolate propagation from features, messages, or
updates. Indeed, adjusted homophily is more comparable across class counts and class imbalance
than common alternatives, while label informativeness agrees more closely with GNN performance
in controlled experiments~\citep{platonov2023characterizing}. Model rankings on these datasets
also change with data splits, hyperparameters, and training procedures
\citep{shchur2018pitfalls}.

Geom-GCN evaluated geometric aggregation on the Texas, Wisconsin, Cornell, Actor,
Chameleon, and Squirrel suite~\citep{pei2020geom}. The suite cannot be treated uniformly as a
clean test of heterophilous propagation because Chameleon and Squirrel contain duplicate nodes
that leak across splits. Removing the duplicates changes model performance substantially
\citep{Platonov2023Benchmarks}. The replacement suite introduced by
\citet{Platonov2023Benchmarks} contains graphs with different structural properties, and standard
GNNs almost always outperform the specialized models evaluated there. The larger LINKX datasets
extend non-homophilous evaluation to substantially larger graphs and show that existing scalable
and mini-batch methods can degrade in this regime~\citep{lim2021linkx}. Within the unified
equation, methods responding to these settings commonly change $\mathbf P_k$, separate channels
through $\mathcal K$, or separate the ego state through $\mathcal B_\ell$. The benchmarks do not
isolate any one of these choices.

\paragraph{Benchmark collections and controlled protocols.}
OGB standardizes large, realistic datasets, application-specific splits, metrics, and baseline
pipelines across node, link, and graph prediction~\citep{hu2020ogb}. Its datasets differ in
domain, graph type, and task, so \texttt{ogbn} and \texttt{ogbl} do not share a single
component interpretation. Their common documented challenges are scale and generalization under
realistic splits. TUDataset collects more than 120 graph-classification and regression datasets
with loaders, baselines, and standardized evaluation~\citep{morris2020tudataset}. It is a
dataset repository rather than a controlled test of one layer component. Benchmarking-GNNs
instead fixes parameter budgets and experimental infrastructure across real and synthetic tasks.
ZINC is molecular graph regression, PATTERN is graph-pattern node classification, and CLUSTER is
semi-supervised clustering on stochastic-block-model graphs
\citep{dwivedi2023benchmarking}. These collections support controlled model comparison, but a
comparison changes whichever components differ between the evaluated models.

\paragraph{Long-range and distribution-shift benchmarks.}
LRGB contains five graph-learning datasets selected because their task construction and graph
statistics suggest dependence on long-range interactions~\citep{Dwivedi2022LRGB}. Its authors
explicitly do not claim a provable long-range requirement. The relevant architectural choices
include the support and reach of $\mathbf P_k$, the channel structure $\mathcal K$, and depth.
However, later reevaluation showed that stronger tuning can greatly reduce or eliminate the
reported gap between message-passing GNNs and Graph Transformers. It also identified missing
feature normalization and an erroneous link-prediction metric in the original evaluation
\citep{Tonshoff2023GapGo}. LRGB therefore evaluates complete model and training configurations,
not oversquashing or a propagation component in isolation.

GOOD provides 11 datasets, 17 domain selections, and 51 splits that distinguish covariate,
concept, and no-shift settings for node and graph prediction~\citep{gui2022good}. A feature-based
domain selection concerns $\mathbf H^{(0)}$, while a structural domain selection concerns the
graph information available to $\mathbf P_k$. Concept shift changes the input--label
relationship and is not itself a layer component. The relevant entry therefore depends on how
each GOOD split is constructed rather than on the benchmark name alone.

\paragraph{Industrial and temporal evaluation.}
GraphLand contributes 14 industrial node-property datasets with diverse sizes, structures,
features, and realistic temporal splits~\citep{Bazhenov2025GraphLand}. Its experiments show
that graph-augmented gradient-boosted trees can be strong baselines, particularly for regression,
and that temporal shift and inductive evaluation can substantially reduce performance. These
findings jointly involve input representation $\mathbf H^{(0)}$, propagation $\mathbf P_k$, and
evaluation conditions outside the layer equation rather than one component.

TGB provides large temporal datasets and realistic protocols for node-property and future-link
prediction~\citep{huang2023tgb}. Performance varies substantially across datasets, and simple
methods can outperform temporal graph models on dynamic node-property tasks. For dynamic link
prediction, EdgeBank shows that repeated edges and easy negative samples can make a
memorization baseline unexpectedly strong and can alter method rankings
\citep{poursafaei2022edgebank}. Temporal state can be represented by an extension
$\mathcal B(t)$, time-varying support by $\mathbf P_k(t)$, and event messages by $\Psi_k$, while
negative sampling and the prediction head remain outside the layer equation.
\colorlet{benchBar}{googleblue}
\colorlet{benchRow}{googleblue!9}
\begin{table}[t]
\centering
\footnotesize
\setlength{\tabcolsep}{4pt}
\renewcommand{\arraystretch}{1.3}
\caption{\textbf{Benchmark evidence in the component vocabulary.} The documented
design or finding column summarizes claims made by the cited benchmark papers. The component
reading is our translation into Eq.~\eqref{eq:unified-notation} and does not attribute performance
to a component. The symbol $\mathbf H^{(0)}$ denotes the input representation. A $\dagger$ marks
a task head or readout, scale or sampling concern, distribution split, or temporal extension outside
the seven base-layer components.}
\label{tab:benchmark-slot-map}
\resizebox{\linewidth}{!}{
\begin{tabular}{@{}L{3.0cm} L{2.6cm} L{6.0cm} L{3.2cm} L{2.5cm}@{}}
\toprule
\rowcolor{benchBar}
{\color{white}\textbf{Benchmark family}} & {\color{white}\textbf{Task / domain}} &
{\color{white}\textbf{Documented design or finding}} &
{\color{white}\textbf{Component-level reading}} & {\color{white}\textbf{Key refs}} \\
\midrule
Citation networks (Cora, Citeseer, Pubmed; Planetoid splits) & Citation graphs &
Node classification on homophilous graphs. Adjusted homophily is preferable for cross-dataset comparison, label informativeness agrees more closely with GNN performance, and rankings depend on splits and tuning &
$\mathbf P_k$ encodes topology, but no layer component is isolated &
\citealp{yang2016planetoid,shchur2018pitfalls,platonov2023characterizing,luo2024classic} \\
\rowcolor{benchRow}
OGB node / link (\texttt{ogbn}, \texttt{ogbl}) & Multiple domains and tasks &
Large datasets with application-specific splits, metrics, and standardized pipelines. Scale and out-of-distribution generalization are common challenges &
Dataset- and task-dependent; scale, split, and prediction head$^{\dagger}$ & \citealp{hu2020ogb} \\
Heterophily suite, legacy (Geom-GCN: Texas, Wisconsin, Cornell, Actor, Chameleon, Squirrel) &
Node classification & Geom-GCN evaluated geometric aggregation on this suite. Duplicate nodes in Chameleon and Squirrel create train--test leakage and materially change results when removed &
No reliable component attribution; $\mathbf P_k,\mathcal K,$ and $\mathcal B_\ell$ are relevant design axes &
\citealp{pei2020geom,Platonov2023Benchmarks} \\
\rowcolor{benchRow}
Heterophily suite, corrected (Roman-empire, Amazon-ratings, Minesweeper, Tolokers, Questions) &
Node classification & Diverse heterophilous graphs without the identified duplicate-node leakage. Standard GNNs are strong baselines, and connectivity is better characterized jointly by adjusted homophily and label informativeness &
$\mathbf P_k,\mathcal K,\mathcal B_\ell,$ and $\mathbf H^{(0)}$ are relevant but not isolated &
\citealp{Platonov2023Benchmarks,platonov2023characterizing} \\
Large non-homophilous (LINKX datasets) & Social and web node classification &
Substantially larger non-homophilous graphs. Existing scalable and mini-batch methods degrade on these datasets &
$\mathbf P_k,\mathcal K,\mathbf H^{(0)}$; scale and mini-batching$^{\dagger}$ & \citealp{lim2021linkx} \\
\rowcolor{benchRow}
TUDataset & Graph classification and regression across more than 120 datasets &
A repository with standardized loaders, baselines, and evaluation rather than a controlled test of one architectural mechanism &
Dataset-dependent; graph-level readout $\rho^{\dagger}$ & \citealp{morris2020tudataset} \\
Benchmarking-GNNs (ZINC, PATTERN, CLUSTER) & Molecular regression and synthetic node classification &
Fixed parameter budgets and shared infrastructure. ZINC predicts a molecular property, PATTERN detects an embedded graph pattern, and CLUSTER predicts SBM community labels &
Multiple components vary across compared models; readout $\rho^{\dagger}$ for ZINC &
\citealp{dwivedi2023benchmarking} \\
\rowcolor{benchRow}
Long Range Graph Benchmark (five datasets) & Molecules and superpixel graphs &
Graph and node tasks selected for evidence suggestive of long-range dependence, not a provable requirement. Tuning closes much of the reported GNN--Transformer gap &
$\mathbf P_k,\mathcal K,$ and stack depth$^{\dagger}$; readout $\rho^{\dagger}$; no isolated oversquashing test &
\citealp{Dwivedi2022LRGB,Tonshoff2023GapGo} \\
Out-of-distribution (GOOD) & Mixed node and graph prediction &
Eleven datasets and 51 splits separating covariate, concept, and no-shift settings across 17 domain selections &
$\mathbf H^{(0)}$ for feature domains, $\mathbf P_k$ for structural domains; concept shift and readout$^{\dagger}$ & \citealp{gui2022good} \\
\rowcolor{benchRow}
GraphLand & Industrial node classification and regression &
Fourteen datasets with diverse features and structures, temporal splits, and inductive settings. Graph-augmented GBDTs are strong baselines in some tasks &
$\mathbf H^{(0)},\mathbf P_k$; scale and temporal split$^{\dagger}$ & \citealp{Bazhenov2025GraphLand} \\
Temporal Graph Benchmark (TGB) & Temporal node-property and future-link prediction &
Large temporal datasets with realistic protocols. Simple methods can outperform temporal models, while EdgeBank exposes repeated-edge and negative-sampling effects &
$\mathcal B(t),\mathbf P_k(t),\Psi_k$; prediction head and negative sampling$^{\dagger}$ &
\citealp{huang2023tgb,poursafaei2022edgebank} \\
\bottomrule
\end{tabular}
}
\end{table}

\paragraph{Boundaries of the component reading.}
Graph-level tasks require the post-stack readout $\rho$, while link prediction also requires a
pairwise scoring head. Both lie downstream of the seven-component layer. Scale, neighborhood
sampling, and mini-batching are systems and approximation choices rather than additional layer
components. Sampling may approximate $\mathbf P_k$, and attribute-guided variants may condition
that approximation on graph properties~\citep{das2024agsgnn}, but the sampling algorithm is not
$\mathbf P_k$ itself. Temporal models extend the static layer through $\mathcal B(t)$,
$\mathbf P_k(t)$, and event-dependent $\Psi_k$. Data splits, negative sampling, metrics, and
training budgets remain evaluation choices. Section~\ref{sec:outlook} treats graph-level readout,
scalable propagation, and temporal carry as open extensions in
Sections~\ref{subsec:op-readout}, \ref{subsec:op-scaling}, and \ref{subsec:op-temporal}.

\paragraph{What benchmark evidence permits component-level conclusions.}
First, protocol validity is logically prior to component attribution. Data splits and training
choices can reverse model rankings~\citep{shchur2018pitfalls}, duplicate nodes can invalidate a
benchmark comparison~\citep{Platonov2023Benchmarks}, and missing normalization or metric errors can
distort an apparent architectural gap. Basic tuning can substantially reduce or eliminate that
gap~\citep{Tonshoff2023GapGo}. Repeated edges and negative
sampling can likewise make a parameter-free memorization baseline competitive in dynamic link
prediction~\citep{poursafaei2022edgebank}. Evaluation must therefore establish dataset and
protocol validity before comparing models, and it must establish a fair model comparison before
attributing a difference to any component.

Second, mechanistic isolation and deployment realism answer different questions. Controlled
synthetic and semi-synthetic data can vary connectivity while holding other factors fixed, as in
the experiments that separate label informativeness from homophily
\citep{platonov2023characterizing}. Realistic benchmarks instead combine multiple sources of
variation. LRGB selects real tasks whose construction and statistics suggest long-range
dependence but explicitly avoids claiming a provable long-range requirement
\citep{Dwivedi2022LRGB}. OGB combines domains, scales, tasks, and application-specific
splits~\citep{hu2020ogb}, while GraphLand's temporal splits can jointly change feature, label, and
structural distributions~\citep{Bazhenov2025GraphLand}. Component validation should therefore
pair controlled tasks that isolate a mechanism with realistic datasets that test whether the
result survives deployment-relevant variation.

Third, equal-resource comparison and best-attainable performance are distinct evaluation
targets. Benchmarking-GNNs fixes parameter budgets and infrastructure to compare models under
matched resources~\citep{dwivedi2023benchmarking}. By contrast, architecture-specific tuning
changes rankings~\citep{shchur2018pitfalls,Tonshoff2023GapGo}, and tuned GCN, GAT, and GraphSAGE
match or exceed recent Graph Transformers on 17 of 18 node-classification datasets studied by
\citet{luo2024classic}. A component study should report both a matched-resource comparison and a
comparison with equal tuning effort. The former measures behavior under a common constraint,
whereas the latter estimates the performance each model can attain. Neither result should be
presented as the other.

Fourth, the relation between benchmark properties and components is generally many-to-many.
Heterophilous node classification can involve $\mathbf P_k$, $\mathcal K$, $\mathcal B_\ell$,
and $\mathbf H^{(0)}$. Long-range tasks involve $\mathbf P_k$, $\mathcal K$, stack depth, and
possibly a downstream readout. Temporal prediction can involve $\mathcal B(t)$,
$\mathbf P_k(t)$, $\Psi_k$, the prediction head, and negative sampling. The seven-component
decomposition is useful here because it identifies the smallest interacting component set while
separating external evaluation choices. It does not justify assigning each benchmark to one
component. Component attribution requires changing one filling at a time while holding the
remaining architecture, training budget, data split, task head, and evaluation protocol fixed.
Claims about component interactions additionally require controlled factorial comparisons.
Section~\ref{subsec:op-evaluation} treats the remaining attribution problem.

\FloatBarrier

\section{Open Problems and Challenges}
\label{sec:outlook}

The preceding sections address the forward problem by representing architectures through $(\mathcal{X},\mathcal{K},\{\mathbf{P}_k\},\{\Psi_k\},\boxplus,\mathcal B_\ell,\phi_\ell)$ (Section~\ref{sec:unified-view}) and using these components as design choices (Section~\ref{sec:generation}). The inverse problem asks which fillings suit measurable dataset, task, or deployment properties, how the components should extend to new domains, and what trade-offs result. No generally validated solution exists. Table~\ref{tab:open-problems} summarizes the research targets examined below. A separate thematic cluster applies the component vocabulary to surrounding systems that remain outside the formal scope of the base-layer equation.

\begingroup
\footnotesize
\setlength{\tabcolsep}{4pt}
\renewcommand{\arraystretch}{1.25}
\setlength{\LTcapwidth}{\textwidth}
\colorlet{opBar}{googlered}
\colorlet{opRow}{googlered!9}
\begin{longtable}{@{}
  >{\raggedright\arraybackslash}p{0.145\textwidth}
  >{\raggedright\arraybackslash}p{0.245\textwidth}
  >{\raggedright\arraybackslash}p{0.245\textwidth}
  >{\raggedright\arraybackslash}p{0.090\textwidth}
  >{\raggedright\arraybackslash}p{0.135\textwidth}
  @{}}
\caption{\textbf{Open problems in graph neural network design.} Each row names a problem,
why it resists current methods, and what would count as progress. The slot column marks the
part of the unified equation~\eqref{eq:unified-notation} (or the named extension) the problem stresses. The key-reference column gives a curated,
representative subset of the citations discussed in the corresponding subsection, not the
full set. Rules separate the thematic clusters.}
\label{tab:open-problems}\\
\toprule
\rowcolor{opBar}
{\color{white}\textbf{Open problem}} & {\color{white}\textbf{Why it is hard}} & {\color{white}\textbf{What would count as progress}} &
{\color{white}\textbf{Slot}} & {\color{white}\textbf{Key refs}} \\
\midrule
\endfirsthead
\multicolumn{5}{c}{\tablename~\thetable\ -- \textit{continued}}\\
\toprule
\rowcolor{opBar}
{\color{white}\textbf{Open problem}} & {\color{white}\textbf{Why it is hard}} & {\color{white}\textbf{What would count as progress}} &
{\color{white}\textbf{Slot}} & {\color{white}\textbf{Key refs}} \\
\midrule
\endhead
\midrule
\multicolumn{5}{r}{\textit{continued on next page}}\\
\endfoot
\bottomrule
\endlastfoot
Generalization beyond expressivity & More expressive fillings provably inflate the
complexity term (WL classes bound Rademacher complexity), and no theory ties SGD implicit bias
to generalization & Non-vacuous, structure-aware bounds naming which statistics control
error at a reasonable computational cost, valid for sparse unbounded-size graphs, and connected
to scaling the operator (Section~\ref{subsec:op-scaling}) & $\mathbf{P}_k,\Psi_k,\boxplus,\phi_\ell$ &
\citealp{morris2024future,vasileiou2025survey,carrasco2025rademacher,amran2026graphop,rauchwerger2025generalization} \\
\addlinespace
\rowcolor{opRow}
Distribution / size shift & Failure is a shift over local structure. Causal-invariance
remedies rest on unverifiable assumptions, and rankings reverse across shift types & A
measurable predictor of which statistic governs transfer and validated environment-free
causal-subgraph discovery & $\mathbf{P}_k,\mathcal{X}$ &
\citealp{yehudai2021size,li2025sizespectral,loveland2024localhomophily,chen2022ciga,gui2022good} \\
\addlinespace
Dataset-adaptive slot selection & Homophily is sensitive to class count/balance and
unreliable for model selection. No mechanistic property-to-slot map exists & A robust measured
property predicting the right $\mathbf{P}_k/\Psi_k$ across datasets &
$\mathbf{P}_k,\Psi_k,\mathcal{X}$ &
\citealp{you2020design,platonov2023characterizing,luan2024heterophily,kormann2026dontbeafraid} \\
\addlinespace
\rowcolor{opRow}
Slots as a search space (graph neural architecture search, graph-NAS) & Graph-NAS learns shift-fragile non-causal maps,
is non-reproducible without a shared space, and uses language-model controllers that lack guarantees & Causal,
data-conditioned search whose discovered tuple generalizes under shift &
all seven components & \citealp{wang2022automated,qin2022nasbenchgraph,li2024carnas,gao2025llm4gnas} \\
\addlinespace
Scaling the operator $\mathbf{P}_k$ & Whether dense attention is even needed is unproven. It is
unknown how much expressivity sub-quadratic attention
sacrifices, sparsification and condensation guarantees do not transfer, and no clean long-range
benchmark exists & A characterized expressivity-efficiency frontier and a structured $O(n)$ operator
provably matching dense attention & $\mathbf{P}_k,\mathcal{X}$ &
\citealp{Deng2024Polynormer,sancak2024geco,chen2024demystifyingsparsification,Tonshoff2023GapGo,gupta2026condensationreset} \\
\addlinespace
\rowcolor{opRow}
Temporal / time-indexed carry slot & Dominant temporal families are incomparable, standard
messages cannot express moving averages, and component-to-gain links are untested & A temporal
WL/counting ladder and a principled continuous-time memory operator with known limits &
$\mathcal{B}(t),\mathbf{P}_k(t),\Psi_k$ &
\citealp{souza2022pint,tjandra2024tgnv2,cong2023graphmixer,li2025dygmamba,tieu2025invariantlink} \\
\addlinespace
Graph-level readout / pooling & Learned local pooling barely beats random or complement-graph
baselines on standard benchmarks, so its value is unproven. A readout can also discard expressivity gained upstream & A
readout with a proven expressivity guarantee that also beats a tuned global-sum baseline on
leakage-free data & readout $\rho$ &
\citealp{mesquita2020rethinking,wang2024poolingbenchmark,vonpichowski2026nodefeatures,bianchilachi2023expressive,ying2018hierarchical,xu2019powerful} \\
\addlinespace
\rowcolor{opRow}
Graph foundation models & Problems include no shared vocabulary, negative transfer from
cross-domain pre-training, no predictive data-scaling law, and possible shortcuts in language-model routes & A transferable
structural vocabulary, a priori compatibility prediction, and structure-grounded transfer &
$\mathbf{H}^{(0)},\mathcal{X},\mathbf{P}_k,\Psi_k$ &
\citealp{mao2024position,liu2025gfmsurvey,zhao2024gcope,tang2025uniaug,liu2024ofa} \\
\addlinespace
Trustworthy GNNs & Robustness, privacy, fairness, and uncertainty are entangled through propagation.
Certified robustness destroys accuracy, aggregation resists forgetting, and ensembles collapse
& Architecture-aware certificates that retain clean accuracy, topology-aware debiasing,
verifiable forgetting, and graph-native uncertainty quantification & $\mathbf{P}_k$ &
\citealp{dai2024trustworthy,scholten2022interception,lai2025auditvotes,dong2023fairness,fan2025opengu,vieira2026ensembles} \\
\addlinespace
\rowcolor{opRow}
Equivariance level \& generation & Strict equivariance cannot break input symmetry,
choosing the level requires task geometry, and generation metrics reverse rankings & A way to
measure task geometry and set or relax the symmetry, valid-by-construction scalable
generation, and a stable metric & symmetry on $\mathbf{P}_k,\Psi_k$ &
\citealp{duval2023hitchhiker,kaba2023symmetry,hofgard2024relaxed,vadgama2025probing,obray2022evaluation,yan2024swingnn} \\
\midrule
\addlinespace
\multicolumn{5}{@{}l}{\textit{Broader outlook beyond the base-layer equation}} \\
\addlinespace
Retrieval-grounded generation & There is no theory for selecting a subgraph or conditioning
a frozen decoder through its propagation, and the retrieval target is fixed heuristically & A principled
query-to-subgraph map and a propagation operator that minimizes hallucination &
$\mathcal{X},\mathbf{P}_k$ &
\citealp{edge2024graphrag,he2024gretriever,mavromatis2024gnnrag,luo2024rog,pan2024unifyingkg,wang2024knowledge} \\
\addlinespace
\rowcolor{opRow}
Personalization & Per-user conditioning under extreme sparsity and drift is difficult without
retraining the operator per user, as is fusing a user model with a decoder & Per-user adaptation surviving
sparsity and drift and personalized generation rather than only ranking & $\mathbf{P}_k,\Psi_k,\mathcal{X}$ &
\citealp{ying2018pinsage,he2020lightgcn,wu2022gnnrec,shi2025cfrag} \\
\addlinespace
Graphs as a reasoning substrate & The reasoning graph is hand-designed, no theory identifies which
structure a task needs, and gains are not attributable to a task property & Task-inferred reasoning
topology and a measured link from structure to reasoning quality & $\mathcal{X},\mathbf{P}_k,\boxplus$ &
\citealp{besta2024got,wang2023nlgraph,wang2026graphshelpllms} \\
\addlinespace
\rowcolor{opRow}
Language agents as graphs & Propagation carries language rather than vectors, with no expressivity ladder
and no credit assignment over agent edges & A task-to-communication-graph map with assignable
credit over an optimized topology & $\mathbf{P}_k,\Psi_k,\mathcal{X}$ &
\citealp{zhuge2024gptswarm,zhang2025gdesigner,qian2025macnet} \\
\addlinespace
The model as a graph (metanetworks) & Permutation-correct weight-space operators are shown
only on small networks, and scaling to edit-relevant sizes is open & A symmetry-correct
$\mathbf{P}_k$ over weights that edits or generates parameters at scale & $\mathbf{P}_k,\Psi_k,\mathcal{X}$ &
\citealp{kofinas2024neuralgraphs,lim2024graphmetanetworks} \\
\midrule
\addlinespace
Evaluation crisis (meta) & Gains evaporate under fair tuning. Other problems include benchmark leakage,
easy-negative and metric artifacts, molecule-dominated datasets, and a possible collapse of
Family~3 (Spectral) into Family~1 (Spatial) for node classification & Leakage-free,
shift-aware, application-grounded benchmarks with non-accuracy metrics &
all seven components (attribution) &
\citealp{bechlerspeicher2025benchmarks,errica2020fair,Platonov2023Benchmarks,luo2024classic,huang2023tgb,jiang2026spectralneither} \\
\end{longtable}
\endgroup

\subsection{Generalization and optimization theory beyond expressivity}
\label{subsec:op-generalization}

\begin{tcolorbox}[
  colback=openQBack,
  colframe=openQFrame,
  boxrule=0.6pt,
  arc=1mm,
  left=2mm, right=2mm, top=1.5mm, bottom=1.5mm,
  boxsep=1mm,
  before skip=4pt, after skip=4pt,
  breakable
]
\textbf{\textcolor{openQFrame}{Open question.}} Is there a joint, non-vacuous, structure-aware
theory that predicts performance and, in slot terms, says which fillers buy task-useful
expressivity without paying the complexity penalty?
\end{tcolorbox}

The Weisfeiler--Leman and counting ladder in Section~\ref{subsec:prop-expressivity}
characterizes what message passing and higher-order extensions can distinguish, but not how
trained GNNs generalize. \citet{morris2024future} therefore call for a theory combining
expressivity, generalization, and the optimization dynamics of stochastic first-order
training. Existing VC-dimension, Rademacher-complexity, PAC-Bayes, and covering-number bounds
are often vacuous and do not identify the structural statistics governing out-of-sample
error~\citep{vasileiou2025survey}. \citet{carrasco2025rademacher} show that the number of
WL-coloring equivalence classes upper-bounds Rademacher complexity. Thus, enriching
$\mathbf{P}_k$, $\Psi_k$, $\boxplus$, or $\phi_\ell$ to gain expressivity inflates the
complexity term in these worst-case bounds. For sparse, unbounded-size graphs,
\citet{amran2026graphop} establish equicontinuity of message maps under the graphop metric,
extending prior limit-based guarantees beyond dense or bounded-size graphs. Whether these
bounds are tight enough to guide slot selection remains open.

This worst-case inflation is not fundamental. \citet{rauchwerger2025generalization}
construct a hierarchical-optimal-transport pseudometric under which message passers are
Lipschitz, $1$-WL-separating, and defined on a relatively compact space, yielding both a
Stone--Weierstrass universal-approximation result and distribution-free generalization
bounds. However, this result covers one metric and does not determine which arbitrary slot
fillings admit coexistence rather than an inflating Rademacher bound. It is also unknown
whether the guarantee survives replacing a dense $\mathbf{P}_k$ with a cheaper operator,
linking the problem to Section~\ref{subsec:op-scaling}.

\subsection{Distribution shift and size generalization}
\label{subsec:op-shift}

\begin{tcolorbox}[
  colback=openQBack,
  colframe=openQFrame,
  boxrule=0.6pt,
  arc=1mm,
  left=2mm, right=2mm, top=1.5mm, bottom=1.5mm,
  boxsep=1mm,
  before skip=4pt, after skip=4pt,
  breakable
]
\textbf{\textcolor{openQFrame}{Open question.}} Which structural statistics measurably control
transfer across graphs of different size or local structure?
\end{tcolorbox}

Distribution shift sharpens this gap. \citet{yehudai2021size} formalize size generalization
through local ``$d$-patterns'' and prove that zero-training-loss solutions can fail on larger
graphs, with gradient descent empirically converging to them. The failure therefore reflects
a shift in local structure rather than a capacity limit. On biological graphs,
\citet{li2025sizespectral} attribute degradation at $2$--$10\times$ larger sizes to spectral
differences induced by subgraph patterns such as average cycle length, although this account
has not been established beyond that domain. Global homophily is likewise unreliable because
local homophily varies within a graph~\citep{loveland2024localhomophily}. Recovering an
environment-invariant causal subgraph~\citep{chen2022ciga} relies on structural-causal-model
assumptions challenged by \citet{xu2024beyondcausality}, while method rankings reverse
between covariate and concept shift~\citep{gui2022good}. The slot-level problem is therefore
to identify which $\mathbf{P}_k$ and receptive-field domain $\mathcal{X}$ remain invariant
across relevant shift types, including size-induced spectral change.

\subsection{Dataset-adaptive slot selection and the fragility of homophily}
\label{subsec:op-adaptive}

\begin{tcolorbox}[
  colback=openQBack,
  colframe=openQFrame,
  boxrule=0.6pt,
  arc=1mm,
  left=2mm, right=2mm, top=1.5mm, bottom=1.5mm,
  boxsep=1mm,
  before skip=4pt, after skip=4pt,
  breakable
]
\textbf{\textcolor{openQFrame}{Open question.}} How can the distribution of task-relevant
information across a dataset be characterized statistically?
\end{tcolorbox}

No principled map connects measured graph properties to slot choices, and homophily is too
fragile to provide one. \citet{you2020design} search a $\sim$315K-design space and transfer
designs using task similarity, but the design--task relation remains empirical. Standard
homophily measures depend on class count and balance and are not comparable across
datasets~\citep{platonov2023characterizing,Mironov2024UnbiasedHomophily}. Adjusted homophily
restores comparability, while label informativeness tracks GNN performance more closely.
Under proper tuning, traditional homophily metrics remain unreliable for model
selection~\citep{luan2024heterophily,Wang2024UnderstandingHeterophily}. A data-centric account
instead attributes degradation to under-informative receptive fields rather than smoothing
itself~\citep{kormann2026dontbeafraid}. Their information distribution is precisely what a
selection map for $\mathbf{P}_k$, $\Psi_k$, and $\mathcal{X}$ must characterize.

\subsection{Automated and causal architecture generation over the slot search space}
\label{subsec:op-nas}

\begin{tcolorbox}[
  colback=openQBack,
  colframe=openQFrame,
  boxrule=0.6pt,
  arc=1mm,
  left=2mm, right=2mm, top=1.5mm, bottom=1.5mm,
  boxsep=1mm,
  before skip=4pt, after skip=4pt,
  breakable
]
\textbf{\textcolor{openQFrame}{Open question.}} Can the unified equation, as a structured
search space, be searched reproducibly, in a way that generalizes under shift and is
conditioned on data properties rather than spurious correlations?
\end{tcolorbox}

The unified equation defines a structured search space. Automated graph learning still
lacks tractable graph-specific spaces, scalable search, and a principled map from data
properties to architectures~\citep{wang2022automated}. Without a shared space, graph neural
architecture search (graph-NAS) results are not comparable, and preferred architectures are
dataset-specific~\citep{qin2022nasbenchgraph}. Graph-NAS can also learn non-causal,
distribution-specific graph--architecture correlations that fail under shift.
Causal-subgraph conditioning addresses this failure but leaves environment-free selection
open~\citep{li2024carnas}. Language-model controllers can adapt search from problem and
space descriptions, but provide no correctness or optimality guarantees and remain limited
by coverage and search cost~\citep{gao2025llm4gnas}. What is missing is a causal,
data-conditioned, and language-steerable search procedure over the full tuple
$(\mathcal{X},\mathcal{K},\{\mathbf{P}_k\},\{\Psi_k\},\boxplus,\mathcal B_\ell,\phi_\ell)$ that
generalizes under shift.

\subsection{Scaling the propagation operator and the cost of global attention}
\label{subsec:op-scaling}

\begin{tcolorbox}[
  colback=openQBack,
  colframe=openQFrame,
  boxrule=0.6pt,
  arc=1mm,
  left=2mm, right=2mm, top=1.5mm, bottom=1.5mm,
  boxsep=1mm,
  before skip=4pt, after skip=4pt,
  breakable
]
\textbf{\textcolor{openQFrame}{Open question.}} Is dense, quadratic all-pairs attention
necessary to capture global context, or can a structured sub-quadratic operator match it
within an $O(n)$ budget?
\end{tcolorbox}

Scaling a fixed architecture raises an expressivity--efficiency problem primarily at
$\mathbf{P}_k$. Polynomial linear attention and global convolution reduce cost, with the
latter reporting a $169\times$ speedup over optimized dense attention on a two-million-node
synthetic graph, but neither characterizes the expressivity sacrificed by
linearity~\citep{Deng2024Polynormer,sancak2024geco}. Million-node training also requires
model--system co-design, so asymptotic complexity understates practical
cost~\citep{zhang2024torchgt}. No universal sparsifier exists, and distance-preserving
sparsifiers can harm downstream GNNs~\citep{chen2024demystifyingsparsification}.
Graph condensation compresses $\mathcal{X}$ but can require model-dependent full-graph
training~\citep{gupta2026condensationreset}, and no scalable method dominates the
accuracy--efficiency frontier~\citep{duan2022largescalebenchmark}. Offline propagation
improves throughput but cannot adapt to inference-time input~\citep{Liao2025HubGT}.
Moreover, fair message-passing tuning largely closes the Long Range Graph Benchmark gap,
leaving the need for global $\mathbf{P}_k$ unestablished~\citep{Tonshoff2023GapGo}. The open
problem is whether dense all-pairs coupling is necessary and, if so, which sub-quadratic
operator preserves its useful expressivity. Compression of $\mathcal{X}$ is a secondary
axis.

\subsection{Temporal graphs and the carry slot as a memory operator}
\label{subsec:op-temporal}

\begin{tcolorbox}[
  colback=openQBack,
  colframe=openQFrame,
  boxrule=0.6pt,
  arc=1mm,
  left=2mm, right=2mm, top=1.5mm, bottom=1.5mm,
  boxsep=1mm,
  before skip=4pt, after skip=4pt,
  breakable
]
\textbf{\textcolor{openQFrame}{Open question.}} What is the temporal analogue of the static
expressivity ladder, what is the right continuous-time form for $\mathcal{B}(t)$, and which
temporal ingredients earn their cost?
\end{tcolorbox}

The static carry slot $\mathcal{B}_\ell$ in Eq.~\eqref{eq:unified-notation} seeds an update
from $\mathbf{H}^{(\ell)}$ and $\mathbf{H}^{(0)}$. Dynamic graphs require a time-indexed
memory operator $\mathcal{B}(t)$ that retains and selectively forgets information across
irregular events. The temporal problem is therefore an under-specified slot rather than a
missing one. Walk-aggregating and memory-based temporal GNNs are mutually incomparable, and
neither computes girth under a temporal WL analogue~\citep{souza2022pint}. Standard temporal
messages also lose event source and target identity, preventing persistent forecasting and
moving-average representation. Restoring that identity recovers the capability in one
architecture but is not standard across temporal-GNN families~\citep{tjandra2024tgnv2}.
Strong temporal link prediction may require neither recurrent memory nor self-attention,
leaving component-level attribution unresolved~\citep{cong2023graphmixer}. Selective
continuous-time state-space memory keyed by inter-event gaps provides one
candidate~\citep{li2025dygmamba}. However, spatial-temporal distribution shift also requires
$\mathcal{B}(t)$ to specify which temporal structure remains
invariant~\citep{tieu2025invariantlink}. The seven-component tuple remains intact over time, with
$\mathbf{P}_k(t)$ carrying time-varying support, $\Psi_k$ preserving event identity, and
$\mathcal B_\ell$ generalizing to $\mathcal{B}(t)$.

\subsection{Graph-level readout as the missing pooling slot}
\label{subsec:op-readout}

\begin{tcolorbox}[
  colback=openQBack,
  colframe=openQFrame,
  boxrule=0.6pt,
  arc=1mm,
  left=2mm, right=2mm, top=1.5mm, bottom=1.5mm,
  boxsep=1mm,
  before skip=4pt, after skip=4pt,
  breakable
]
\textbf{\textcolor{openQFrame}{Open question.}} Does a learned pooling operator help at all?
\end{tcolorbox}

The seven-component equation updates node states but does not include the
permutation-invariant readout $\rho$ that maps the final node-state multiset to a graph
vector. Its hierarchical counterpart $\mathbf{Q}_\ell$ coarsens the graph between layers
(Section~\ref{sec:unified-view}). Both discard node-level structure, so pooling benchmarks
test $\mathbf{Q}_\ell$ while the same expressivity concern applies to $\rho$. Replacing
learned local pooling with random assignments or complement-graph clustering barely changes
standard graph-classification accuracy~\citep{mesquita2020rethinking}. Across 17 pooling
methods and 28 datasets, no method consistently dominates and unpooled baselines match most
of them~\citep{wang2024poolingbenchmark}. Feature--topology mismatch explains these marginal
or inconsistent gains~\citep{vonpichowski2026nodefeatures}.

A readout can discard expressivity gained through $\Psi_k$ and $\phi_\ell$.
\citet{bianchilachi2023expressive} provide sufficient preservation conditions, but not an
operator that also outperforms global summation. Differentiable soft
clustering~\citep{ying2018hierarchical} and the injective sum readout that reaches $1$-WL with
injective aggregation~\citep{xu2019powerful} bracket the design space. Closing this gap needs an
expressivity-preserving readout that beats a tuned global-sum baseline on leakage-free data.
Because $\rho$ lies downstream of the seven layer components, Section~\ref{sec:benchmarks}
correctly marks graph classification as out-of-slot.

\subsection{The missing transferable vocabulary for graph foundation models}
\label{subsec:op-gfm}

\begin{tcolorbox}[
  colback=openQBack,
  colframe=openQFrame,
  boxrule=0.6pt,
  arc=1mm,
  left=2mm, right=2mm, top=1.5mm, bottom=1.5mm,
  boxsep=1mm,
  before skip=4pt, after skip=4pt,
  breakable
]
\textbf{\textcolor{openQFrame}{Open question.}} Which slot fillings should be shared versus
adapted across domains?
\end{tcolorbox}

A versatile graph foundation model requires shared primitives that transfer across
feature- and structure-heterogeneous graphs, together with positive transfer and predictable
data scaling~\citep{mao2024position}. Open challenges include structural alignment, semantic
heterogeneity, scalable pre-training, and trustworthy
evaluation~\citep{liu2025gfmsurvey,wang2025gfmsurvey}. A unified text-attributed feature and label space
currently covers only classification on text-rich graphs~\citep{liu2024ofa}. Cross-domain
pre-training can cause negative transfer, and compatibility cannot yet be predicted a
priori~\citep{zhao2024gcope}. Structural augmentation provides initial evidence of
cross-domain data scaling but not a predictive scaling model~\citep{tang2025uniaug}.
Graph--language alignment may also exploit shortcuts rather than genuine structural
transfer~\citep{tang2024graphgpt,pan2024integrating}, paralleling
Section~\ref{subsec:op-reasoning}. The required vocabulary spans
$\mathbf{H}^{(0)}$, $\mathcal{X}$, $\mathbf{P}_k$, and $\Psi_k$.

\subsection{Trustworthy GNNs and entanglement through the propagation slot}
\label{subsec:op-trust}

\begin{tcolorbox}[
  colback=openQBack,
  colframe=openQFrame,
  boxrule=0.6pt,
  arc=1mm,
  left=2mm, right=2mm, top=1.5mm, bottom=1.5mm,
  boxsep=1mm,
  before skip=4pt, after skip=4pt,
  breakable
]
\textbf{\textcolor{openQFrame}{Open question.}} Why does trustworthy graph learning resist any
per-axis fix?
\end{tcolorbox}

Message passing propagates sensitive attributes, adversarial perturbations, and structural
bias through the same operator. Privacy, robustness, fairness, uncertainty, and accuracy
therefore interact rather than forming independent axes~\citep{dai2024trustworthy}. A single
adversarial edge can affect an entire receptive field, while neighboring attributes leak and
heterophily may amplify or mask the effect~\citep{Zhu2022HeterophilyRobustness}. Gray-box
message-interception certificates exploiting $\mathbf{P}_k$ are tighter than conservative
black-box certificates~\citep{scholten2022interception}, yet randomized smoothing can require
enough noise to collapse clean accuracy~\citep{lai2025auditvotes}. Propagation also amplifies
sensitive-attribute correlations~\citep{dong2023fairness}, and recursive aggregation retains
deleted information through neighbors, obstructing unlearning~\citep{fan2025opengu}. Deep
ensembles provide little benefit because independently trained GNNs can undergo epistemic
collapse~\citep{vieira2026ensembles}. The resulting targets are architecture-aware
certification, topology-aware debiasing, verifiable forgetting, and graph-native uncertainty
quantification.

\subsection{Equivariance level and symmetry relaxation}
\label{subsec:op-equivariance}

\begin{tcolorbox}[
  colback=openQBack,
  colframe=openQFrame,
  boxrule=0.6pt,
  arc=1mm,
  left=2mm, right=2mm, top=1.5mm, bottom=1.5mm,
  boxsep=1mm,
  before skip=4pt, after skip=4pt,
  breakable
]
\textbf{\textcolor{openQFrame}{Open question.}} Which symmetry constraint should $\mathbf{P}_k$
and $\Psi_k$ satisfy, and when should it be relaxed?
\end{tcolorbox}

Geometric GNNs span a spectrum of symmetry constraints, leaving an
expressivity--scalability--symmetry trade-off~\citep{duval2023hitchhiker}. Because an
equivariant function cannot break an input's symmetry, exact equivariance cannot select
among symmetry-equivalent outcomes~\citep{kaba2023symmetry}. Learnable relaxation for $E(3)$
GNNs provides one candidate filler~\citep{hofgard2024relaxed}. At matched capacity in
convolutional image networks, equivariance helps only when aligned with task geometry, and
reference-frame symmetry breaking can improve performance~\citep{vadgama2025probing}. This
finding has not been tested directly for graphs.

\paragraph{Graph generation.}
Permutation-invariant diffusion models are harder to train than non-invariant models whose
target distributions have fewer modes, although post-hoc permutation can recover
invariance~\citep{yan2024swingnn}. Evaluation metrics are themselves unstable and can reverse model
rankings~\citep{obray2022evaluation}. What remains is a valid-by-construction, scalable
generator and a stable metric. At the slot level, task geometry must determine symmetry
constraints on $\mathbf{P}_k$ and $\Psi_k$ that guarantee equivariance without preventing
the distinction of symmetry-inequivalent configurations.

\subsection{Broader Outlook Beyond the Base-Layer Equation}

The following questions concern systems surrounding a graph layer. We use the component
vocabulary to locate their graph-side design choices, but the seven-component equation does
not represent the complete decoder, retrieval pipeline, or multi-agent controller.

\subsubsection{Subgraph selection and propagation for retrieval-grounded generation}
\label{subsec:op-retrieval}
\begin{tcolorbox}[
  colback=openQBack,
  colframe=openQFrame,
  boxrule=0.6pt,
  arc=1mm,
  left=2mm, right=2mm, top=1.5mm, bottom=1.5mm,
  boxsep=1mm,
  before skip=4pt, after skip=4pt,
  breakable
]
\textbf{\textcolor{openQFrame}{Open question.}} Which subgraph should be retrieved, and how
should its propagation condition the decoder, when neither the retrieval target nor the
coupling has a theory yet?
\end{tcolorbox}

Graph retrieval-augmented generation inserts a graph operator between a corpus and a frozen
language model to ground generation in explicit
structure~\citep{han2024retrieval,peng2025graph}. Community-structured indexing improves corpus-global
queries but fixes graph granularity~\citep{edge2024graphrag}. Question-specific subgraph
retrieval with propagated embeddings~\citep{he2024gretriever} and GNN reasoning over dense
subgraphs with verbalized paths~\citep{mavromatis2024gnnrag} likewise choose the retrieved
domain $\mathcal{X}$ heuristically. Agentic methods instead let an LLM beam-search
knowledge-graph relations~\citep{sun2024thinkongraph} or select tools while updating
memory~\citep{jiang2025kgagent}. However, prompting or supervised imitation still fixes the
exploration policy rather than learning an operator jointly with the decoder. Faithful,
interpretable knowledge-graph grounding remains the
target~\citep{luo2024rog,pan2024unifyingkg,wang2024knowledge}. The unresolved slot-level problem is
joint query-conditioned selection of $\mathcal{X}$ and $\mathbf{P}_k$ to minimize
hallucination for a fixed decoder.

\subsubsection{Personalization by conditioning the operator on the individual}
\label{subsec:op-personalization}
\begin{tcolorbox}[
  colback=openQBack,
  colframe=openQFrame,
  boxrule=0.6pt,
  arc=1mm,
  left=2mm, right=2mm, top=1.5mm, bottom=1.5mm,
  boxsep=1mm,
  before skip=4pt, after skip=4pt,
  breakable
]
\textbf{\textcolor{openQFrame}{Open question.}} How can per-user conditioning be done under
extreme sparsity and drift without retraining the operator per user?
\end{tcolorbox}

Personalization conditions propagation on user profiles and interactions, commonly represented
by a user--item graph~\citep{wang2023collaboration}. Web-scale graph
convolution~\citep{ying2018pinsage} and minimal neighborhood aggregation for collaborative
filtering~\citep{he2020lightgcn} still learn one shared operator and derive personalization from static
user embeddings. This design struggles with cold starts, preference drift, and mismatch
between a fixed operator and non-stationary $\mathcal{X}$~\citep{wu2022gnnrec}. Language-model
personalization further requires fusing user representations with a decoder so that generation,
not only ranking, is personalized~\citep{zhang2024personalization}. Collaborative-filtering
retrieval can condition generation on a user's history and those of similar users, but the
fusion remains retrieval-mediated rather than co-designed with the
decoder~\citep{shi2025cfrag}. The open task is a user-conditioned $\mathbf{P}_k$ and $\Psi_k$ over
$\mathcal{X}$ that survive sparsity and drift while supporting personalized generation.

\subsubsection{Graphs as a reasoning substrate for language models}
\label{subsec:op-reasoning}
\begin{tcolorbox}[
  colback=openQBack,
  colframe=openQFrame,
  boxrule=0.6pt,
  arc=1mm,
  left=2mm, right=2mm, top=1.5mm, bottom=1.5mm,
  boxsep=1mm,
  before skip=4pt, after skip=4pt,
  breakable
]
\textbf{\textcolor{openQFrame}{Open question.}} What theory would determine which structure a
given task actually needs?
\end{tcolorbox}

A graph over a language model's intermediate states can structure multi-step reasoning, but
its topology is usually hand-designed. Graph-of-thought methods generalize chain and tree
prompting through aggregation and refinement, yet user-specified topology prevents
attributing gains to measured task properties~\citep{besta2024got}. Performance also
degrades with graph-problem size~\citep{wang2023nlgraph}, and genuine reasoning remains hard
to distinguish from shortcut exploitation~\citep{wang2026graphshelpllms}. This calls for a
task-inferred domain $\mathcal{X}$ and propagation topology $\mathbf{P}_k$, with intermediate
states mixed by $\boxplus$ into $\mixmsg$ and measurable structure--quality links.

\subsubsection{Language agents as optimizable graphs}
\label{subsec:op-agents}
\begin{tcolorbox}[
  colback=openQBack,
  colframe=openQFrame,
  boxrule=0.6pt,
  arc=1mm,
  left=2mm, right=2mm, top=1.5mm, bottom=1.5mm,
  boxsep=1mm,
  before skip=4pt, after skip=4pt,
  breakable
]
\textbf{\textcolor{openQFrame}{Open question.}} What principled map takes a task to a
communication graph, together with credit assignment over it?
\end{tcolorbox}

In graph-structured agent systems, nodes are operations and edges are message channels.
Jointly optimizing prompts and edges remains empirical and does not explain which topology a
task requires or why an edge helps~\citep{zhuge2024gptswarm}. A variational graph
auto-encoder can instead generate task-adaptive communication
topology~\citep{zhang2025gdesigner}. A reported collaborative scaling law finds logistic performance
growth with agent count and an advantage for irregular, small-world topologies over regular
ones~\citep{qian2025macnet}. Yet propagation carries language rather than vectors, leaving no
expressivity ladder or account of when denser routing earns its cost. The open target is a
task-conditioned $\mathbf{P}_k$ and $\Psi_k$ over the operation domain $\mathcal{X}$ with
assignable credit.

\subsubsection{Editing and generating networks by treating the model as a graph}
\label{subsec:op-metanetworks}
\begin{tcolorbox}[
  colback=openQBack,
  colframe=openQFrame,
  boxrule=0.6pt,
  arc=1mm,
  left=2mm, right=2mm, top=1.5mm, bottom=1.5mm,
  boxsep=1mm,
  before skip=4pt, after skip=4pt,
  breakable
]
\textbf{\textcolor{openQFrame}{Open question.}} What permutation-correct operator can edit,
merge, or generate weights while scaling to large models?
\end{tcolorbox}

Representing neurons as nodes and weights as edges lets a GNN read and write another model's
parameters. Existing operators respect neuron-permutation symmetry and process diverse
architectures~\citep{kofinas2024neuralgraphs,lim2024graphmetanetworks}, but are demonstrated
only on small networks. This applies Section~\ref{subsec:op-equivariance} to a weight-space
domain $\mathcal{X}$ and leaves unresolved whether symmetry supplies useful structure at
scale or merely constrains capacity. Progress here needs $\mathbf{P}_k$ and $\Psi_k$ over
weights that support symmetry-correct parameter editing and generation at language-model
scale.

\subsection{The evaluation crisis and whether progress can be measured}
\label{subsec:op-evaluation}

\begin{tcolorbox}[
  colback=openQBack,
  colframe=openQFrame,
  boxrule=0.6pt,
  arc=1mm,
  left=2mm, right=2mm, top=1.5mm, bottom=1.5mm,
  boxsep=1mm,
  before skip=4pt, after skip=4pt,
  breakable
]
\textbf{\textcolor{openQFrame}{Open question.}} Can reported gains be reliably attributed to a
specific slot filling, rather than to tuning effort, benchmark leakage, or metric artifacts?
\end{tcolorbox}

Section~\ref{sec:benchmarks} shows that protocol validity, fair model comparison, and component
attribution are separate requirements. The open problem is to turn that hierarchy into an
evaluation design that isolates one filling while preserving realistic data and equal tuning
effort. Existing benchmarks also overrepresent small molecules and marginal accuracy, which
limits application-grounded and foundation-model evaluation
\citep{bechlerspeicher2025benchmarks}. Structure-agnostic baselines match or exceed GNNs on
several graph-classification benchmarks~\citep{errica2020fair}, while self-explainable GNNs lack
formal faithfulness guarantees~\citep{azzolin2025beyond}. Standard fidelity and sparsity metrics
do not test recovery of causal variables, motivating an out-of-distribution-generalization score
as a causal-validity proxy~\citep{zhang2026egs}.

Progress therefore requires leakage-free, shift-aware, application-grounded comparisons that
vary one component or a specified interaction while controlling the remaining architecture and
evaluation pipeline. The concern also reaches the taxonomy. Spectral propagation may not be theoretically distinct from
spatial message passing for node classification~\citep{jiang2026spectralneither}. If this
critique holds broadly, Family~3 (Spectral) collapses into Family~1 (Spatial) for that setting,
requiring a task-conditional taxonomy or a distinction beyond eigenspace representability.

\newpage
\section{Conclusion}
\label{sec:conclusion}

Graph neural network architectures have multiplied faster than the language used to compare
them. We unify covered graph neural layers through seven components consisting of an update
domain, channel set, propagation bank, per-channel message maps, channel-mixing operator,
ego/residual map, and update map. Function-valued fillings allow the same equation to express
local message passing, attention, spectral filtering, global communication, relation-specific
channels, higher-order domains, and geometric messages. The mixing operator $\boxplus$ is a
component, while $\mixmsg$ denotes only its output. Worked reductions of four canonical
layers and representative fillings across all seven families make this coverage checkable.

The equation defines a unified component system. Equivalent factorizations remain
possible, while the slot discipline provides consistent assignments for component-level
analysis. Under this discipline, differing components identify where the framework records
architectural differences. Functional equivalence additionally requires agreement among the
internal mechanisms and component interactions that determine the resulting inductive bias. The seven primary-axis
families are therefore nonexclusive and remain distinct from reusable operations and problem
settings.

The unified system supports analysis as well as comparison. Under endpoint-local messages
and node-local updates, operator support bounds one-layer dependencies and yields a necessary
full-row condition for one-layer global mixing. In the linear additive regime, raw channel
count is non-identifiable, whereas minimum Kronecker channel rank is invariant. These
attribution and identifiability results depend on explicitly separating the propagation bank,
messages, channels, mixing, and update. More broadly, spectral and topological properties of
$\mathbf P_k$ connect propagation to oversmoothing, oversquashing, heterophilic failure, and
expressivity limits. The same vocabulary identifies which component a remedy changes,
translates documented benchmark evidence descriptively into component choices, and exposes
structured combinations for architecture generation. These generated architectures provide formally
specified candidates for empirical evaluation.

The contribution establishes representation and component-level analysis, while selecting the
best filling for a dataset remains an empirical inverse problem. Establishing a validated map from
measurable graph and task properties to component choices
requires controlled evidence and robust descriptors, particularly because common statistics
such as homophily are fragile. The unified equation, families, and comparison tables provide a
shared language for pursuing that problem and for stating exactly what a new layer changes.

\newpage
\bibliographystyle{tmlr}
\bibliography{main}

@book{chung1997spectral,
  author    = {Fan R. K. Chung},
  title     = {Spectral Graph Theory},
  series    = {CBMS Regional Conference Series in Mathematics},
  number    = {92},
  publisher = {American Mathematical Society},
  year      = {1997}
}

@article{shuman2013emerging,
  author  = {David I. Shuman and Sunil K. Narang and Pascal Frossard and Antonio Ortega and Pierre Vandergheynst},
  title   = {The Emerging Field of Signal Processing on Graphs: Extending High-Dimensional Data Analysis to Networks and Other Irregular Domains},
  journal = {IEEE Signal Processing Magazine},
  volume  = {30},
  number  = {3},
  pages   = {83--98},
  year    = {2013},
  doi     = {10.1109/MSP.2012.2235192}
}

@article{scarselli2009graph,
  author  = {Franco Scarselli and Marco Gori and Ah Chung Tsoi and Markus Hagenbuchner and Gabriele Monfardini},
  title   = {The Graph Neural Network Model},
  journal = {IEEE Transactions on Neural Networks},
  volume  = {20},
  number  = {1},
  pages   = {61--80},
  year    = {2009},
  doi     = {10.1109/TNN.2008.2005605}
}

@inproceedings{bruna2014spectral,
  author    = {Joan Bruna and Wojciech Zaremba and Arthur Szlam and Yann LeCun},
  title     = {Spectral Networks and Locally Connected Networks on Graphs},
  booktitle = {International Conference on Learning Representations},
  year      = {2014},
  url       = {https://arxiv.org/abs/1312.6203}
}

@inproceedings{duvenaud2015convolutional,
  author    = {David K. Duvenaud and Dougal Maclaurin and Jorge Aguilera-Iparraguirre and Rafael G{\'o}mez-Bombarelli and Timothy Hirzel and Al{\'a}n Aspuru-Guzik and Ryan P. Adams},
  title     = {Convolutional Networks on Graphs for Learning Molecular Fingerprints},
  booktitle = {Advances in Neural Information Processing Systems},
  volume    = {28},
  pages     = {2224--2232},
  year      = {2015},
  url       = {https://arxiv.org/abs/1509.09292}
}

@inproceedings{defferrard2016convolutional,
  author    = {Micha{\"e}l Defferrard and Xavier Bresson and Pierre Vandergheynst},
  title     = {Convolutional Neural Networks on Graphs with Fast Localized Spectral Filtering},
  booktitle = {Advances in Neural Information Processing Systems},
  volume    = {29},
  pages     = {3837--3845},
  year      = {2016},
  url       = {https://proceedings.neurips.cc/paper/2016/hash/04df4d434d481c5bb723be1b6df1ee65-Abstract.html}
}

@inproceedings{li2016gated,
  author    = {Yujia Li and Daniel Tarlow and Marc Brockschmidt and Richard S. Zemel},
  title     = {Gated Graph Sequence Neural Networks},
  booktitle = {International Conference on Learning Representations},
  year      = {2016},
  url       = {https://arxiv.org/abs/1511.05493}
}

@inproceedings{yang2016planetoid,
  author        = {Zhilin Yang and William W. Cohen and Ruslan Salakhutdinov},
  title         = {Revisiting Semi-Supervised Learning with Graph Embeddings},
  booktitle     = {Proceedings of the 33rd International Conference on Machine Learning},
  series        = {Proceedings of Machine Learning Research},
  volume        = {48},
  pages         = {40--48},
  year          = {2016},
  eprint        = {1603.08861},
  archivePrefix = {arXiv},
  url           = {https://proceedings.mlr.press/v48/yanga16.html}
}

@article{bresson2017residual,
  author        = {Xavier Bresson and Thomas Laurent},
  title         = {Residual Gated Graph {ConvNets}},
  journal       = {CoRR},
  volume        = {abs/1711.07553},
  year          = {2017},
  eprint        = {1711.07553},
  archivePrefix = {arXiv},
  primaryClass  = {cs.LG},
  url           = {https://arxiv.org/abs/1711.07553}
}

@inproceedings{gilmer2017neural,
  author    = {Justin Gilmer and Samuel S. Schoenholz and Patrick F. Riley and Oriol Vinyals and George E. Dahl},
  title     = {Neural Message Passing for Quantum Chemistry},
  booktitle = {Proceedings of the 34th International Conference on Machine Learning},
  series    = {Proceedings of Machine Learning Research},
  volume    = {70},
  pages     = {1263--1272},
  publisher = {PMLR},
  year      = {2017},
  url       = {https://proceedings.mlr.press/v70/gilmer17a.html}
}

@inproceedings{hamilton2017inductive,
  author    = {William L. Hamilton and Rex Ying and Jure Leskovec},
  title     = {Inductive Representation Learning on Large Graphs},
  booktitle = {Advances in Neural Information Processing Systems},
  volume    = {30},
  pages     = {1024--1034},
  year      = {2017},
  url       = {https://papers.nips.cc/paper/2017/hash/5dd9db5e033da9c6fb5ba83c7a7ebea9-Abstract.html}
}

@inproceedings{khalil2017learning,
  author    = {Elias B. Khalil and Hanjun Dai and Yuyu Zhang and Bistra Dilkina and Le Song},
  title     = {Learning Combinatorial Optimization Algorithms over Graphs},
  booktitle = {Advances in Neural Information Processing Systems},
  volume    = {30},
  pages     = {6348--6358},
  year      = {2017},
  url       = {https://papers.nips.cc/paper/7214-learning-combinatorial-optimization-algorithms-over-graphs}
}

@inproceedings{kipf2017semi,
  author        = {Thomas N. Kipf and Max Welling},
  title         = {Semi-Supervised Classification with Graph Convolutional Networks},
  booktitle     = {International Conference on Learning Representations},
  year          = {2017},
  eprint        = {1609.02907},
  archivePrefix = {arXiv},
  url           = {https://openreview.net/forum?id=SJU4ayYgl}
}

@inproceedings{monti2017geometric,
  author    = {Federico Monti and Davide Boscaini and Jonathan Masci and Emanuele Rodol{\`a} and Jan Svoboda and Michael M. Bronstein},
  title     = {Geometric Deep Learning on Graphs and Manifolds Using Mixture Model {CNNs}},
  booktitle = {Proceedings of the {IEEE} Conference on Computer Vision and Pattern Recognition ({CVPR})},
  pages     = {5425--5434},
  year      = {2017},
  doi       = {10.1109/CVPR.2017.576},
  url       = {https://doi.org/10.1109/CVPR.2017.576}
}

@inproceedings{schutt2017schnet,
  author    = {Kristof T. Sch{\"u}tt and Pieter-Jan Kindermans and Huziel Enoc Sauceda Felix and Stefan Chmiela and Alexandre Tkatchenko and Klaus-Robert M{\"u}ller},
  title     = {{SchNet}: A Continuous-Filter Convolutional Neural Network for Modeling Quantum Interactions},
  booktitle = {Advances in Neural Information Processing Systems},
  volume    = {30},
  pages     = {991--1001},
  year      = {2017},
  url       = {https://papers.nips.cc/paper/6700-schnet-a-continuous-filter-convolutional-neural-network-for-modeling-quantum-interactions}
}

@inproceedings{simonovsky2017dynamic,
  author    = {Martin Simonovsky and Nikos Komodakis},
  title     = {Dynamic Edge-Conditioned Filters in Convolutional Neural Networks on Graphs},
  booktitle = {Proceedings of the {IEEE} Conference on Computer Vision and Pattern Recognition ({CVPR})},
  pages     = {29--38},
  year      = {2017},
  doi       = {10.1109/CVPR.2017.11},
  url       = {https://arxiv.org/abs/1704.02901}
}

@article{battaglia2018relational,
  author        = {Peter W. Battaglia and Jessica B. Hamrick and Victor Bapst and Alvaro Sanchez-Gonzalez and Vinicius Zambaldi and Mateusz Malinowski and Andrea Tacchetti and David Raposo and Adam Santoro and Ryan Faulkner and Caglar Gulcehre and Francis Song and Andrew Ballard and Justin Gilmer and George Dahl and Ashish Vaswani and Kelsey Allen and Charles Nash and Victoria Langston and Chris Dyer and Nicolas Heess and Daan Wierstra and Pushmeet Kohli and Matt Botvinick and Oriol Vinyals and Yujia Li and Razvan Pascanu},
  title         = {Relational inductive biases, deep learning, and graph networks},
  journal       = {CoRR},
  volume        = {abs/1806.01261},
  year          = {2018},
  eprint        = {1806.01261},
  archivePrefix = {arXiv},
  url           = {https://arxiv.org/abs/1806.01261}
}

@inproceedings{chen2018fastgcn,
  author        = {Jie Chen and Tengfei Ma and Cao Xiao},
  title         = {{FastGCN}: Fast Learning with Graph Convolutional Networks via Importance Sampling},
  booktitle     = {International Conference on Learning Representations},
  year          = {2018},
  eprint        = {1801.10247},
  archivePrefix = {arXiv},
  url           = {https://openreview.net/forum?id=rytstxWAW}
}

@inproceedings{chen2018stochastic,
  author        = {Jianfei Chen and Jun Zhu and Le Song},
  title         = {Stochastic Training of Graph Convolutional Networks with Variance Reduction},
  booktitle     = {Proceedings of the 35th International Conference on Machine Learning},
  series        = {Proceedings of Machine Learning Research},
  volume        = {80},
  pages         = {942--950},
  publisher     = {PMLR},
  year          = {2018},
  eprint        = {1710.10568},
  archivePrefix = {arXiv},
  url           = {https://proceedings.mlr.press/v80/chen18p.html}
}

@inproceedings{li2018deeper,
  author        = {Qimai Li and Zhichao Han and Xiao-Ming Wu},
  title         = {Deeper Insights Into Graph Convolutional Networks for Semi-Supervised Learning},
  booktitle     = {Proceedings of the AAAI Conference on Artificial Intelligence},
  volume        = {32},
  pages         = {3538--3545},
  year          = {2018},
  doi           = {10.1609/aaai.v32i1.11604},
  eprint        = {1801.07606},
  archivePrefix = {arXiv},
  url           = {https://ojs.aaai.org/index.php/AAAI/article/view/11604}
}

@inproceedings{schlichtkrull2018modeling,
  author        = {Michael Sejr Schlichtkrull and Thomas N. Kipf and Peter Bloem and Rianne van den Berg and Ivan Titov and Max Welling},
  title         = {Modeling Relational Data with Graph Convolutional Networks},
  booktitle     = {European semantic web conference},
  series        = {Lecture Notes in Computer Science},
  volume        = {10843},
  pages         = {593--607},
  publisher     = {Springer},
  year          = {2018},
  doi           = {10.1007/978-3-319-93417-4_38},
  eprint        = {1703.06103},
  archivePrefix = {arXiv}
}

@inproceedings{shchur2018pitfalls,
  author        = {Oleksandr Shchur and Maximilian Mumme and Aleksandar Bojchevski and Stephan G{\"u}nnemann},
  title         = {Pitfalls of Graph Neural Network Evaluation},
  booktitle     = {Relational Representation Learning Workshop, NeurIPS},
  year          = {2018},
  eprint        = {1811.05868},
  archivePrefix = {arXiv},
  url           = {https://arxiv.org/abs/1811.05868}
}

@article{thekumparampil2018attention,
  author        = {Kiran K. Thekumparampil and Chong Wang and Sewoong Oh and Li-Jia Li},
  title         = {Attention-based Graph Neural Network for Semi-supervised Learning},
  journal       = {CoRR},
  volume        = {abs/1803.03735},
  year          = {2018},
  eprint        = {1803.03735},
  archivePrefix = {arXiv},
  url           = {https://arxiv.org/abs/1803.03735}
}

@inproceedings{velickovic2018gat,
  author        = {Petar Veli{\v{c}}kovi{\'c} and Guillem Cucurull and Arantxa Casanova and Adriana Romero and Pietro Li{\`o} and Yoshua Bengio},
  title         = {Graph Attention Networks},
  booktitle     = {International Conference on Learning Representations},
  year          = {2018},
  eprint        = {1710.10903},
  archivePrefix = {arXiv},
  url           = {https://openreview.net/forum?id=rJXMpikCZ}
}

@inproceedings{xu2018representation,
  author        = {Keyulu Xu and Chengtao Li and Yonglong Tian and Tomohiro Sonobe and Ken-ichi Kawarabayashi and Stefanie Jegelka},
  title         = {Representation Learning on Graphs with Jumping Knowledge Networks},
  booktitle     = {Proceedings of the 35th International Conference on Machine Learning},
  series        = {Proceedings of Machine Learning Research},
  volume        = {80},
  pages         = {5453--5462},
  publisher     = {PMLR},
  year          = {2018},
  eprint        = {1806.03536},
  archivePrefix = {arXiv},
  url           = {https://proceedings.mlr.press/v80/xu18c.html}
}

@inproceedings{ying2018hierarchical,
  author        = {Rex Ying and Jiaxuan You and Christopher Morris and Xiang Ren and William L. Hamilton and Jure Leskovec},
  title         = {Hierarchical Graph Representation Learning with Differentiable Pooling},
  booktitle     = {Advances in Neural Information Processing Systems},
  volume        = {31},
  pages         = {4805--4815},
  year          = {2018},
  eprint        = {1806.08804},
  archivePrefix = {arXiv},
  url           = {https://proceedings.neurips.cc/paper/2018/hash/e77dbaf6759253c7c6d0efc5690369c7-Abstract.html}
}

@inproceedings{ying2018pinsage,
  author    = {Rex Ying and Ruining He and Kaifeng Chen and Pong Eksombatchai and William L. Hamilton and Jure Leskovec},
  title     = {Graph Convolutional Neural Networks for Web-Scale Recommender Systems},
  booktitle = {Proceedings of the 24th ACM SIGKDD International Conference on Knowledge Discovery and Data Mining},
  pages     = {974--983},
  publisher = {ACM},
  year      = {2018},
  doi       = {10.1145/3219819.3219890}
}

@inproceedings{zhang2018gaan,
  author        = {Jiani Zhang and Xingjian Shi and Junyuan Xie and Hao Ma and Irwin King and Dit-Yan Yeung},
  title         = {{GaAN}: Gated Attention Networks for Learning on Large and Spatiotemporal Graphs},
  booktitle     = {Proceedings of the Thirty-Fourth Conference on Uncertainty in Artificial Intelligence},
  pages         = {339--349},
  year          = {2018},
  eprint        = {1803.07294},
  archivePrefix = {arXiv},
  url           = {https://www.auai.org/uai2018/proceedings/papers/139.pdf}
}

@article{busbridge2019relational,
  author        = {Dan Busbridge and Dane Sherburn and Pietro Cavallo and Nils Y. Hammerla},
  title         = {Relational Graph Attention Networks},
  journal       = {CoRR},
  volume        = {abs/1904.05811},
  year          = {2019},
  eprint        = {1904.05811},
  archivePrefix = {arXiv},
  url           = {https://arxiv.org/abs/1904.05811}
}

@inproceedings{chen2019ring,
  author        = {Zhengdao Chen and Soledad Villar and Lei Chen and Joan Bruna},
  title         = {On the Equivalence between Graph Isomorphism Testing and Function Approximation with {GNNs}},
  booktitle     = {Advances in Neural Information Processing Systems},
  volume        = {32},
  pages         = {15868--15876},
  year          = {2019},
  eprint        = {1905.12560},
  archivePrefix = {arXiv},
  url           = {https://proceedings.neurips.cc/paper/2019/hash/71ee911dd06428a96c143a0b135041a4-Abstract.html}
}

@inproceedings{chiang2019cluster,
  author        = {Wei-Lin Chiang and Xuanqing Liu and Si Si and Yang Li and Samy Bengio and Cho-Jui Hsieh},
  title         = {{Cluster-GCN}: An Efficient Algorithm for Training Deep and Large Graph Convolutional Networks},
  booktitle     = {Proceedings of the 25th ACM SIGKDD International Conference on Knowledge Discovery and Data Mining},
  pages         = {257--266},
  publisher     = {ACM},
  year          = {2019},
  doi           = {10.1145/3292500.3330925},
  eprint        = {1905.07953},
  archivePrefix = {arXiv},
  url           = {https://dl.acm.org/doi/10.1145/3292500.3330925}
}

@inproceedings{feng2019hypergraph,
  author        = {Yifan Feng and Haoxuan You and Zizhao Zhang and Rongrong Ji and Yue Gao},
  title         = {Hypergraph Neural Networks},
  booktitle     = {Proceedings of the AAAI Conference on Artificial Intelligence},
  volume        = {33},
  pages         = {3558--3565},
  year          = {2019},
  doi           = {10.1609/aaai.v33i01.33013558},
  eprint        = {1809.09401},
  archivePrefix = {arXiv},
  url           = {https://ojs.aaai.org/index.php/AAAI/article/view/4235}
}

@inproceedings{gao2019graph,
  author        = {Hongyang Gao and Shuiwang Ji},
  title         = {Graph {U-Nets}},
  booktitle     = {Proceedings of the 36th International Conference on Machine Learning},
  series        = {Proceedings of Machine Learning Research},
  volume        = {97},
  pages         = {2083--2092},
  publisher     = {PMLR},
  year          = {2019},
  eprint        = {1905.05178},
  archivePrefix = {arXiv},
  url           = {https://proceedings.mlr.press/v97/gao19a.html}
}

@inproceedings{klicpera2019predict,
  author        = {Johannes Gasteiger and Aleksandar Bojchevski and Stephan G{\"u}nnemann},
  title         = {Predict then Propagate: Graph Neural Networks meet Personalized {PageRank}},
  booktitle     = {International Conference on Learning Representations},
  year          = {2019},
  eprint        = {1810.05997},
  archivePrefix = {arXiv},
  url           = {https://openreview.net/forum?id=H1gL-2A9Ym}
}

@inproceedings{lee2019self,
  author        = {Junhyun Lee and Inyeop Lee and Jaewoo Kang},
  title         = {Self-Attention Graph Pooling},
  booktitle     = {Proceedings of the 36th International Conference on Machine Learning},
  series        = {Proceedings of Machine Learning Research},
  volume        = {97},
  pages         = {3734--3743},
  publisher     = {PMLR},
  year          = {2019},
  eprint        = {1904.08082},
  archivePrefix = {arXiv},
  url           = {https://proceedings.mlr.press/v97/lee19c.html}
}

@article{levie2019cayleynets,
  author        = {Ron Levie and Federico Monti and Xavier Bresson and Michael M. Bronstein},
  title         = {{CayleyNets}: Graph Convolutional Neural Networks with Complex Rational Spectral Filters},
  journal       = {IEEE Transactions on Signal Processing},
  volume        = {67},
  number        = {1},
  pages         = {97--109},
  year          = {2019},
  doi           = {10.1109/TSP.2018.2879624},
  eprint        = {1705.07664},
  archivePrefix = {arXiv},
  url           = {https://ieeexplore.ieee.org/document/8521593/}
}

@inproceedings{liu2019geniepath,
  author        = {Ziqi Liu and Chaochao Chen and Longfei Li and Jun Zhou and Xiaolong Li and Le Song and Yuan Qi},
  title         = {{GeniePath}: Graph Neural Networks with Adaptive Receptive Paths},
  booktitle     = {Proceedings of the AAAI Conference on Artificial Intelligence},
  volume        = {33},
  pages         = {4424--4431},
  year          = {2019},
  doi           = {10.1609/aaai.v33i01.33014424},
  eprint        = {1802.00910},
  archivePrefix = {arXiv},
  url           = {https://ojs.aaai.org/index.php/AAAI/article/view/4354}
}

@inproceedings{maron2019invariant,
  author        = {Haggai Maron and Heli Ben-Hamu and Nadav Shamir and Yaron Lipman},
  title         = {Invariant and Equivariant Graph Networks},
  booktitle     = {International Conference on Learning Representations},
  year          = {2019},
  eprint        = {1812.09902},
  archivePrefix = {arXiv},
  url           = {https://openreview.net/forum?id=Syx72jC9tm}
}

@inproceedings{maron2019provably,
  author        = {Haggai Maron and Heli Ben-Hamu and Hadar Serviansky and Yaron Lipman},
  title         = {Provably Powerful Graph Networks},
  booktitle     = {Advances in Neural Information Processing Systems},
  volume        = {32},
  pages         = {2153--2164},
  year          = {2019},
  eprint        = {1905.11136},
  archivePrefix = {arXiv},
  url           = {https://proceedings.neurips.cc/paper/2019/hash/bb04af0f7ecaee4aae62035497da1387-Abstract.html}
}

@inproceedings{morris2019weisfeiler,
  author        = {Christopher Morris and Martin Ritzert and Matthias Fey and William L. Hamilton and Jan Eric Lenssen and Gaurav Rattan and Martin Grohe},
  title         = {Weisfeiler and {Leman} Go Neural: Higher-Order Graph Neural Networks},
  booktitle     = {Proceedings of the AAAI Conference on Artificial Intelligence},
  volume        = {33},
  pages         = {4602--4609},
  year          = {2019},
  doi           = {10.1609/aaai.v33i01.33014602},
  eprint        = {1810.02244},
  archivePrefix = {arXiv},
  url           = {https://ojs.aaai.org/index.php/AAAI/article/view/4384}
}

@inproceedings{murphy2019relational,
  author        = {Ryan L. Murphy and Balasubramaniam Srinivasan and Vinayak A. Rao and Bruno Ribeiro},
  title         = {Relational Pooling for Graph Representations},
  booktitle     = {Proceedings of the 36th International Conference on Machine Learning},
  series        = {Proceedings of Machine Learning Research},
  volume        = {97},
  pages         = {4663--4673},
  publisher     = {PMLR},
  year          = {2019},
  eprint        = {1903.02541},
  archivePrefix = {arXiv},
  url           = {https://proceedings.mlr.press/v97/murphy19a.html}
}

@misc{nt2019revisiting,
  author        = {Hoang NT and Takanori Maehara},
  title         = {Revisiting Graph Neural Networks: All We Have is Low-Pass Filters},
  year          = {2019},
  eprint        = {1905.09550},
  archivePrefix = {arXiv},
  url           = {https://arxiv.org/abs/1905.09550}
}

@inproceedings{wang2019daegc,
  author        = {Chun Wang and Shirui Pan and Ruiqi Hu and Guodong Long and Jing Jiang and Chengqi Zhang},
  title         = {Attributed Graph Clustering: A Deep Attentional Embedding Approach},
  booktitle     = {Proceedings of the Twenty-Eighth International Joint Conference on Artificial Intelligence},
  pages         = {3670--3676},
  year          = {2019},
  doi           = {10.24963/ijcai.2019/509},
  eprint        = {1906.06532},
  archivePrefix = {arXiv},
  url           = {https://www.ijcai.org/proceedings/2019/509}
}

@inproceedings{wang2019heterogeneous,
  author        = {Xiao Wang and Houye Ji and Chuan Shi and Bai Wang and Peng Cui and Philip S. Yu and Yanfang Ye},
  title         = {Heterogeneous Graph Attention Network},
  booktitle     = {The World Wide Web Conference},
  pages         = {2022--2032},
  publisher     = {ACM},
  year          = {2019},
  doi           = {10.1145/3308558.3313562},
  eprint        = {1903.07293},
  archivePrefix = {arXiv},
  url           = {https://doi.org/10.1145/3308558.3313562}
}

@inproceedings{wu2019simplifying,
  author        = {Felix Wu and Amauri H. Souza and Tianyi Zhang and Christopher Fifty and Tao Yu and Kilian Q. Weinberger},
  title         = {Simplifying Graph Convolutional Networks},
  booktitle     = {Proceedings of the 36th International Conference on Machine Learning},
  series        = {Proceedings of Machine Learning Research},
  volume        = {97},
  pages         = {6861--6871},
  publisher     = {PMLR},
  year          = {2019},
  eprint        = {1902.07153},
  archivePrefix = {arXiv},
  url           = {https://proceedings.mlr.press/v97/wu19e.html}
}

@inproceedings{xu2019graph,
  author        = {Bingbing Xu and Huawei Shen and Qi Cao and Yunqi Qiu and Xueqi Cheng},
  title         = {Graph Wavelet Neural Network},
  booktitle     = {International Conference on Learning Representations},
  year          = {2019},
  eprint        = {1904.07785},
  archivePrefix = {arXiv},
  url           = {https://openreview.net/forum?id=H1ewdiR5tQ}
}

@inproceedings{xu2019powerful,
  author        = {Keyulu Xu and Weihua Hu and Jure Leskovec and Stefanie Jegelka},
  title         = {How Powerful are Graph Neural Networks?},
  booktitle     = {International Conference on Learning Representations},
  year          = {2019},
  eprint        = {1810.00826},
  archivePrefix = {arXiv},
  url           = {https://openreview.net/forum?id=ryGs6iA5Km}
}

@inproceedings{yadati2019hypergcn,
  author        = {Naganand Yadati and Madhav Nimishakavi and Prateek Yadav and Vikram Nitin and Anand Louis and Partha P. Talukdar},
  title         = {{HyperGCN}: A New Method For Training Graph Convolutional Networks on Hypergraphs},
  booktitle     = {Advances in Neural Information Processing Systems},
  volume        = {32},
  pages         = {1509--1520},
  year          = {2019},
  eprint        = {1809.02589},
  archivePrefix = {arXiv},
  url           = {https://proceedings.neurips.cc/paper/2019/hash/1efa39bcaec6f3900149160693694536-Abstract.html}
}

@inproceedings{zou2019ladies,
  author        = {Difan Zou and Ziniu Hu and Yewen Wang and Song Jiang and Yizhou Sun and Quanquan Gu},
  title         = {Layer-Dependent Importance Sampling for Training Deep and Large Graph Convolutional Networks},
  booktitle     = {Advances in Neural Information Processing Systems},
  volume        = {32},
  pages         = {11247--11256},
  year          = {2019},
  eprint        = {1911.07323},
  archivePrefix = {arXiv},
  url           = {https://proceedings.neurips.cc/paper/2019/hash/91ba4a4478a66bee9812b0804b6f9d1b-Abstract.html}
}

@inproceedings{bianchi2020mincut,
  author    = {Filippo Maria Bianchi and Daniele Grattarola and Cesare Alippi},
  title     = {Spectral Clustering with Graph Neural Networks for Graph Pooling},
  booktitle = {Proceedings of the 37th International Conference on Machine Learning},
  series    = {Proceedings of Machine Learning Research},
  volume    = {119},
  pages     = {874--883},
  publisher = {PMLR},
  year      = {2020},
  url       = {https://proceedings.mlr.press/v119/bianchi20a.html}
}

@misc{cai2020note,
  author        = {Chen Cai and Yusu Wang},
  title         = {A Note on Over-Smoothing for Graph Neural Networks},
  year          = {2020},
  eprint        = {2006.13318},
  archivePrefix = {arXiv},
  url           = {https://arxiv.org/abs/2006.13318},
  note          = {Presented at the ICML 2020 Workshop on Graph Representation Learning and Beyond}
}

@inproceedings{chen2020measuring,
  author        = {Deli Chen and Yankai Lin and Wei Li and Peng Li and Jie Zhou and Xu Sun},
  title         = {Measuring and Relieving the Over-Smoothing Problem for Graph Neural Networks from the Topological View},
  booktitle     = {Proceedings of the AAAI Conference on Artificial Intelligence},
  volume        = {34},
  pages         = {3438--3445},
  year          = {2020},
  doi           = {10.1609/aaai.v34i04.5747},
  eprint        = {1909.03211},
  archivePrefix = {arXiv},
  url           = {https://ojs.aaai.org/index.php/AAAI/article/view/5747}
}

@inproceedings{chen2020simple,
  author        = {Ming Chen and Zhewei Wei and Zengfeng Huang and Bolin Ding and Yaliang Li},
  title         = {Simple and Deep Graph Convolutional Networks},
  booktitle     = {Proceedings of the 37th International Conference on Machine Learning},
  series        = {Proceedings of Machine Learning Research},
  volume        = {119},
  pages         = {1725--1735},
  publisher     = {PMLR},
  year          = {2020},
  eprint        = {2007.02133},
  archivePrefix = {arXiv},
  url           = {https://proceedings.mlr.press/v119/chen20v.html}
}

@inproceedings{chen2020substructures,
  author        = {Zhengdao Chen and Lei Chen and Soledad Villar and Joan Bruna},
  title         = {Can Graph Neural Networks Count Substructures?},
  booktitle     = {Advances in Neural Information Processing Systems},
  volume        = {33},
  pages         = {10383--10395},
  year          = {2020},
  eprint        = {2002.04025},
  archivePrefix = {arXiv},
  url           = {https://proceedings.neurips.cc/paper_files/paper/2020/hash/75877cb75154206c4e65e76b88a12712-Abstract.html}
}

@inproceedings{corso2020principal,
  author        = {Gabriele Corso and Luca Cavalleri and Dominique Beaini and Pietro Li{\`o} and Petar Veli{\v{c}}kovi{\'c}},
  title         = {Principal Neighbourhood Aggregation for Graph Nets},
  booktitle     = {Advances in Neural Information Processing Systems},
  volume        = {33},
  pages         = {13260--13271},
  year          = {2020},
  eprint        = {2004.05718},
  archivePrefix = {arXiv},
  url           = {https://proceedings.neurips.cc/paper/2020/hash/99cad265a1768cc2dd013f0e740300ae-Abstract.html}
}

@inproceedings{errica2020fair,
  author        = {Federico Errica and Marco Podda and Davide Bacciu and Alessio Micheli},
  title         = {A Fair Comparison of Graph Neural Networks for Graph Classification},
  booktitle     = {International Conference on Learning Representations},
  year          = {2020},
  eprint        = {1912.09893},
  archivePrefix = {arXiv},
  primaryClass  = {cs.LG},
  url           = {https://openreview.net/forum?id=HygDF6NFPB}
}

@inproceedings{fuchs2020se3,
  author        = {Fabian B. Fuchs and Daniel E. Worrall and Volker Fischer and Max Welling},
  title         = {{SE(3)}-Transformers: {3D} Roto-Translation Equivariant Attention Networks},
  booktitle     = {Advances in Neural Information Processing Systems},
  volume        = {33},
  pages         = {1970--1981},
  year          = {2020},
  eprint        = {2006.10503},
  archivePrefix = {arXiv},
  url           = {https://proceedings.neurips.cc/paper/2020/hash/15231a7ce4ba789d13b722cc5c955834-Abstract.html}
}

@inproceedings{garg2020generalization,
  author        = {Vikas K. Garg and Stefanie Jegelka and Tommi S. Jaakkola},
  title         = {Generalization and Representational Limits of Graph Neural Networks},
  booktitle     = {Proceedings of the 37th International Conference on Machine Learning},
  series        = {Proceedings of Machine Learning Research},
  volume        = {119},
  pages         = {3419--3430},
  publisher     = {PMLR},
  year          = {2020},
  eprint        = {2002.06157},
  archivePrefix = {arXiv},
  url           = {https://proceedings.mlr.press/v119/garg20c.html}
}

@inproceedings{gasteiger2020directional,
  author        = {Johannes Gasteiger and Janek Gro{\ss} and Stephan G{\"u}nnemann},
  title         = {Directional Message Passing for Molecular Graphs},
  booktitle     = {International Conference on Learning Representations},
  year          = {2020},
  eprint        = {2003.03123},
  archivePrefix = {arXiv},
  url           = {https://openreview.net/forum?id=B1eWbxStPH}
}

@inproceedings{gu2020implicit,
  author        = {Fangda Gu and Heng Chang and Wenwu Zhu and Somayeh Sojoudi and Laurent El Ghaoui},
  title         = {Implicit Graph Neural Networks},
  booktitle     = {Advances in Neural Information Processing Systems},
  volume        = {33},
  pages         = {11984--11995},
  year          = {2020},
  eprint        = {2009.06211},
  archivePrefix = {arXiv},
  url           = {https://proceedings.neurips.cc/paper/2020/hash/8b5c8441a8ff8e151b191c53c1842a38-Abstract.html}
}

@inproceedings{he2020lightgcn,
  author        = {Xiangnan He and Kuan Deng and Xiang Wang and Yan Li and Yongdong Zhang and Meng Wang},
  title         = {{LightGCN}: Simplifying and Powering Graph Convolution Network for Recommendation},
  booktitle     = {Proceedings of the 43rd International ACM SIGIR Conference on Research and Development in Information Retrieval},
  pages         = {639--648},
  publisher     = {ACM},
  year          = {2020},
  doi           = {10.1145/3397271.3401063},
  eprint        = {2002.02126},
  archivePrefix = {arXiv},
  url           = {https://dl.acm.org/doi/10.1145/3397271.3401063}
}

@inproceedings{hu2020heterogeneous,
  author        = {Ziniu Hu and Yuxiao Dong and Kuansan Wang and Yizhou Sun},
  title         = {Heterogeneous Graph Transformer},
  booktitle     = {Proceedings of The Web Conference 2020},
  pages         = {2704--2710},
  publisher     = {ACM},
  year          = {2020},
  doi           = {10.1145/3366423.3380027},
  eprint        = {2003.01332},
  archivePrefix = {arXiv},
  url           = {https://dl.acm.org/doi/10.1145/3366423.3380027}
}

@inproceedings{hu2020ogb,
  author        = {Weihua Hu and Matthias Fey and Marinka Zitnik and Yuxiao Dong and Hongyu Ren and Bowen Liu and Michele Catasta and Jure Leskovec},
  title         = {Open Graph Benchmark: Datasets for Machine Learning on Graphs},
  booktitle     = {Advances in Neural Information Processing Systems},
  volume        = {33},
  pages         = {22118--22133},
  year          = {2020},
  eprint        = {2005.00687},
  archivePrefix = {arXiv},
  primaryClass  = {cs.LG},
  url           = {https://proceedings.neurips.cc/paper/2020/hash/fb60d411a5c5b72b2e7d3527cfc84fd0-Abstract.html}
}

@article{li2020deepergcn,
  author        = {Guohao Li and Chenxin Xiong and Guocheng Qian and Ali Thabet and Bernard Ghanem},
  title         = {{DeeperGCN}: Training Deeper {GCN}s with Generalized Aggregation Functions},
  journal       = {IEEE Transactions on Pattern Analysis and Machine Intelligence},
  volume        = {45},
  number        = {11},
  pages         = {13024--13034},
  year          = {2023},
  doi           = {10.1109/TPAMI.2023.3306930},
  eprint        = {2006.07739},
  archivePrefix = {arXiv},
  url           = {https://arxiv.org/abs/2006.07739},
  note          = {Extended and retitled journal version of the 2020 preprint ``DeeperGCN: All You Need to Train Deeper GCNs'' (arXiv:2006.07739); adds Guocheng Qian as a co-author.}
}

@inproceedings{mesquita2020rethinking,
  author    = {Diego Mesquita and Amauri H. Souza and Samuel Kaski},
  title     = {Rethinking Pooling in Graph Neural Networks},
  booktitle = {Advances in Neural Information Processing Systems},
  volume    = {33},
  pages     = {2220--2231},
  year      = {2020},
  url       = {https://proceedings.neurips.cc/paper/2020/hash/1764183ef03fc7324eb58c3842bd9a57-Abstract.html}
}

@inproceedings{morris2020sparse,
  author        = {Christopher Morris and Gaurav Rattan and Petra Mutzel},
  title         = {Weisfeiler and {Leman} go sparse: Towards scalable higher-order graph embeddings},
  booktitle     = {Advances in Neural Information Processing Systems},
  volume        = {33},
  pages         = {21824--21840},
  year          = {2020},
  eprint        = {1904.01543},
  archivePrefix = {arXiv},
  url           = {https://proceedings.neurips.cc/paper/2020/hash/f81dee42585b3814de199b2e88757f5c-Abstract.html}
}

@misc{morris2020tudataset,
  author        = {Christopher Morris and Nils M. Kriege and Franka Bause and Kristian Kersting and Petra Mutzel and Marion Neumann},
  title         = {{TUDataset}: A Collection of Benchmark Datasets for Learning with Graphs},
  year          = {2020},
  doi           = {10.48550/arXiv.2007.08663},
  eprint        = {2007.08663},
  archivePrefix = {arXiv},
  primaryClass  = {cs.LG},
  url           = {https://arxiv.org/abs/2007.08663},
  note          = {Presented at the ICML 2020 Workshop on Graph Representation Learning and Beyond}
}

@inproceedings{oono2020expressive,
  author        = {Kenta Oono and Taiji Suzuki},
  title         = {Graph Neural Networks Exponentially Lose Expressive Power for Node Classification},
  booktitle     = {International Conference on Learning Representations},
  year          = {2020},
  eprint        = {1905.10947},
  archivePrefix = {arXiv},
  url           = {https://openreview.net/forum?id=S1ldO2EFPr}
}

@inproceedings{pei2020geom,
  author        = {Hongbin Pei and Bingzhe Wei and Kevin Chen-Chuan Chang and Yu Lei and Bo Yang},
  title         = {{Geom-GCN}: Geometric Graph Convolutional Networks},
  booktitle     = {International Conference on Learning Representations},
  year          = {2020},
  eprint        = {2002.05287},
  archivePrefix = {arXiv},
  url           = {https://openreview.net/forum?id=S1e2agrFvS}
}

@inproceedings{rong2020dropedge,
  author        = {Yu Rong and Wenbing Huang and Tingyang Xu and Junzhou Huang},
  title         = {{DropEdge}: Towards Deep Graph Convolutional Networks on Node Classification},
  booktitle     = {International Conference on Learning Representations},
  year          = {2020},
  eprint        = {1907.10903},
  archivePrefix = {arXiv},
  url           = {https://openreview.net/forum?id=Hkx1qkrKPr}
}

@article{rossi2020deep,
  author  = {Ryan A. Rossi and Rong Zhou and Nesreen K. Ahmed},
  title   = {Deep Inductive Graph Representation Learning},
  journal = {IEEE Transactions on Knowledge and Data Engineering},
  volume  = {32},
  number  = {3},
  pages   = {438--452},
  year    = {2020},
  doi     = {10.1109/TKDE.2018.2878247},
  url     = {https://doi.org/10.1109/TKDE.2018.2878247}
}

@inproceedings{rossi2020temporal,
  author        = {Emanuele Rossi and Benjamin Paul Chamberlain and Fabrizio Frasca and Davide Eynard and Federico Monti and Michael M. Bronstein},
  title         = {Temporal Graph Networks for Deep Learning on Dynamic Graphs},
  booktitle     = {ICML 2020 Workshop on Graph Representation Learning and Beyond},
  year          = {2020},
  eprint        = {2006.10637},
  archivePrefix = {arXiv},
  url           = {https://arxiv.org/abs/2006.10637},
  note          = {Workshop/arXiv paper; not main ICML proceedings.}
}

@inproceedings{vashishth2020compgcn,
  author        = {Shikhar Vashishth and Soumya Sanyal and Vikram Nitin and Partha P. Talukdar},
  title         = {Composition-based Multi-Relational Graph Convolutional Networks},
  booktitle     = {International Conference on Learning Representations},
  year          = {2020},
  eprint        = {1911.03082},
  archivePrefix = {arXiv},
  url           = {https://openreview.net/forum?id=BylA_C4tPr}
}

@inproceedings{xhonneux2020continuous,
  author        = {Louis-Pascal A. C. Xhonneux and Meng Qu and Jian Tang},
  title         = {Continuous Graph Neural Networks},
  booktitle     = {Proceedings of the 37th International Conference on Machine Learning},
  series        = {Proceedings of Machine Learning Research},
  volume        = {119},
  pages         = {10432--10441},
  publisher     = {PMLR},
  year          = {2020},
  eprint        = {1912.00967},
  archivePrefix = {arXiv},
  url           = {https://proceedings.mlr.press/v119/xhonneux20a.html}
}

@inproceedings{you2020design,
  author        = {Jiaxuan You and Rex Ying and Jure Leskovec},
  title         = {Design Space for Graph Neural Networks},
  booktitle     = {Advances in Neural Information Processing Systems},
  volume        = {33},
  pages         = {17009--17021},
  year          = {2020},
  eprint        = {2011.08843},
  archivePrefix = {arXiv},
  url           = {https://proceedings.neurips.cc/paper/2020/hash/c5c3d4fe6b2cc463c7d7ecba17cc9de7-Abstract.html}
}

@inproceedings{zeng2020graphsaint,
  author        = {Hanqing Zeng and Hongkuan Zhou and Ajitesh Srivastava and Rajgopal Kannan and Viktor Prasanna},
  title         = {{GraphSAINT}: Graph Sampling Based Inductive Learning Method},
  booktitle     = {International Conference on Learning Representations},
  year          = {2020},
  eprint        = {1907.04931},
  archivePrefix = {arXiv},
  url           = {https://openreview.net/forum?id=BJe8pkHFwS}
}

@inproceedings{zhang2020cardinality,
  author    = {Shuo Zhang and Lei Xie},
  title     = {Improving Attention Mechanism in Graph Neural Networks via Cardinality Preservation},
  booktitle = {Proceedings of the Twenty-Ninth International Joint Conference on Artificial Intelligence},
  pages     = {1395--1402},
  year      = {2020},
  doi       = {10.24963/ijcai.2020/194},
  url       = {https://www.ijcai.org/proceedings/2020/194}
}

@article{zhang2020graphbert,
  author        = {Jiawei Zhang and Haopeng Zhang and Congying Xia and Li Sun},
  title         = {{Graph-Bert}: Only Attention is Needed for Learning Graph Representations},
  journal       = {CoRR},
  volume        = {abs/2001.05140},
  year          = {2020},
  eprint        = {2001.05140},
  archivePrefix = {arXiv},
  url           = {https://arxiv.org/abs/2001.05140}
}

@inproceedings{zhao2020pairnorm,
  author        = {Lingxiao Zhao and Leman Akoglu},
  title         = {{PairNorm}: Tackling Oversmoothing in {GNNs}},
  booktitle     = {International Conference on Learning Representations},
  year          = {2020},
  eprint        = {1909.12223},
  archivePrefix = {arXiv},
  url           = {https://openreview.net/forum?id=rkecl1rtwB}
}

@inproceedings{zhu2020beyond,
  author        = {Jiong Zhu and Yujun Yan and Lingxiao Zhao and Mark Heimann and Leman Akoglu and Danai Koutra},
  title         = {Beyond Homophily in Graph Neural Networks: Current Limitations and Effective Designs},
  booktitle     = {Advances in Neural Information Processing Systems},
  volume        = {33},
  pages         = {7793--7804},
  year          = {2020},
  eprint        = {2006.11468},
  archivePrefix = {arXiv},
  url           = {https://proceedings.neurips.cc/paper/2020/hash/58ae23d878a47004366189884c2f8440-Abstract.html}
}

@inproceedings{Alon2021Bottleneck,
  author        = {Uri Alon and Eran Yahav},
  title         = {On the Bottleneck of Graph Neural Networks and its Practical Implications},
  booktitle     = {International Conference on Learning Representations},
  year          = {2021},
  eprint        = {2006.05205},
  archivePrefix = {arXiv},
  url           = {https://openreview.net/forum?id=i80OPhOCVH2}
}

@inproceedings{balcilar2021analyzing,
  author    = {Muhammet Balcilar and Guillaume Renton and Pierre H{\'e}roux and Benoit Ga{\"u}z{\`e}re and S{\'e}bastien Adam and Paul Honeine},
  title     = {Analyzing the Expressive Power of Graph Neural Networks in a Spectral Perspective},
  booktitle = {International Conference on Learning Representations},
  year      = {2021},
  url       = {https://openreview.net/forum?id=-qh0M9XWxnv}
}

@inproceedings{balcilar2021breaking,
  author        = {Muhammet Balcilar and Pierre H{\'e}roux and Benoit Ga{\"u}z{\`e}re and Pascal Vasseur and S{\'e}bastien Adam and Paul Honeine},
  title         = {Breaking the Limits of Message Passing Graph Neural Networks},
  booktitle     = {Proceedings of the 38th International Conference on Machine Learning},
  series        = {Proceedings of Machine Learning Research},
  volume        = {139},
  pages         = {599--608},
  publisher     = {PMLR},
  year          = {2021},
  eprint        = {2106.04319},
  archivePrefix = {arXiv},
  url           = {https://proceedings.mlr.press/v139/balcilar21a.html}
}

@inproceedings{Barcelo2021LocalGraphParams,
  author        = {Pablo Barcel{\'o} and Floris Geerts and Juan L. Reutter and Maksimilian Ryschkov},
  title         = {Graph Neural Networks with Local Graph Parameters},
  booktitle     = {Advances in Neural Information Processing Systems},
  volume        = {34},
  pages         = {25280--25293},
  year          = {2021},
  eprint        = {2106.06707},
  archivePrefix = {arXiv},
  url           = {https://proceedings.neurips.cc/paper/2021/hash/d4d8d1ac7e00e9105775a6b660dd3cbb-Abstract.html}
}

@inproceedings{Beaini2021DGN,
  author        = {Dominique Beaini and Saro Passaro and Vincent L{\'e}tourneau and William L. Hamilton and Gabriele Corso and Pietro Li{\`o}},
  title         = {Directional Graph Networks},
  booktitle     = {Proceedings of the 38th International Conference on Machine Learning},
  series        = {Proceedings of Machine Learning Research},
  volume        = {139},
  pages         = {748--758},
  publisher     = {PMLR},
  year          = {2021},
  eprint        = {2010.02863},
  archivePrefix = {arXiv},
  url           = {https://proceedings.mlr.press/v139/beaini21a.html}
}

@inproceedings{bo2021beyond,
  author        = {Deyu Bo and Xiao Wang and Chuan Shi and Huawei Shen},
  title         = {Beyond Low-frequency Information in Graph Convolutional Networks},
  booktitle     = {Proceedings of the AAAI Conference on Artificial Intelligence},
  volume        = {35},
  pages         = {3950--3957},
  year          = {2021},
  doi           = {10.1609/aaai.v35i5.16514},
  eprint        = {2101.00797},
  archivePrefix = {arXiv},
  url           = {https://ojs.aaai.org/index.php/AAAI/article/view/16514}
}

@inproceedings{Bodnar2021CWNetworks,
  author        = {Cristian Bodnar and Fabrizio Frasca and Nina Otter and Yuguang Wang and Pietro Li{\`o} and Guido F. Mont{\'u}far and Michael M. Bronstein},
  title         = {Weisfeiler and {Lehman} Go Cellular: {CW} Networks},
  booktitle     = {Advances in Neural Information Processing Systems},
  volume        = {34},
  pages         = {2625--2640},
  year          = {2021},
  eprint        = {2106.12575},
  archivePrefix = {arXiv},
  url           = {https://proceedings.neurips.cc/paper/2021/hash/157792e4abb490f99dbd738483e0d2d4-Abstract.html}
}

@inproceedings{bodnar2021simplicial,
  author        = {Cristian Bodnar and Fabrizio Frasca and Yuguang Wang and Nina Otter and Guido F. Mont{\'u}far and Pietro Li{\`o} and Michael M. Bronstein},
  title         = {Weisfeiler and {Lehman} Go Topological: Message Passing Simplicial Networks},
  booktitle     = {Proceedings of the 38th International Conference on Machine Learning},
  series        = {Proceedings of Machine Learning Research},
  volume        = {139},
  pages         = {1026--1037},
  publisher     = {PMLR},
  year          = {2021},
  eprint        = {2103.03212},
  archivePrefix = {arXiv},
  url           = {https://proceedings.mlr.press/v139/bodnar21a.html}
}

@inproceedings{Cai2021GraphNorm,
  author        = {Tianle Cai and Shengjie Luo and Keyulu Xu and Di He and Tie-Yan Liu and Liwei Wang},
  title         = {{GraphNorm}: A Principled Approach to Accelerating Graph Neural Network Training},
  booktitle     = {Proceedings of the 38th International Conference on Machine Learning},
  series        = {Proceedings of Machine Learning Research},
  volume        = {139},
  pages         = {1204--1215},
  publisher     = {PMLR},
  year          = {2021},
  eprint        = {2009.03294},
  archivePrefix = {arXiv},
  url           = {https://proceedings.mlr.press/v139/cai21e.html}
}

@inproceedings{Chamberlain2021BLEND,
  author        = {Benjamin Paul Chamberlain and James Rowbottom and Davide Eynard and Francesco Di Giovanni and Xiaowen Dong and Michael M. Bronstein},
  title         = {{Beltrami} Flow and Neural Diffusion on Graphs},
  booktitle     = {Advances in Neural Information Processing Systems},
  volume        = {34},
  pages         = {1594--1609},
  year          = {2021},
  eprint        = {2110.09443},
  archivePrefix = {arXiv},
  url           = {https://proceedings.neurips.cc/paper/2021/hash/0cbed40c0d920b94126eaf5e707be1f5-Abstract.html}
}

@inproceedings{Chamberlain2021GRAND,
  author        = {Benjamin Paul Chamberlain and James Rowbottom and Maria I. Gorinova and Michael M. Bronstein and Stefan Webb and Emanuele Rossi},
  title         = {{GRAND}: Graph Neural Diffusion},
  booktitle     = {Proceedings of the 38th International Conference on Machine Learning},
  series        = {Proceedings of Machine Learning Research},
  volume        = {139},
  pages         = {1407--1418},
  publisher     = {PMLR},
  year          = {2021},
  eprint        = {2106.10934},
  archivePrefix = {arXiv},
  url           = {https://proceedings.mlr.press/v139/chamberlain21a.html}
}

@inproceedings{chien2021adaptive,
  author        = {Eli Chien and Jianhao Peng and Pan Li and Olgica Milenkovic},
  title         = {Adaptive Universal Generalized {PageRank} Graph Neural Network},
  booktitle     = {International Conference on Learning Representations},
  year          = {2021},
  eprint        = {2006.07988},
  archivePrefix = {arXiv},
  url           = {https://openreview.net/forum?id=n6jl7fLxrP}
}

@inproceedings{cotta2021reconstruction,
  author        = {Leonardo Cotta and Christopher Morris and Bruno Ribeiro},
  title         = {Reconstruction for Powerful Graph Representations},
  booktitle     = {Advances in Neural Information Processing Systems},
  volume        = {34},
  pages         = {1713--1726},
  year          = {2021},
  eprint        = {2110.00577},
  archivePrefix = {arXiv},
  url           = {https://proceedings.neurips.cc/paper/2021/hash/0d8080853a54f8985276b0130266a657-Abstract.html}
}

@inproceedings{dwivedi2021generalization,
  author        = {Vijay Prakash Dwivedi and Xavier Bresson},
  title         = {A Generalization of {Transformer} Networks to Graphs},
  booktitle     = {AAAI Workshop on Deep Learning on Graphs: Methods and Applications},
  year          = {2021},
  eprint        = {2012.09699},
  archivePrefix = {arXiv},
  url           = {https://arxiv.org/abs/2012.09699},
  note          = {Workshop paper; not AAAI main proceedings}
}

@inproceedings{he2021bernnet,
  author        = {Mingguo He and Zhewei Wei and Zengfeng Huang and Hongteng Xu},
  title         = {{BernNet}: Learning Arbitrary Graph Spectral Filters via {Bernstein} Approximation},
  booktitle     = {Advances in Neural Information Processing Systems},
  volume        = {34},
  pages         = {14239--14251},
  year          = {2021},
  eprint        = {2106.10994},
  archivePrefix = {arXiv},
  url           = {https://proceedings.neurips.cc/paper/2021/hash/76f1cfd7754a6e4fc3281bcccb3d0902-Abstract.html}
}

@inproceedings{kreuzer2021rethinking,
  author        = {Devin Kreuzer and Dominique Beaini and William L. Hamilton and Vincent L{\'e}tourneau and Prudencio Tossou},
  title         = {Rethinking Graph {Transformers} with Spectral Attention},
  booktitle     = {Advances in Neural Information Processing Systems},
  volume        = {34},
  pages         = {21618--21629},
  year          = {2021},
  eprint        = {2106.03893},
  archivePrefix = {arXiv},
  url           = {https://proceedings.neurips.cc/paper/2021/hash/b4fd1d2cb085390fbbadae65e07876a7-Abstract.html}
}

@inproceedings{Li2021RevGNN,
  author        = {Guohao Li and Matthias M{\"u}ller and Bernard Ghanem and Vladlen Koltun},
  title         = {Training Graph Neural Networks with 1000 Layers},
  booktitle     = {Proceedings of the 38th International Conference on Machine Learning},
  series        = {Proceedings of Machine Learning Research},
  volume        = {139},
  pages         = {6437--6449},
  publisher     = {PMLR},
  year          = {2021},
  eprint        = {2106.07476},
  archivePrefix = {arXiv},
  url           = {https://proceedings.mlr.press/v139/li21o.html}
}

@inproceedings{liao2021pacbayesian,
  author        = {Renjie Liao and Raquel Urtasun and Richard Zemel},
  title         = {A {PAC}-{B}ayesian Approach to Generalization Bounds for Graph Neural Networks},
  booktitle     = {International Conference on Learning Representations},
  year          = {2021},
  eprint        = {2012.07690},
  archivePrefix = {arXiv},
  url           = {https://openreview.net/forum?id=TR-Nj6nFx42}
}

@inproceedings{lim2021linkx,
  author        = {Derek Lim and Felix Hohne and Xiuyu Li and Sijia Linda Huang and Vaishnavi Gupta and Omkar Bhalerao and Ser-Nam Lim},
  title         = {Large Scale Learning on Non-Homophilous Graphs: New Benchmarks and Strong Simple Methods},
  booktitle     = {Advances in Neural Information Processing Systems},
  volume        = {34},
  pages         = {20887--20902},
  year          = {2021},
  eprint        = {2110.14446},
  archivePrefix = {arXiv},
  url           = {https://proceedings.neurips.cc/paper/2021/hash/ae816a80e4c1c56caa2eb4e1819cbb2f-Abstract.html}
}

@inproceedings{liu2021eignn,
  author        = {Juncheng Liu and Kenji Kawaguchi and Bryan Hooi and Yiwei Wang and Xiaokui Xiao},
  title         = {{EIGNN}: Efficient Infinite-Depth Graph Neural Networks},
  booktitle     = {Advances in Neural Information Processing Systems},
  volume        = {34},
  pages         = {18762--18773},
  year          = {2021},
  eprint        = {2202.10720},
  archivePrefix = {arXiv},
  url           = {https://proceedings.neurips.cc/paper/2021/hash/9bd5ee6fe55aaeb673025dbcb8f939c1-Abstract.html}
}

@inproceedings{ma2021gatpos,
  author        = {Liheng Ma and Reihaneh Rabbany and Adriana Romero-Soriano},
  title         = {Graph Attention Networks with Positional Embeddings},
  booktitle     = {Advances in Knowledge Discovery and Data Mining},
  series        = {Lecture Notes in Computer Science},
  volume        = {12712},
  pages         = {514--527},
  publisher     = {Springer},
  year          = {2021},
  doi           = {10.1007/978-3-030-75762-5_41},
  eprint        = {2105.04037},
  archivePrefix = {arXiv},
  url           = {https://link.springer.com/chapter/10.1007/978-3-030-75762-5_41}
}

@inproceedings{Papp2021DropGNN,
  author        = {P{\'a}l Andr{\'a}s Papp and Karolis Martinkus and Lukas Faber and Roger Wattenhofer},
  title         = {{DropGNN}: Random Dropouts Increase the Expressiveness of Graph Neural Networks},
  booktitle     = {Advances in Neural Information Processing Systems},
  volume        = {34},
  pages         = {21997--22009},
  year          = {2021},
  eprint        = {2111.06283},
  archivePrefix = {arXiv},
  url           = {https://proceedings.neurips.cc/paper/2021/hash/b8b2926bd27d4307569ad119b6025f94-Abstract.html}
}

@article{rossi2021knowledge,
  author    = {Andrea Rossi and Denilson Barbosa and Donatella Firmani and Antonio Matinata and Paolo Merialdo},
  title     = {Knowledge Graph Embedding for Link Prediction: A Comparative Analysis},
  journal   = {ACM Transactions on Knowledge Discovery from Data},
  volume    = {15},
  number    = {2},
  pages     = {14:1--14:49},
  articleno = {14},
  year      = {2021},
  doi       = {10.1145/3424672},
  url       = {https://doi.org/10.1145/3424672}
}

@inproceedings{satorras2021en,
  author        = {V{\'i}ctor Garcia Satorras and Emiel Hoogeboom and Max Welling},
  title         = {{E(n)} Equivariant Graph Neural Networks},
  booktitle     = {Proceedings of the 38th International Conference on Machine Learning},
  series        = {Proceedings of Machine Learning Research},
  volume        = {139},
  pages         = {9323--9332},
  publisher     = {PMLR},
  year          = {2021},
  eprint        = {2102.09844},
  archivePrefix = {arXiv},
  url           = {https://proceedings.mlr.press/v139/satorras21a.html}
}

@inproceedings{thost2021directed,
  author        = {Veronika Thost and Jie Chen},
  title         = {Directed Acyclic Graph Neural Networks},
  booktitle     = {International Conference on Learning Representations},
  year          = {2021},
  eprint        = {2101.07965},
  archivePrefix = {arXiv},
  url           = {https://openreview.net/forum?id=JbuYF437WB6}
}

@inproceedings{wang2021magna,
  author        = {Guangtao Wang and Rex Ying and Jing Huang and Jure Leskovec},
  title         = {Multi-hop Attention Graph Neural Networks},
  booktitle     = {Proceedings of the Thirtieth International Joint Conference on Artificial Intelligence},
  pages         = {3089--3096},
  year          = {2021},
  doi           = {10.24963/ijcai.2021/425},
  eprint        = {2009.14332},
  archivePrefix = {arXiv},
  url           = {https://www.ijcai.org/proceedings/2021/425}
}

@inproceedings{yang2021diverse,
  author    = {Liang Yang and Mengzhe Li and Liyang Liu and Bingxin Niu and Chuan Wang and Xiaochun Cao and Yuanfang Guo},
  title     = {Diverse Message Passing for Attribute with Heterophily},
  booktitle = {Advances in Neural Information Processing Systems},
  volume    = {34},
  pages     = {4751--4763},
  year      = {2021},
  url       = {https://proceedings.neurips.cc/paper/2021/hash/253614bbac999b38b5b60cae531c4969-Abstract.html}
}

@inproceedings{yehudai2021size,
  author        = {Gilad Yehudai and Ethan Fetaya and Eli Meirom and Gal Chechik and Haggai Maron},
  title         = {From Local Structures to Size Generalization in Graph Neural Networks},
  booktitle     = {Proceedings of the 38th International Conference on Machine Learning},
  series        = {Proceedings of Machine Learning Research},
  volume        = {139},
  pages         = {11975--11986},
  publisher     = {PMLR},
  year          = {2021},
  eprint        = {2010.08853},
  archivePrefix = {arXiv},
  url           = {https://proceedings.mlr.press/v139/yehudai21a.html}
}

@inproceedings{ying2021transformers,
  author        = {Chengxuan Ying and Tianle Cai and Shengjie Luo and Shuxin Zheng and Guolin Ke and Di He and Yanming Shen and Tie-Yan Liu},
  title         = {Do Transformers Really Perform Badly for Graph Representation?},
  booktitle     = {Advances in Neural Information Processing Systems},
  volume        = {34},
  pages         = {28877--28888},
  year          = {2021},
  eprint        = {2106.05234},
  archivePrefix = {arXiv},
  url           = {https://proceedings.neurips.cc/paper/2021/hash/f1c1592588411002af340cbaedd6fc33-Abstract.html}
}

@inproceedings{you2021identity,
  author        = {Jiaxuan You and Jonathan M. Gomes-Selman and Rex Ying and Jure Leskovec},
  title         = {Identity-aware Graph Neural Networks},
  booktitle     = {Proceedings of the AAAI Conference on Artificial Intelligence},
  volume        = {35},
  number        = {12},
  pages         = {10737--10745},
  year          = {2021},
  doi           = {10.1609/aaai.v35i12.17283},
  eprint        = {2101.10320},
  archivePrefix = {arXiv},
  url           = {https://ojs.aaai.org/index.php/AAAI/article/view/17283}
}

@inproceedings{zhang2021nested,
  author        = {Muhan Zhang and Pan Li},
  title         = {Nested Graph Neural Networks},
  booktitle     = {Advances in Neural Information Processing Systems},
  volume        = {34},
  pages         = {15734--15747},
  year          = {2021},
  eprint        = {2110.13197},
  archivePrefix = {arXiv},
  url           = {https://proceedings.neurips.cc/paper/2021/hash/8462a7c229aea03dde69da754c3bbcc4-Abstract.html}
}

@inproceedings{bevilacqua2022equivariant,
  author        = {Beatrice Bevilacqua and Fabrizio Frasca and Derek Lim and Balasubramaniam Srinivasan and Chen Cai and Gopinath Balamurugan and Michael M. Bronstein and Haggai Maron},
  title         = {Equivariant Subgraph Aggregation Networks},
  booktitle     = {International Conference on Learning Representations},
  year          = {2022},
  eprint        = {2110.02910},
  archivePrefix = {arXiv},
  url           = {https://openreview.net/forum?id=dFbKQaRk15w}
}

@inproceedings{Bodnar2022DiagNSD,
  author        = {Cristian Bodnar and Francesco Di Giovanni and Benjamin Paul Chamberlain and Pietro Li{\`o} and Michael M. Bronstein},
  title         = {Neural Sheaf Diffusion: A Topological Perspective on Heterophily and Oversmoothing in {GNNs}},
  booktitle     = {Advances in Neural Information Processing Systems},
  volume        = {35},
  pages         = {18527--18541},
  year          = {2022},
  eprint        = {2202.04579},
  archivePrefix = {arXiv},
  url           = {https://proceedings.neurips.cc/paper_files/paper/2022/hash/75c45fca2aa416ada062b26cc4fb7641-Abstract-Conference.html}
}

@inproceedings{brody2022attentive,
  author        = {Shaked Brody and Uri Alon and Eran Yahav},
  title         = {How Attentive are Graph Attention Networks?},
  booktitle     = {International Conference on Learning Representations},
  year          = {2022},
  eprint        = {2105.14491},
  archivePrefix = {arXiv},
  url           = {https://openreview.net/forum?id=F72ximsx7C1}
}

@inproceedings{chen2022ciga,
  author        = {Yongqiang Chen and Yonggang Zhang and Yatao Bian and Han Yang and Kaili Ma and Binghui Xie and Tongliang Liu and Bo Han and James Cheng},
  title         = {Learning Causally Invariant Representations for Out-of-Distribution Generalization on Graphs},
  booktitle     = {Advances in Neural Information Processing Systems},
  volume        = {35},
  pages         = {22131--22148},
  year          = {2022},
  eprint        = {2202.05441},
  archivePrefix = {arXiv},
  url           = {https://proceedings.neurips.cc/paper_files/paper/2022/hash/8b21a7ea42cbcd1c29a7a88c444cce45-Abstract-Conference.html}
}

@inproceedings{chen2022optimization,
  author        = {Qi Chen and Yifei Wang and Yisen Wang and Jiansheng Yang and Zhouchen Lin},
  title         = {Optimization-Induced Graph Implicit Nonlinear Diffusion},
  booktitle     = {Proceedings of the 39th International Conference on Machine Learning},
  series        = {Proceedings of Machine Learning Research},
  volume        = {162},
  pages         = {3648--3661},
  publisher     = {PMLR},
  year          = {2022},
  eprint        = {2206.14418},
  archivePrefix = {arXiv},
  url           = {https://proceedings.mlr.press/v162/chen22z.html}
}

@inproceedings{chien2022allset,
  author        = {Eli Chien and Chao Pan and Jianhao Peng and Olgica Milenkovic},
  title         = {You are {AllSet}: A Multiset Function Framework for Hypergraph Neural Networks},
  booktitle     = {International Conference on Learning Representations},
  year          = {2022},
  eprint        = {2106.13264},
  archivePrefix = {arXiv},
  url           = {https://openreview.net/forum?id=hpBTIv2uy_E}
}

@inproceedings{duan2022largescalebenchmark,
  author        = {Keyu Duan and Zirui Liu and Peihao Wang and Wenqing Zheng and Kaixiong Zhou and Tianlong Chen and Xia Hu and Zhangyang Wang},
  title         = {A Comprehensive Study on Large-Scale Graph Training: Benchmarking and Rethinking},
  booktitle     = {Advances in Neural Information Processing Systems},
  volume        = {35},
  pages         = {5376--5389},
  year          = {2022},
  eprint        = {2210.07494},
  archivePrefix = {arXiv},
  url           = {https://proceedings.neurips.cc/paper_files/paper/2022/hash/23ee05bf1f4ade71c0f8f5ca722df601-Abstract-Datasets_and_Benchmarks.html},
  note          = {Datasets and Benchmarks Track}
}

@inproceedings{dwivedi2022graph,
  author        = {Vijay Prakash Dwivedi and Anh Tuan Luu and Thomas Laurent and Yoshua Bengio and Xavier Bresson},
  title         = {Graph Neural Networks with Learnable Structural and Positional Representations},
  booktitle     = {International Conference on Learning Representations},
  year          = {2022},
  eprint        = {2110.07875},
  archivePrefix = {arXiv},
  url           = {https://openreview.net/forum?id=wTTjnvGphYj}
}

@inproceedings{Dwivedi2022LRGB,
  author        = {Vijay Prakash Dwivedi and Ladislav Ramp{\'a}{\v{s}}ek and Mikhail Galkin and Ali Parviz and Guy Wolf and Anh Tuan Luu and Dominique Beaini},
  title         = {Long Range Graph Benchmark},
  booktitle     = {Advances in Neural Information Processing Systems},
  volume        = {35},
  pages         = {22326--22340},
  year          = {2022},
  doi           = {10.52202/068431-1622},
  eprint        = {2206.08164},
  archivePrefix = {arXiv},
  url           = {https://proceedings.neurips.cc/paper_files/paper/2022/hash/8c3c666820ea055a77726d66fc7d447f-Abstract-Datasets_and_Benchmarks.html},
  note          = {Datasets and Benchmarks Track}
}

@inproceedings{Frasca2022SubgraphSymmetries,
  author        = {Fabrizio Frasca and Beatrice Bevilacqua and Michael M. Bronstein and Haggai Maron},
  title         = {Understanding and Extending Subgraph {GNNs} by Rethinking Their Symmetries},
  booktitle     = {Advances in Neural Information Processing Systems},
  volume        = {35},
  pages         = {31376--31390},
  year          = {2022},
  doi           = {10.52202/068431-2275},
  eprint        = {2206.11140},
  archivePrefix = {arXiv},
  url           = {https://proceedings.neurips.cc/paper_files/paper/2022/hash/cb2a4cc70db72ea779abd01107782c7b-Abstract-Conference.html}
}

@inproceedings{gui2022good,
  author        = {Shurui Gui and Xiner Li and Limei Wang and Shuiwang Ji},
  title         = {{GOOD}: A Graph Out-of-Distribution Benchmark},
  booktitle     = {Advances in Neural Information Processing Systems},
  volume        = {35},
  pages         = {2059--2073},
  year          = {2022},
  doi           = {10.52202/068431-0150},
  eprint        = {2206.08452},
  archivePrefix = {arXiv},
  url           = {https://proceedings.neurips.cc/paper_files/paper/2022/hash/0dc91de822b71c66a7f54fa121d8cbb9-Abstract-Datasets_and_Benchmarks.html},
  note          = {Datasets and Benchmarks Track}
}

@inproceedings{he2022chebnetii,
  author        = {Mingguo He and Zhewei Wei and Ji-Rong Wen},
  title         = {Convolutional Neural Networks on Graphs with {Chebyshev} Approximation, Revisited},
  booktitle     = {Advances in Neural Information Processing Systems},
  volume        = {35},
  pages         = {7264--7276},
  year          = {2022},
  doi           = {10.52202/068431-0527},
  eprint        = {2202.03580},
  archivePrefix = {arXiv},
  url           = {https://proceedings.neurips.cc/paper_files/paper/2022/hash/2f9b3ee2bcea04b327c09d7e3145bd1e-Abstract-Conference.html}
}

@inproceedings{keriven2022nottoo,
  author        = {Nicolas Keriven},
  title         = {Not Too Little, Not Too Much: A Theoretical Analysis of Graph (Over)Smoothing},
  booktitle     = {Advances in Neural Information Processing Systems},
  volume        = {35},
  pages         = {2268--2281},
  year          = {2022},
  doi           = {10.52202/068431-0165},
  eprint        = {2205.12156},
  archivePrefix = {arXiv},
  url           = {https://proceedings.neurips.cc/paper_files/paper/2022/hash/0f956ca6f667c62e0f71511773c86a59-Abstract-Conference.html}
}

@article{liu2021non,
  author        = {Meng Liu and Zhengyang Wang and Shuiwang Ji},
  title         = {Non-Local Graph Neural Networks},
  journal       = {IEEE Transactions on Pattern Analysis and Machine Intelligence},
  volume        = {44},
  number        = {12},
  pages         = {10270--10276},
  year          = {2022},
  doi           = {10.1109/TPAMI.2021.3134200},
  eprint        = {2005.14612},
  archivePrefix = {arXiv},
  url           = {https://doi.org/10.1109/TPAMI.2021.3134200}
}

@inproceedings{luan2022revisiting,
  author        = {Sitao Luan and Chenqing Hua and Qincheng Lu and Jiaqi Zhu and Mingde Zhao and Shuyuan Zhang and Xiao-Wen Chang and Doina Precup},
  title         = {Revisiting Heterophily For Graph Neural Networks},
  booktitle     = {Advances in Neural Information Processing Systems},
  volume        = {35},
  pages         = {1362--1375},
  year          = {2022},
  doi           = {10.52202/068431-0100},
  eprint        = {2210.07606},
  archivePrefix = {arXiv},
  url           = {https://proceedings.neurips.cc/paper_files/paper/2022/hash/092359ce5cf60a80e882378944bf1be4-Abstract-Conference.html}
}

@inproceedings{Ma2022HomophilyNecessary,
  author        = {Yao Ma and Xiaorui Liu and Neil Shah and Jiliang Tang},
  title         = {Is Homophily a Necessity for Graph Neural Networks?},
  booktitle     = {International Conference on Learning Representations},
  year          = {2022},
  eprint        = {2106.06134},
  archivePrefix = {arXiv},
  url           = {https://openreview.net/forum?id=ucASPPD9GKN}
}

@inproceedings{morris2022speqnets,
  author        = {Christopher Morris and Gaurav Rattan and Sandra Kiefer and Siamak Ravanbakhsh},
  title         = {{SpeqNets}: Sparsity-aware Permutation-equivariant Graph Networks},
  booktitle     = {Proceedings of the 39th International Conference on Machine Learning},
  series        = {Proceedings of Machine Learning Research},
  volume        = {162},
  pages         = {16017--16042},
  publisher     = {PMLR},
  year          = {2022},
  eprint        = {2203.13913},
  archivePrefix = {arXiv},
  url           = {https://proceedings.mlr.press/v162/morris22a.html}
}

@inproceedings{obray2022evaluation,
  author        = {Leslie O'Bray and Max Horn and Bastian Rieck and Karsten Borgwardt},
  title         = {Evaluation Metrics for Graph Generative Models: Problems, Pitfalls, and Practical Solutions},
  booktitle     = {International Conference on Learning Representations},
  year          = {2022},
  eprint        = {2106.01098},
  archivePrefix = {arXiv},
  primaryClass  = {cs.LG},
  url           = {https://openreview.net/forum?id=tBtoZYKd9n}
}

@inproceedings{park2022deformable,
  author        = {Jinyoung Park and Sungdong Yoo and Jihwan Park and Hyunwoo J. Kim},
  title         = {Deformable Graph Convolutional Networks},
  booktitle     = {Proceedings of the AAAI Conference on Artificial Intelligence},
  volume        = {36},
  number        = {7},
  pages         = {7949--7956},
  year          = {2022},
  doi           = {10.1609/aaai.v36i7.20765},
  eprint        = {2112.14438},
  archivePrefix = {arXiv},
  url           = {https://ojs.aaai.org/index.php/AAAI/article/view/20765}
}

@inproceedings{poursafaei2022edgebank,
  author        = {Farimah Poursafaei and Shenyang Huang and Kellin Pelrine and Reihaneh Rabbany},
  title         = {Towards Better Evaluation for Dynamic Link Prediction},
  booktitle     = {Advances in Neural Information Processing Systems},
  volume        = {35},
  pages         = {32928--32941},
  year          = {2022},
  doi           = {10.52202/068431-2386},
  eprint        = {2207.10128},
  archivePrefix = {arXiv},
  url           = {https://proceedings.neurips.cc/paper_files/paper/2022/hash/d49042a5d49818711c401d34172f9900-Abstract-Datasets_and_Benchmarks.html},
  note          = {Datasets and Benchmarks Track}
}

@inproceedings{qian2022ordered,
  author        = {Chendi Qian and Gaurav Rattan and Floris Geerts and Mathias Niepert and Christopher Morris},
  title         = {Ordered Subgraph Aggregation Networks},
  booktitle     = {Advances in Neural Information Processing Systems},
  volume        = {35},
  pages         = {21030--21045},
  year          = {2022},
  doi           = {10.52202/068431-1529},
  eprint        = {2206.11168},
  archivePrefix = {arXiv},
  url           = {https://proceedings.neurips.cc/paper_files/paper/2022/hash/8471dc3f5180df17e2fa84f106f1ee8e-Abstract-Conference.html}
}

@inproceedings{qin2022nasbenchgraph,
  author        = {Yijian Qin and Ziwei Zhang and Xin Wang and Zeyang Zhang and Wenwu Zhu},
  title         = {{NAS-Bench-Graph}: Benchmarking Graph Neural Architecture Search},
  booktitle     = {Advances in Neural Information Processing Systems},
  volume        = {35},
  pages         = {54--69},
  year          = {2022},
  doi           = {10.52202/068431-0005},
  eprint        = {2206.09166},
  archivePrefix = {arXiv},
  url           = {https://proceedings.neurips.cc/paper_files/paper/2022/hash/004bed4e186fdd7ebb73aad6e97c2332-Abstract-Datasets_and_Benchmarks.html}
}

@inproceedings{rampavsek2022recipe,
  author        = {Ladislav Ramp{\'a}{\v{s}}ek and Mikhail Galkin and Vijay Prakash Dwivedi and Anh Tuan Luu and Guy Wolf and Dominique Beaini},
  title         = {Recipe for a General, Powerful, Scalable Graph Transformer},
  booktitle     = {Advances in Neural Information Processing Systems},
  volume        = {35},
  pages         = {14501--14515},
  year          = {2022},
  doi           = {10.52202/068431-1054},
  eprint        = {2205.12454},
  archivePrefix = {arXiv},
  url           = {https://proceedings.neurips.cc/paper_files/paper/2022/hash/5d4834a159f1547b267a05a4e2b7cf5e-Abstract-Conference.html}
}

@inproceedings{rusch2022graphcon,
  author        = {T. Konstantin Rusch and Benjamin P. Chamberlain and James Rowbottom and Siddhartha Mishra and Michael M. Bronstein},
  title         = {Graph-Coupled Oscillator Networks},
  booktitle     = {Proceedings of the 39th International Conference on Machine Learning},
  series        = {Proceedings of Machine Learning Research},
  volume        = {162},
  pages         = {18888--18909},
  publisher     = {PMLR},
  year          = {2022},
  eprint        = {2202.02296},
  archivePrefix = {arXiv},
  url           = {https://proceedings.mlr.press/v162/rusch22a.html}
}

@inproceedings{scholten2022interception,
  author        = {Yan Scholten and Jan Schuchardt and Simon Geisler and Aleksandar Bojchevski and Stephan G{\"u}nnemann},
  title         = {Randomized Message-Interception Smoothing: Gray-box Certificates for Graph Neural Networks},
  booktitle     = {Advances in Neural Information Processing Systems},
  volume        = {35},
  pages         = {33146--33158},
  year          = {2022},
  doi           = {10.52202/068431-2402},
  eprint        = {2301.02039},
  archivePrefix = {arXiv},
  primaryClass  = {cs.LG},
  url           = {https://proceedings.neurips.cc/paper_files/paper/2022/hash/d66d8164cfbf012cac2866edbb375035-Abstract-Conference.html}
}

@inproceedings{souza2022pint,
  author        = {Amauri H. Souza and Diego Mesquita and Samuel Kaski and Vikas Garg},
  title         = {Provably Expressive Temporal Graph Networks},
  booktitle     = {Advances in Neural Information Processing Systems},
  volume        = {35},
  pages         = {32257--32269},
  year          = {2022},
  doi           = {10.52202/068431-2337},
  eprint        = {2209.15059},
  archivePrefix = {arXiv},
  url           = {https://proceedings.neurips.cc/paper_files/paper/2022/hash/d029c97ee0db162c60f2ebc9cb93387e-Abstract-Conference.html}
}

@inproceedings{Topping2022Curvature,
  author        = {Jake Topping and Francesco Di Giovanni and Benjamin Paul Chamberlain and Xiaowen Dong and Michael M. Bronstein},
  title         = {Understanding Over-Squashing and Bottlenecks on Graphs via Curvature},
  booktitle     = {International Conference on Learning Representations},
  year          = {2022},
  eprint        = {2111.14522},
  archivePrefix = {arXiv},
  url           = {https://openreview.net/forum?id=7UmjRGzp-A}
}

@misc{wang2022automated,
  author        = {Xin Wang and Ziwei Zhang and Haoyang Li and Wenwu Zhu},
  title         = {Automated Graph Machine Learning: Approaches, Libraries, Benchmarks and Directions},
  year          = {2022},
  eprint        = {2201.01288},
  archivePrefix = {arXiv},
  url           = {https://arxiv.org/abs/2201.01288}
}

@inproceedings{wang2022how,
  author        = {Xiyuan Wang and Muhan Zhang},
  title         = {How Powerful are Spectral Graph Neural Networks},
  booktitle     = {Proceedings of the 39th International Conference on Machine Learning},
  series        = {Proceedings of Machine Learning Research},
  volume        = {162},
  pages         = {23341--23362},
  publisher     = {PMLR},
  year          = {2022},
  eprint        = {2205.11172},
  archivePrefix = {arXiv},
  url           = {https://proceedings.mlr.press/v162/wang22am.html}
}

@inproceedings{wijesinghe2022graphsnn,
  author    = {Asiri Wijesinghe and Qing Wang},
  title     = {A New Perspective on ``How Graph Neural Networks Go Beyond {Weisfeiler--Lehman}?''},
  booktitle = {International Conference on Learning Representations},
  year      = {2022},
  url       = {https://openreview.net/forum?id=uxgg9o7bI_3}
}

@article{wu2022gnnrec,
  author  = {Shiwen Wu and Fei Sun and Wentao Zhang and Xu Xie and Bin Cui},
  title   = {Graph Neural Networks in Recommender Systems: A Survey},
  journal = {ACM Computing Surveys},
  volume  = {55},
  number  = {5},
  pages   = {97:1--97:37},
  year    = {2023},
  doi     = {10.1145/3535101}
}

@inproceedings{Wu2022NodeFormer,
  author        = {Qitian Wu and Wentao Zhao and Zenan Li and David P. Wipf and Junchi Yan},
  title         = {{NodeFormer}: A Scalable Graph Structure Learning Transformer for Node Classification},
  booktitle     = {Advances in Neural Information Processing Systems},
  volume        = {35},
  pages         = {27387--27401},
  year          = {2022},
  doi           = {10.52202/068431-1986},
  eprint        = {2306.08385},
  archivePrefix = {arXiv},
  url           = {https://proceedings.neurips.cc/paper_files/paper/2022/hash/af790b7ae573771689438bbcfc5933fe-Abstract-Conference.html}
}

@inproceedings{yang2022graph,
  author        = {Tianmeng Yang and Yujing Wang and Zhihan Yue and Yaming Yang and Yunhai Tong and Jing Bai},
  title         = {Graph Pointer Neural Networks},
  booktitle     = {Proceedings of the AAAI Conference on Artificial Intelligence},
  volume        = {36},
  pages         = {8832--8839},
  year          = {2022},
  doi           = {10.1609/aaai.v36i8.20864},
  eprint        = {2110.00973},
  archivePrefix = {arXiv},
  url           = {https://ojs.aaai.org/index.php/AAAI/article/view/20864}
}

@inproceedings{zhao2022from,
  author        = {Lingxiao Zhao and Wei Jin and Leman Akoglu and Neil Shah},
  title         = {From Stars to Subgraphs: Uplifting Any {GNN} with Local Structure Awareness},
  booktitle     = {International Conference on Learning Representations},
  year          = {2022},
  eprint        = {2110.03753},
  archivePrefix = {arXiv},
  url           = {https://openreview.net/forum?id=Mspk_WYKoEH}
}

@inproceedings{zheng2022meta,
  author    = {Xin Zheng and Miao Zhang and Chunyang Chen and Chaojie Li and Chuan Zhou and Shirui Pan},
  title     = {Multi-Relational Graph Neural Architecture Search with Fine-grained Message Passing},
  booktitle = {2022 IEEE International Conference on Data Mining},
  pages     = {783--792},
  publisher = {IEEE},
  year      = {2022},
  doi       = {10.1109/ICDM54844.2022.00089}
}

@inproceedings{Zhu2022HeterophilyRobustness,
  author        = {Jiong Zhu and Junchen Jin and Donald Loveland and Michael T. Schaub and Danai Koutra},
  title         = {How Does Heterophily Impact the Robustness of Graph Neural Networks? Theoretical Connections and Practical Implications},
  booktitle     = {Proceedings of the 28th ACM SIGKDD Conference on Knowledge Discovery and Data Mining},
  pages         = {2637--2647},
  publisher     = {ACM},
  year          = {2022},
  doi           = {10.1145/3534678.3539418},
  eprint        = {2106.07767},
  archivePrefix = {arXiv},
  url           = {https://dl.acm.org/doi/10.1145/3534678.3539418}
}

@inproceedings{bianchilachi2023expressive,
  author        = {Filippo Maria Bianchi and Veronica Lachi},
  title         = {The Expressive Power of Pooling in Graph Neural Networks},
  booktitle     = {Advances in Neural Information Processing Systems},
  volume        = {36},
  pages         = {71603--71618},
  year          = {2023},
  doi           = {10.52202/075280-3135},
  eprint        = {2304.01575},
  archivePrefix = {arXiv},
  url           = {https://proceedings.neurips.cc/paper_files/paper/2023/hash/e26f31de8b13ec569bf507e6ae2cd952-Abstract-Conference.html}
}

@inproceedings{black2023resistance,
  author        = {Mitchell Black and Zhengchao Wan and Amir Nayyeri and Yusu Wang},
  title         = {Understanding Oversquashing in {GNNs} through the Lens of Effective Resistance},
  booktitle     = {Proceedings of the 40th International Conference on Machine Learning},
  series        = {Proceedings of Machine Learning Research},
  volume        = {202},
  pages         = {2528--2547},
  publisher     = {PMLR},
  year          = {2023},
  month         = {23--29 Jul},
  eprint        = {2302.06835},
  archivePrefix = {arXiv},
  url           = {https://proceedings.mlr.press/v202/black23a.html}
}

@article{bouritsas2023improving,
  author        = {Giorgos Bouritsas and Fabrizio Frasca and Stefanos Zafeiriou and Michael M. Bronstein},
  title         = {Improving Graph Neural Network Expressivity via Subgraph Isomorphism Counting},
  journal       = {IEEE Transactions on Pattern Analysis and Machine Intelligence},
  volume        = {45},
  number        = {1},
  pages         = {657--668},
  year          = {2023},
  doi           = {10.1109/TPAMI.2022.3154319},
  eprint        = {2006.09252},
  archivePrefix = {arXiv}
}

@inproceedings{Cai2023MPNNGTConnection,
  author        = {Chen Cai and Truong Son Hy and Rose Yu and Yusu Wang},
  title         = {On the Connection Between {MPNN} and Graph Transformer},
  booktitle     = {Proceedings of the 40th International Conference on Machine Learning},
  series        = {Proceedings of Machine Learning Research},
  volume        = {202},
  pages         = {3408--3430},
  publisher     = {PMLR},
  year          = {2023},
  eprint        = {2301.11956},
  archivePrefix = {arXiv},
  url           = {https://proceedings.mlr.press/v202/cai23b.html}
}

@article{Chen2023GDAMNs,
  author        = {Jie Chen and Shouzhen Chen and Mingyuan Bai and Jian Pu and Junping Zhang and Junbin Gao},
  title         = {Graph Decoupling Attention {Markov} Networks for Semisupervised Graph Node Classification},
  journal       = {IEEE Transactions on Neural Networks and Learning Systems},
  volume        = {34},
  number        = {12},
  pages         = {9859--9873},
  year          = {2023},
  doi           = {10.1109/TNNLS.2022.3161453},
  eprint        = {2104.13718},
  archivePrefix = {arXiv},
  url           = {https://ieeexplore.ieee.org/document/9744550}
}

@article{chen2024demystifyingsparsification,
  author        = {Yuhan Chen and Haojie Ye and Sanketh Vedula and Alex Bronstein and Ronald Dreslinski and Trevor Mudge and Nishil Talati},
  title         = {Demystifying Graph Sparsification Algorithms in Graph Properties Preservation},
  journal       = {Proceedings of the VLDB Endowment},
  volume        = {17},
  number        = {3},
  pages         = {427--440},
  year          = {2023},
  doi           = {10.14778/3632093.3632106},
  eprint        = {2311.12314},
  archivePrefix = {arXiv}
}

@inproceedings{choi2023gread,
  author        = {Jeongwhan Choi and Seoyoung Hong and Noseong Park and Sung-Bae Cho},
  title         = {{GREAD}: Graph Neural Reaction-Diffusion Networks},
  booktitle     = {Proceedings of the 40th International Conference on Machine Learning},
  series        = {Proceedings of Machine Learning Research},
  volume        = {202},
  pages         = {5722--5747},
  publisher     = {PMLR},
  year          = {2023},
  eprint        = {2211.14208},
  archivePrefix = {arXiv},
  url           = {https://proceedings.mlr.press/v202/choi23a.html}
}

@inproceedings{cong2023graphmixer,
  author        = {Weilin Cong and Si Zhang and Jian Kang and Baichuan Yuan and Hao Wu and Xin Zhou and Hanghang Tong and Mehrdad Mahdavi},
  title         = {Do We Really Need Complicated Model Architectures for Temporal Networks?},
  booktitle     = {International Conference on Learning Representations},
  year          = {2023},
  eprint        = {2302.11636},
  archivePrefix = {arXiv},
  url           = {https://openreview.net/forum?id=ayPPc0SyLv1}
}

@inproceedings{DiGiovanni2023OverSquashing,
  author        = {Francesco Di Giovanni and Lorenzo Giusti and Federico Barbero and Giulia Luise and Pietro Li{\`o} and Michael M. Bronstein},
  title         = {On Over-Squashing in Message Passing Neural Networks: The Impact of Width, Depth, and Topology},
  booktitle     = {Proceedings of the 40th International Conference on Machine Learning},
  series        = {Proceedings of Machine Learning Research},
  volume        = {202},
  pages         = {7865--7885},
  publisher     = {PMLR},
  year          = {2023},
  eprint        = {2302.02941},
  archivePrefix = {arXiv},
  url           = {https://proceedings.mlr.press/v202/di-giovanni23a.html}
}

@article{digiovanni2023understanding,
  author        = {Francesco Di Giovanni and James Rowbottom and Benjamin P. Chamberlain and Thomas Markovich and Michael M. Bronstein},
  title         = {Understanding Convolution on Graphs via Energies},
  journal       = {Transactions on Machine Learning Research},
  year          = {2023},
  eprint        = {2206.10991},
  archivePrefix = {arXiv},
  url           = {https://openreview.net/forum?id=v5ew3FPTgb}
}

@article{dong2023fairness,
  author        = {Yushun Dong and Jing Ma and Song Wang and Chen Chen and Jundong Li},
  title         = {Fairness in Graph Mining: A Survey},
  journal       = {IEEE Transactions on Knowledge and Data Engineering},
  volume        = {35},
  number        = {10},
  pages         = {10583--10602},
  year          = {2023},
  doi           = {10.1109/TKDE.2023.3265598},
  eprint        = {2204.09888},
  archivePrefix = {arXiv},
  primaryClass  = {cs.LG}
}

@misc{duval2023hitchhiker,
  author        = {Alexandre Duval and Simon V. Mathis and Chaitanya K. Joshi and Victor Schmidt and Santiago Miret and Fragkiskos D. Malliaros and Taco Cohen and Pietro Li{\`o} and Yoshua Bengio and Michael M. Bronstein},
  title         = {A Hitchhiker's Guide to Geometric {GNN}s for {3D} Atomic Systems},
  year          = {2023},
  doi           = {10.48550/arXiv.2312.07511},
  eprint        = {2312.07511},
  archivePrefix = {arXiv},
  primaryClass  = {cs.LG},
  url           = {https://arxiv.org/abs/2312.07511}
}

@article{dwivedi2023benchmarking,
  author  = {Vijay Prakash Dwivedi and Chaitanya K. Joshi and Anh Tuan Luu and Thomas Laurent and Yoshua Bengio and Xavier Bresson},
  title   = {Benchmarking Graph Neural Networks},
  journal = {Journal of Machine Learning Research},
  volume  = {24},
  number  = {43},
  pages   = {1--48},
  year    = {2023},
  url     = {https://jmlr.org/papers/v24/22-0567.html}
}

@inproceedings{Eliasof2023omegaGNN,
  author        = {Moshe Eliasof and Lars Ruthotto and Eran Treister},
  title         = {Improving Graph Neural Networks with Learnable Propagation Operators},
  booktitle     = {Proceedings of the 40th International Conference on Machine Learning},
  series        = {Proceedings of Machine Learning Research},
  volume        = {202},
  pages         = {9224--9245},
  publisher     = {PMLR},
  year          = {2023},
  eprint        = {2210.17224},
  archivePrefix = {arXiv},
  url           = {https://proceedings.mlr.press/v202/eliasof23b.html}
}

@article{Gao2021SAGAT,
  author  = {Jianliang Gao and Jun Gao and Xiaoting Ying and Mingming Lu and Jian-Xin Wang},
  title   = {Higher-Order Interaction Goes Neural: A Substructure Assembling Graph Attention Network for Graph Classification},
  journal = {IEEE Transactions on Knowledge and Data Engineering},
  volume  = {35},
  number  = {2},
  pages   = {1594--1608},
  year    = {2023},
  doi     = {10.1109/TKDE.2021.3105544},
  url     = {https://doi.org/10.1109/TKDE.2021.3105544}
}

@inproceedings{giraldo2023tradeoff,
  author        = {Jhony H. Giraldo and Konstantinos Skianis and Thierry Bouwmans and Fragkiskos D. Malliaros},
  title         = {On the Trade-off between Over-smoothing and Over-squashing in Deep Graph Neural Networks},
  booktitle     = {Proceedings of the 32nd ACM International Conference on Information and Knowledge Management},
  pages         = {566--576},
  publisher     = {ACM},
  year          = {2023},
  doi           = {10.1145/3583780.3614997},
  eprint        = {2212.02374},
  archivePrefix = {arXiv}
}

@inproceedings{Gutteridge2023DRew,
  author        = {Benjamin Gutteridge and Xiaowen Dong and Michael M. Bronstein and Francesco Di Giovanni},
  title         = {{DRew}: Dynamically Rewired Message Passing with Delay},
  booktitle     = {Proceedings of the 40th International Conference on Machine Learning},
  series        = {Proceedings of Machine Learning Research},
  volume        = {202},
  pages         = {12252--12267},
  publisher     = {PMLR},
  year          = {2023},
  eprint        = {2305.08018},
  archivePrefix = {arXiv},
  url           = {https://proceedings.mlr.press/v202/gutteridge23a.html}
}

@inproceedings{huang2023i2gnn,
  author        = {Yinan Huang and Xingang Peng and Jianzhu Ma and Muhan Zhang},
  title         = {Boosting the Cycle Counting Power of Graph Neural Networks with {$I^2$}-{GNNs}},
  booktitle     = {International Conference on Learning Representations},
  year          = {2023},
  eprint        = {2210.13978},
  archivePrefix = {arXiv},
  url           = {https://openreview.net/forum?id=kDSmxOspsXQ}
}

@inproceedings{Huang2023MidGCN,
  author        = {Jincheng Huang and Lun Du and Xu Chen and Qiang Fu and Shi Han and Dongmei Zhang},
  title         = {Robust Mid-Pass Filtering Graph Convolutional Networks},
  booktitle     = {Proceedings of the ACM Web Conference 2023},
  pages         = {328--338},
  publisher     = {ACM},
  year          = {2023},
  doi           = {10.1145/3543507.3583335},
  eprint        = {2302.08048},
  archivePrefix = {arXiv},
  url           = {https://dl.acm.org/doi/10.1145/3543507.3583335}
}

@inproceedings{huang2023tgb,
  author        = {Shenyang Huang and Farimah Poursafaei and Jacob Danovitch and Matthias Fey and Weihua Hu and Emanuele Rossi and Jure Leskovec and Michael M. Bronstein and Guillaume Rabusseau and Reihaneh Rabbany},
  title         = {Temporal Graph Benchmark for Machine Learning on Temporal Graphs},
  booktitle     = {Advances in Neural Information Processing Systems},
  volume        = {36},
  pages         = {2056--2073},
  year          = {2023},
  eprint        = {2307.01026},
  archivePrefix = {arXiv},
  url           = {https://proceedings.neurips.cc/paper_files/paper/2023/hash/066b98e63313162f6562b35962671288-Abstract-Datasets_and_Benchmarks.html},
  note          = {Datasets and Benchmarks Track}
}

@inproceedings{kaba2023symmetry,
  author        = {S{\'e}kou-Oumar Kaba and Siamak Ravanbakhsh},
  title         = {Symmetry Breaking and Equivariant Neural Networks},
  booktitle     = {NeurIPS 2023 Workshop on Symmetry and Geometry in Neural Representations},
  year          = {2023},
  eprint        = {2312.09016},
  archivePrefix = {arXiv},
  primaryClass  = {cs.LG},
  url           = {https://openreview.net/forum?id=d55JaRL9wh},
  note          = {Extended abstract}
}

@inproceedings{Karhadkar2023FoSR,
  author        = {Kedar Karhadkar and Pradeep Kr. Banerjee and Guido Mont{\'u}far},
  title         = {{FoSR}: First-Order Spectral Rewiring for Addressing Oversquashing in {GNNs}},
  booktitle     = {International Conference on Learning Representations},
  year          = {2023},
  eprint        = {2210.11790},
  archivePrefix = {arXiv},
  url           = {https://openreview.net/forum?id=3YjQfCLdrzz}
}

@inproceedings{Liu2023CurvDrop,
  author    = {Yang Liu and Chuan Zhou and Shirui Pan and Jia Wu and Zhao Li and Hongyang Chen and Peng Zhang},
  title     = {{CurvDrop}: A {Ricci} Curvature Based Approach to Prevent Graph Neural Networks from Over-Smoothing and Over-Squashing},
  booktitle = {Proceedings of the ACM Web Conference 2023},
  pages     = {221--230},
  publisher = {ACM},
  year      = {2023},
  doi       = {10.1145/3543507.3583269},
  url       = {https://dl.acm.org/doi/10.1145/3543507.3583269}
}

@inproceedings{luan2023whenhelp,
  author        = {Sitao Luan and Chenqing Hua and Minkai Xu and Qincheng Lu and Jiaqi Zhu and Xiao-Wen Chang and Jie Fu and Jure Leskovec and Doina Precup},
  title         = {When Do Graph Neural Networks Help with Node Classification? Investigating the Homophily Principle on Node Distinguishability},
  booktitle     = {Advances in Neural Information Processing Systems},
  volume        = {36},
  pages         = {28748--28760},
  year          = {2023},
  eprint        = {2304.14274},
  archivePrefix = {arXiv},
  url           = {https://proceedings.neurips.cc/paper_files/paper/2023/hash/5ba11de4c74548071899cf41dec078bf-Abstract-Conference.html}
}

@inproceedings{morris2023wlmeetvc,
  author        = {Christopher Morris and Floris Geerts and Jan T{\"o}nshoff and Martin Grohe},
  title         = {{WL} meet {VC}},
  booktitle     = {Proceedings of the 40th International Conference on Machine Learning},
  series        = {Proceedings of Machine Learning Research},
  volume        = {202},
  pages         = {25275--25302},
  publisher     = {PMLR},
  year          = {2023},
  month         = {23--29 Jul},
  eprint        = {2301.11039},
  archivePrefix = {arXiv},
  url           = {https://proceedings.mlr.press/v202/morris23a.html}
}

@inproceedings{Nguyen2023BORF,
  author        = {Khang Nguyen and Nong Minh Hieu and Vinh Duc Nguyen and Nhat Ho and Stanley Osher and Tan Minh Nguyen},
  title         = {Revisiting Over-smoothing and Over-squashing Using {Ollivier-Ricci} Curvature},
  booktitle     = {Proceedings of the 40th International Conference on Machine Learning},
  series        = {Proceedings of Machine Learning Research},
  volume        = {202},
  pages         = {25956--25979},
  publisher     = {PMLR},
  year          = {2023},
  eprint        = {2211.15779},
  archivePrefix = {arXiv},
  url           = {https://proceedings.mlr.press/v202/nguyen23c.html}
}

@inproceedings{Platonov2023Benchmarks,
  author        = {Oleg Platonov and Denis Kuznedelev and Michael Diskin and Artem Babenko and Liudmila Prokhorenkova},
  title         = {A Critical Look at the Evaluation of {GNNs} under Heterophily: Are We Really Making Progress?},
  booktitle     = {International Conference on Learning Representations},
  year          = {2023},
  eprint        = {2302.11640},
  archivePrefix = {arXiv},
  url           = {https://openreview.net/forum?id=tJbbQfw-5wv}
}

@inproceedings{platonov2023characterizing,
  author        = {Oleg Platonov and Denis Kuznedelev and Artem Babenko and Liudmila Prokhorenkova},
  title         = {Characterizing Graph Datasets for Node Classification: Homophily-Heterophily Dichotomy and Beyond},
  booktitle     = {Advances in Neural Information Processing Systems},
  volume        = {36},
  pages         = {523--548},
  year          = {2023},
  eprint        = {2209.06177},
  archivePrefix = {arXiv},
  url           = {https://proceedings.neurips.cc/paper_files/paper/2023/hash/01b681025fdbda8e935a66cc5bb6e9de-Abstract-Conference.html}
}

@misc{rusch2023survey,
  author        = {T. Konstantin Rusch and Michael M. Bronstein and Siddhartha Mishra},
  title         = {A Survey on Oversmoothing in Graph Neural Networks},
  year          = {2023},
  doi           = {10.48550/arXiv.2303.10993},
  eprint        = {2303.10993},
  archivePrefix = {arXiv},
  url           = {https://arxiv.org/abs/2303.10993}
}

@inproceedings{tahmasebi2023recursion,
  author        = {Behrooz Tahmasebi and Derek Lim and Stefanie Jegelka},
  title         = {The Power of Recursion in Graph Neural Networks for Counting Substructures},
  booktitle     = {Proceedings of The 26th International Conference on Artificial Intelligence and Statistics},
  series        = {Proceedings of Machine Learning Research},
  volume        = {206},
  pages         = {11023--11042},
  publisher     = {PMLR},
  year          = {2023},
  month         = {25--27 Apr},
  eprint        = {2012.03174},
  archivePrefix = {arXiv},
  url           = {https://proceedings.mlr.press/v206/tahmasebi23a.html}
}

@inproceedings{wang2023collaboration,
  author        = {Yu Wang and Yuying Zhao and Yi Zhang and Tyler Derr},
  title         = {Collaboration-Aware Graph Convolutional Network for Recommender Systems},
  booktitle     = {Proceedings of the ACM Web Conference 2023},
  pages         = {91--101},
  year          = {2023},
  doi           = {10.1145/3543507.3583229},
  eprint        = {2207.06221},
  archivePrefix = {arXiv},
  url           = {https://doi.org/10.1145/3543507.3583229}
}

@inproceedings{Wang2023EDHNN,
  author        = {Peihao Wang and Shenghao Yang and Yunyu Liu and Zhangyang Wang and Pan Li},
  title         = {Equivariant Hypergraph Diffusion Neural Operators},
  booktitle     = {International Conference on Learning Representations},
  year          = {2023},
  eprint        = {2207.06680},
  archivePrefix = {arXiv},
  url           = {https://openreview.net/forum?id=RiTjKoscnNd}
}

@inproceedings{wang2023nlgraph,
  author        = {Heng Wang and Shangbin Feng and Tianxing He and Zhaoxuan Tan and Xiaochuang Han and Yulia Tsvetkov},
  title         = {Can Language Models Solve Graph Problems in Natural Language?},
  booktitle     = {Advances in Neural Information Processing Systems},
  volume        = {36},
  pages         = {30840--30861},
  year          = {2023},
  eprint        = {2305.10037},
  archivePrefix = {arXiv},
  url           = {https://proceedings.neurips.cc/paper_files/paper/2023/hash/622afc4edf2824a1b6aaf5afe153fa93-Abstract-Conference.html}
}

@inproceedings{Wu2023DemystifyingOversmoothing,
  author        = {Xinyi Wu and Amir Ajorlou and Zihui Wu and Ali Jadbabaie},
  title         = {Demystifying Oversmoothing in Attention-Based Graph Neural Networks},
  booktitle     = {Advances in Neural Information Processing Systems},
  volume        = {36},
  pages         = {35084--35106},
  year          = {2023},
  eprint        = {2305.16102},
  archivePrefix = {arXiv},
  url           = {https://proceedings.neurips.cc/paper_files/paper/2023/hash/6e4cdfdd909ea4e34bfc85a12774cba0-Abstract-Conference.html}
}

@inproceedings{Zhang2023Biconnectivity,
  author        = {Bohang Zhang and Shengjie Luo and Liwei Wang and Di He},
  title         = {Rethinking the Expressive Power of {GNNs} via Graph Biconnectivity},
  booktitle     = {International Conference on Learning Representations},
  year          = {2023},
  eprint        = {2301.09505},
  archivePrefix = {arXiv},
  url           = {https://openreview.net/forum?id=r9hNv76KoT3}
}

@inproceedings{Zhang2023SubgraphWLHierarchy,
  author        = {Bohang Zhang and Guhao Feng and Yiheng Du and Di He and Liwei Wang},
  title         = {A Complete Expressiveness Hierarchy for Subgraph {GNNs} via Subgraph {Weisfeiler-Lehman} Tests},
  booktitle     = {Proceedings of the 40th International Conference on Machine Learning},
  series        = {Proceedings of Machine Learning Research},
  volume        = {202},
  pages         = {41019--41077},
  publisher     = {PMLR},
  year          = {2023},
  eprint        = {2302.07090},
  archivePrefix = {arXiv},
  url           = {https://proceedings.mlr.press/v202/zhang23k.html}
}

@inproceedings{Zhao2023CDEGRAND,
  author        = {Kai Zhao and Qiyu Kang and Yang Song and Rui She and Sijie Wang and Wee Peng Tay},
  title         = {Graph Neural Convection-Diffusion with Heterophily},
  booktitle     = {Proceedings of the Thirty-Second International Joint Conference on Artificial Intelligence},
  pages         = {4656--4664},
  year          = {2023},
  doi           = {10.24963/ijcai.2023/518},
  eprint        = {2305.16780},
  archivePrefix = {arXiv},
  url           = {https://www.ijcai.org/proceedings/2023/518}
}

@inproceedings{Zhou2023RelationalPooling,
  author        = {Cai Zhou and Xiyuan Wang and Muhan Zhang},
  title         = {From Relational Pooling to Subgraph {GNNs}: A Universal Framework for More Expressive Graph Neural Networks},
  booktitle     = {Proceedings of the 40th International Conference on Machine Learning},
  series        = {Proceedings of Machine Learning Research},
  volume        = {202},
  pages         = {42742--42768},
  publisher     = {PMLR},
  year          = {2023},
  eprint        = {2305.04963},
  archivePrefix = {arXiv},
  url           = {https://proceedings.mlr.press/v202/zhou23n.html}
}

@inproceedings{barshalom2024subgraphormer,
  author        = {Guy Bar-Shalom and Beatrice Bevilacqua and Haggai Maron},
  title         = {Subgraphormer: Unifying Subgraph {GNNs} and Graph Transformers via Graph Products},
  booktitle     = {Proceedings of the 41st International Conference on Machine Learning},
  series        = {Proceedings of Machine Learning Research},
  volume        = {235},
  pages         = {2959--2989},
  publisher     = {PMLR},
  year          = {2024},
  eprint        = {2402.08450},
  archivePrefix = {arXiv},
  url           = {https://proceedings.mlr.press/v235/bar-shalom24a.html}
}

@inproceedings{besta2024got,
  author        = {Maciej Besta and Nils Blach and Ales Kubicek and Robert Gerstenberger and Micha{\l} Podstawski and Lukas Gianinazzi and Joanna Gajda and Tomasz Lehmann and Hubert Niewiadomski and Piotr Nyczyk and Torsten Hoefler},
  title         = {{Graph of Thoughts}: Solving Elaborate Problems with Large Language Models},
  booktitle     = {Proceedings of the AAAI Conference on Artificial Intelligence},
  volume        = {38},
  number        = {16},
  pages         = {17682--17690},
  year          = {2024},
  doi           = {10.1609/aaai.v38i16.29720},
  eprint        = {2308.09687},
  archivePrefix = {arXiv},
  primaryClass  = {cs.CL},
  url           = {https://ojs.aaai.org/index.php/AAAI/article/view/29720}
}

@article{dai2024trustworthy,
  author        = {Enyan Dai and Tianxiang Zhao and Huaisheng Zhu and Junjie Xu and Zhimeng Guo and Hui Liu and Jiliang Tang and Suhang Wang},
  title         = {A Comprehensive Survey on Trustworthy Graph Neural Networks: Privacy, Robustness, Fairness, and Explainability},
  journal       = {Machine Intelligence Research},
  volume        = {21},
  number        = {6},
  pages         = {1011--1061},
  year          = {2024},
  doi           = {10.1007/s11633-024-1510-8},
  eprint        = {2204.08570},
  archivePrefix = {arXiv}
}

@inproceedings{das2024agsgnn,
  author        = {Siddhartha Shankar Das and S. M. Ferdous and Mahantesh M. Halappanavar and Edoardo Serra and Alex Pothen},
  title         = {{AGS-GNN}: Attribute-guided Sampling for Graph Neural Networks},
  booktitle     = {Proceedings of the 30th ACM SIGKDD Conference on Knowledge Discovery and Data Mining},
  pages         = {538--549},
  publisher     = {ACM},
  year          = {2024},
  doi           = {10.1145/3637528.3671940},
  eprint        = {2405.15218},
  archivePrefix = {arXiv},
  url           = {https://doi.org/10.1145/3637528.3671940}
}

@inproceedings{Deng2024Polynormer,
  author        = {Chenhui Deng and Zichao Yue and Zhiru Zhang},
  title         = {Polynormer: Polynomial-Expressive Graph {Transformer} in Linear Time},
  booktitle     = {International Conference on Learning Representations},
  year          = {2024},
  eprint        = {2403.01232},
  archivePrefix = {arXiv},
  url           = {https://openreview.net/forum?id=hmv1LpNfXa}
}

@article{digiovanni2024oversquashing,
  author        = {Francesco Di Giovanni and T. Konstantin Rusch and Michael M. Bronstein and Andreea Deac and Marc Lackenby and Siddhartha Mishra and Petar Veli{\v{c}}kovi{\'c}},
  title         = {How does over-squashing affect the power of {GNNs}?},
  journal       = {Transactions on Machine Learning Research},
  year          = {2024},
  eprint        = {2306.03589},
  archivePrefix = {arXiv},
  url           = {https://openreview.net/forum?id=KJRoQvRWNs}
}

@misc{edge2024graphrag,
  author        = {Darren Edge and Ha Trinh and Newman Cheng and Joshua Bradley and Alex Chao and Apurva Mody and Steven Truitt and Dasha Metropolitansky and Robert Osazuwa Ness and Jonathan Larson},
  title         = {From Local to Global: A {Graph RAG} Approach to Query-Focused Summarization},
  year          = {2024},
  doi           = {10.48550/arXiv.2404.16130},
  eprint        = {2404.16130},
  archivePrefix = {arXiv},
  primaryClass  = {cs.CL},
  url           = {https://arxiv.org/abs/2404.16130}
}

@inproceedings{Finkelshtein2024CoGNN,
  author        = {Ben Finkelshtein and Xingyue Huang and Michael M. Bronstein and {\.I}smail {\.I}lkan Ceylan},
  title         = {Cooperative Graph Neural Networks},
  booktitle     = {Proceedings of the 41st International Conference on Machine Learning},
  series        = {Proceedings of Machine Learning Research},
  volume        = {235},
  pages         = {13633--13659},
  publisher     = {PMLR},
  year          = {2024},
  eprint        = {2310.01267},
  archivePrefix = {arXiv},
  url           = {https://proceedings.mlr.press/v235/finkelshtein24a.html}
}

@inproceedings{geisler2024spatiospectral,
  author        = {Simon Geisler and Arthur Kosmala and Daniel Herbst and Stephan G{\"u}nnemann},
  title         = {Spatio-Spectral Graph Neural Networks},
  booktitle     = {Advances in Neural Information Processing Systems},
  volume        = {37},
  pages         = {49022--49080},
  year          = {2024},
  doi           = {10.52202/079017-1554},
  eprint        = {2405.19121},
  archivePrefix = {arXiv},
  primaryClass  = {cs.LG},
  url           = {https://papers.neurips.cc/paper_files/paper/2024/hash/580c4ec4738ff61d5862a122cdf139b6-Abstract-Conference.html}
}

@misc{guo2024spatially,
  author        = {Jingwei Guo and Kaizhu Huang and Xinping Yi and Zixian Su and Rui Zhang},
  title         = {Rethinking Spectral Graph Neural Networks with Spatially Adaptive Filtering},
  year          = {2024},
  doi           = {10.48550/arXiv.2401.09071},
  eprint        = {2401.09071},
  archivePrefix = {arXiv},
  primaryClass  = {cs.LG},
  url           = {https://arxiv.org/abs/2401.09071}
}

@misc{han2024retrieval,
  author        = {Haoyu Han and Yu Wang and Harry Shomer and Kai Guo and Jiayuan Ding and Yongjia Lei and Mahantesh Halappanavar and Ryan A. Rossi and Subhabrata Mukherjee and Xianfeng Tang and Qi He and Zhigang Hua and Bo Long and Tong Zhao and Neil Shah and Amin Javari and Yinglong Xia and Jiliang Tang},
  title         = {Retrieval-Augmented Generation with Graphs ({GraphRAG})},
  year          = {2024},
  doi           = {10.48550/arXiv.2501.00309},
  eprint        = {2501.00309},
  archivePrefix = {arXiv},
  primaryClass  = {cs.LG},
  url           = {https://arxiv.org/abs/2501.00309}
}

@inproceedings{he2024gretriever,
  author        = {Xiaoxin He and Yijun Tian and Yifei Sun and Nitesh V. Chawla and Thomas Laurent and Yann LeCun and Xavier Bresson and Bryan Hooi},
  title         = {{G-Retriever}: Retrieval-Augmented Generation for Textual Graph Understanding and Question Answering},
  booktitle     = {Advances in Neural Information Processing Systems},
  volume        = {37},
  pages         = {132876--132907},
  year          = {2024},
  doi           = {10.52202/079017-4224},
  eprint        = {2402.07630},
  archivePrefix = {arXiv},
  primaryClass  = {cs.LG},
  url           = {https://proceedings.neurips.cc/paper_files/paper/2024/hash/efaf1c9726648c8ba363a5c927440529-Abstract-Conference.html}
}

@article{He2024PMPGNN,
  author  = {Tiantian He and Yang Liu and Yew-Soon Ong and Xiaohu Wu and Xin Luo},
  title   = {Polarized message-passing in graph neural networks},
  journal = {Artificial Intelligence},
  volume  = {331},
  pages   = {104129},
  year    = {2024},
  doi     = {10.1016/j.artint.2024.104129},
  url     = {https://doi.org/10.1016/j.artint.2024.104129}
}

@inproceedings{hofgard2024relaxed,
  author        = {Elyssa Hofgard and Rui Wang and Robin Walters and Tess Smidt},
  title         = {Relaxed Equivariant Graph Neural Networks},
  booktitle     = {ICML 2024 Workshop on Geometry-grounded Representation Learning and Generative Modeling},
  year          = {2024},
  eprint        = {2407.20471},
  archivePrefix = {arXiv},
  primaryClass  = {cs.LG},
  url           = {https://openreview.net/forum?id=eVB1fn37Ay},
  note          = {Extended abstract}
}

@article{Huang2024Flow2GNN,
  author  = {Changqin Huang and Yi Wang and Yunliang Jiang and Ming Li and Xiaodi Huang and Shijin Wang and Shirui Pan and Chuan Zhou},
  title   = {{Flow2GNN}: Flexible Two-Way Flow Message Passing for Enhancing {GNNs} Beyond Homophily},
  journal = {IEEE Transactions on Cybernetics},
  volume  = {54},
  number  = {11},
  pages   = {6607--6618},
  year    = {2024},
  doi     = {10.1109/TCYB.2024.3412149},
  url     = {https://doi.org/10.1109/TCYB.2024.3412149}
}

@inproceedings{huang2024spe,
  author        = {Yinan Huang and William Lu and Joshua Robinson and Yu Yang and Muhan Zhang and Stefanie Jegelka and Pan Li},
  title         = {On the Stability of Expressive Positional Encodings for Graphs},
  booktitle     = {International Conference on Learning Representations},
  year          = {2024},
  eprint        = {2310.02579},
  archivePrefix = {arXiv},
  url           = {https://proceedings.iclr.cc/paper_files/paper/2024/hash/ad52f9c2b232a719963a620dcf88fbe5-Abstract-Conference.html}
}

@inproceedings{jamadandi2024spectralpruning,
  author        = {Adarsh Jamadandi and Celia Rubio-Madrigal and Rebekka Burkholz},
  title         = {Spectral Graph Pruning Against Over-Squashing and Over-Smoothing},
  booktitle     = {Advances in Neural Information Processing Systems},
  volume        = {37},
  pages         = {10348--10379},
  year          = {2024},
  doi           = {10.52202/079017-0331},
  eprint        = {2404.04612},
  archivePrefix = {arXiv},
  url           = {https://proceedings.neurips.cc/paper_files/paper/2024/hash/140aac600566125915df7e74ff538f66-Abstract-Conference.html}
}

@inproceedings{jin2024hombasis,
  author        = {Emily Jin and Michael M. Bronstein and {\.I}smail {\.I}lkan Ceylan and Matthias Lanzinger},
  title         = {Homomorphism Counts for Graph Neural Networks: All About That Basis},
  booktitle     = {Proceedings of the 41st International Conference on Machine Learning},
  series        = {Proceedings of Machine Learning Research},
  volume        = {235},
  pages         = {22075--22098},
  publisher     = {PMLR},
  year          = {2024},
  month         = {21--27 Jul},
  eprint        = {2402.08595},
  archivePrefix = {arXiv},
  url           = {https://proceedings.mlr.press/v235/jin24a.html}
}

@inproceedings{kofinas2024neuralgraphs,
  author        = {Miltiadis Kofinas and Boris Knyazev and Yan Zhang and Yunlu Chen and Gertjan J. Burghouts and Efstratios Gavves and Cees G. M. Snoek and David W. Zhang},
  title         = {Graph Neural Networks for Learning Equivariant Representations of Neural Networks},
  booktitle     = {International Conference on Learning Representations},
  year          = {2024},
  eprint        = {2403.12143},
  archivePrefix = {arXiv},
  url           = {https://openreview.net/forum?id=oO6FsMyDBt}
}

@inproceedings{Lee2024FeatureDistribution,
  author        = {Soo Yong Lee and Sunwoo Kim and Fanchen Bu and Jaemin Yoo and Jiliang Tang and Kijung Shin},
  title         = {Feature Distribution on Graph Topology Mediates the Effect of Graph Convolution: Homophily Perspective},
  booktitle     = {Proceedings of the 41st International Conference on Machine Learning},
  series        = {Proceedings of Machine Learning Research},
  volume        = {235},
  pages         = {26686--26714},
  publisher     = {PMLR},
  year          = {2024},
  eprint        = {2402.04621},
  archivePrefix = {arXiv},
  url           = {https://proceedings.mlr.press/v235/lee24m.html}
}

@inproceedings{Lell2024HyperAggregation,
  author        = {Nicolas Lell and Ansgar Scherp},
  title         = {{HyperAggregation}: Aggregating over Graph Edges with Hypernetworks},
  booktitle     = {2024 International Joint Conference on Neural Networks (IJCNN)},
  pages         = {1--9},
  publisher     = {IEEE},
  year          = {2024},
  doi           = {10.1109/IJCNN60899.2024.10650980},
  eprint        = {2407.11596},
  archivePrefix = {arXiv},
  url           = {https://ieeexplore.ieee.org/document/10650980/}
}

@misc{li2024homgen,
  author        = {Shouheng Li and Dongwoo Kim and Qing Wang},
  title         = {Generalization of Graph Neural Networks through the Lens of Homomorphism},
  year          = {2024},
  doi           = {10.48550/arXiv.2403.06079},
  eprint        = {2403.06079},
  archivePrefix = {arXiv},
  primaryClass  = {cs.LG},
  url           = {https://arxiv.org/abs/2403.06079}
}

@inproceedings{lim2024graphmetanetworks,
  author        = {Derek Lim and Haggai Maron and Marc T. Law and Jonathan Lorraine and James Lucas},
  title         = {Graph Metanetworks for Processing Diverse Neural Architectures},
  booktitle     = {International Conference on Learning Representations},
  year          = {2024},
  eprint        = {2312.04501},
  archivePrefix = {arXiv},
  url           = {https://openreview.net/forum?id=ijK5hyxs0n}
}

@inproceedings{liu2024ofa,
  author        = {Hao Liu and Jiarui Feng and Lecheng Kong and Ningyue Liang and Dacheng Tao and Yixin Chen and Muhan Zhang},
  title         = {One for All: Towards Training One Graph Model for All Classification Tasks},
  booktitle     = {International Conference on Learning Representations},
  year          = {2024},
  eprint        = {2310.00149},
  archivePrefix = {arXiv},
  url           = {https://proceedings.iclr.cc/paper_files/paper/2024/hash/57faf5642eb06e0602b95f6aa989b38a-Abstract-Conference.html}
}

@inproceedings{loveland2024localhomophily,
  author        = {Donald Loveland and Jiong Zhu and Mark Heimann and Benjamin Fish and Michael T. Schaub and Danai Koutra},
  title         = {On Performance Discrepancies Across Local Homophily Levels in Graph Neural Networks},
  booktitle     = {Proceedings of the Second Learning on Graphs Conference},
  series        = {Proceedings of Machine Learning Research},
  volume        = {231},
  pages         = {6:1--6:30},
  publisher     = {PMLR},
  year          = {2024},
  eprint        = {2306.05557},
  archivePrefix = {arXiv},
  url           = {https://proceedings.mlr.press/v231/loveland24a.html}
}

@inproceedings{luo2024classic,
  author        = {Yuankai Luo and Lei Shi and Xiao-Ming Wu},
  title         = {Classic {GNN}s are Strong Baselines: Reassessing {GNN}s for Node Classification},
  booktitle     = {Advances in Neural Information Processing Systems},
  volume        = {37},
  pages         = {97650--97669},
  year          = {2024},
  doi           = {10.52202/079017-3098},
  eprint        = {2406.08993},
  archivePrefix = {arXiv},
  primaryClass  = {cs.LG},
  url           = {https://proceedings.neurips.cc/paper_files/paper/2024/hash/b10ed15ff1aa864f1be3a75f1ffc021b-Abstract-Datasets_and_Benchmarks_Track.html},
  note          = {Datasets and Benchmarks Track}
}

@inproceedings{luo2024rog,
  author        = {Linhao Luo and Yuan-Fang Li and Gholamreza Haffari and Shirui Pan},
  title         = {Reasoning on Graphs: Faithful and Interpretable Large Language Model Reasoning},
  booktitle     = {International Conference on Learning Representations},
  year          = {2024},
  eprint        = {2310.01061},
  archivePrefix = {arXiv},
  primaryClass  = {cs.CL},
  url           = {https://openreview.net/forum?id=ZGNWW7xZ6Q}
}

@inproceedings{mao2024position,
  author        = {Haitao Mao and Zhikai Chen and Wenzhuo Tang and Jianan Zhao and Yao Ma and Tong Zhao and Neil Shah and Mikhail Galkin and Jiliang Tang},
  title         = {Position: Graph Foundation Models Are Already Here},
  booktitle     = {Proceedings of the 41st International Conference on Machine Learning},
  series        = {Proceedings of Machine Learning Research},
  volume        = {235},
  pages         = {34670--34692},
  publisher     = {PMLR},
  year          = {2024},
  eprint        = {2402.02216},
  archivePrefix = {arXiv},
  url           = {https://proceedings.mlr.press/v235/mao24a.html}
}

@inproceedings{morris2024future,
  author        = {Christopher Morris and Fabrizio Frasca and Nadav Dym and Haggai Maron and {\.I}smail {\.I}lkan Ceylan and Ron Levie and Derek Lim and Michael M. Bronstein and Martin Grohe and Stefanie Jegelka},
  title         = {Position: Future Directions in the Theory of Graph Machine Learning},
  booktitle     = {Proceedings of the 41st International Conference on Machine Learning},
  series        = {Proceedings of Machine Learning Research},
  volume        = {235},
  pages         = {36294--36307},
  publisher     = {PMLR},
  year          = {2024},
  month         = {21--27 Jul},
  eprint        = {2402.02287},
  archivePrefix = {arXiv},
  url           = {https://proceedings.mlr.press/v235/morris24a.html}
}

@inproceedings{Muller2024AligningTransformers,
  author        = {Luis M{\"u}ller and Christopher Morris},
  title         = {Aligning Transformers with {Weisfeiler-Leman}},
  booktitle     = {Proceedings of the 41st International Conference on Machine Learning},
  series        = {Proceedings of Machine Learning Research},
  volume        = {235},
  pages         = {36654--36704},
  publisher     = {PMLR},
  year          = {2024},
  eprint        = {2406.03148},
  archivePrefix = {arXiv},
  url           = {https://proceedings.mlr.press/v235/muller24c.html}
}

@article{pan2024integrating,
  author        = {Shirui Pan and Yizhen Zheng and Yixin Liu},
  title         = {Integrating Graphs with Large Language Models: Methods and Prospects},
  journal       = {IEEE Intelligent Systems},
  volume        = {39},
  number        = {1},
  pages         = {64--68},
  year          = {2024},
  doi           = {10.1109/MIS.2023.3332242},
  eprint        = {2310.05499},
  archivePrefix = {arXiv}
}

@article{pan2024unifyingkg,
  author  = {Shirui Pan and Linhao Luo and Yufei Wang and Chen Chen and Jiapu Wang and Xindong Wu},
  title   = {Unifying Large Language Models and Knowledge Graphs: A Roadmap},
  journal = {IEEE Transactions on Knowledge and Data Engineering},
  volume  = {36},
  number  = {7},
  pages   = {3580--3599},
  year    = {2024},
  doi     = {10.1109/TKDE.2024.3352100},
  url     = {https://doi.org/10.1109/TKDE.2024.3352100}
}

@inproceedings{paolino2024loopy,
  author        = {Raffaele Paolino and Sohir Maskey and Pascal Welke and Gitta Kutyniok},
  title         = {Weisfeiler and {Leman} Go Loopy: A New Hierarchy for Graph Representational Learning},
  booktitle     = {Advances in Neural Information Processing Systems},
  volume        = {37},
  pages         = {120780--120831},
  year          = {2024},
  doi           = {10.52202/079017-3838},
  eprint        = {2403.13749},
  archivePrefix = {arXiv},
  primaryClass  = {cs.LG},
  url           = {https://proceedings.neurips.cc/paper_files/paper/2024/hash/dad28e90cd2c8caedf362d49c4d99e70-Abstract-Conference.html}
}

@misc{park2024taming,
  author        = {Moonjeong Park and Dongwoo Kim},
  title         = {Taming Gradient Oversmoothing and Expansion in Graph Neural Networks},
  year          = {2024},
  eprint        = {2410.04824},
  archivePrefix = {arXiv},
  url           = {https://arxiv.org/abs/2410.04824}
}

@inproceedings{qian2024prmpnn,
  author        = {Chendi Qian and Andrei Manolache and Kareem Ahmed and Zhe Zeng and Guy Van den Broeck and Mathias Niepert and Christopher Morris},
  title         = {Probabilistically Rewired Message-Passing Neural Networks},
  booktitle     = {International Conference on Learning Representations},
  year          = {2024},
  eprint        = {2310.02156},
  archivePrefix = {arXiv},
  url           = {https://proceedings.iclr.cc/paper_files/paper/2024/hash/876b45367d9069f0e91e359c57155ab1-Abstract-Conference.html}
}

@inproceedings{rossi2024edge,
  author        = {Emanuele Rossi and Bertrand Charpentier and Francesco Di Giovanni and Fabrizio Frasca and Stephan G{\"u}nnemann and Michael M. Bronstein},
  title         = {Edge Directionality Improves Learning on Heterophilic Graphs},
  booktitle     = {Proceedings of the Second Learning on Graphs Conference},
  series        = {Proceedings of Machine Learning Research},
  volume        = {231},
  pages         = {25:1--25:27},
  publisher     = {PMLR},
  year          = {2024},
  eprint        = {2305.10498},
  archivePrefix = {arXiv},
  url           = {https://proceedings.mlr.press/v231/rossi24a.html}
}

@inproceedings{roth2024rank,
  author        = {Andreas Roth and Thomas Liebig},
  title         = {Rank Collapse Causes Over-Smoothing and Over-Correlation in Graph Neural Networks},
  booktitle     = {Proceedings of the Second Learning on Graphs Conference},
  series        = {Proceedings of Machine Learning Research},
  volume        = {231},
  pages         = {35:1--35:23},
  publisher     = {PMLR},
  year          = {2024},
  eprint        = {2308.16800},
  archivePrefix = {arXiv},
  url           = {https://proceedings.mlr.press/v231/roth24a.html}
}

@article{Shi2024VRGNN,
  author        = {Fengzhao Shi and Yanan Cao and Ren Li and Xixun Lin and Yanmin Shang and Chuan Zhou and Jia Wu and Shirui Pan},
  title         = {{VR-GNN}: Variational Relation Vector Graph Neural Network for Modeling Homophily and Heterophily},
  journal       = {World Wide Web},
  volume        = {27},
  number        = {3},
  articleno     = {32},
  year          = {2024},
  doi           = {10.1007/s11280-024-01261-8},
  eprint        = {2211.14523},
  archivePrefix = {arXiv},
  url           = {https://doi.org/10.1007/s11280-024-01261-8}
}

@inproceedings{sun2024thinkongraph,
  author        = {Jiashuo Sun and Chengjin Xu and Lumingyuan Tang and Saizhuo Wang and Chen Lin and Yeyun Gong and Lionel M. Ni and Heung-Yeung Shum and Jian Guo},
  title         = {{Think-on-Graph}: Deep and Responsible Reasoning of Large Language Model on Knowledge Graph},
  booktitle     = {International Conference on Learning Representations},
  year          = {2024},
  eprint        = {2307.07697},
  archivePrefix = {arXiv},
  url           = {https://openreview.net/forum?id=nnVO1PvbTv}
}

@inproceedings{tang2024graphgpt,
  author        = {Jiabin Tang and Yuhao Yang and Wei Wei and Lei Shi and Lixin Su and Suqi Cheng and Dawei Yin and Chao Huang},
  title         = {{GraphGPT}: Graph Instruction Tuning for Large Language Models},
  booktitle     = {Proceedings of the 47th International ACM SIGIR Conference on Research and Development in Information Retrieval},
  pages         = {491--500},
  year          = {2024},
  doi           = {10.1145/3626772.3657775},
  eprint        = {2310.13023},
  archivePrefix = {arXiv}
}

@inproceedings{tjandra2024tgnv2,
  author        = {Benedict Aaron Tjandra and Federico Barbero and Michael M. Bronstein},
  title         = {Enhancing the Expressivity of Temporal Graph Networks through Source-Target Identification},
  booktitle     = {NeurIPS 2024 Workshop on Symmetry and Geometry in Neural Representations (NeurReps)},
  year          = {2024},
  eprint        = {2411.03596},
  archivePrefix = {arXiv},
  url           = {https://arxiv.org/abs/2411.03596}
}

@article{Tonshoff2023GapGo,
  author        = {Jan T{\"o}nshoff and Martin Ritzert and Eran Rosenbluth and Martin Grohe},
  title         = {Where Did the Gap Go? Reassessing the {Long-Range Graph Benchmark}},
  journal       = {Transactions on Machine Learning Research},
  year          = {2024},
  eprint        = {2309.00367},
  archivePrefix = {arXiv},
  url           = {https://openreview.net/forum?id=Nm0WX86sKv}
}

@inproceedings{wang2024knowledge,
  author        = {Yu Wang and Nedim Lipka and Ryan A. Rossi and Alexa Siu and Ruiyi Zhang and Tyler Derr},
  title         = {Knowledge Graph Prompting for Multi-Document Question Answering},
  booktitle     = {Proceedings of the AAAI Conference on Artificial Intelligence},
  volume        = {38},
  number        = {17},
  pages         = {19206--19214},
  year          = {2024},
  doi           = {10.1609/aaai.v38i17.29889},
  eprint        = {2308.11730},
  archivePrefix = {arXiv},
  url           = {https://ojs.aaai.org/index.php/AAAI/article/view/29889}
}

@misc{wang2024poolingbenchmark,
  author        = {Pengyun Wang and Junyu Luo and Yanxin Shen and Ming Zhang and Shaoen Qin and Hanwen Xing and Siyu Heng and Xiao Luo},
  title         = {A Comprehensive Graph Pooling Benchmark: Effectiveness, Robustness and Generalizability},
  year          = {2024},
  doi           = {10.48550/arXiv.2406.09031},
  eprint        = {2406.09031},
  archivePrefix = {arXiv},
  primaryClass  = {cs.LG},
  url           = {https://arxiv.org/abs/2406.09031}
}

@inproceedings{Wang2024UnderstandingHeterophily,
  author        = {Junfu Wang and Yuanfang Guo and Liang Yang and Yunhong Wang},
  title         = {Understanding Heterophily for Graph Neural Networks},
  booktitle     = {Proceedings of the 41st International Conference on Machine Learning},
  series        = {Proceedings of Machine Learning Research},
  volume        = {235},
  pages         = {50489--50529},
  publisher     = {PMLR},
  year          = {2024},
  eprint        = {2401.09125},
  archivePrefix = {arXiv},
  url           = {https://proceedings.mlr.press/v235/wang24u.html}
}

@article{Wu2023RFAGNN,
  author  = {Lirong Wu and Haitao Lin and Bozhen Hu and Cheng Tan and Zhangyang Gao and Zicheng Liu and Stan Z. Li},
  title   = {Beyond Homophily and Homogeneity Assumption: Relation-Based Frequency Adaptive Graph Neural Networks},
  journal = {IEEE Transactions on Neural Networks and Learning Systems},
  volume  = {35},
  number  = {6},
  pages   = {8497--8509},
  year    = {2024},
  doi     = {10.1109/TNNLS.2022.3230417},
  url     = {https://ieeexplore.ieee.org/document/10011169}
}

@misc{xu2024beyondcausality,
  author        = {Can Xu and Yao Cheng and Jianxiang Yu and Haosen Wang and Jingsong Lv and Yao Liu and Xiang Li},
  title         = {Improving Graph Out-of-distribution Generalization Beyond Causality},
  year          = {2024},
  eprint        = {2407.10204},
  archivePrefix = {arXiv},
  url           = {https://arxiv.org/abs/2407.10204}
}

@article{yan2024swingnn,
  author        = {Qi Yan and Zhengyang Liang and Yang Song and Renjie Liao and Lele Wang},
  title         = {{SwinGNN}: Rethinking Permutation Invariance in Diffusion Models for Graph Generation},
  journal       = {Transactions on Machine Learning Research},
  year          = {2024},
  eprint        = {2307.01646},
  archivePrefix = {arXiv},
  primaryClass  = {cs.LG},
  url           = {https://openreview.net/forum?id=abfi5plvQ4}
}

@article{Yang2024NCGNN,
  author        = {Rui Yang and Wenrui Dai and Chenglin Li and Junni Zou and Hongkai Xiong},
  title         = {{NCGNN}: Node-Level Capsule Graph Neural Network for Semisupervised Classification},
  journal       = {IEEE Transactions on Neural Networks and Learning Systems},
  volume        = {35},
  number        = {1},
  pages         = {1025--1039},
  year          = {2024},
  doi           = {10.1109/TNNLS.2022.3179306},
  eprint        = {2012.03476},
  archivePrefix = {arXiv},
  url           = {https://ieeexplore.ieee.org/document/9792205}
}

@inproceedings{zhang2024torchgt,
  author        = {Meng Zhang and Jie Sun and Qinghao Hu and Peng Sun and Zeke Wang and Yonggang Wen and Tianwei Zhang},
  title         = {{TorchGT}: A Holistic System for Large-Scale Graph {Transformer} Training},
  booktitle     = {Proceedings of the International Conference for High Performance Computing, Networking, Storage, and Analysis (SC)},
  articleno     = {77},
  year          = {2024},
  doi           = {10.1109/SC41406.2024.00083},
  eprint        = {2407.14106},
  archivePrefix = {arXiv},
  url           = {https://arxiv.org/abs/2407.14106}
}

@inproceedings{zhao2024gcope,
  author        = {Haihong Zhao and Aochuan Chen and Xiangguo Sun and Hong Cheng and Jia Li},
  title         = {All in One and One for All: A Simple yet Effective Method towards Cross-domain Graph Pretraining},
  booktitle     = {Proceedings of the 30th ACM SIGKDD Conference on Knowledge Discovery and Data Mining},
  pages         = {4443--4454},
  year          = {2024},
  doi           = {10.1145/3637528.3671913},
  eprint        = {2402.09834},
  archivePrefix = {arXiv}
}

@inproceedings{Zheng2024DisentanglingHomophily,
  author        = {Yilun Zheng and Sitao Luan and Lihui Chen},
  title         = {What Is Missing For Graph Homophily? Disentangling Graph Homophily For Graph Neural Networks},
  booktitle     = {Advances in Neural Information Processing Systems},
  volume        = {37},
  pages         = {68406--68452},
  year          = {2024},
  doi           = {10.52202/079017-2185},
  eprint        = {2406.18854},
  archivePrefix = {arXiv},
  url           = {https://proceedings.neurips.cc/paper_files/paper/2024/hash/7e810b2c75d69be186cadd2fe3febeab-Abstract-Conference.html}
}

@misc{zhou2024finegrained,
  author        = {Junru Zhou and Muhan Zhang},
  title         = {Fine-Grained Expressive Power of {Weisfeiler-Leman}: A Homomorphism Counting Perspective},
  year          = {2024},
  doi           = {10.48550/arXiv.2410.03517},
  eprint        = {2410.03517},
  archivePrefix = {arXiv},
  primaryClass  = {cs.LG},
  url           = {https://arxiv.org/abs/2410.03517}
}

@inproceedings{zhuge2024gptswarm,
  author    = {Mingchen Zhuge and Wenyi Wang and Louis Kirsch and Francesco Faccio and Dmitrii Khizbullin and J{\"u}rgen Schmidhuber},
  title     = {{GPTSwarm}: Language Agents as Optimizable Graphs},
  booktitle = {Proceedings of the 41st International Conference on Machine Learning},
  series    = {Proceedings of Machine Learning Research},
  volume    = {235},
  pages     = {62743--62767},
  publisher = {PMLR},
  year      = {2024},
  url       = {https://proceedings.mlr.press/v235/zhuge24a.html}
}

@inproceedings{azzolin2025beyond,
  author        = {Steve Azzolin and Sagar Malhotra and Andrea Passerini and Stefano Teso},
  title         = {Beyond Topological Self-Explainable {GNN}s: A Formal Explainability Perspective},
  booktitle     = {Proceedings of the 42nd International Conference on Machine Learning},
  series        = {Proceedings of Machine Learning Research},
  volume        = {267},
  pages         = {2045--2080},
  publisher     = {PMLR},
  year          = {2025},
  eprint        = {2502.02719},
  archivePrefix = {arXiv},
  primaryClass  = {cs.LG},
  url           = {https://proceedings.mlr.press/v267/azzolin25a.html}
}

@inproceedings{Bazhenov2025GraphLand,
  author        = {Gleb Bazhenov and Oleg Platonov and Liudmila Prokhorenkova},
  title         = {{GraphLand}: Evaluating Graph Machine Learning Models on Diverse Industrial Data},
  booktitle     = {Advances in Neural Information Processing Systems},
  volume        = {38},
  year          = {2025},
  eprint        = {2409.14500},
  archivePrefix = {arXiv},
  url           = {https://papers.nips.cc/paper_files/paper/2025/hash/8f54de761abc405145d8b4df8bef3f18-Abstract-Datasets_and_Benchmarks_Track.html},
  note          = {Datasets and Benchmarks Track}
}

@inproceedings{bechlerspeicher2025benchmarks,
  author        = {Maya Bechler-Speicher and Ben Finkelshtein and Fabrizio Frasca and Luis M{\"u}ller and Jan T{\"o}nshoff and Antoine Siraudin and Viktor Zaverkin and Michael M. Bronstein and Mathias Niepert and Bryan Perozzi and Mikhail Galkin and Christopher Morris},
  title         = {Position: Graph Learning Will Lose Relevance Due To Poor Benchmarks},
  booktitle     = {Proceedings of the 42nd International Conference on Machine Learning},
  series        = {Proceedings of Machine Learning Research},
  volume        = {267},
  pages         = {81067--81089},
  publisher     = {PMLR},
  year          = {2025},
  eprint        = {2502.14546},
  archivePrefix = {arXiv},
  primaryClass  = {cs.LG},
  url           = {https://proceedings.mlr.press/v267/bechler-speicher25a.html}
}

@misc{carrasco2025rademacher,
  author        = {Martin Carrasco and Caio F. Deberaldini Netto and Vahan A. Martirosyan and Ehimare Okoyomon and Caterina Graziani},
  title         = {On the {Rademacher} Complexity of Graph Neural Networks: Unifying Expressivity and Geometry},
  year          = {2025},
  eprint        = {2510.10101},
  archivePrefix = {arXiv},
  url           = {https://arxiv.org/abs/2510.10101},
  note          = {NeurIPS 2025 Workshop on New Perspectives in Advancing Graph Machine Learning}
}

@inproceedings{Chen2025NeuralWalker,
  author        = {Dexiong Chen and Till Hendrik Schulz and Karsten Borgwardt},
  title         = {Learning Long Range Dependencies on Graphs via Random Walks},
  booktitle     = {International Conference on Learning Representations},
  year          = {2025},
  eprint        = {2406.03386},
  archivePrefix = {arXiv},
  primaryClass  = {cs.LG},
  url           = {https://proceedings.iclr.cc/paper_files/paper/2025/hash/f039f5b34b2c2ae27808cf2a01ac3306-Abstract-Conference.html}
}

@misc{chen2025residual,
  author        = {Ziang Chen and Zhengjiang Lin and Shi Chen and Yury Polyanskiy and Philippe Rigollet},
  title         = {Residual Connections Provably Mitigate Oversmoothing in Graph Neural Networks},
  year          = {2025},
  eprint        = {2501.00762},
  archivePrefix = {arXiv},
  url           = {https://arxiv.org/abs/2501.00762}
}

@inproceedings{Errica2025AMP,
  author        = {Federico Errica and Henrik Christiansen and Viktor Zaverkin and Takashi Maruyama and Mathias Niepert and Francesco Alesiani},
  title         = {Adaptive Message Passing: A General Framework to Mitigate Oversmoothing, Oversquashing, and Underreaching},
  booktitle     = {Proceedings of the 42nd International Conference on Machine Learning},
  series        = {Proceedings of Machine Learning Research},
  volume        = {267},
  pages         = {15490--15515},
  publisher     = {PMLR},
  year          = {2025},
  eprint        = {2312.16560},
  archivePrefix = {arXiv},
  url           = {https://proceedings.mlr.press/v267/errica25a.html}
}

@inproceedings{fan2025opengu,
  author        = {Bowen Fan and Yuming Ai and Xunkai Li and Zhilin Guo and Lei Zhu and Guang Zeng and Rong-Hua Li and Guoren Wang},
  title         = {{OpenGU}: A Comprehensive Benchmark for Graph Unlearning},
  booktitle     = {Advances in Neural Information Processing Systems},
  volume        = {38},
  year          = {2025},
  eprint        = {2501.02728},
  archivePrefix = {arXiv},
  url           = {https://proceedings.neurips.cc/paper_files/paper/2025/hash/45d7dc2e7d8db0dbc13a88a4d6c08c5a-Abstract-Datasets_and_Benchmarks_Track.html},
  note          = {Datasets and Benchmarks Track}
}

@inproceedings{gai2025spectralhom,
  author        = {Jingchu Gai and Yiheng Du and Bohang Zhang and Haggai Maron and Liwei Wang},
  title         = {Homomorphism Expressivity of Spectral Invariant Graph Neural Networks},
  booktitle     = {International Conference on Learning Representations},
  year          = {2025},
  eprint        = {2503.00485},
  archivePrefix = {arXiv},
  url           = {https://openreview.net/forum?id=rdv6yeMFpn}
}

@misc{gao2025llm4gnas,
  author        = {Yang Gao and Hong Yang and Yizhi Chen and Junxian Wu and Peng Zhang and Haishuai Wang},
  title         = {{LLM4GNAS}: A Large Language Model Based Toolkit for Graph Neural Architecture Search},
  year          = {2025},
  eprint        = {2502.10459},
  archivePrefix = {arXiv},
  url           = {https://arxiv.org/abs/2502.10459}
}

@inproceedings{hariri2025chebnet,
  author        = {Ali Hariri and {\'A}lvaro Arroyo and Alessio Gravina and Moshe Eliasof and Carola-Bibiane Sch{\"o}nlieb and Davide Bacciu and Xiaowen Dong and Kamyar Azizzadenesheli and Pierre Vandergheynst},
  title         = {Return of {ChebNet}: Understanding and Improving an Overlooked {GNN} on Long Range Tasks},
  booktitle     = {Advances in Neural Information Processing Systems},
  volume        = {38},
  pages         = {136166--136196},
  year          = {2025},
  eprint        = {2506.07624},
  archivePrefix = {arXiv},
  url           = {https://proceedings.neurips.cc/paper_files/paper/2025/hash/c6de943558fe0b1bf4ea8f09fbcede44-Abstract-Conference.html}
}

@inproceedings{Ito2025LLPE,
  author        = {Michael Ito and Jiong Zhu and Dexiong Chen and Danai Koutra and Jenna Wiens},
  title         = {Learning {Laplacian} Positional Encodings for Heterophilous Graphs},
  booktitle     = {Proceedings of The 28th International Conference on Artificial Intelligence and Statistics},
  series        = {Proceedings of Machine Learning Research},
  volume        = {258},
  pages         = {2755--2763},
  publisher     = {PMLR},
  year          = {2025},
  eprint        = {2504.20430},
  archivePrefix = {arXiv},
  url           = {https://proceedings.mlr.press/v258/ito25a.html}
}

@inproceedings{jiang2025kgagent,
  author        = {Jinhao Jiang and Kun Zhou and Wayne Xin Zhao and Yang Song and Chen Zhu and Hengshu Zhu and Ji-Rong Wen},
  title         = {{KG-Agent}: An Efficient Autonomous Agent Framework for Complex Reasoning over Knowledge Graph},
  booktitle     = {Proceedings of the 63rd Annual Meeting of the Association for Computational Linguistics},
  pages         = {9505--9523},
  publisher     = {Association for Computational Linguistics},
  year          = {2025},
  doi           = {10.18653/v1/2025.acl-long.468},
  eprint        = {2402.11163},
  archivePrefix = {arXiv},
  url           = {https://aclanthology.org/2025.acl-long.468/}
}

@inproceedings{Kanatsoulis2025PEARL,
  author        = {Charilaos I. Kanatsoulis and Evelyn Choi and Stefanie Jegelka and Jure Leskovec and Alejandro Ribeiro},
  title         = {Learning Efficient Positional Encodings with Graph Neural Networks},
  booktitle     = {International Conference on Learning Representations},
  year          = {2025},
  eprint        = {2502.01122},
  archivePrefix = {arXiv},
  url           = {https://proceedings.iclr.cc/paper_files/paper/2025/hash/59883a1524bf58cf2dd86d0fc8d499e2-Abstract-Conference.html}
}

@inproceedings{li2024carnas,
  author        = {Peiwen Li and Xin Wang and Zeyang Zhang and Ziwei Zhang and Fang Shen and Jialong Wang and Yang Li and Wenwu Zhu},
  title         = {Causal-aware Graph Neural Architecture Search under Distribution Shifts},
  booktitle     = {Proceedings of the 31st ACM SIGKDD Conference on Knowledge Discovery and Data Mining},
  pages         = {1458--1469},
  year          = {2025},
  doi           = {10.1145/3711896.3736873},
  eprint        = {2405.16489},
  archivePrefix = {arXiv}
}

@inproceedings{li2025dygmamba,
  author        = {Dongyuan Li and Shiyin Tan and Ying Zhang and Ming Jin and Shirui Pan and Manabu Okumura and Renhe Jiang},
  title         = {{DyG-Mamba}: Continuous State Space Modeling on Dynamic Graphs},
  booktitle     = {Advances in Neural Information Processing Systems},
  volume        = {38},
  pages         = {129101--129130},
  year          = {2025},
  eprint        = {2408.06966},
  archivePrefix = {arXiv},
  url           = {https://proceedings.neurips.cc/paper_files/paper/2025/hash/bbba680acd2826f23928c6675a19f0e7-Abstract-Conference.html}
}

@inproceedings{li2025sizespectral,
  author        = {Gaotang Li and Danai Koutra and Yujun Yan},
  title         = {Tackling Size Generalization of Graph Neural Networks on Biological Data from a Spectral Perspective},
  booktitle     = {Proceedings of the 31st ACM SIGKDD Conference on Knowledge Discovery and Data Mining},
  pages         = {1341--1352},
  year          = {2025},
  doi           = {10.1145/3711896.3737144},
  eprint        = {2305.15611},
  archivePrefix = {arXiv}
}

@inproceedings{Liang2025SpectrumSparsification,
  author        = {Langzhang Liang and Fanchen Bu and Zixing Song and Zenglin Xu and Shirui Pan and Kijung Shin},
  title         = {Mitigating Over-Squashing in Graph Neural Networks by Spectrum-Preserving Sparsification},
  booktitle     = {Proceedings of the 42nd International Conference on Machine Learning},
  series        = {Proceedings of Machine Learning Research},
  volume        = {267},
  pages         = {37051--37070},
  publisher     = {PMLR},
  year          = {2025},
  eprint        = {2506.16110},
  archivePrefix = {arXiv},
  url           = {https://proceedings.mlr.press/v267/liang25a.html}
}

@inproceedings{Liao2025HubGT,
  author        = {Ningyi Liao and Zihao Yu and Siqiang Luo and Gao Cong},
  title         = {{HubGT}: Fast Graph {Transformer} with Decoupled Hierarchy Labeling},
  booktitle     = {Advances in Neural Information Processing Systems},
  volume        = {38},
  pages         = {22346--22376},
  year          = {2025},
  eprint        = {2412.04738},
  archivePrefix = {arXiv},
  url           = {https://proceedings.neurips.cc/paper_files/paper/2025/hash/20593d156ddc92277a234bfe065569fe-Abstract-Conference.html}
}

@article{liu2025gfmsurvey,
  author        = {Jiawei Liu and Cheng Yang and Zhiyuan Lu and Junze Chen and Yibo Li and Mengmei Zhang and Ting Bai and Yuan Fang and Lichao Sun and Philip S. Yu and Chuan Shi},
  title         = {Graph Foundation Models: Concepts, Opportunities and Challenges},
  journal       = {IEEE Transactions on Pattern Analysis and Machine Intelligence},
  volume        = {47},
  number        = {6},
  pages         = {5023--5044},
  year          = {2025},
  doi           = {10.1109/TPAMI.2025.3548729},
  eprint        = {2310.11829},
  archivePrefix = {arXiv}
}

@inproceedings{luo2025graphlevel,
  author        = {Yuankai Luo and Lei Shi and Xiao-Ming Wu},
  title         = {Can Classic {GNN}s Be Strong Baselines for Graph-level Tasks? {S}imple Architectures Meet Excellence},
  booktitle     = {Proceedings of the 42nd International Conference on Machine Learning},
  editor        = {Aarti Singh and Maryam Fazel and Daniel Hsu and Simon Lacoste-Julien and Felix Berkenkamp and Tegan Maharaj and Kiri Wagstaff and Jerry Zhu},
  series        = {Proceedings of Machine Learning Research},
  volume        = {267},
  pages         = {41290--41310},
  publisher     = {PMLR},
  year          = {2025},
  eprint        = {2502.09263},
  archivePrefix = {arXiv},
  primaryClass  = {cs.LG},
  url           = {https://proceedings.mlr.press/v267/luo25h.html}
}

@inproceedings{mavromatis2024gnnrag,
  author        = {Costas Mavromatis and George Karypis},
  title         = {{GNN-RAG}: Graph Neural Retrieval for Efficient Large Language Model Reasoning on Knowledge Graphs},
  booktitle     = {Findings of the Association for Computational Linguistics: {ACL} 2025},
  pages         = {16682--16699},
  publisher     = {Association for Computational Linguistics},
  year          = {2025},
  doi           = {10.18653/v1/2025.findings-acl.856},
  eprint        = {2405.20139},
  archivePrefix = {arXiv},
  url           = {https://aclanthology.org/2025.findings-acl.856/}
}

@inproceedings{Mironov2024UnbiasedHomophily,
  author        = {Mikhail Mironov and Liudmila Prokhorenkova},
  title         = {Revisiting Graph Homophily Measures},
  booktitle     = {Proceedings of the Third Learning on Graphs Conference},
  series        = {Proceedings of Machine Learning Research},
  volume        = {269},
  pages         = {1:1--1:22},
  publisher     = {PMLR},
  year          = {2025},
  eprint        = {2412.09663},
  archivePrefix = {arXiv},
  url           = {https://proceedings.mlr.press/v269/mironov25a.html}
}

@inproceedings{mishayev2025shortrange,
  author        = {Yaaqov Mishayev and Yonatan Sverdlov and Tal Amir and Nadav Dym},
  title         = {Short-Range Oversquashing},
  booktitle     = {Proceedings of the Fourth Learning on Graphs Conference},
  series        = {Proceedings of Machine Learning Research},
  publisher     = {PMLR},
  year          = {2025},
  eprint        = {2511.20406},
  archivePrefix = {arXiv},
  url           = {https://openreview.net/forum?id=rmX8Jamnyg}
}

@inproceedings{Pellizzoni2025SubstructureCounting,
  author    = {Paolo Pellizzoni and Till Hendrik Schulz and Karsten Borgwardt},
  title     = {Graph Neural Networks Can (Often) Count Substructures},
  booktitle = {International Conference on Learning Representations},
  year      = {2025},
  url       = {https://proceedings.iclr.cc/paper_files/paper/2025/hash/ef72fa6579401ffff9da246a5014f055-Abstract-Conference.html}
}

@article{peng2025graph,
  author    = {Boci Peng and Yun Zhu and Yongchao Liu and Xiaohe Bo and Haizhou Shi and Chuntao Hong and Yan Zhang and Siliang Tang},
  title     = {Graph Retrieval-Augmented Generation: A Survey},
  journal   = {ACM Transactions on Information Systems},
  volume    = {44},
  number    = {2},
  pages     = {1--52},
  publisher = {ACM},
  year      = {2025},
  doi       = {10.1145/3777378}
}

@inproceedings{qian2025macnet,
  author        = {Chen Qian and Zihao Xie and YiFei Wang and Wei Liu and Kunlun Zhu and Hanchen Xia and Yufan Dang and Zhuoyun Du and Weize Chen and Cheng Yang and Zhiyuan Liu and Maosong Sun},
  title         = {Scaling Large Language Model-based Multi-Agent Collaboration},
  booktitle     = {International Conference on Learning Representations},
  year          = {2025},
  eprint        = {2406.07155},
  archivePrefix = {arXiv},
  url           = {https://openreview.net/forum?id=K3n5jPkrU6}
}

@inproceedings{rauchwerger2025generalization,
  author        = {Levi Rauchwerger and Stefanie Jegelka and Ron Levie},
  title         = {Generalization, Expressivity, and Universality of Graph Neural Networks on Attributed Graphs},
  booktitle     = {International Conference on Learning Representations},
  year          = {2025},
  eprint        = {2411.05464},
  archivePrefix = {arXiv},
  url           = {https://openreview.net/forum?id=qKgd7RaAem}
}

@inproceedings{sancak2024geco,
  author        = {Kaan Sancak and Zhigang Hua and Jin Fang and Yan Xie and Andrey Malevich and Bo Long and Muhammed Fatih Balin and {\"U}mit V. {\c{C}}ataly{\"u}rek},
  title         = {A Scalable and Effective Alternative to Graph {Transformers}},
  booktitle     = {Proceedings of the AAAI Conference on Artificial Intelligence},
  volume        = {39},
  number        = {19},
  pages         = {20255--20263},
  publisher     = {AAAI Press},
  year          = {2025},
  doi           = {10.1609/aaai.v39i19.34231},
  eprint        = {2406.12059},
  archivePrefix = {arXiv},
  url           = {https://ojs.aaai.org/index.php/AAAI/article/view/34231}
}

@inproceedings{shi2025cfrag,
  author        = {Teng Shi and Jun Xu and Xiao Zhang and Xiaoxue Zang and Kai Zheng and Yang Song and Han Li},
  title         = {Retrieval Augmented Generation with Collaborative Filtering for Personalized Text Generation},
  booktitle     = {Proceedings of the 48th International ACM SIGIR Conference on Research and Development in Information Retrieval},
  pages         = {1294--1304},
  year          = {2025},
  doi           = {10.1145/3726302.3730075},
  eprint        = {2504.05731},
  archivePrefix = {arXiv}
}

@inproceedings{Southern2025HyMN,
  author    = {Joshua Southern and Yam Eitan and Guy Bar-Shalom and Michael M. Bronstein and Haggai Maron and Fabrizio Frasca},
  title     = {Balancing Efficiency and Expressiveness: Subgraph {GNN}s with Walk-Based Centrality},
  booktitle = {Proceedings of the 42nd International Conference on Machine Learning},
  series    = {Proceedings of Machine Learning Research},
  volume    = {267},
  pages     = {56615--56649},
  publisher = {PMLR},
  year      = {2025},
  url       = {https://proceedings.mlr.press/v267/southern25a.html}
}

@inproceedings{southern2025virtualnodes,
  author        = {Joshua Southern and Francesco Di Giovanni and Michael M. Bronstein and Johannes F. Lutzeyer},
  title         = {Understanding Virtual Nodes: Oversquashing and Node Heterogeneity},
  booktitle     = {International Conference on Learning Representations},
  year          = {2025},
  eprint        = {2405.13526},
  archivePrefix = {arXiv},
  url           = {https://openreview.net/forum?id=NmcOAwRyH5}
}

@inproceedings{tang2025uniaug,
  author        = {Wenzhuo Tang and Haitao Mao and Danial Dervovic and Ivan Brugere and Saumitra Mishra and Yuying Xie and Jiliang Tang},
  title         = {Cross-Domain Graph Data Scaling: A Showcase with Diffusion Models},
  booktitle     = {Advances in Neural Information Processing Systems},
  volume        = {38},
  pages         = {123619--123648},
  year          = {2025},
  eprint        = {2406.01899},
  archivePrefix = {arXiv},
  url           = {https://proceedings.neurips.cc/paper_files/paper/2025/hash/b2fa07e58da29247ddea5a12f472bb0e-Abstract-Conference.html}
}

@inproceedings{tieu2025invariantlink,
  author        = {Katherine Tieu and Dongqi Fu and Jun Wu and Jingrui He},
  title         = {Invariant Link Selector for Spatial-Temporal Out-of-Distribution Problem},
  booktitle     = {Proceedings of The 28th International Conference on Artificial Intelligence and Statistics},
  series        = {Proceedings of Machine Learning Research},
  volume        = {258},
  pages         = {4753--4761},
  year          = {2025},
  eprint        = {2505.24178},
  archivePrefix = {arXiv},
  url           = {https://proceedings.mlr.press/v258/tieu25a.html}
}

@inproceedings{vadgama2025probing,
  author        = {Sharvaree Vadgama and Mohammad Mohaiminul Islam and Domas Buracas and Christian A. Shewmake and Artem Moskalev and Erik J. Bekkers},
  title         = {Probing Equivariance and Symmetry Breaking in Convolutional Networks},
  booktitle     = {Advances in Neural Information Processing Systems},
  volume        = {38},
  pages         = {121445--121481},
  year          = {2025},
  eprint        = {2501.01999},
  archivePrefix = {arXiv},
  primaryClass  = {cs.LG},
  url           = {https://proceedings.neurips.cc/paper_files/paper/2025/hash/afb4b81632998eed0357939db825aa0d-Abstract-Conference.html}
}

@misc{vasileiou2025survey,
  author        = {Antonis Vasileiou and Stefanie Jegelka and Ron Levie and Christopher Morris},
  title         = {Survey on Generalization Theory for Graph Neural Networks},
  year          = {2025},
  eprint        = {2503.15650},
  archivePrefix = {arXiv},
  url           = {https://arxiv.org/abs/2503.15650}
}

@inproceedings{Wang2024DeGTA,
  author        = {Xiaotang Wang and Yun Zhu and Haizhou Shi and Yongchao Liu and Chuntao Hong},
  title         = {Graph Triple Attention Networks: A Decoupled Perspective},
  booktitle     = {Proceedings of the 31st ACM SIGKDD Conference on Knowledge Discovery and Data Mining},
  pages         = {1527--1538},
  publisher     = {ACM},
  year          = {2025},
  doi           = {10.1145/3690624.3709223},
  eprint        = {2408.07654},
  archivePrefix = {arXiv},
  url           = {https://doi.org/10.1145/3690624.3709223}
}

@misc{wang2025gfmsurvey,
  author        = {Zehong Wang and Zheyuan Liu and Tianyi Ma and Jiazheng Li and Zheyuan Zhang and Xingbo Fu and Yiyang Li and Zhengqing Yuan and Wei Song and Yijun Ma and Qingkai Zeng and Xiusi Chen and Jianan Zhao and Jundong Li and Meng Jiang and Pietro Li{\`o} and Nitesh Chawla and Chuxu Zhang and Yanfang Ye},
  title         = {Graph Foundation Models: A Comprehensive Survey},
  year          = {2025},
  eprint        = {2505.15116},
  archivePrefix = {arXiv},
  url           = {https://arxiv.org/abs/2505.15116}
}

@inproceedings{wang2025signed,
  author        = {Jiaqi Wang and Xinyi Wu and James Cheng and Yifei Wang},
  title         = {A Signed Graph Approach to Understanding and Mitigating Oversmoothing},
  booktitle     = {Advances in Neural Information Processing Systems},
  volume        = {38},
  pages         = {130778--130813},
  year          = {2025},
  eprint        = {2502.11394},
  archivePrefix = {arXiv},
  url           = {https://proceedings.neurips.cc/paper_files/paper/2025/hash/bda88ed2892f5e61c9a9bf215c566913-Abstract-Conference.html}
}

@article{zhang2024personalization,
  author  = {Zhehao Zhang and Ryan A. Rossi and Branislav Kveton and Yijia Shao and Diyi Yang and Hamed Zamani and Franck Dernoncourt and Joe Barrow and Tong Yu and Sungchul Kim and Ruiyi Zhang and Jiuxiang Gu and Tyler Derr and Hongjie Chen and Junda Wu and Xiang Chen and Zichao Wang and Subrata Mitra and Nedim Lipka and Nesreen K. Ahmed and Yu Wang},
  title   = {Personalization of Large Language Models: A Survey},
  journal = {Transactions on Machine Learning Research},
  year    = {2025},
  issn    = {2835-8856},
  url     = {https://openreview.net/forum?id=tf6A9EYMo6},
  note    = {Survey Certification}
}

@inproceedings{zhang2025gdesigner,
  author        = {Guibin Zhang and Yanwei Yue and Xiangguo Sun and Guancheng Wan and Miao Yu and Junfeng Fang and Kun Wang and Tianlong Chen and Dawei Cheng},
  title         = {{G-Designer}: Architecting Multi-agent Communication Topologies via Graph Neural Networks},
  booktitle     = {Proceedings of the 42nd International Conference on Machine Learning},
  series        = {Proceedings of Machine Learning Research},
  volume        = {267},
  pages         = {76678--76692},
  publisher     = {PMLR},
  year          = {2025},
  eprint        = {2410.11782},
  archivePrefix = {arXiv},
  url           = {https://proceedings.mlr.press/v267/zhang25cu.html}
}

@inproceedings{amran2026graphop,
  author        = {Ofek Amran and Tom Gilat and Ron Levie},
  title         = {A Graphop Analysis of Graph Neural Networks on Sparse Graphs: Generalization and Universal Approximation},
  booktitle     = {International Conference on Machine Learning},
  year          = {2026},
  eprint        = {2602.08785},
  archivePrefix = {arXiv},
  url           = {https://arxiv.org/abs/2602.08785},
  note          = {Accepted at ICML 2026 (co-author-confirmed); PMLR proceedings volume/pages not yet assigned}
}

@inproceedings{arnaiz2026demystifying,
  author        = {Adri{\'a}n Arnaiz-Rodr{\'i}guez and Federico Errica},
  title         = {Oversmoothing, ``Oversquashing'', Heterophily, Long-Range, and more: Demystifying Common Beliefs in Graph Machine Learning},
  booktitle     = {International Conference on Learning Representations},
  year          = {2026},
  eprint        = {2505.15547},
  archivePrefix = {arXiv},
  url           = {https://openreview.net/forum?id=Q3MisVkuTu}
}

@inproceedings{blayney2026glstm,
  author        = {Hugh Blayney and {\'A}lvaro Arroyo and Xiaowen Dong and Michael M. Bronstein},
  title         = {{gLSTM}: Mitigating Over-Squashing by Increasing Storage Capacity},
  booktitle     = {International Conference on Learning Representations},
  year          = {2026},
  eprint        = {2510.08450},
  archivePrefix = {arXiv},
  url           = {https://openreview.net/forum?id=c4mXEOXlAL}
}

@inproceedings{Du2026LGAN,
  author        = {Lin Du and Lu Bai and Jincheng Li and Lixin Cui and Hangyuan Du and Lichi Zhang and Yuting Chen and Zhao Li},
  title         = {{LGAN}: An Efficient High-Order Graph Neural Network via the Line Graph Aggregation},
  booktitle     = {Proceedings of the AAAI Conference on Artificial Intelligence},
  volume        = {40},
  number        = {25},
  pages         = {20914--20922},
  year          = {2026},
  doi           = {10.1609/aaai.v40i25.39232},
  eprint        = {2512.10735},
  archivePrefix = {arXiv},
  url           = {https://ojs.aaai.org/index.php/AAAI/article/view/39232}
}

@inproceedings{gupta2026condensationreset,
  author        = {Mridul Gupta and Samyak Jain and Vansh Ramani and Hariprasad Kodamana and Sayan Ranu},
  title         = {Position: Graph Condensation Needs a Reset --- Move Beyond Full-dataset Training and Model-Dependence},
  booktitle     = {Proceedings of the 43rd International Conference on Machine Learning},
  series        = {Proceedings of Machine Learning Research},
  publisher     = {PMLR},
  year          = {2026},
  eprint        = {2605.18893},
  archivePrefix = {arXiv},
  url           = {https://arxiv.org/abs/2605.18893}
}

@misc{jiang2026spectralneither,
  author        = {Qin Jiang and Chengjia Wang and Michael Lones and Dongdong Chen and Wei Pang},
  title         = {Position: Spectral {GNN}s Are Neither Spectral Nor Superior for Node Classification},
  year          = {2026},
  doi           = {10.48550/arXiv.2603.19091},
  eprint        = {2603.19091},
  archivePrefix = {arXiv},
  primaryClass  = {cs.LG},
  url           = {https://arxiv.org/abs/2603.19091}
}

@inproceedings{keriven2025backward,
  author        = {Nicolas Keriven},
  title         = {Backward Oversmoothing: why is it hard to train deep Graph Neural Networks?},
  booktitle     = {Proceedings of the 43rd International Conference on Machine Learning},
  series        = {Proceedings of Machine Learning Research},
  publisher     = {PMLR},
  year          = {2026},
  eprint        = {2505.16736},
  archivePrefix = {arXiv},
  url           = {https://arxiv.org/abs/2505.16736}
}

@misc{kormann2026dontbeafraid,
  author        = {Niklas Kormann and Benjamin Doerr and Johannes F. Lutzeyer},
  title         = {Position: Don't be Afraid of Over-Smoothing And Over-Squashing},
  year          = {2026},
  eprint        = {2601.07419},
  archivePrefix = {arXiv},
  url           = {https://arxiv.org/abs/2601.07419}
}

@inproceedings{lai2025auditvotes,
  author        = {Yuni Lai and Yulin Zhu and Yixuan Sun and Yulun Wu and Bin Xiao and Gaolei Li and Jianhua Li and Qi Xie and Kai Zhou},
  title         = {{AuditVotes}: Elevating Provable Defense for {GNNs} with Efficient Augmentation and Conditional Smoothing},
  booktitle     = {Proceedings of the ACM SIGSAC Conference on Computer and Communications Security},
  year          = {2026},
  eprint        = {2503.22998},
  archivePrefix = {arXiv},
  primaryClass  = {cs.LG},
  url           = {https://arxiv.org/abs/2503.22998},
  note          = {Accepted at ACM CCS 2026}
}

@inproceedings{luan2024heterophily,
  author        = {Sitao Luan and Qincheng Lu and Chenqing Hua and Xinyu Wang and Jiaqi Zhu and Xiao-Wen Chang},
  title         = {Revealing the Pitfalls and Re-Evaluating the Advancement of Heterophilic Graph Learning},
  booktitle     = {35th International Conference on Artificial Neural Networks (ICANN)},
  year          = {2026},
  eprint        = {2409.05755},
  archivePrefix = {arXiv},
  url           = {https://arxiv.org/abs/2409.05755}
}

@inproceedings{RubioMadrigal2026FAF,
  author        = {Celia Rubio-Madrigal and Rebekka Burkholz},
  title         = {Fixed Aggregation Features Can Rival {GNNs}},
  booktitle     = {Proceedings of the 43rd International Conference on Machine Learning},
  series        = {Proceedings of Machine Learning Research},
  publisher     = {PMLR},
  year          = {2026},
  eprint        = {2601.19449},
  archivePrefix = {arXiv},
  primaryClass  = {cs.LG},
  url           = {https://arxiv.org/abs/2601.19449}
}

@misc{vieira2026ensembles,
  author        = {Pedro C. Vieira and Pedro Ribeiro and Viacheslav Borovitskiy},
  title         = {Do Deep Ensembles Actually Capture Uncertainty in Graph Neural Networks?},
  year          = {2026},
  doi           = {10.48550/arXiv.2605.22593},
  eprint        = {2605.22593},
  archivePrefix = {arXiv},
  primaryClass  = {cs.LG},
  url           = {https://arxiv.org/abs/2605.22593}
}

@misc{vonpichowski2026nodefeatures,
  author        = {Jan von Pichowski and Al{\v{z}}beta Hrabo{\v{s}}ov{\'a} and Ingo Scholtes and Christopher Bl{\"o}cker},
  title         = {The Role of Node Features in Graph Pooling},
  year          = {2026},
  doi           = {10.48550/arXiv.2605.06250},
  eprint        = {2605.06250},
  archivePrefix = {arXiv},
  primaryClass  = {cs.LG},
  url           = {https://arxiv.org/abs/2605.06250}
}

@inproceedings{Wang2026FSpecGNN,
  author        = {Xiaohan Wang and Deyu Bo and Longlong Li and Kelin Xia},
  title         = {Full-Spectrum Graph Neural Networks: Expressive and Scalable},
  booktitle     = {Proceedings of the 43rd International Conference on Machine Learning},
  series        = {Proceedings of Machine Learning Research},
  publisher     = {PMLR},
  year          = {2026},
  eprint        = {2605.05759},
  archivePrefix = {arXiv},
  primaryClass  = {cs.LG},
  url           = {https://arxiv.org/abs/2605.05759}
}

@article{wang2026graphshelpllms,
  author        = {Xiyuan Wang and Yi Hu and Yanbo Wang and Chuan Shi and Muhan Zhang},
  title         = {Position: How can Graphs Help Large Language Models?},
  journal       = {Frontiers of Computer Science},
  year          = {2026},
  doi           = {10.1007/s11704-026-51651-6},
  eprint        = {2605.02452},
  archivePrefix = {arXiv},
  url           = {https://arxiv.org/abs/2605.02452},
  note          = {Just Accepted, published online 24 Apr 2026; volume/issue/pages not yet assigned}
}

@misc{zhang2026egs,
  author        = {Ding Zhang and Siddharth Betala and Chirag Agarwal},
  title         = {Quantifying Explanation Quality in Graph Neural Networks using Out-of-Distribution Generalization},
  year          = {2026},
  doi           = {10.48550/arXiv.2602.07708},
  eprint        = {2602.07708},
  archivePrefix = {arXiv},
  primaryClass  = {cs.LG},
  url           = {https://arxiv.org/abs/2602.07708}
}

@inproceedings{zhang2026measuring,
  author        = {Kaicheng Zhang and Piero Deidda and Desmond J. Higham and Francesco Tudisco},
  title         = {Are We Measuring Oversmoothing in Graph Neural Networks Correctly?},
  booktitle     = {International Conference on Learning Representations},
  year          = {2026},
  eprint        = {2502.04591},
  archivePrefix = {arXiv},
  url           = {https://openreview.net/forum?id=r3DOISnvHD}
}

@misc{bronstein2021geometric,
  author        = {Michael M. Bronstein and Joan Bruna and Taco Cohen and Petar Veli{\v{c}}kovi{\'c}},
  title         = {Geometric Deep Learning: Grids, Groups, Graphs, Geodesics, and Gauges},
  year          = {2021},
  eprint        = {2104.13478},
  archivePrefix = {arXiv},
  url           = {https://arxiv.org/abs/2104.13478},
  note          = {arXiv preprint arXiv:2104.13478}
}

@misc{velickovic2022message,
  author        = {Petar Veli{\v{c}}kovi{\'c}},
  title         = {Message passing all the way up},
  year          = {2022},
  eprint        = {2202.11097},
  archivePrefix = {arXiv},
  url           = {https://openreview.net/forum?id=Bc8GiEZkTe5},
  note          = {ICLR 2022 Workshop on Geometrical and Topological Representation Learning}
}

@article{chami2022machine,
  author  = {Ines Chami and Sami Abu-El-Haija and Bryan Perozzi and Christopher R{\'e} and Kevin Murphy},
  title   = {Machine Learning on Graphs: A Model and Comprehensive Taxonomy},
  journal = {Journal of Machine Learning Research},
  volume  = {23},
  number  = {89},
  pages   = {1--64},
  year    = {2022},
  url     = {https://jmlr.org/papers/v23/20-852.html}
}
\newpage
\appendix
\clearpage
\phantomsection
\addcontentsline{toc}{section}{Appendix}
\appendixsection{Notation}
\label{sec:notation}

\begingroup
\small
\setlength{\tabcolsep}{3pt}
\renewcommand{\arraystretch}{1.08}
\captionsetup{font=small,skip=2pt}

\begin{longtable}{L{0.30\textwidth}L{0.64\textwidth}}
\caption{\textbf{Notation used throughout the paper.} Part~I lists standard graph, spectral,
and learning symbols; Part~II lists the unified layer-slot notation of Eq.~\eqref{eq:unified-notation}.}
\label{tab:notation}\\
\toprule
\textbf{Symbol} & \textbf{Meaning} \\
\midrule
\endfirsthead

\toprule
\textbf{Symbol} & \textbf{Meaning} \\
\midrule
\endhead

\bottomrule
\endlastfoot

\rowcolor{meDomain!10}
\multicolumn{2}{>{\raggedright\arraybackslash}p{0.94\textwidth}}{\textbf{\medomain{Part~I\;\textbar\;Standard notation.}} Graph objects, matrices, spectral quantities, features, and learning conventions following established usage in spectral graph theory and graph neural networks.} \\

\addlinespace[3pt]
\multicolumn{2}{l}{\textit{Graph structure}} \\
\addlinespace[1pt]
$\mathcal{G}=(\mathcal{V},\mathcal{E})$ & Graph with node set $\mathcal{V}$ and edge set $\mathcal{E}$ \\
$n=|\mathcal{V}|$,\; $m=|\mathcal{E}|$ & Number of nodes and edges \\
$\mathcal{N}(v)$ & Neighborhood of node $v$ \\
$d_v$ & Degree of node $v$ \\

\addlinespace[3pt]
\multicolumn{2}{l}{\textit{Adjacency and degree matrices}} \\
\addlinespace[1pt]
$\mathbf{A}\in\mathbb{R}^{n\times n}$ & Adjacency matrix; $A_{uv}=1$ (unweighted) or $A_{uv}=w_{uv}$ (weighted) \\
$\mathbf{D}$ & Degree matrix, diagonal with $D_{vv}=d_v$ \\
$\tilde{\mathbf{A}}=\mathbf{A}+\mathbf{I}$ & Self-loop-augmented adjacency \\
$\tilde{\mathbf{D}}$ & Degree matrix of $\tilde{\mathbf{A}}$ \\
$\hat{\mathbf{A}}=\tilde{\mathbf{D}}^{-1/2}\tilde{\mathbf{A}}\tilde{\mathbf{D}}^{-1/2}$ & Symmetric normalized adjacency with self-loops \\
$\bar{\mathbf{A}}=\mathbf{D}^{-1/2}\mathbf{A}\mathbf{D}^{-1/2}$ & Symmetric normalized adjacency without self-loops (self-loop-free) \\

\addlinespace[3pt]
\multicolumn{2}{l}{\textit{Laplacians}} \\
\addlinespace[1pt]
$\mathbf{L}=\mathbf{D}-\mathbf{A}$ & Combinatorial (unnormalized) graph Laplacian \\
$\mathbf{L}_{\mathrm{sym}}=\mathbf{I}-\bar{\mathbf{A}}=\mathbf{I}-\mathbf{D}^{-1/2}\mathbf{A}\mathbf{D}^{-1/2}$ & Symmetric normalized Laplacian; spectrum in $[0,2]$ for undirected graphs with nonnegative weights \\
$\mathbf{L}_{\mathrm{ref}}=\mathbf{I}-\hat{\mathbf{A}}$ & Reference propagation Laplacian (self-loop normalized); own orthonormal eigenbasis, spectrum in $[0,2)$ (self-loops exclude the bipartite eigenvalue $2$) \\
$\tilde{\mathbf{L}}_{\mathrm{ch}}=2\mathbf{L}_{\mathrm{sym}}/\lambda_{\max}-\mathbf{I}$ & $[-1,1]$-rescaled Laplacian for Chebyshev polynomial filters \\

\addlinespace[3pt]
\multicolumn{2}{l}{\textit{Spectral quantities}} \\
\addlinespace[1pt]
$\mathbf{x}\in\mathbb{R}^{n}$ & Graph signal: node features treated as a vertex-indexed signal, one column of $\mathbf{X}$ (Section~\ref{subsec:spectral-spatial-prelim}) \\
$\mathbf{L}_{\mathrm{sym}}=\mathbf{U}\boldsymbol{\Lambda}\mathbf{U}^{\top}$ & Eigendecomposition of $\mathbf{L}_{\mathrm{sym}}$ \\
$\mathbf{U}=[\mathbf{u}_1,\ldots,\mathbf{u}_n]$ & Graph Fourier basis; orthonormal eigenvectors of $\mathbf{L}_{\mathrm{sym}}$ \\
$\widehat{\mathbf{x}}=\mathbf{U}^{\top}\mathbf{x}$,\; $\mathbf{x}=\mathbf{U}\widehat{\mathbf{x}}$ & Graph Fourier transform and its inverse, mapping a signal between node and frequency domains \\
$\boldsymbol{\Lambda}=\mathrm{diag}(\lambda_1,\ldots,\lambda_n)$ & Eigenvalues, $0=\lambda_1\leq\cdots\leq\lambda_n$ \\
$\lambda_{\max}=\lambda_n$ & Largest eigenvalue of $\mathbf{L}_{\mathrm{sym}}$ (used to rescale for Chebyshev filters) \\
$g_{\theta}(\mathbf{L}_{\mathrm{sym}})\mathbf{x}=\mathbf{U}\,g_{\theta}(\boldsymbol{\Lambda})\,\mathbf{U}^{\top}\mathbf{x}$ & Spectral graph filter \\
$\sum_{k=0}^{K}\theta_k\mathbf{L}_{\mathrm{sym}}^{k}\mathbf{x}$ & Degree-$K$ polynomial filter; localizes filtering to $K$-hop neighborhoods and avoids eigendecomposition \\
$\mathbf{P}=\mathbf{U}g(\boldsymbol{\Lambda})\mathbf{U}^{\top}=\sum_k\theta_k\mathbf{S}_k$ & Spectral propagation operator; the two forms agree only when $\mathbf{P}$ acts scalarly on each $\mathbf{L}_{\mathrm{sym}}$ eigenspace, otherwise only the spatial (support-based) form applies \\
$(\mu_i,\mathbf{v}_i)$,\; $\mu_\star=\max_{i\ge 2}|\mu_i|$ & Eigenpairs of $\hat{\mathbf{A}}$ (eigenvectors $\mathbf{v}_i$ distinct from the $\mathbf{L}_{\mathrm{sym}}$ basis $\mathbf{u}_i$) and the largest magnitude among the non-principal eigenvalues; the ordering $1=\mu_1>\cdots>-1$ and $\mu_\star<1$ use the connected-graph assumption of Section~\ref{sec:analysis}, where the self-loops in $\hat{\mathbf{A}}$ exclude the $\mu=-1$ eigenvalue (so $\mu_\star<1$ holds even on bipartite graphs) \\
$\boldsymbol{\Gamma}_j$ & Spectral projector onto the $j$-th eigenspace (used in Sections~\ref{subsec:spectral-spatial-prelim} and \ref{sec:analysis}) \\
$E(\mathbf{H})=\operatorname{tr}(\mathbf{H}^\top\mathbf{L}_{\mathrm{ref}}\mathbf{H})$ & Dirichlet energy of node states (defined with $\mathbf{L}_{\mathrm{ref}}$, not $\mathbf{L}$ or $\mathbf{L}_{\mathrm{sym}}$; see Section~\ref{sec:analysis}) \\

\addlinespace[3pt]
\multicolumn{2}{l}{\textit{Features, states, and labels}} \\
\addlinespace[1pt]
$\mathbf{X}\in\mathbb{R}^{n\times d}$ & Input node feature matrix; row $\vx_v$ is the feature vector of node $v$ \\
$\ve_{uv}$ or $\mathbf{E}$ & Edge features \\
$\mathbf{H}^{(\ell)}\in\mathbb{R}^{n\times d_\ell}$ & Hidden node representations at layer $\ell$; $\mathbf{H}^{(0)}=\mathbf{X}$ or an embedding of $\mathbf{X}$ \\
$\vh_v^{(\ell)}$ & Hidden representation of node $v$ at layer $\ell$ \\
$\mathbf{Z}$ & Final representations before prediction \\
$\hat{\mathbf{Y}}$,\; $\mathbf{Y}\in\mathbb{R}^{n\times c}$ & Predicted and ground-truth labels \\

\addlinespace[3pt]
\multicolumn{2}{l}{\textit{Homophily measures} (Section~\ref{subsec:homophily-prelim})} \\
\addlinespace[1pt]
$\mathcal{V}_{+}=\{v:|\mathcal{N}(v)|>0\}$ & Non-isolated nodes, used to average node-level homophily \\
$h_{\mathrm{edge}}$ & Edge homophily: fraction of edges connecting same-label nodes \\
$h_{\mathrm{node}}$ & Node homophily: edge homophily averaged locally over $\mathcal{V}_{+}$ \\
$h_{\mathrm{feat}}$ & Feature homophily: mean neighbor feature similarity (e.g.\ cosine) \\

\addlinespace[3pt]
\multicolumn{2}{l}{\textit{Learnable maps and operations}} \\
\addlinespace[1pt]
$\mathbf{W}^{(\ell)}$ & Weight matrix at layer $\ell$ \\
$\mathbf{a}^{(\ell)}$ & Attention parameter vector \\
$\alpha_{ij}^{(\ell)}$ & Attention coefficient from node $j$ to node $i$ \\
$\sigma(\cdot)$ & Element-wise nonlinearity \\
$\Vert$ & Concatenation \\
$\odot$ & Hadamard (element-wise) product \\

\midrule
\rowcolor{meNeighborhood!10}
\multicolumn{2}{>{\raggedright\arraybackslash}p{0.94\textwidth}}{\textbf{\meneigh{Part~II\;\textbar\;Unified layer slots.}} Eq.~\eqref{eq:unified-notation} writes a layer as $\bar{\mathbf{H}}^{(\ell+1)}=\phi_\ell(\mathcal{B}_\ell(\mathbf{H}^{(\ell)},\mathbf{H}^{(0)}),\mixmsg^{(\ell)})$ with $\mixmsg^{(\ell)}=\boxplus_{k\in\mathcal K}\mathbf{C}_k^{(\ell)}$ and message map $\Psi_k^{(\ell)}=\Psi_k^{(\ell)}(\mathbf{H}^{(\ell)},\mathbf{H}^{(0)},\mathbf{E})$. Scalar channel coefficients are absorbed into $\mathbf{P}_k^{(\ell)}$ ($\mathbf{P}_k^{(\ell)}\leftarrow\gamma_k^{(\ell)}\mathbf{P}_k^{(\ell)}$), so the mix carries no separate coefficient.} \\

\addlinespace[3pt]
$\mathcal X$ & Update domain: the object a layer updates (nodes by default) \\
$k,\;\mathcal K,\;K_\ell$ & Channel index, channel set, and channel count \\
$\mathbf{P}_{k}^{(\ell)}$ & Channel-$k$ propagation operator on the current domain \\
$\Psi_k^{(\ell)}$ & Channel-$k$ message map / value before propagation; in the linear case $\Psi_k^{(\ell)}=\mathbf{H}^{(\ell)}\mathbf{W}_k^{(\ell)}$ \\
$\mathbf{C}_k^{(\ell)}$ & Channel-$k$ contribution: row $i$ is $[\mathbf{P}_k^{(\ell)}\Psi_k^{(\ell)}]_{i:}$ for a linear-value channel, or $\operatorname{Agg}_{j}([\mathbf{P}_k^{(\ell)}]_{ij}\Psi_k^{(\ell)}(\vh_i,\vh_j,\ve_{ij}))$ for a pairwise channel \\
$n_\ell$ & Node count at layer $\ell$; equals the fixed $n$ of Part~I unless pooling coarsens the domain \\
$\gamma_k^{(\ell)}$ & Scalar channel coefficient, absorbed into $\mathbf{P}_k^{(\ell)}$ rather than carried as a separate slot \\
$\boxplus$ & Channel-fusion operator (combines channel contributions; equivariant under node relabeling) \\
$\mixmsg^{(\ell)}$ & Mixed message $\boxplus_{k\in\mathcal K}\mathbf{C}_k^{(\ell)}$; the intermediate output of the fusion operator, with no additive ego/residual term \\
$\mathcal{B}_\ell$ & Ego/residual map carrying the object's own current or initial state; jointly displayed in the update cell of the family tables \\
$\phi_\ell$ & Update map producing the node-preserving intermediate state $\bar{\mathbf{H}}^{(\ell+1)}$ (before optional pooling) from the ego/residual term and $\mixmsg^{(\ell)}$ \\
$\mathbf{Q}_\ell$ & Node-to-cluster pooling map, $\mathbf{H}^{(\ell+1)}=\mathbf{Q}_\ell\bar{\mathbf{H}}^{(\ell+1)}$; the identity unless an external pooling operation coarsens the domain \\
$\rho$ & Post-stack readout producing the task representation $\mathbf{Z}=\rho(\mathbf{H}^{(0)},\dots,\mathbf{H}^{(L)})\mathbf{W}_{\mathrm{out}}$ from the layer states (the last state for most node-level models, the full stack for jumping-knowledge designs) \\

\addlinespace[3pt]
\rowcolor{meReadout!8}
\multicolumn{2}{>{\raggedright\arraybackslash}p{0.94\textwidth}}{\textit{Compact family-table form.} The per-method tables display the seven components in six columns $(\mathcal X,\mathcal K,\mathbf P_k,\Psi_k,\boxplus,(\mathcal B_\ell,\phi_\ell))$. Scalar coefficients are folded into $\mathbf P_k$, and the final column jointly reports the distinct ego/residual and update components. The Mixing column reports $\mixmsg=\boxplus_k\mathbf C_k$. For a pairwise channel, neighbor aggregation defines $\mathbf C_k$ before $\boxplus$ acts across channels. When $\mathcal K=\{1\}$, $\boxplus=\mathrm{id}$. Wherever it appears in a cell, $\mixmsg$ denotes only the output of $\boxplus$ and is not an additional component or column.} \\

\end{longtable}

\endgroup

\appendixsection{Operations: Composable Modules}
\label{sec:operations}

The seven families of Appendix~\ref{app:family-comparison-tables} are distinguished by
\emph{which slot of the unified equation~\eqref{eq:unified-notation} carries their defining mechanism};
most touch several slots, but one is the locus of what makes the family distinct. The
modules in this section are different in kind. Each is a reusable edit with a
\emph{primary} component or external-location assignment, applicable across families rather
than defining one. A module's primary location is where it chiefly acts, not
a guarantee that it touches nothing else. This is the operative
content of the architecture/operation distinction drawn in
Section~\ref{subsec:scope}. A family answers ``what does a layer compute?''; an
operation answers ``what is bolted onto a layer, regardless of which family it
belongs to?''

We make the distinction precise using the slot vocabulary of
Section~\ref{sec:unified-view}. These seven components comprise the update domain $\mathcal{X}$, the channel set
$\mathcal{K}$, the operator bank $\{\mathbf{P}_k^{(\ell)}\}$ (with the scalar
coefficients $\gamma_k^{(\ell)}$ folded in), the message map $\Psi_k^{(\ell)}$, the
mixing $\boxplus$, the ego/residual map $\mathcal{B}_\ell$, and the update
$\phi_\ell$. The initial state $\mathbf H^{(0)}$, optional between-layer coarsening map
$\mathbf Q_\ell$, and post-stack readout $\rho$ are related locations but are not additional
members of the seven-component per-layer tuple. Each operation below names its primary
component or external location using the vocabulary and placement rules of
Section~\ref{sec:unified-view}.
As a bookkeeping rule, operations with \emph{different} primary component assignments suggest compatibility,
while two that edit the \emph{same} component (e.g.\ two residual schemes both writing
$\mathcal{B}_\ell$) must be reconciled by an explicit combination rule in that component. The
qualifier is that the components are not fully independent. An operation nominally aimed at one component
can reach another (a pairwise positional encoding can enter the attention operator $\mathbf{P}_k$,
and a state-dependent normalization changes the operator it feeds), so the map is a heuristic
predicting which add-ons are \emph{prima facie} compatible, not a theorem that different-component
edits never interact.
Table~\ref{tab:operations-slots} summarizes the mapping.

\appendixsubsection{Positional and Structural Encodings}
\label{subsec:op-encodings}

Message passing on an unattributed graph is blind to position. In an $L$-layer MPNN, two nodes
with isomorphic rooted attributed $L$-hop neighborhoods receive identical representations, which
is the combinatorial root of the expressiveness ceiling discussed in
Section~\ref{subsec:prop-expressivity}. Positional encodings (PE) and structural
encodings (SE) can break this symmetry by injecting a per-node feature that depends on
where the node sits in the graph (PE) or on the local subgraph around it (SE); whether a given
encoding actually separates two such nodes depends on the descriptor.

\paragraph{Primary location.} The message map $\Psi_k$, often through the external initial
state $\mathbf{H}^{(0)}$ that feeds it. Following the fold-in convention of
Section~\ref{sec:unified-view}, an encoding is not a new slot. A \emph{node-level} encoding
contributes a feature matrix $\mathbf{E}_{\mathrm{pos}}\in\mathbb{R}^{n\times p}$ that enters
the layer either by concatenation into the initial state,
$\mathbf{H}^{(0)}\leftarrow[\mathbf{X}\,\Vert\,\mathbf{E}_{\mathrm{pos}}]$, or by addition into
the message map $\Psi_k$; its primary component is $\Psi_k$, with $\mathbf H^{(0)}$ recorded
as the external input location when the encoding is applied before the stack. It leaves a
\emph{fixed} structural operator untouched, since the graph over which messages flow is unchanged
and only the node descriptors are augmented. It can, however, change a \emph{state-dependent} operator
$\mathbf{P}_k=\mathbf{P}_k(\mathbf{H})$ indirectly. Enriching the input feeds new descriptors into
the attention score, so the encoding reaches $\mathbf{P}_k$ through the state even though it is
introduced at the input.

A \emph{pairwise} encoding edits the operator more directly. A relative-distance or edge-structural bias that modulates the attention score
enters the operator $\mathbf{P}_k$ itself (Graphormer's spatial and edge biases~\citep{ying2021transformers},
Table~\ref{tab:family-transformer}), and a pairwise feature added on the value side enters the
pairwise message $\Psi_k$. Only a node-level encoding paired with a fixed structural operator is
operator-preserving; with a state-dependent $\mathbf{P}_k(\mathbf{H})$ even a node-level encoding
reaches the operator indirectly, as above.

\paragraph{Representative forms.} Laplacian eigenvectors give a spectral PE. The
$p$ lowest \emph{nontrivial} eigenvectors of $\mathbf{L}_{\mathrm{sym}}$
(Section~\ref{subsec:spectral-spatial-prelim}) furnish coordinates that are smooth on the graph.
These exclude the nullspace of $\mathbf{L}_{\mathrm{sym}}$, which for a graph with $c$ components is
$c$-dimensional and spanned by the degree-weighted component indicators
$\mathbf{D}^{1/2}\mathbf{1}_{C}$, reducing to the single vector $\mathbf{D}^{1/2}\mathbf{1}$ in the
connected case. The
eigenvectors carry two distinct ambiguities with distinct remedies. A simple eigenvector has a
sign ambiguity, usually mitigated by random sign flipping as a training augmentation, and resolved
structurally by a sign-invariant map. A repeated eigenvalue additionally has a basis ambiguity.
Any orthogonal rotation within its eigenspace is valid, which sign flipping cannot address, and it
requires a basis-invariant or equivariant map over the whole eigenspace.
Random-walk SE~\citep{dwivedi2022graph} uses the
diagonal return probabilities $[(\mathbf{D}^{-1}\mathbf{A})^k]_{ii}$ for
$k=1,\dots,K$, encoding the probability-weighted return over closed walks at a node (a weighted
sum, not an unweighted count). These closed walks include backtracking walks, not only simple cycles. The walk operator matters here. For a simple graph (no self-loops in $\mathbf{A}$), the bare
$\mathbf{D}^{-1}\mathbf{A}$ (defined on positive degrees, with an explicit convention for isolated
nodes) has zero diagonal, so it assigns no probability to staying in place after a single step,
whereas the augmented $\tilde{\mathbf{D}}^{-1}(\mathbf{A}+\mathbf{I})$ has a positive one-step
self-transition probability $1/(1+d_i)$ at every node. (At $k\ge2$ both operators can assign
positive return probability via a walk out and back; the distinction is specifically about the
one-step diagonal.) Distance
encodings record shortest-path or resistance distances to a set of anchor nodes, which requires
an equivariant anchor-selection rule, since arbitrarily indexed anchors would break permutation
equivariance.
Learned and graph-adaptive variants replace fixed eigenvectors with trainable
encodings~\citep{dwivedi2022graph,kreuzer2021rethinking,Ito2025LLPE,Kanatsoulis2025PEARL}.

\paragraph{Why it is an operation, not a family.} A PE/SE module composes with GCN,
with a graph Transformer, or with an equivariant network alike; in its input-augmentation
form it changes the values supplied to the components rather than introducing a new component.
Graph Transformers are prominent consumers because their dense attention has no built-in locality to
recover position from, but the module itself is family-agnostic.

\paragraph{Link to generated architectures.} Because a PE/SE module's primary edit is a filler
for $\Psi_k$, or for the external $\mathbf{H}^{(0)}$ that feeds it, adopting one reads directly
onto the three generation levels of Section~\ref{sec:generation}. Concatenating a cataloged
encoding, such as a Laplacian PE, into a family that does not natively carry one recombines an
existing filler across families and is a \textcolor{googleblue}{\bf \scshape level~1} move.
Supplying a learned or graph-adaptive encoding not yet in the inventory~\citep{Ito2025LLPE,Kanatsoulis2025PEARL}
as the $\Psi_k$ filler is instead \textcolor{googlegreen}{\bf level~2}, provided the resulting
layer does not reduce to an existing method. A pairwise encoding that reaches $\mathbf{P}_k$
rather than $\Psi_k$ is classified the same way, just against the operator slot.

\appendixsubsection{Graph Rewiring and Virtual Nodes}
\label{subsec:op-rewiring}

The motivation is oversquashing (Section~\ref{subsec:prop-oversquashing}). When many long-range
signals must pass through a structural bottleneck, message passing on the original edges
compresses them into a fixed-width vector and loses information. Rewiring addresses this by
changing the communication operator, \emph{which nodes communicate} and with what weight,
adding, removing, or reweighting edges to ease the bottleneck. It does not necessarily relieve
the bottleneck, though, and can trade against faithfulness to the input graph. This is not purely a support change.
Edge reweighting and the degree renormalization that follows adding or dropping edges also change
the operator's entries, not only its sparsity pattern.

\paragraph{Primary location.} The operator bank $\{\mathbf{P}_k^{(\ell)}\}$, by changing the
support and/or the scalar weights it is built on (and, when the rewiring adds nodes, the domain
$\mathcal{X}$), is exactly the fold-in Section~\ref{sec:unified-view} specifies for rewiring. A rewiring is
\emph{static} when the support is modified once as preprocessing (the operator bank is
rebuilt on the rewired graph and then fixed) and \emph{dynamic} when the support is
modified per layer.

\paragraph{Representative forms.} Curvature-based methods add edges at negatively
curved bottlenecks and optionally drop edges in dense
regions~\citep{Topping2022Curvature,Nguyen2023BORF}. Spectral methods rewire to
increase the spectral gap. The relevant quantity is the lazy or absolute spectral gap that
governs mixing time and the expansion-related bounds constraining oversquashing (the ordinary
$\lambda_2$-gap alone is insufficient when negative eigenvalues dominate the
spectral radius)~\citep{Karhadkar2023FoSR}. The gap is not a sole determinant, though, since a
larger gap also accelerates smoothing, so the two pathologies trade off rather than improving
together. Delay-based and dynamic schemes let the
effective support grow with depth so that distant nodes connect only in later
layers~\citep{Gutteridge2023DRew}. A \emph{virtual node} is the limiting case. It is a
single node connected to all others, which in the terms of
Section~\ref{sec:unified-view} enlarges the domain $\mathcal{X}$ and gives every pair a
two-hop path $j\to v\to i$. This is distinct from adding a dense global channel directly in
$\mathbf{P}_k$, which couples all pairs in one hop. The virtual node instead routes through an extra
vertex, so $\mathbf{h}_j$ reaches $i$ only after two synchronous layers (unless the virtual state
was already aggregated), as the one-layer mixing analysis of Appendix~\ref{sec:analysis} notes. It is also the cleanest bridge to graph
Transformers. The virtual node opens a global communication channel that can simulate global
interaction under suitable depth, width, and update capacity, though a single token is not in
general equivalent to arbitrary pairwise attention~\citep{Cai2023MPNNGTConnection}.

\paragraph{Caution.} Rewiring trades faithfulness to the input graph for
information flow. When the task actually depends on the original topology, rewiring
can hurt; the operation is justified by a diagnosed bottleneck, not applied by
default. Edge-dropping variants carry a further, more basic risk. Removing edges in
dense regions without a connectivity safeguard can disconnect a component or isolate
a node, which the addition side of curvature- and spectral-based rewiring does not
by itself prevent.

\paragraph{Link to generated architectures.} Rewiring already supplies Section~\ref{sec:generation}'s
clearest worked example. Curvature-Gated Propagation (Table~\ref{tab:generated}) lifts the discrete
curvature $\kappa_{ij}$ introduced above for edge addition and removal~\citep{Topping2022Curvature,Nguyen2023BORF}
and reuses it instead as a gate inside the propagation filler $\mathbf{P}_k$ of an otherwise
ordinary spatial layer, a \textcolor{googlegreen}{\bf level~2} move because that gated operator
is not yet in the catalogued inventory. Importing an already-cataloged rewired operator, say a
spectral-gap-increasing rewiring native to one family, as the $\{\mathbf{P}_k^{(\ell)}\}$ filler
of a different family is instead \textcolor{googleblue}{\bf \scshape level~1}. A virtual node
only fits this same recombination reading when the host layer's domain and channel set can
absorb the extra vertex without changing type. Treating it as a first-class second domain still
does not by itself produce a \textcolor{googlered}{\bf \scshape level~3} expansion because the
graded extension already admits multiple typed domains and inter-rank channels. The
Sequential Co-Lifted Node--Edge Layer in Section~\ref{subsec:generated-architectures}
reaches Level~3 only because its added schedule
makes the node update consume the newly written edge state within the same macro-layer.

\appendixsubsection{Skip and Residual Connections}
\label{subsec:op-residual}

Residual connections let a layer's output retain a copy of an earlier state. They
are the standard remedy for the vanishing-signal and oversmoothing problems that
limit depth (Section~\ref{subsec:prop-oversmoothing}). Under the contraction conditions of
Theorem~\ref{thm:oversmoothing} (notably $\bar s\,\mu_\star<1$), repeated propagation without a
residual drives all node states toward the degree-dominated subspace. An initial residual can
keep each node anchored away from that limit. The energy floor of Theorem~\ref{thm:floor} is
proved for the initial-residual (PPR) iteration specifically, and a generic residual need not by
itself guarantee escape from the smoothing subspace.

\paragraph{Primary location.} The ego/residual map $\mathcal{B}_\ell$, which
Section~\ref{sec:unified-view} routes into the update $\phi_\ell$ as the ego term it
receives. In the common case
$\mathcal{B}_\ell(\mathbf{H}^{(\ell)})=\mathbf{H}^{(\ell)}$ (identity residual). A
\emph{cross-layer skip} generalizes this to read earlier layers,
$\mathcal{B}_\ell(\mathbf H^{(0)},\ldots,\mathbf H^{(\ell)})=
\sum_{\ell'<\ell}\mathbf{S}_{\ell'}^{(\ell)}
\mathbf{H}^{(\ell')}\mathbf{W}_{\ell'}^{(\ell)}$, where
$\mathbf{W}_{\ell'}^{(\ell)}\in\mathbb{R}^{d_{\ell'}\times d_{\mathcal B,\ell}}$
maps the earlier width to the width $d_{\mathcal B,\ell}$ expected by the ego-input argument
of $\phi_\ell$ (for a common additive update, $d_{\mathcal B,\ell}=d_{\ell+1}$), and
$\mathbf{S}_{\ell'}^{(\ell)}\in\mathbb{R}^{n_\ell\times n_{\ell'}}$
is a per-layer mixing coefficient selecting which earlier states are re-injected (a scalar
$s_{\ell'}^{(\ell)}\mathbf{I}$ in the common node-preserving case, where $n_\ell=n_{\ell'}$). To
preserve permutation equivariance, $\mathbf{S}_{\ell'}^{(\ell)}$ must itself be equivariant
(in practice a scalar multiple of the identity, or a declared coarsening/alignment map when the
node set changes), since an arbitrary fixed matrix would encode node identities.
Reading every earlier state exceeds the single-step signature $\mathcal{B}_\ell(\mathbf{H}^{(\ell)},\mathbf{H}^{(0)})$
that Section~\ref{sec:unified-view} declares. The identity and initial residual fit it directly,
whereas a full multi-layer skip either requires the extended ego signature
$\mathcal{B}_\ell(\mathbf{H}^{(0)},\dots,\mathbf{H}^{(\ell)})$ or, for dense jumping-knowledge
aggregation, is deferred to the readout $\rho$ (below), where reading the whole stack is native.

\paragraph{Representative forms.} An \emph{initial residual} re-injects a fixed,
width-aligned initial representation $\mathbf{G}^{(0)}$ at every layer,
$\mathcal{B}_\ell(\mathbf H^{(\ell)},\mathbf H^{(0)})=\alpha\,\mathbf{G}^{(0)}$,
as in APPNP and GCNII (Appendix~\ref{app:family-comparison-tables}, the spectral and
polynomial-filter family). The anchor is the \emph{transformed} input, not the raw features.
APPNP teleports to $\mathbf{G}^{(0)}=f_\theta(\mathbf{X})$ (an MLP of the input, distinct from
the framework's $\mathbf{H}^{(0)}=\mathbf{X}$, and written $\mathbf{G}$ rather than the
framework's reserved final-representation $\mathbf{Z}$), and GCNII likewise anchors to an initial hidden
representation. Keeping every node tied to its own input contributes to GCNII's ability to train
deeply, alongside its identity mapping in the weight path.
A \emph{dense skip} (jumping-knowledge style) makes $\mathcal{B}_\ell$ read all
previous layers and is most naturally folded into the readout $\rho$ when the
combination is deferred to the end~\citep{xu2018representation}. \emph{Reversible}
residuals constrain $\mathcal{B}_\ell$ and $\phi_\ell$ so the forward map is
invertible, trading recomputation for memory and enabling very deep
stacks~\citep{Li2021RevGNN}.

\paragraph{Composition note.} The initial residual and the dense skip both write
$\mathcal{B}_\ell$, so they do not compose independently the way different-component operations do.
Stacking them requires an explicit combination rule in that component, for example
$\mathcal{B}_\ell(\mathbf H^{(0)},\ldots,\mathbf H^{(\ell)})
=\alpha\mathbf{G}^{(0)}+\sum_{\ell'<\ell}\mathbf{S}_{\ell'}^{(\ell)}\mathbf{H}^{(\ell')}\mathbf{W}_{\ell'}^{(\ell)}$,
rather than two edits that pass through each other untouched. This is the same-component interaction the
factorization flags. Same slot means reconcile, not necessarily choose one.

\paragraph{Link to generated architectures.} Because $\mathcal{B}_\ell$ is one of the seven
recombinable components, moving an existing residual scheme into a family that does not
natively use it is \textcolor{googleblue}{\bf \scshape level~1}: Table~\ref{tab:generated} marks
Spectrally-Gated Propagation and Curvature-Gated Propagation with the same
linear-specialization ego/update pair ($^\dagger$) precisely because both borrow the plain
identity residual for $\mathcal{B}_\ell$ rather than proposing a new one. A genuinely new ego
map, for instance a combination rule for stacked initial and dense skips (the Composition note
above) that is not already catalogued, would instead be \textcolor{googlegreen}{\bf level~2}.

\appendixsubsection{Normalization}
\label{subsec:op-normalization}

Normalization rescales or recenters representations to stabilize training and, in
the graph setting, to counter oversmoothing by preventing node states from
collapsing together.

\paragraph{Primary location.} Most commonly the update map $\phi_\ell$, but the placement
depends on where the normalization map $\nu$ is inserted. \emph{Post-update} normalization wraps the output,
$\phi_\ell\mapsto\nu\circ\phi_\ell$, and leaves the operator bank and message map untouched.
\emph{Pre-activation} normalization sits inside the update, $\phi_\ell(a,b)=\sigma(\nu(a+b))$,
still within $\phi_\ell$. \emph{Pre-normalization} of the state before propagation,
$\bar{\mathbf{H}}=\nu(\mathbf{H})$ with $\mathbf{P}_k=\mathbf{P}_k(\bar{\mathbf{H}})$ and
$\Psi_k=\Psi_k(\bar{\mathbf{H}})$, as in pre-norm transformer blocks, instead feeds a
state-dependent operator and message, so it reaches $\mathbf{P}_k$ and $\Psi_k$ rather than
$\phi_\ell$ alone. Normalization is therefore a clear case of the primary-component caveat. Its primary
component is $\phi_\ell$, but the pre-norm variant is not operator-preserving.

PairNorm, GraphNorm, and graph-level batch/instance normalizations generally couple several
node rows. When they are folded into $\phi_\ell$, they therefore use the generalized
permutation-equivariant update allowed by Theorem~\ref{thm:equivariance}, not the stricter
node-local row-wise specialization used by some results in Appendix~\ref{sec:analysis}.
Those row-wise results do not automatically extend to a cross-node normalization unless the
normalization's equivariance and the particular proof assumptions are checked separately.

\paragraph{Representative forms.} Generic batch and layer normalization carry over
from deep learning unchanged. Graph-specific variants exploit graph structure.
PairNorm centers and rescales node representations so their aggregate pairwise distance is fixed
to a prescribed scale, opposing the contraction that causes oversmoothing rather than conserving
the incoming distance across the layer~\citep{zhao2020pairnorm}. GraphNorm normalizes each graph
separately and learns a feature-wise mean-shift parameter, acting as a trainable preconditioner
that retains useful graph-level information~\citep{Cai2021GraphNorm}. DropEdge is a closely
related regularizer. It randomly removes edges during training, a stochastic perturbation of
the operator bank rather than a normalization in any exact sense~\citep{rong2020dropedge}. It
only loosely rescales propagation intensity: because the degree-normalization map $\eta$ is
nonlinear, the expected normalized operator of the dropped graph need not be a rescaling of the
clean one. $\mathbb{E}[\eta(\mathbf{A}_{\mathrm{drop}})]$ need not equal $c\,\eta(\mathbf{A})$
for any constant $c$ (such as the edge keep-probability), though special graphs or distributions
may coincide, so the ``normalization'' reading is heuristic, not an identity in expectation.

\paragraph{Link to generated architectures.} Reusing an existing graph-specific normalization as the
$\phi_\ell$ filler, or for pre-normalization the $\mathbf{P}_k/\Psi_k$ filler, of a family that
does not natively normalize that way is again \textcolor{googleblue}{\bf \scshape level~1}
recombination. A normalization rule outside the current inventory would instead be
\textcolor{googlegreen}{\bf level~2} if folded into $\phi_\ell$ under the generalized
permutation-equivariant update Theorem~\ref{thm:equivariance} allows, since PairNorm and
GraphNorm already show that a cross-node normalization can satisfy that broader hypothesis
without being row-wise.

\appendixsubsection{Pooling and Readout}
\label{subsec:op-pooling}

Graph-level tasks require a single vector per graph. Pooling produces it, either in
one shot (global readout) or hierarchically (coarsening the graph in stages).

\paragraph{Primary location.} For flat, global pooling, this is the post-stack readout $\rho$ alone, a
permutation-invariant aggregate over the final node states (Section~\ref{subsec:symmetry-prelim}).
For hierarchical pooling, the affected core components are the per-layer domain $\mathcal{X}$
and operator bank $\{\mathbf{P}_k^{(\ell)}\}$. They are edited between layers through the
auxiliary coarsening map $\mathbf{Q}_\ell$ below. This map is outside the seven-component
per-layer tuple and is also distinct from the post-stack readout $\rho$; $\rho$ still acts only once, at the end, on whatever
domain the coarsening has produced by the final layer.

\paragraph{Representative forms.} Flat readouts sum, average, or max node states; the
sum is the form that can be made injective on multisets. The raw sum is not itself injective
(distinct multisets can share a total), and an injective pointwise encoding does not by itself
fix this (its summed values can still collide). For bounded-size multisets over a countable state
space, however, there is a \emph{specially chosen} encoding $f$ for which
$S\mapsto\sum_{x\in S}f(x)$ is injective, and an injective post-map then preserves the
distinction. This is the property GIN~\citep{xu2019powerful} relies on for expressiveness.
Learned hierarchical poolings include differentiable soft cluster assignment, which
learns a soft map of nodes to a smaller node set~\citep{ying2018hierarchical};
score-based node selection, which keeps a top-$k$ subset by a learned
score~\citep{lee2019self,gao2019graph}; and spectral methods that pool by a relaxed
min-cut objective~\citep{bianchi2020mincut}. The choice interacts with the readout.
Hierarchical pooling changes the domain on which $\rho$ ultimately acts.

\paragraph{Formal mechanics.} Hierarchical pooling is written in the unified equation as a
node-to-cluster map $\mathbf{Q}_\ell\in\mathbb{R}^{n_{\ell+1}\times n_\ell}$ applied to the
node-preserving intermediate state, $\mathbf{H}^{(\ell+1)}=\mathbf{Q}_\ell\bar{\mathbf{H}}^{(\ell+1)}$
(Section~\ref{subsec:setup}). For assignment-based pooling a common choice is
$\mathbf{A}^{(\ell+1)}=\mathbf{Q}_\ell\mathbf{A}^{(\ell)}\mathbf{Q}_\ell^\top$, followed by
reconstruction of $\{\mathbf{P}_k^{(\ell+1)}\}$; selection-based and topology-aware poolings
instead define the next graph directly. For the layer to stay equivariant, $\mathbf{Q}_\ell$
must itself be computed equivariantly. Under a source-node relabeling $\boldsymbol{\Pi}$ and
the induced target-cluster relabeling $\boldsymbol{\Pi}_{\mathrm{out}}$,
$\mathbf{Q}_\ell(\boldsymbol{\Pi}\mathbf{A}\boldsymbol{\Pi}^\top,\boldsymbol{\Pi}\mathbf{H})=\boldsymbol{\Pi}_{\mathrm{out}}\,\mathbf{Q}_\ell(\mathbf{A},\mathbf{H})\,\boldsymbol{\Pi}^\top$;
a fixed arbitrary $\mathbf{Q}_\ell$ would instead encode node identities and break the symmetry.
When an initial residual continues across pooling, the teleport target is realigned to the
current domain by the composition of the coarsening maps applied so far.
This composition is $\mathbf{Q}_{\ell-1}\cdots\mathbf{Q}_0\mathbf{H}^{(0)}$, where each $\mathbf{Q}_t$ maps layer
$t$'s output domain to layer $t{+}1$'s. This realignment carries a feature projection when
$d_0\neq d_{\ell+1}$.

\paragraph{Link to generated architectures.} Flat pooling acts only through the external readout
$\rho$, which none of the three generation levels of Section~\ref{sec:generation} constrain, so
choosing or designing $\rho$ is orthogonal to architecture generation. It can be paired with a Level~1,
Level~2, or Level~3 layer without itself counting as a slot-filling move. Hierarchical pooling
is the partial exception. Because it edits $\mathcal{X}$ and $\{\mathbf{P}_k^{(\ell)}\}$ between
layers through $\mathbf{Q}_\ell$, adopting an existing coarsening scheme in a family that does
not natively pool is the same \textcolor{googleblue}{\bf \scshape level~1} recombination as for
any other per-layer component, while a genuinely new equivariant $\mathbf{Q}_\ell$ would be
\textcolor{googlegreen}{\bf level~2}.

\appendixsubsection{Continuous-Depth and ODE Views}
\label{subsec:op-continuous}

A residual layer $\mathbf{H}^{(\ell+1)}=\mathbf{H}^{(\ell)}+h\,f(\mathbf{H}^{(\ell)})$
is one explicit Euler step of size $h$ of the ODE
$\dot{\mathbf{H}}(t)=f(\mathbf{H}(t))$ (the standard residual is the unit step $h=1$). The
continuous-depth view takes this
seriously. Depth becomes integration time, and the layer update is read as a
discretized dynamical system rather than a fixed stack.

\paragraph{Primary location.} In the narrow case, a pure discrete-to-continuous reinterpretation, this is the
update $\phi_\ell$. The \emph{same} propagation operators $\{\mathbf{P}_k^{(\ell)}\}$ supply the
right-hand-side dynamics, and only the integrator that assembles them into a depth-varying map
changes. This is why we class it as an operation rather than a family. Broader ODE architectures go further than
the integrator and edit other slots too. A second-order or otherwise augmented dynamics enlarges
the per-object \emph{state} (a velocity channel appended to $\mathbf{H}$, on the same node domain
$\mathcal{X}$, not a new set of graph objects). Such dynamics can also change the message $\Psi_k$, the update
$\phi_\ell$, or the operator $\mathbf{P}_k$, the components that define the vector field. The domain $\mathcal{X}$ changes only
if new objects are introduced. The classification is clean only for the
reinterpretation; the augmented case spans several components.

\paragraph{Representative forms.} Diffusion-based models set $f$ to a graph diffusion
and integrate it, recovering smoothing as heat
flow~\citep{Chamberlain2021GRAND,xhonneux2020continuous,Chamberlain2021BLEND,choi2023gread,Zhao2023CDEGRAND}; oscillator-based models add
a second-order term so the dynamics do not collapse to the diffusion fixed point,
which is a continuous-depth answer to oversmoothing~\citep{rusch2022graphcon}.

\paragraph{Scope caveat.} We catalog the continuous-depth \emph{reinterpretation} as
an operation on $\phi_\ell$. We do not claim the resulting models are equivalent to
their discrete counterparts in expressiveness or stability; those are properties of
the specific vector field and integrator, established in the cited works. This is
also the one place where our operation/family line is a modeling choice rather than a
fact. A reader who regards the integrator as the defining computation may prefer to
read these as a family. We flag the choice rather than hide it.

\paragraph{Link to generated architectures.} The narrow reinterpretation changes only the
integrator, not any filler, so by itself it does not correspond to any of the three generation
levels of Section~\ref{sec:generation}. The same catalogued $\{\mathbf{P}_k^{(\ell)}\}$ and
$\Psi_k$ are reused, just assembled by a different depth-varying map. The augmented dynamics are
the case that does connect. Appending a velocity channel to the state and letting it drive
$\Psi_k$, $\mathbf{P}_k$, or $\phi_\ell$ is \textcolor{googlegreen}{\bf level~2} if the resulting
filler is new and does not reduce to an existing method, since the domain $\mathcal{X}$ itself
is unchanged. We do not classify it as \textcolor{googlered}{\bf \scshape level~3}, because
Level~3 is reserved for an added, typed, separately removable component, not an enlarged
per-node state.

\appendixsubsection{Mini-Batch Sampling}
\label{subsec:op-sampling}

Full-batch training materializes the entire graph each step, which is infeasible at
scale. Sampling restricts each step to a subgraph, trading exact propagation for
tractable memory and runtime.

\paragraph{Primary location.} The operator bank $\{\mathbf{P}_k^{(\ell)}\}$, restricted to
a sampled support. Sampling does not change the layer's functional form; it changes the
support the function is evaluated on, and it is usually a \emph{training-time} operation, though
neighbor sampling is also used at inference on graphs too large to propagate in full. An
appropriately reweighted estimator on the sampled support (importance weights correcting for the
sampling distribution, not the bare sampler) makes the propagated pre-activation
$\widehat{\mathbf{m}}$ an unbiased estimate of the full aggregate. This does not make the
layer output unbiased. For a nonlinear update,
$\mathbb{E}[\sigma(\widehat{\mathbf{m}})]\neq\sigma(\mathbb{E}[\widehat{\mathbf{m}}])$ in general,
so neighbor and cluster sampling approximate rather than reproduce the full-graph forward pass.

\paragraph{Representative forms.} Node-wise sampling draws a fixed number of
neighbors per node and per layer~\citep{hamilton2017inductive,chen2018stochastic}; layer-wise sampling
draws a shared set of nodes per layer to control the neighbor-explosion of stacking
node-wise samplers~\citep{chen2018fastgcn,zou2019ladies}; subgraph sampling trains on
induced subgraphs, restricting the support to a sampled piece that is connected for some
samplers (cluster-based) but need not be in general (random-node and random-edge
samplers)~\citep{chiang2019cluster,zeng2020graphsaint}. The axis that separates these is
which estimator they use and whether it is unbiased, a property of the sampling distribution,
estimator, and importance/normalization weights jointly rather than of any single component.

\paragraph{Link to generated architectures.} Sampling restricts the support on which
$\{\mathbf{P}_k^{(\ell)}\}$ is evaluated without changing its functional form, so sampling
itself is not a generation move. A generated architecture at any of the three levels can be
trained under node-wise, layer-wise, or subgraph sampling without changing its level.

\colorlet{opsBar}{googleyellow}
\colorlet{opsRow}{googleyellow!9}
\begin{table}[t]
\centering
\caption{Operations mapped to their \emph{primary location}: a component of
Eq.~\eqref{eq:unified-notation}, an external input/between-layer map, or the post-stack readout
$\rho$. Different primary component assignments suggest compatibility; edits sharing a
component require reconciliation. ``Time''
marks whether the edit is typically active at inference or mainly during training.}
\label{tab:operations-slots}
\small
\renewcommand{\arraystretch}{1.2}
\begin{tabularx}{\textwidth}{L{0.20\textwidth} L{0.20\textwidth} c X}
\toprule
\rowcolor{opsBar}
{\color{white}\textbf{Operation}} & {\color{white}\textbf{Primary location}} & {\color{white}\textbf{Time}} & {\color{white}\textbf{Effect}} \\
\midrule
Positional / structural encoding
& $\Psi_k$; external input $\mathbf{H}^{(0)}$ for pre-stack encoding; $\mathbf{P}_k$ for pairwise score biases
& infer
& Adds position/structure descriptors; can break neighborhood-isomorphism symmetry. \\
\addlinespace[2pt]
\rowcolor{opsRow}
Rewiring / virtual node
& $\{\mathbf{P}_k^{(\ell)}\}$ (support/weights; $\mathcal{X}$ if nodes added)
& infer
& Changes which nodes communicate; can relieve oversquashing bottlenecks. \\
\addlinespace[2pt]
Skip / residual
& $\mathcal{B}_\ell$
& infer
& Re-injects earlier state; enables depth, can counter oversmoothing. \\
\addlinespace[2pt]
\rowcolor{opsRow}
Normalization
& $\phi_\ell$ primarily; $\mathbf{P}_k,\Psi_k$ for state pre-norm
& infer
& Rescales/recenters states; stabilizes training, can oppose collapse. \\
\addlinespace[2pt]
Pooling / readout
& external $\rho$ for flat readout; $\mathcal{X}$/$\{\mathbf{P}_k^{(\ell)}\}$ via external $\mathbf Q_\ell$ if hierarchical
& infer
& Produces graph-level vector; coarsens support in stages. \\
\addlinespace[2pt]
\rowcolor{opsRow}
Continuous-depth / ODE
& $\phi_\ell$ primarily; $\Psi_k,\mathbf{P}_k$ for augmented dynamics; $\mathcal{X}$ only if objects are added
& infer
& Reads depth as integration time over a chosen graph vector field. \\
\addlinespace[2pt]
Mini-batch sampling
& $\{\mathbf{P}_k^{(\ell)}\}$ (restricted support)
& usually train; optionally infer
& Restricts propagation to a sampled subgraph for scalability. \\
\bottomrule
\end{tabularx}
\end{table}

\paragraph{Reading the table.} Table~\ref{tab:operations-slots} gives a quick compatibility
heuristic from the primary-location assignments. A model can simultaneously add a Laplacian PE
(primary component $\Psi_k$), use an initial residual ($\mathcal{B}_\ell$), and normalize ($\phi_\ell$),
and because these primary components differ they are \emph{prima facie} compatible. Operations that share a
primary component must instead be reconciled in that component. Two residual schemes both claim
$\mathcal{B}_\ell$, and a virtual node, graph rewiring, and mini-batch sampling all act on
$\{\mathbf{P}_k\}$, so combining them means defining the sampler \emph{on} the rewired or
augmented operator rather than treating them as independent edits. The heuristic is not a
guarantee even across components. As the caveats above note, a
pairwise PE can reach $\mathbf{P}_k$ and a pre-norm reaches $\mathbf{P}_k,\Psi_k$, so distinct
primary components indicate \emph{prima facie} compatibility, which a model with such cross-component
operations should still check. The same factorization that localizes architectural differences
in the family tables thus also organizes the operations.

\appendixsection{Theoretical Analysis}
\label{sec:analysis}

\paragraph{Overview.} The four subsections that follow answer four different
questions about the same object, the propagation bank $\{\mathbf{P}_k\}$ together with
the slots that surround it. Section~\ref{subsec:subsumption} asks which named
architectures are exact special cases of the unified equation, and shows that
spectral filters, multi-hop models, and attention layers are recovered as particular
fillings, differing only along a small number of axes localized to specific slots.
Sections~\ref{subsec:locality} and~\ref{subsec:smoothing} ask what the propagation
bank can and cannot do. The first bounds reach (what a node can see, and
what it takes for one layer to mix everything), and the second bounds
behavior under depth (when repeated propagation collapses information, and two
component-level mechanisms analyzed here that can prevent it: the ego/residual map
$\mathcal{B}_\ell$ and the propagation bank $\mathbf{P}_k$). Section~\ref{subsec:wl-symmetry}
closes with two conditional structural boundaries: a ceiling on what structure-blind
components can distinguish, and the equivariance conditions the component maps and
operators must satisfy to define a permutation-consistent graph layer. Several results restate
established facts, including oversmoothing's geometric decay, polynomial-filter universality,
and the $1$-WL ceiling. Their role is to verify that the component vocabulary preserves known
theory across architectural families. The results specific to the decomposition are the
slot-level attribution and identifiability consequences. They include the fact that channel
count is not identifiable while channel rank is, that
one-layer global mixing has a precise necessary condition on the support of
$\mathbf{P}_k$, and that oversmoothing is escaped through more than one slot-level
mechanism, chiefly the ego/residual map $\mathcal{B}_\ell$ and the spectral filling of
$\mathbf{P}_k$.

\paragraph{Notation/Conventions.} Each result names the slots and assumptions it constrains, so a design choice can be traced to the relevant columns rather than only to a named architecture. Table~\ref{tab:theory-results} groups results by the property they establish, following the four-part organization above. Because the grouping is thematic rather than sequential, the Theorem/Lemma/Proposition numbers do not always run consecutively within a subsection below. Lemma~\ref{lem:return}, Proposition~\ref{prop:wl}, and Theorem~\ref{thm:equivariance} are each numbered earlier than the results they are grouped with, since their thematic home is a later subsection. Unless stated otherwise, the displayed linear specialization is node-preserving, with no intervening pooling operation ($\mathbf{Q}_\ell=\mathbf{I}$):
\[
\mathbf{H}^{(\ell+1)} \;=\; \sigma\!\Big(\textstyle\sum_{k} \mathbf{P}_k^{(\ell)} \mathbf{H}^{(\ell)} \mathbf{W}_k^{(\ell)} \;+\; \mathcal{B}_\ell\big(\mathbf{H}^{(\ell)},\mathbf{H}^{(0)}\big)\Big).
\]
This specialization uses additive mixing
$\mixmsg^{(\ell)}=\boxplus_k\mathbf C_k^{(\ell)}=\sum_k\mathbf P_k^{(\ell)}\mathbf H^{(\ell)}\mathbf W_k^{(\ell)}$
and the update $\phi_\ell(a,b)=\sigma(a+b)$; $\mixmsg^{(\ell)}$ is the output of
$\boxplus$, not an additional component.
For all spectral and Dirichlet-energy statements, the graph is undirected, connected, and has nonnegative edge weights. We write $\tilde{\mathbf{A}}=\mathbf{A}+\mathbf{I}$ for the self-loop-augmented adjacency and use the self-loop normalized adjacency $\hat{\mathbf{A}} = \tilde{\mathbf{D}}^{-1/2}\tilde{\mathbf{A}}\tilde{\mathbf{D}}^{-1/2}$ with eigenpairs $(\mu_i,\mathbf{v}_i)$ (the $\mathbf{v}_i$ are eigenvectors of $\hat{\mathbf{A}}$, distinct in general from the columns of the $\mathbf{L}_{\mathrm{sym}}$ Fourier matrix $\mathbf{U}$ of Section~\ref{sec:preliminaries}) ordered $1=\mu_1>\mu_2\ge\cdots\ge\mu_n>-1$, the second-largest eigenvalue magnitude $\mu_\star=\max_{i\ge 2}|\mu_i|<1$, the augmented normalized Laplacian $\tilde{\mathbf{L}} = \mathbf{I}-\hat{\mathbf{A}}$ (this is the reference propagation Laplacian $\mathbf{L}_{\mathrm{ref}}$ of Section~\ref{sec:preliminaries}; we keep the shorthand $\tilde{\mathbf{L}}$ within this section), and the Dirichlet energy $E(\mathbf{H})=\operatorname{tr}(\mathbf{H}^\top \tilde{\mathbf{L}} \mathbf{H})$. The energy is defined with respect to $\mathbf{L}_{\mathrm{ref}}=\tilde{\mathbf{L}}$ rather than the unnormalized $\mathbf{L}$ or $\mathbf{L}_{\mathrm{sym}}$ because the proofs work in the eigenbasis of $\hat{\mathbf{A}}$; the quadratic form $\mathbf{x}^\top\mathbf{L}\mathbf{x}$ introduced in Section~\ref{sec:preliminaries} serves only as intuition for graph variation and refers to a different operator. The symbol $\tilde{\mathbf{L}}$ denotes this augmented Laplacian throughout the section and is \emph{not} the Chebyshev operator; the distinct, $[-1,1]$-rescaled operator that Chebyshev polynomials act on is written $\tilde{\mathbf{L}}_{\mathrm{ch}}=2\mathbf{L}_{\mathrm{sym}}/\lambda_{\max}-\mathbf{I}$, and the ego/residual map is the $\mathcal{B}_\ell$ of Eq.~\eqref{eq:unified-notation-linear}. A permutation is written $\boldsymbol{\Pi}$. $\mathcal{B}_\ell$ and $\phi_\ell$ are taken node-local and row-wise throughout, an additional
hypothesis on top of the slot discipline of Section~\ref{subsec:slot-discipline} rather than a
consequence of it; Propositions~\ref{prop:receptive}, \ref{prop:global}, \ref{prop:wl}, and
Theorem~\ref{thm:equivariance} state the relevant half of it explicitly because their arguments
turn on it, and Theorem~\ref{thm:equivariance} closes by noting how the conclusion extends when
row-wiseness is relaxed to equivariance.

\begingroup
\small
\setlength{\tabcolsep}{5pt}
\renewcommand{\arraystretch}{1.2}
\setlength{\LTcapwidth}{\textwidth}
\colorlet{theoryBar}{googlegreen}
\colorlet{theoryRow}{googlegreen!9}
\begin{longtable}{@{}
  >{\raggedright\arraybackslash}p{0.200\textwidth}
  >{\raggedright\arraybackslash}p{0.125\textwidth}
  >{\raggedright\arraybackslash}p{0.565\textwidth}
  @{}}
\caption{Summary of the main theoretical results.}
\label{tab:theory-results}\\
\toprule
\rowcolor{theoryBar}
\textbf{Result} & \textbf{Reference} & \textbf{Description} \\
\midrule
\endfirsthead
\multicolumn{3}{c}{\tablename~\thetable\ -- \textit{continued}}\\
\toprule
\rowcolor{theoryBar}
\textbf{Result} & \textbf{Reference} & \textbf{Description} \\
\midrule
\endhead
\midrule
\multicolumn{3}{r}{\textit{continued on next page}}\\
\endfoot
\bottomrule
\endlastfoot
Spectral subsumption & (Lemma~\ref{lem:spectral-subsumption}) &
A layer applying scalar functions of a fixed shift operator through its spectral calculus is the unified equation with $\mathbf{P}_k=g_k(\mathbf{S})$; ChebNet, SGC, APPNP, GPR-GNN, and BernNet are each an exact special case, separated by the operator/basis, the coefficient constraint, and the coupled-versus-decoupled message slot. \\
\rowcolor{theoryRow}
Concatenation is additive mixing & (Prop.~\ref{prop:concat}) &
Multi-head concatenation equals a sum of head outputs embedded into disjoint output blocks, introducing no separate mixing mechanism beyond summation. \\
Concatenation expressivity & (Cor.~\ref{cor:concat-fusion}) &
In the linear regime concatenation matches additive mixing only after projection to a common width; concatenation itself creates explicit head separation in disjoint coordinates, and downstream weights (or a blockwise nonlinearity) decide whether it is exploited or erased. \\
\rowcolor{theoryRow}
Attention subsumption & (Lemma~\ref{lem:attention-subsumption}) &
A single-head layer whose neighbor weight is a scalar function of node states times a linear value is the unified equation with a state-dependent operator; restricted to the single-head, edge-feature-free case, GAT and GATv2 differ only in the score inside $\mathbf{P}_k$. \\
Channel count not identifiable & (Prop.~\ref{prop:channel-rank}) &
The linear pre-activation depends only on $\mathbf{T}=\sum_k\mathbf{W}_k^\top\otimes\mathbf{P}_k$, so the channel count is not recoverable and the invariant is the channel rank. \\
\rowcolor{theoryRow}
Receptive field and underreaching & (Prop.~\ref{prop:receptive}) &
Locally computed one-hop channels confine the depth-$L$ receptive field to the $L$-hop ball, so a node output is independent of inputs farther than $L$ hops; a globally state-dependent operator escapes this even at one-hop support. \\
One-layer global mixing & (Prop.~\ref{prop:global}) &
For fixed operators, a full off-diagonal support row is necessary, and generically sufficient, for a single layer to mix all \emph{other} nodes, so one-layer globality needs a genuinely full effective row (a dense operator, or a rewiring that fills the row; a virtual node needs two layers); it is not necessary for globally state-dependent operators. \\
\rowcolor{theoryRow}
Return probabilities as structural self-weights & (Lemma~\ref{lem:return}) &
In the $k$-hop channel the self-weight $[\hat{\mathbf{A}}^k]_{ii}$ is the $k$-step return probability, so multi-hop channels inject a structural self-weight set by the $k$-step closed walks (self-loops and backtracking included). \\
One-step contraction & (Lemma~\ref{lem:contraction}) &
A single application of $\hat{\mathbf{A}}$ contracts Dirichlet energy by at most $\mu_\star^2$. \\
\rowcolor{theoryRow}
One-layer energy bound & (Thm.~\ref{thm:oversmoothing}) &
A nonlinear layer scales energy by at most $(s\mu_\star)^2$; when $s\mu_\star<1$, this is a contraction and depth drives geometric collapse toward the degree-dominated subspace. \\
Initial-residual energy floor & (Thm.~\ref{thm:floor}) &
The initial-residual iteration converges to the personalized-PageRank fixed point, whose energy stays strictly above a fraction of the anchor energy when the anchor carries nonzero non-DC content. \\
\rowcolor{theoryRow}
Filter realizability & (Lemma~\ref{lem:realizability}) &
A polynomial channel of degree at most $q-1$ realizes any frequency response over the $q$ distinct eigenvalues of $\hat{\mathbf{A}}$. \\
Heterophilic configuration & (Cor.~\ref{cor:heterophily}) &
A high-pass response that zeros the smoothing direction is realizable inside $\mathbf{P}_k$ at degree at most $q-1$, retaining (with nonzero response) any class signal that the discriminative band carries (high-frequency under heterophily, but not necessarily all of it). \\
\rowcolor{theoryRow}
Spectral universality & (Thm.~\ref{thm:universality}) &
When $\hat{\mathbf{A}}$ has distinct eigenvalues and the signal misses no frequency, polynomial filters of $\hat{\mathbf{A}}$ reach every vector in $\mathbb{R}^n$. \\
$1$-WL ceiling and escape mechanisms & (Prop.~\ref{prop:wl}) &
A node-domain stack with no identifiers, structure-blind node-wise maps, and one-hop operators whose weights depend only on incident $1$-WL colors is bounded by $1$-WL; three structural mechanisms can lift the ceiling (the domain, the input message, and a structure-aware operator, where an uninformative filling need not help), though a structure-aware $\phi/\boxplus/$readout would too, which is why those are restricted. \\
\rowcolor{theoryRow}
Neighborhood-aggregation injectivity & (Lemma~\ref{lem:nbr-agg-inj}) &
Neighbor-sum aggregation (realized by $\mathbf{P}=\mathbf{A}$, not by channel mixing) of an injectively encoded message, paired with an injective combination of the aggregate with the root state, attains the $1$-WL ceiling (the raw sum is not injective on multisets, and injective neighbor aggregation alone does not suffice), while the mean and max aggregators cannot be made injective. \\
Permutation equivariance & (Thm.~\ref{thm:equivariance}) &
Equivariant node or inter-rank operators and messages (including shared row-wise or pairwise specializations), permutation-commuting mixing, ego, and update maps, and an equivariant optional pooling map give an equivariant stack and, with an invariant readout, an invariant graph function. \\
\end{longtable}
\endgroup

\appendixsubsection{Subsumption: families as slot choices} \label{subsec:subsumption}

Named spectral and attention models are exact fillings of $\mathbf{P}_k$ and $\Psi_k$, recovered as single points of the linear specialization. The factorization also makes precise what the mixing component $\boxplus$ does. Multi-head concatenation is additive mixing (Prop.~\ref{prop:concat} and Cor.~\ref{cor:concat-fusion}), and the channel count is not identifiable from the layer map (Prop.~\ref{prop:channel-rank}).

The formal statement of Lemma~\ref{lem:spectral-subsumption} (Spectral subsumption) is given in
Section~\ref{subsec:setup}, where it is first cited.

\begin{proof}[Proof of Lemma~\ref{lem:spectral-subsumption}]\label{proof:lem-spectral-subsumption}
Identify $\mathbf{P}_k\leftarrow g_k(\mathbf{S})$ and take $\Psi_k$ as the model's linear message. Any scalar coefficient of channel $k$, when present, is absorbed by $\mathbf{P}_k\leftarrow\gamma_k \mathbf{P}_k$ (ChebNet has none in this placement, its freedom being the message matrices $\boldsymbol{\Theta}_k$), so the action of the mixing component, $\boxplus_k \mathbf{P}_k\Psi_k$, is a pure sum; $\mathcal{B}_\ell$ forms the separate ego/residual input and $\phi_\ell$ combines the two. This is exactly the slot discipline. The three differences are read off the components after substitution: axis~(i) from the choice of $\mathbf{S}$ and $g_k$ in $\mathbf{P}_k$, axis~(ii) from the constraint on the absorbed coefficient, and axis~(iii) from whether $\Psi_k$ reads the evolving state $\mathbf{H}^{(\ell)}$ (coupled) or a layer-independent signal $f_\theta(\mathbf{X})$ (decoupled). Each named model fixes these three, and substituting the fixed values recovers its defining layer exactly: ChebNet's axis~(i)/(iii) choices give $\sum_k T_k(\tilde{\mathbf{L}}_{\mathrm{ch}})\mathbf{H}^{(\ell)}\boldsymbol{\Theta}_k$, its own defining sum; SGC's single decoupled channel gives $\hat{\mathbf{A}}^{K}\mathbf{X}\mathbf{W}$ directly; GPR-GNN's signed $\gamma_k$ on the decoupled channel gives $\sum_k\gamma_k\hat{\mathbf{A}}^{k}f_\theta(\mathbf{X})$, its defining propagation; and BernNet's nonnegative Bernstein weights on the decoupled channel give its Bernstein-basis polynomial filter output; APPNP's substitution is worked out in full above. In each case the right-hand side is the model's own equation with no relabeling beyond the component names. The free framework admits every listed filling and also admits hybrid fillings outside these named definitions, for example a layer combining a coupled channel $\Psi_1=\mathbf{H}^{(\ell)}\boldsymbol{\Theta}_1$ with a decoupled channel $\Psi_2=f_\theta(\mathbf{X})$; each listed model fixes axis~(iii) to one mode.
\end{proof}

The formal statements of Proposition~\ref{prop:concat} (Concatenation is additive mixing) and its
Corollary~\ref{cor:concat-fusion} are given in Section~\ref{subsec:setup}, where they are first cited.
Only the proposition is proved below. The corollary's proof is the block-partition argument
already given in its own statement in Section~\ref{subsec:setup}, so it is not repeated here.

\begin{proof}[Proof of Proposition~\ref{prop:concat}]\label{proof:prop-concat}
$\mathbf{P}_h\mathbf{H}\mathbf{W}_h\mathbf{J}_h$ is supported on the columns of block $h$ and equals $\mathbf{P}_h\mathbf{H}\mathbf{W}_h$ there. Because the blocks are disjoint, the sum over $h$ writes each block once, which is the concatenation.
\end{proof}

The formal statement of Lemma~\ref{lem:attention-subsumption} (Attention subsumption) is given in
Section~\ref{subsec:worked-reductions}, where it is first cited.

\begin{proof}[Proof of Lemma~\ref{lem:attention-subsumption}]\label{proof:lem-attention-subsumption}
The per-node update
$\vh_i^{(\ell+1)}=\sigma\big(\sum_{j}\alpha_{ij}(\mathbf{H})\,\vh_j^{(\ell)}\mathbf{W}_V\big)$
is row $i$ of $\sigma(\mathbf{P}_1 \mathbf{H} \mathbf{W}_V)$ with
$[\mathbf{P}_1]_{ij}=\alpha_{ij}(\mathbf{H})$ and
$\Psi_1=\mathbf{H}\mathbf{W}_V$. The scalar $\alpha_{ij}$ is an entry of $\mathbf{P}_1$,
the value is the message, and the support/value separation is preserved. For $M$ heads,
each an instance of this single-head identification with its own $\mathbf{P}_m,\Psi_m$,
concatenation is exactly the mixing $\Vert_m\mathbf{P}_m\Psi_m$ of
Proposition~\ref{prop:concat}; a final-layer head average
$\tfrac1M\sum_m\mathbf{P}_m\Psi_m$ is the same additive mixing with each head's
contribution rescaled by $1/M$, which can be absorbed into $\mathbf{P}_m$. When an edge
feature $\ve_{ij}$ enters the score, $\alpha_{ij}(\mathbf{H},\ve_{ij})$, it modulates
$\mathbf{P}_1$'s entries directly; when it is instead added on the value side,
$\Psi_1(\vh_i,\vh_j,\ve_{ij})=\vh_j\mathbf{W}_V+f(\ve_{ij})$, it enters the message;
both routes can be used at once, since they modify different components. Restricted to the
single-head, edge-feature-free case, GAT and GATv2 both take the identified form with the
same linear-value message class and differ in the score producing $\alpha_{ij}$. Using
$\mathbf W_S$ for the score projection, GAT uses
$e_{ij}=\operatorname{LReLU}\!\bigl([\vh_i\mathbf W_S\Vert\vh_j\mathbf W_S]\mathbf a\bigr)$,
whereas GATv2 changes the order to
$e_{ij}=\operatorname{LReLU}\!\bigl([\vh_i\Vert\vh_j]\widetilde{\mathbf W}_S\bigr)\mathbf a$.
Thus the learned vector $\mathbf a$ is applied after the nonlinearity in GATv2, while
$\widetilde{\mathbf W}_S$ is applied after concatenating the endpoints; this yields the dynamic attention
scoring class established for GATv2. Either way the difference is confined to the functional
form of $[\mathbf{P}_1]_{ij}$.
\end{proof}

The formal statement of Proposition~\ref{prop:channel-rank} (Channel count is not identifiable
in the linear regime) is given in Section~\ref{subsec:setup}, where it is first cited.

\begin{proof}[Proof of Proposition~\ref{prop:channel-rank}]\label{proof:prop-channel-rank}
The map $\mathbf{H}\mapsto\sum_k \mathbf{P}_k\mathbf{H}\mathbf{W}_k$ depends on the terms only through their sum $\mathbf{T}$, by linearity. Two channel decompositions with the same $\mathbf{T}$ are therefore the same function. The number of summands is not unique: a linear split $\mathbf{W}_k=\mathbf{W}_k'+\mathbf{W}_k''$ replaces one term by two with the same $\mathbf{P}_k$ and the same total, and a canceling pair $(\mathbf{P},+\mathbf{W}),(\mathbf{P},-\mathbf{W})$ adds two terms summing to zero, both raising $K$ at fixed $\mathbf{T}$. Hence $K$ is not recoverable, while the minimum number of Kronecker summands over all decompositions of $\mathbf{T}$ is well defined and is the channel rank.
\end{proof}

\appendixsubsection{Locality and globality: properties of the support of $\mathbf{P}_k$} \label{subsec:locality}

Under the local-computation or fixed-operator hypotheses stated by the results below, the
support pattern of $\mathbf{P}_k$, rather than the numerical values of its present entries,
determines which node states can directly reach a target. Locality bounds the receptive field to
the $L$-hop ball at depth $L$ (Prop.~\ref{prop:receptive}); a single fixed-operator layer requires
a full off-diagonal row to mix every other node (Prop.~\ref{prop:global}); and a multi-hop channel
injects return-probability mass on the diagonal (Lemma~\ref{lem:return}). Globally
state-dependent operator entries fall outside this support-only reading, as both propositions
state explicitly.

The formal statement of Proposition~\ref{prop:receptive} (Receptive field and underreaching) is
given in Section~\ref{subsec:prop-oversquashing}, where it is first cited.

\begin{proof}[Proof of Proposition~\ref{prop:receptive}]\label{proof:prop-receptive}
Let $R_\ell(i)\subseteq \mathcal{V}$ denote the set of input nodes $j$ such that $\vh_i^{(\ell)}$ depends on $\vh_j^{(0)}$; we show, by induction on the layer index $\ell$, the statement \emph{for every node} $v$: $R_\ell(v)\subseteq \bar{B}_\ell(v)$ (writing the closed hop-ball with a bar to keep it visually distinct from the ego/residual map $\mathcal{B}_\ell$, which appears in the same argument below; the inductive step applies this hypothesis at nodes $j\in \bar{B}_1(i)$ other than $i$ itself, so the induction must be on the universally-quantified statement, not on a single fixed node), with no appeal to the conclusion. \emph{Base case} $\ell=0$: $\vh_i^{(0)}$ depends only on itself, so $R_0(i)=\{i\}=\bar{B}_0(i)$. \emph{One-step support.} Row $i$ of $\mathbf{P}_k^{(\ell)}\mathbf{H}^{(\ell)}\mathbf{W}_k^{(\ell)}$ is $\sum_{j}[\mathbf{P}_k^{(\ell)}]_{ij}(\mathbf{H}^{(\ell)}\mathbf{W}_k^{(\ell)})_j$, summed over $j\in \bar{B}_1(i)$ by the support hypothesis; each surviving coefficient $[\mathbf{P}_k^{(\ell)}]_{ij}$ reads only $\vh_i^{(\ell)},\vh_j^{(\ell)},\ve_{ij}$ (local-computation hypothesis) and each summed message reads $\vh_j^{(\ell)}$ with $j\in \bar{B}_1(i)$; so row $i$ of the linear-value channel is a function of $\{\vh_j^{(\ell)}:j\in \bar{B}_1(i)\}$ alone. For a pairwise-message channel, every term $[\mathbf{P}_k^{(\ell)}]_{ij}\Psi_k^{(\ell)}(\vh_i^{(\ell)},\vh_j^{(\ell)},\ve_{ij})$ has the same endpoint-local dependence and is aggregated only over $j\in \bar{B}_1(i)$, so its channel contribution obeys the identical bound. (This is what blocks a zero-hop but globally state-dependent operator such as $\mathbf{P}(\mathbf{H})=\mathrm{sum}(\mathbf{H})\,\mathbf{I}$ with $\mathrm{sum}(\mathbf{H})=\sum_{j,c}H_{jc}$, whose diagonal support is one-hop yet whose single entry reads every node.) By the per-node assumption $\mathcal{B}_\ell,\phi_\ell$ add no dependence beyond row $i$ (the teleport term touches only $\vh_i^{(0)}$), so $\vh_i^{(\ell+1)}$ is a function of $\{\vh_j^{(\ell)}:j\in \bar{B}_1(i)\}$. \emph{Inductive step.} Composing with the inductive hypothesis, $R_{\ell+1}(i)\subseteq\bigcup_{j\in \bar{B}_1(i)}R_\ell(j)\subseteq\bigcup_{j\in \bar{B}_1(i)}\bar{B}_\ell(j)=\bar{B}_{\ell+1}(i)$, the last equality because a node reachable in $\le\ell$ hops from a one-hop neighbor of $i$ is reachable in $\le\ell+1$ hops from $i$, and conversely. Hence $R_L(i)\subseteq \bar{B}_L(i)$; if $d(i,j)>L$ then $j\notin \bar{B}_L(i)$, so $\vh_i^{(L)}$ is independent of $\vh_j^{(0)}$.
\end{proof}

The formal statement of Proposition~\ref{prop:global} (One-layer global mixing) is given in
Section~\ref{subsec:prop-transformer}, where it is first cited. (Proposition~\ref{prop:wl},
numbered in between, is grouped with the expressivity results below in
Section~\ref{subsec:wl-symmetry}, where it fits thematically.)

\begin{proof}[Proof of Proposition~\ref{prop:global}]\label{proof:prop-global}
Fix $j\neq i$. \emph{Necessity.} With the operators fixed, row $i$ of the mixed message $\boxplus_k \mathbf{P}_k\Psi_k$ is $\sum_k\sum_{j'}[\mathbf{P}_k]_{ij'}(\mathbf{H}\mathbf{W}_k)_{j'}$, a function of $\{\vh_{j'}:\exists k,\,[\mathbf{P}_k]_{ij'}\neq 0\}$ alone. If $[\mathbf{P}_k]_{ij}=0$ for all $k$, then $\vh_j$ enters no term of the mixed message (the message $\Psi_k$ is node-wise and $\boxplus$ acts over channels), and since $j\neq i$ the node-local ego map reads only $\vh_i,\vh_i^{(0)}$ and the node-local $\phi_\ell$ adds no further cross-node dependence; so $\vh_i^{(\ell+1)}$ is independent of $\vh_j$. Hence dependence on all $j\neq i$ forces every off-diagonal entry of row $i$. \emph{Generic sufficiency.} For $j\neq i$, the ego input $\mathbf{a}=\mathcal{B}_\ell(\mathbf{H})_i$ does not depend on $\vh_j$, so it is genuinely fixed as $\vh_j$ varies. The dependence of the mixed message at row $i$ on the row vector $\vh_j$ is right multiplication by $\sum_k[\mathbf{P}_k]_{ij}\mathbf{W}_k$; this matrix is nonzero for $\{\mathbf{W}_k\}$ outside a measure-zero set whenever some $[\mathbf{P}_k]_{ij}\neq 0$. A nonzero linear map already suffices: evaluating at $\vh_j=0$ and at a row vector $\vh_j=\mathbf v$ satisfying $\mathbf v(\sum_k[\mathbf{P}_k]_{ij}\mathbf{W}_k)\neq0$ exhibits the dependence, so injectivity is not required. Thus generically every such $j$ enters the mixed message. Message-faithfulness of $\phi_\ell$ ($\mathbf{b}\mapsto\phi_\ell(\mathbf{a},\mathbf{b})$ injective at this fixed $\mathbf{a}$) then transmits the dependence to the output: varying $\vh_j$ changes the mixed message and hence $\vh_i^{(\ell+1)}$. The exceptional cancellations (e.g.\ two channels with opposite weights on the same edge) are non-generic, so a full off-diagonal row is the operative condition for one-layer globality. (When $\mathbf{P}_k=\mathbf{P}_k(\mathbf{H})$ the dependence picks up the omitted Jacobian terms $\sum_k\partial_{\vh_j}[\mathbf{P}_k(\mathbf{H})]_{i\cdot}\,(\mathbf{H}\mathbf{W}_k)_\cdot$, derivatives of \emph{present} entries such as $[\mathbf{P}_k]_{ii}(\mathbf{H})$, which can carry dependence on a node $j$ that occupies no support position $(i,j)$; the necessity direction then fails, as the diagonal example shows.)
\end{proof}

The formal statement of Lemma~\ref{lem:return} (Return probabilities as structural
self-weights) is given in Section~\ref{subsec:setup}, where it is first cited.

\begin{proof}[Proof of Lemma~\ref{lem:return}]\label{proof:lem-return}
Row $i$ of $\mathbf{P}_k\mathbf{H}\mathbf{W}_k$ is $\sum_j [\mathbf{P}_k]_{ij}(\mathbf{H}\mathbf{W}_k)_j=[\mathbf{P}_k]_{ii}(\mathbf{H}\mathbf{W}_k)_i+\sum_{j\neq i}[\mathbf{P}_k]_{ij}(\mathbf{H}\mathbf{W}_k)_j$. Since $\hat{\mathbf{A}}^{k}=\tilde{\mathbf{D}}^{1/2}(\tilde{\mathbf{D}}^{-1}\tilde{\mathbf{A}})^{k}\tilde{\mathbf{D}}^{-1/2}$ is a conjugation by the diagonal $\tilde{\mathbf{D}}^{1/2}$, which leaves diagonal entries unchanged, $[\hat{\mathbf{A}}^{k}]_{ii}=[(\tilde{\mathbf{D}}^{-1}\tilde{\mathbf{A}})^{k}]_{ii}$, which is the $k$-step return probability of the self-loop random walk at $i$.
\end{proof}

\appendixsubsection{Smoothing, homophily, and heterophily: spectral filling of $\mathbf{P}_k$} \label{subsec:smoothing}

The problem is smoothing. Repeated low-pass propagation by $\hat{\mathbf{A}}$ contracts Dirichlet energy (Lemma~\ref{lem:contraction}), and a stack of such layers drives geometric collapse toward the degree-dominated subspace (Thm.~\ref{thm:oversmoothing}). We analyze two principal component-level escape mechanisms. The ego/residual map $\mathcal{B}_\ell$ holds the energy above a fixed floor through the initial residual (Thm.~\ref{thm:floor}), and a spectral filling of $\mathbf{P}_k$ replaces the low-pass response outright, realizing any band (Lemma~\ref{lem:realizability}), retaining the discriminative signal under heterophily (Cor.~\ref{cor:heterophily}), and reaching every output in the simple-spectrum limit (Thm.~\ref{thm:universality}).

The formal statement of Lemma~\ref{lem:contraction} (One-step contraction) is given in
Section~\ref{subsec:prop-oversmoothing}, where it is first cited.

\begin{proof}[Proof of Lemma~\ref{lem:contraction}]\label{proof:lem-contraction}
Write $\mathbf{H}=\sum_i \mathbf{v}_i \mathbf{c}_i^\top$ in the eigenbasis of $\hat{\mathbf{A}}$ (orthonormal since $\hat{\mathbf{A}}$ is real symmetric; $\mathbf{c}_i^\top=\mathbf{v}_i^\top\mathbf{H}$). Since $\tilde{\mathbf{L}}\mathbf{v}_i=(1-\mu_i)\mathbf{v}_i$ and $\mathbf{v}_i^\top\mathbf{v}_j=\delta_{ij}$, the cross terms in $\operatorname{tr}(\mathbf{H}^\top\tilde{\mathbf{L}}\mathbf{H})$ vanish, giving $E(\mathbf{H})=\sum_i (1-\mu_i)\Vert \mathbf{c}_i\Vert^2$ and $E(\hat{\mathbf{A}} \mathbf{H})=\sum_i (1-\mu_i)\mu_i^2\Vert \mathbf{c}_i\Vert^2$. The $i=1$ term vanishes since $\mu_1=1$. For $i\ge 2$ we have $1-\mu_i\ge 0$ and $\mu_i^2\le\mu_\star^2$, so $E(\hat{\mathbf{A}} \mathbf{H})\le \mu_\star^2\sum_{i\ge 2}(1-\mu_i)\Vert \mathbf{c}_i\Vert^2=\mu_\star^2 E(\mathbf{H})$.
\end{proof}

The formal statement of Theorem~\ref{thm:oversmoothing} (One-layer energy bound) is given in
Section~\ref{subsec:prop-oversmoothing}, where it is first cited.

\begin{proof}[Proof of Theorem~\ref{thm:oversmoothing}]\label{proof:thm-oversmoothing}
With $\tilde d_i=1+d_i$ and the identity $E(\mathbf{M})=\tfrac12\sum_{i,j}[\tilde{\mathbf{A}}]_{ij}\Vert \mathbf{m}_i/\sqrt{\tilde d_i}-\mathbf{m}_j/\sqrt{\tilde d_j}\Vert^2$ (the sum over \emph{ordered} pairs $(i,j)$, so the $\tfrac12$ corrects the double count; equivalently $\sum_{\{i,j\}\in E}$ over unordered edges without the $\tfrac12$), positive homogeneity gives $\sigma(\mathbf{m}_i)/\sqrt{\tilde d_i}=\sigma(\mathbf{m}_i/\sqrt{\tilde d_i})$; the $1$-Lipschitz property then contracts each per-edge difference, $\|\sigma(\mathbf{m}_i/\sqrt{\tilde d_i})-\sigma(\mathbf{m}_j/\sqrt{\tilde d_j})\|\le\|\mathbf{m}_i/\sqrt{\tilde d_i}-\mathbf{m}_j/\sqrt{\tilde d_j}\|$, so summing over edges gives $E(\sigma(\mathbf{M}))\le E(\mathbf{M})$. For the weight, $E(\mathbf{M}\mathbf{W})=\operatorname{tr}(\mathbf{W}\mathbf{W}^\top \mathbf{M}^\top\tilde{\mathbf{L}} \mathbf{M})\le\Vert \mathbf{W}\Vert_2^2\operatorname{tr}(\mathbf{M}^\top\tilde{\mathbf{L}} \mathbf{M})=\Vert \mathbf{W}\Vert_2^2 E(\mathbf{M})$, using $\operatorname{tr}(\mathbf{B}\mathbf{C})\le\Vert \mathbf{B}\Vert_2\operatorname{tr}(\mathbf{C})$ for $\mathbf{B}\succeq 0$ and $\mathbf{C}\succeq 0$. Chaining the three steps with Lemma~\ref{lem:contraction},
$E(\sigma(\hat{\mathbf{A}} \mathbf{H} \mathbf{W}))\le E(\hat{\mathbf{A}} \mathbf{H} \mathbf{W})\le s^2 E(\hat{\mathbf{A}} \mathbf{H})\le s^2\mu_\star^2 E(\mathbf{H})$. Applying this bound layer by layer, $E(\mathbf{H}^{(\ell+1)})\le (s_\ell\mu_\star)^2 E(\mathbf{H}^{(\ell)})$, and composing over $\ell=0,\dots,L-1$ with $\bar s=\max_\ell s_\ell$ yields $E(\mathbf{H}^{(L)})\le(\bar s\,\mu_\star)^{2L}E(\mathbf{H}^{(0)})$.
\end{proof}

The formal statement of Theorem~\ref{thm:floor} (Initial-residual energy floor) is given in
Section~\ref{subsec:prop-oversmoothing}, where it is first cited.

\begin{proof}[Proof of Theorem~\ref{thm:floor}]\label{proof:thm-floor}
The linear part $(1-\eta)\hat{\mathbf{A}}$ is symmetric with operator norm $\Vert(1-\eta)\hat{\mathbf{A}}\Vert_2=(1-\eta)\max_i|\mu_i|=1-\eta<1$ (using $\mu_1=1$), so the affine map is a Euclidean contraction with a unique fixed point, given by solving $(\mathbf{I}-(1-\eta)\hat{\mathbf{A}})\mathbf{G}^\star=\eta \mathbf{G}^{(0)}$; the Neumann series converges. Expand $\mathbf{G}^{(0)}=\sum_i \mathbf{v}_i (\mathbf{c}_i^{(0)})^\top$ in the eigenbasis of $\hat{\mathbf{A}}$. The fixed point is $\mathbf{G}^\star=\sum_i \mathbf{v}_i (\mathbf{c}_i^\star)^\top$ with $\mathbf{c}_i^\star=\rho_i\,\mathbf{c}_i^{(0)}$ and scalar $\rho_i=\eta/(1-(1-\eta)\mu_i)$. For $i\ge 2$, the section-wide $\mu_i\in(-1,1)$ gives $1-(1-\eta)\mu_i\in(\eta,2-\eta)$, hence $\rho_i>\eta/(2-\eta)$; the strictness of this lower bound is exactly what the open left end $\mu_i>-1$ secures (automatic here, since the self-loops in $\hat{\mathbf{A}}=\tilde{\mathbf{D}}^{-1/2}(\mathbf{A}+\mathbf{I})\tilde{\mathbf{D}}^{-1/2}$ preclude a bipartite $\mu_i=-1$, which would instead give $\rho_i=\eta/(2-\eta)$ and only a non-strict floor). Note the contraction above used only $\mu_1=1$ and $\eta\in(0,1)$, so convergence to $\mathbf{G}^\star$ holds regardless of $\mu_i=-1$; the self-loop fact is needed only for the \emph{strict} energy floor. Therefore
\[
E(\mathbf{G}^\star)=\sum_{i\ge 2}(1-\mu_i)\,\rho_i^2\,\Vert \mathbf{c}_i^{(0)}\Vert^2>\Big(\tfrac{\eta}{2-\eta}\Big)^2\sum_{i\ge 2}(1-\mu_i)\Vert \mathbf{c}_i^{(0)}\Vert^2=\Big(\tfrac{\eta}{2-\eta}\Big)^2 E(\mathbf{G}^{(0)}),
\]
where the strict inequality holds because $E(\mathbf{G}^{(0)})>0$ forces some $i\ge 2$ with $(1-\mu_i)\Vert \mathbf{c}_i^{(0)}\Vert^2>0$.
\end{proof}

The formal statements of Lemma~\ref{lem:realizability} (Filter realizability) and its
Corollary~\ref{cor:heterophily} (Heterophilic configuration) are given in
Section~\ref{subsec:prop-heterophily}, where they are first cited.

\begin{proof}[Proof of Lemma~\ref{lem:realizability}]\label{proof:lem-realizability}
The algebra of polynomials in $\hat{\mathbf{A}}$ has dimension $q$ with basis $\{\boldsymbol{\Gamma}_j\}$, and $\{\mathbf{I},\hat{\mathbf{A}},\dots,\hat{\mathbf{A}}^{q-1}\}$ is also a basis because the Vandermonde matrix in the distinct $\nu_j$ is invertible. Hence $\sum_j r_j\boldsymbol{\Gamma}_j$ is a polynomial of degree at most $q-1$ in $\hat{\mathbf{A}}$, and matching $g(\nu_j)=r_j$ recovers $\gamma$ by interpolation.
\end{proof}

\begin{proof}[Proof of Corollary~\ref{cor:heterophily}]\label{proof:cor-heterophily}
On a connected graph with self-loops, $\nu=1$ is always one of the $q$ distinct eigenvalues of $\hat{\mathbf{A}}$ (Section~\ref{sec:analysis}'s standing convention gives $\mu_1=1$ with a one-dimensional eigenspace), and any discriminative non-DC eigenvalue is distinct from it by definition. Apply Lemma~\ref{lem:realizability} with $r_j=0$ at the index $j$ for which $\nu_j=1$, and $r_j$ equal to any nonzero target response at every index carrying discriminative signal; the interpolation is unconstrained across indices, so both assignments hold simultaneously. The resulting degree-$(q{-}1)$ polynomial channel therefore has zero response on the smoothing direction while retaining a nonzero response wherever the discriminative signal lies, regardless of which band that is.
\end{proof}

The formal statement of Theorem~\ref{thm:universality} (Spectral universality) is given in
Section~\ref{subsec:prop-expressivity}, where it is first cited.

\begin{proof}[Proof of Theorem~\ref{thm:universality}]\label{proof:thm-universality}
Under the distinct-spectrum hypothesis the $n$ eigenvalues $\mu_1,\dots,\mu_n$ of $\hat{\mathbf{A}}$ are exactly the $q=n$ distinct values $\nu_j$ of Lemma~\ref{lem:realizability}, so we index by $\mu_j$ here. Write $\mathbf{x}=\sum_j \hat x_j \mathbf{v}_j$ with all $\hat x_j\neq 0$. A polynomial filter produces $\sum_j g(\mu_j)\hat x_j \mathbf{v}_j$. As $g$ ranges over polynomials of degree at most $n-1$, the values $g(\mu_1),\dots,g(\mu_n)$ are arbitrary because the $\mu_j$ are distinct (the $n\times n$ Vandermonde in the $\mu_j$ is invertible), so $g(\mu_j)\hat x_j$ attains every coefficient vector. The image is $\operatorname{span}\{\mathbf{v}_j\}=\mathbb{R}^n$.
\end{proof}

\appendixsubsection{Expressivity boundary and symmetry} \label{subsec:wl-symmetry}

One boundary is an upper bound on power. A node-domain stack whose relevant components stay
structure-blind cannot exceed $1$-WL (Prop.~\ref{prop:wl}), and attaining that ceiling demands
an injective neighbor aggregation realized through the message/channel construction together
with an injective root--aggregate update (Lemma~\ref{lem:nbr-agg-inj}); this is not the
cross-channel mixing component $\boxplus$. The other boundary is permutation symmetry.
The domain and channel sets are relabeled/indexed consistently, $\mathbf P_k$ must transform
equivariantly, $\Psi_k$ must be shared or equivariant, and $\boxplus$, $\mathcal B_\ell$, and
$\phi_\ell$ must commute with node relabeling (Thm.~\ref{thm:equivariance}). Shared row maps
are a sufficient special case for the map-valued components, not a necessary description of
all seven components.

The formal statement of Proposition~\ref{prop:wl} ($1$-WL ceiling and the escape mechanisms) is given in
Section~\ref{subsec:prop-expressivity}, where it is first cited.

\begin{proof}[Proof of Proposition~\ref{prop:wl}]\label{proof:prop-wl}
Under the stated hypothesis, each layer computes $\vh_i^{(\ell+1)}$ as a shared node-wise function of $\vh_i^{(\ell)}$ and the per-neighbor-tuple multiset $\{\!\{\big((\,[\mathbf{P}_k]_{ij}\,)_{k},\,\vh_j^{(\ell)}\big)\}\!\}_{j\in \mathcal{N}(i)}$, where the weight vector $(\,[\mathbf{P}_k]_{ij}\,)_{k}$ across channels is itself a function of the colors $(c_i^{(\ell)},c_j^{(\ell)})$; this pair-and-multiset is refinable by one round of color refinement, and the readout is invariant. The base case $\ell=0$ holds by the standard $1$-WL convention that colors are initialized from $\mathbf{H}^{(0)}$ itself, so $\vh_i^{(0)}$ is already a function of the initial color $c_i^{(0)}$; by induction the node colors at every subsequent layer are coarsenings of the $1$-WL colors, and the structure-blind readout sees only their multiset, so two graphs with identical $1$-WL color histograms produce identical outputs. The ceiling can be exceeded by changing the object being colored (lifting $\mathcal{X}$), by breaking neighborhood-isomorphism symmetry in the input (enriching $\mathbf{H}^{(0)}$), or by making $\mathbf{P}_k$ depend on structure not determined by the incident colors, the last of which is exactly the failure of the refinability hypothesis. The necessity of the hypothesis is shown by the connected pair: the triangular prism $K_3\,\square\,K_2$ versus the complete bipartite graph $K_{3,3}$. Both are $3$-regular on six nodes with constant features, hence $1$-WL-indistinguishable, and both are connected (consistent with the section-wide convention). A one-hop operator weighted by per-edge triangle counts $[\mathbf{P}]_{ij}=A_{ij}(\mathbf{A}^2)_{ij}$ is nonzero on the prism's triangle edges (the six edges of the two triangular faces, each lying on one triangle, $(\mathbf{A}^2)_{ij}=1$; the three connecting edges lie on no triangle, $(\mathbf{A}^2)_{ij}=0$), so $\mathbf{P}$ is not identically zero on the prism, while it vanishes identically on the triangle-free bipartite $K_{3,3}$ ($(\mathbf{A}^2)_{ij}=0$ on every edge), so a single such layer separates them.
\end{proof}

The formal statement of Lemma~\ref{lem:nbr-agg-inj} (Neighborhood-aggregation injectivity) is
given in Section~\ref{subsec:prop-expressivity}, where it is first cited.

\begin{proof}[Proof of Lemma~\ref{lem:nbr-agg-inj}]\label{proof:lem-nbr-agg-inj}
Injectivity of the raw sum fails directly: over $\mathbb{Z}$ the distinct multisets $\{\!\{1,3\}\!\}$ and $\{\!\{2,2\}\!\}$ have equal sum $4$. The repair is the standard construction of \citet{xu2019powerful}: since the state set is countable, there is a map $f$ (for multisets of size at most $B$, take $f(x)=N^{-Z(x)}$ for an injective code $Z$ into $\mathbb{N}_0$ and $N>B$, so distinct multisets yield distinct finite base-$N$ digit vectors) with $S\mapsto\sum_{x\in S}f(x)$ injective on neighbor multisets. This is necessary but not sufficient for the $1$-WL ceiling: $1$-WL refines on the pair $(c_i,\{\!\{c_j\}\!\}_{j\in \mathcal{N}(i)})$, so an update that maps two roots with equal neighbor multisets but \emph{different} root colors to the same output is strictly coarser. Concretely, a symmetric aggregate over $\{\text{root}\}\cup \mathcal{N}(i)$ collapses the $1$-WL-distinct rooted neighborhoods $(a;\{\!\{b,c\}\!\})$ and $(b;\{\!\{a,c\}\!\})$, which share the combined multiset $\{\!\{a,b,c\}\!\}$. The remedy is the GIN update $g(\vh_i,\sum_{j\in \mathcal{N}(i)}f(\vh_j))$ with $g$ injective in the (root, aggregate) pair (realized in the framework by a separate ego term $\mathcal{B}_\ell$ and an injective $\phi_\ell$), which is injective on rooted neighborhoods and therefore matches $1$-WL. The $(1+\epsilon)$ self-weight of \citet{xu2019powerful} is one such injective pair map under suitable encoding and parameter conditions (an irrational $\epsilon$ keeping the root and aggregate scales separable), not a canonical stand-in for every injective $g$. For the mean and max aggregators no encoding suffices: $\mathrm{mean}\{\!\{a\}\!\}=\mathrm{mean}\{\!\{a,a\}\!\}$ for any $a$, and $\max\{\!\{a,b\}\!\}=\max\{\!\{a,b,b\}\!\}$ for $a<b$, so both collapse distinct multisets regardless of the encoding applied beforehand.
\end{proof}

The formal statement of Theorem~\ref{thm:equivariance} (Permutation equivariance) is given in
Section~\ref{subsec:framework-scope}, where it is first cited.

\begin{proof}[Proof of Theorem~\ref{thm:equivariance}]\label{proof:thm-equivariance}
Write the relabeled inputs as $\mathbf{H}'=\boldsymbol{\Pi}\mathbf{H}$, $\mathbf{A}'=\boldsymbol{\Pi}\mathbf{A}\boldsymbol{\Pi}^\top$, $\mathbf{E}'=\boldsymbol{\Pi}\!\cdot\!\mathbf{E}$, so that entrywise $\vh'_{\pi(i)}=\vh_i$ and $\mathbf{E}'_{\pi(i)\pi(j):}=\mathbf{E}_{ij:}$. For a linear-value channel, $\mathbf{P}_k(\mathbf{A}',\mathbf{H}')\Psi_k(\mathbf{H}')= \boldsymbol{\Pi}\mathbf{P}_k(\mathbf{A},\mathbf{H})\boldsymbol{\Pi}^\top\,\boldsymbol{\Pi}\Psi_k(\mathbf{H})=\boldsymbol{\Pi}\big(\mathbf{P}_k(\mathbf{A},\mathbf{H})\Psi_k(\mathbf{H})\big)$, using $\boldsymbol{\Pi}^\top\boldsymbol{\Pi}=\mathbf{I}$ and $\Psi_k(\boldsymbol{\Pi}\mathbf H)=\boldsymbol{\Pi}\Psi_k(\mathbf H)$; a shared row-wise $\Psi_k$ is one sufficient way to obtain this identity. The identity is unchanged in the edge-aware case, where $\mathbf{P}_k(\mathbf{A}',\mathbf{H}',\mathbf{E}')=\boldsymbol{\Pi}\mathbf{P}_k(\mathbf{A},\mathbf{H},\mathbf{E})\boldsymbol{\Pi}^\top$ supplies the same conjugation. For a pairwise-message channel, written per node as $\mathrm{Agg}_{j\in \mathcal{N}(i)}\big([\mathbf{P}_k]_{ij}\,\Psi_k(\vh_i,\vh_j,\ve_{ij})\big)$, there is no such matrix product. By the entrywise form of operator equivariance (in the three-argument signature when edge features enter $\mathbf{P}_k$), $[\mathbf{P}_k(\mathbf{A}',\mathbf{H}',\mathbf{E}')]_{\pi(i)\pi(j)}=[\mathbf{P}_k(\mathbf{A},\mathbf{H},\mathbf{E})]_{ij}$, and the pair message $\Psi_k(\vh'_{\pi(i)},\vh'_{\pi(j)},\mathbf{E}'_{\pi(i)\pi(j):})=\Psi_k(\vh_i,\vh_j,\ve_{ij})$ at the relabeled pair equals the message at $(i,j)$; relabeling also carries the neighbor set $\mathcal{N}(i)$ to $\mathcal{N}(\pi(i))$ as a set. A permutation-invariant $\mathrm{Agg}$ over the neighbor multiset is insensitive to the order in which those neighbors are enumerated, so the aggregated channel at $\pi(i)$ equals that at $i$, i.e.\ the channel is equivariant. In both cases the permutation-commuting mixing $\boxplus$, ego/residual map $\mathcal{B}_\ell$, and update $\phi_\ell$ preserve the leading $\boldsymbol{\Pi}$, giving $f(\boldsymbol{\Pi}\mathbf{H},\boldsymbol{\Pi}\mathbf{A}\boldsymbol{\Pi}^\top,\boldsymbol{\Pi}\!\cdot\!\mathbf{E})=\boldsymbol{\Pi} f(\mathbf{H},\mathbf{A},\mathbf{E})$. Shared row maps are the node-local specialization; the same calculation requires only equivariance, so it also covers row-coupled maps such as permutation-equivariant graph normalizations.

For a graded linear-value channel, the typed transformation law gives
$\mathbf{P}'_{k,r\to s}\Psi'_{k,r}=\boldsymbol{\Pi}_s\mathbf{P}_{k,r\to s}\boldsymbol{\Pi}_r^\top\boldsymbol{\Pi}_r\Psi_{k,r}=\boldsymbol{\Pi}_s(\mathbf{P}_{k,r\to s}\Psi_{k,r})$, using $\boldsymbol{\Pi}_r^\top\boldsymbol{\Pi}_r=\mathbf{I}$ exactly as before, now with the source-rank permutation absorbed on the right and the target-rank permutation surviving on the left. For a graded pairwise-message channel, written per target object $i$ of rank $s$ as $\mathrm{Agg}_{j\in N_r(i)}\big([\mathbf{P}_{k,r\to s}]_{ij}\,\Psi_{k,r\to s}(\vh_i,\vh_j,\ve_{ij})\big)$ with $j$ ranging over incident rank-$r$ objects: relabeling carries $i\mapsto\pi_s(i)$ and $j\mapsto\pi_r(j)$, and the entrywise form of the typed operator's equivariance gives $[\mathbf{P}_{k,r\to s}(\ldots')]_{\pi_s(i)\pi_r(j)}=[\mathbf{P}_{k,r\to s}(\ldots)]_{ij}$, the same identity as the single-rank case but now indexed by the pair $(\pi_s,\pi_r)$ rather than one shared $\pi$. The pair message $\Psi_{k,r\to s}(\vh'_{\pi_s(i)},\vh'_{\pi_r(j)},\mathbf{E}'_{\pi_s(i)\pi_r(j):})=\Psi_{k,r\to s}(\vh_i,\vh_j,\ve_{ij})$ likewise equals the unrelabeled value, since each endpoint's relabeled state is exactly its original state under its own rank's permutation, and the incidence set $N_r(i)$ is carried to $N_r(\pi_s(i))$ as a set by the equivariant incidence structure relating the two ranks. A permutation-invariant $\mathrm{Agg}$ over this relabeled multiset is again insensitive to enumeration order, so the aggregated channel at $\pi_s(i)$ equals that at $i$: the same conclusion as the single-rank pairwise case, with $\boldsymbol{\Pi}_r,\boldsymbol{\Pi}_s$ in place of the one shared $\boldsymbol{\Pi}$. The target-rank mix, ego map, and update preserve the leading $\boldsymbol{\Pi}_s$. For pooling, writing $\bar{\mathbf{H}}'=\boldsymbol{\Pi}\bar{\mathbf{H}}$ and using the stated law for $\mathbf{Q}_\ell$ gives $\mathbf{Q}'_\ell\bar{\mathbf{H}}'=\boldsymbol{\Pi}_{\mathrm{out}}\mathbf{Q}_\ell\boldsymbol{\Pi}^\top\boldsymbol{\Pi}\bar{\mathbf{H}}=\boldsymbol{\Pi}_{\mathrm{out}}\mathbf{Q}_\ell\bar{\mathbf{H}}$; equivariant construction of the coarsened graph supplies the corresponding relabeled operator bank at the next layer. A permutation-invariant readout then maps the resulting equivariant features to an invariant graph output.
\end{proof}

Each of the seventeen results is expressed through one or more components of the equation rather than through a named architecture alone. Subsumption identifies whole families with particular fillings of $\mathbf{P}_k$ and $\Psi_k$; locality and globality are read from the support and state dependence of $\mathbf{P}_k$; smoothing and its evasion are read from the spectrum of $\mathbf{P}_k$ together with the ego/residual term $\mathcal{B}_\ell$; and the $1$-WL ceiling states which components must remain structure-blind for the bound to hold. Equivariance instead imposes a typed collection of constraints: consistent relabeling of $\mathcal X$ and $\mathcal K$, conjugation/inter-rank transformation of $\mathbf P_k$, equivariant messages and mixing, and equivariant ego/update maps. Thus a property of a GNN design traces to a framework component or a stated interaction among components rather than to a model name. The recast-standard results (contraction, the energy bound, realizability, universality, the WL ceiling, aggregation injectivity) show that the framework reproduces known facts in a single notation; the additional slot-level consequences developed here include channel-rank identifiability (in the linear regime, where the layer depends on the channels only through $\mathbf{T}=\sum_k\mathbf{W}_k^\top\otimes\mathbf{P}_k$), the full-row condition for one-layer globality under the proposition's hypotheses, and the localization of two principal smoothing escapes in the ego/residual map $\mathcal{B}_\ell$ and the spectral filling of $\mathbf{P}_k$.

\appendixsection{Per-Family Comparison Tables}
\label{app:family-comparison-tables}
\appendixsubsection{Assignment rules for primary families}
\label{app-family-assignment-rules}

\begin{table*}[!ht]
\centering
\begingroup
\scriptsize
\setlength{\tabcolsep}{4pt}
\renewcommand{\arraystretch}{1.25}
\arrayrulecolor{black!30}
\setlength{\arrayrulewidth}{0.5pt}
\caption{\textbf{Primary family assignment rules.}
The rules are listed in family order. A method is placed in the family whose rule best describes
the dominant mechanism of the cited layer update. Other mechanisms are recorded as cross-references
or boundary notes.}
\label{tab-family-assignment-rules}
\begin{tabularx}{\textwidth}{|>{\centering\arraybackslash}p{0.055\textwidth}|>{\raggedright\arraybackslash}p{0.18\textwidth}|>{\raggedright\arraybackslash}X|>{\raggedright\arraybackslash}p{0.28\textwidth}|}
\hline
\rowcolor{googleblue!18}
\textcolor{googleblue!45!black}{\textbf{Order}} &
\textcolor{googleblue!45!black}{\textbf{Primary family}} &
\textcolor{googleblue!45!black}{\textbf{Assignment rule}} &
\textcolor{googleblue!45!black}{\textbf{How secondary mechanisms are handled}} \\
\hline

\rowcolor{repSpatial!10}
1 &
Family~1. Spatial message passing &
Choose this family when the layer is best described as ordinary neighbor aggregation with shared message and update maps, and no more specific rule below is dominant. &
Fixed-point solvers, readouts, sampling procedures, and routing loops are marked as external machinery when they are not one synchronous layer update. \\
\hline

\rowcolor{repAttention!10}
2 &
Family~2. Attention and adaptive weighting &
Choose this family when the main contribution is state-dependent attention, adaptive edge weights, learned masks, gates, routing coefficients, or generated aggregation weights. &
Spatial supports and relation features remain secondary unless another family better describes the cited contribution. \\
\hline

\rowcolor{repSpectral!10}
3 &
Family~3. Spectral and diffusion &
Choose this family when the main contribution is a Laplacian or adjacency polynomial, spectral filter, diffusion kernel, hop basis, signed propagation basis, or learned graph filter. &
Ordinary neighbor aggregation is treated as the implementation of an operator basis rather than the primary family. \\
\hline

\rowcolor{repTransformer!10}
4 &
Family~4. Graph transformer &
Choose this family when the defining layer performs global token mixing, transformer self-attention, structural attention biasing, sparse global attention, or tokenized graph communication. &
Positional encodings, structural encodings, and local message passing are treated as supporting components. \\
\hline

\rowcolor{repHetero!10}
5 &
Family~5. Heterogeneous and multi-relational &
Choose this family when typed nodes, typed edges, relation channels, meta-paths, or knowledge-graph relation states define the update. &
Attention over relations remains secondary when relation typing is the defining axis. \\
\hline

\rowcolor{repHigher!10}
6 &
Family~6. Higher-order and subgraph &
Choose this family when the updated objects are tuples, rooted subgraphs, simplices, cells, hyperedges, motifs, or another lifted domain rather than ordinary nodes alone. &
Attention or spectral operations on the lifted domain are treated as secondary mechanisms. \\
\hline

\rowcolor{repGeometric!10}
7 &
Family~7. Geometric and equivariant &
Choose this family when the defining layer uses coordinates, distances, directions, angles, frames, vector features, tensor features, or a stated Euclidean or rotation equivariance requirement. &
Spatial aggregation, attention, and operator banks are treated as secondary when geometry defines the layer. \\
\hline
\end{tabularx}
\endgroup
\end{table*}

\colorlet{spatbarA}{repSpatial}
\colorlet{spatbarB}{repSpatial!80}
\colorlet{spatbarC}{repSpatial!62}
\colorlet{spatbarD}{repSpatial!46}

\colorlet{spatrowA}{repSpatial!9}
\colorlet{spatrowB}{repSpatial!9}
\colorlet{spatrowC}{repSpatial!9}
\colorlet{spatrowD}{repSpatial!9}

\providecommand{\MLP}{\operatorname{MLP}}
\providecommand{\GRU}{\operatorname{GRU}}
\providecommand{\Agg}{\operatorname{Agg}}
\providecommand{\MsgNorm}{\operatorname{MsgNorm}}
\providecommand{\Diag}{\operatorname{diag}}
\providecommand{\squash}{\operatorname{squash}}
\providecommand{\ReLU}{\operatorname{ReLU}}
\providecommand{\normop}{\operatorname{norm}}

\makeatletter
\@ifundefined{NC@find@Y}{\newcolumntype{Y}[1]{>{\raggedright\arraybackslash}p{#1}}}{}
\@ifundefined{NC@find@Z}{\newcolumntype{Z}[1]{>{\raggedright\arraybackslash}p{#1}}}{}
\@ifundefined{NC@find@Q}{\newcolumntype{Q}[1]{>{\raggedright\arraybackslash$}p{#1}<{$}}}{}
\@ifundefined{NC@find@B}{\newcolumntype{B}[1]{>{\centering\arraybackslash$}p{#1}<{$}}}{}
\@ifundefined{NC@find@T}{\newcolumntype{T}[1]{>{\centering\arraybackslash}p{#1}}}{}
\makeatother

\begingroup
\scriptsize
\renewcommand{\arraystretch}{1.5}
\arrayrulecolor{black!45}
\setlength{\arrayrulewidth}{0.5pt}
\setlength{\tabcolsep}{3.67pt}
\setlength{\LTpre}{6pt}
\setlength{\LTpost}{6pt}
\setlength{\LTcapwidth}{\textwidth}

\begin{longtable}{
@{}
Y{2.22cm}
B{1.04cm}
B{1.48cm}
B{2.67cm}
B{2.37cm}
B{2.67cm}
B{2.37cm}
@{}}

\caption{\scriptsize\textbf{Family 1. Spatial Message-Passing GNNs.}
Unified-component decompositions of local message-passing layers. Neighbor aggregation forms
each pairwise channel contribution before $\boxplus$ mixes channels; scalar degree and scaler
terms are absorbed into $\mathbf P_k$. Markers identify covered slices: $^\dagger$ is one
JKNet base layer and $^\ddagger$ the Graph Network node update. Fixed-point, sampled-depth,
action-selection, routing, and readout machinery is external unless shown.%
}
\label{tab:family1_spatial_slots}
\\[6pt]

\toprule
\rowcolor{spatbarA}
\textcolor{white}{\textbf{Method}} &
\multicolumn{1}{B{1.04cm}}{\textcolor{white}{\boldsymbol{\mathcal X}}} &
\multicolumn{1}{B{1.48cm}}{\textcolor{white}{\boldsymbol{\mathcal K}}} &
\multicolumn{1}{T{2.67cm}}{\textcolor{white}{\makecell[c]{\textbf{Propagation}\\$\boldsymbol{\mathbf P_k^{(\ell)}}$}}} &
\multicolumn{1}{T{2.37cm}}{\textcolor{white}{\makecell[c]{\textbf{Message}\\$\boldsymbol{\Psi_k^{(\ell)}}$}}} &
\multicolumn{1}{T{2.67cm}}{\textcolor{white}{\makecell[c]{\textbf{Mixing}\\$\boldsymbol{\boxplus^{(\ell)}}$}}} &
\multicolumn{1}{T{2.37cm}}{\textcolor{white}{\makecell[c]{\textbf{Ego/Update}\\$\boldsymbol{(\mathcal B_\ell,\phi_\ell)}$}}}
\\[6pt]
\midrule
\endfirsthead

\multicolumn{7}{c}{\tablename\ \thetable\ -- continued} \\[6pt]
\toprule
\rowcolor{spatbarA}
\textcolor{white}{\textbf{Method}} &
\multicolumn{1}{B{1.04cm}}{\textcolor{white}{\boldsymbol{\mathcal X}}} &
\multicolumn{1}{B{1.48cm}}{\textcolor{white}{\boldsymbol{\mathcal K}}} &
\multicolumn{1}{T{2.67cm}}{\textcolor{white}{\makecell[c]{\textbf{Propagation}\\$\boldsymbol{\mathbf P_k^{(\ell)}}$}}} &
\multicolumn{1}{T{2.37cm}}{\textcolor{white}{\makecell[c]{\textbf{Message}\\$\boldsymbol{\Psi_k^{(\ell)}}$}}} &
\multicolumn{1}{T{2.67cm}}{\textcolor{white}{\makecell[c]{\textbf{Mixing}\\$\boldsymbol{\boxplus^{(\ell)}}$}}} &
\multicolumn{1}{T{2.37cm}}{\textcolor{white}{\makecell[c]{\textbf{Ego/Update}\\$\boldsymbol{(\mathcal B_\ell,\phi_\ell)}$}}}
\\[6pt]
\midrule
\endhead

\midrule
\multicolumn{7}{r}{\textit{continued on next page}} \\[6pt]
\endfoot

\bottomrule
\endlastfoot

\rowcolor{spatbarA}
\multicolumn{7}{@{}p{16.38cm}@{}}{%
\textcolor{white}{\textbf{1a\ \ Anchor models}\quad
\emph{fixed one-hop spatial propagation}}}
\\[6pt]

\hline
\rowcolor{spatrowA}
GCN~\cite{kipf2017semi} &
V &
\{1\} &
\hat{\mathbf A} &
\mathbf H^{(\ell)}\mathbf W^{(\ell)} &
\begin{gathered}\mixmsg^{(\ell)}=\mathbf P_1\Psi_1\\{\scriptstyle(\boxplus:\mathrm{id})}\end{gathered} &
\sigma(\mixmsg^{(\ell)})
\\[6pt]

\hline
\rowcolor{spatrowA}
GraphSAGE-Mean~\cite{hamilton2017inductive} &
V &
\{1\} &
\mathbf D^{-1}\mathbf A &
\mathbf H^{(\ell)} &
\begin{gathered}\mixmsg^{(\ell)}=\mathbf P_1\Psi_1\\{\scriptstyle(\boxplus:\mathrm{id})}\end{gathered} &
\begin{gathered}\normop\!\big(\sigma([\mathbf H^{(\ell)}\Vert\\ \mixmsg^{(\ell)}]\\\mathbf W^{(\ell)})\big)\end{gathered}
\\[6pt]

\hline
\rowcolor{spatrowA}
GIN~\cite{xu2019powerful} &
V &
\{1\} &
\mathbf A &
\mathbf H^{(\ell)} &
\begin{gathered}\mixmsg^{(\ell)}=\mathbf P_1\Psi_1\\{\scriptstyle(\boxplus:\mathrm{id})}\end{gathered} &
\begin{gathered}\MLP^{(\ell)}\!\big((1+\epsilon_\ell)\\\mathbf H^{(\ell)}+\mixmsg^{(\ell)}\big)\end{gathered}
\\[6pt]

\hline
\rowcolor{spatrowA}
MPNN~\cite{gilmer2017neural} &
V &
\{1\} &
\mathbf A &
\begin{gathered}\Psi_\ell(\vh_i^{(\ell)},\vh_j^{(\ell)},\\\ve_{ij})\end{gathered} &
\begin{gathered}\mathbf C_{1,i}\\
{}=\textstyle\sum\nolimits_{j\in\mathcal N(i)}\\
\Psi_\ell\bigl(\vh_i^{(\ell)},\\
\vh_j^{(\ell)},\ve_{ij}\bigr),\\
\mixmsg_i=\mathbf C_{1,i}\\
{\scriptstyle(\boxplus:\mathrm{id})}\end{gathered} &
\begin{gathered}\phi_\ell\bigl(\\
\vh_i^{(\ell)},\\
\mixmsg_i^{(\ell)}\bigr)\end{gathered}
\\[6pt]

\rowcolor{spatbarB}
\multicolumn{7}{@{}p{16.38cm}@{}}{%
\textcolor{white}{\textbf{1b\ \ Aggregators, gates, message maps, and implicit fixed points}\quad
\emph{one-hop support with nonlinear message or update maps}}}
\\[6pt]

\hline
\rowcolor{spatrowB}
Original GNN~\cite{scarselli2009graph} &
V &
\{1\} &
\mathbf A &
\begin{gathered}h_w\big(l_i,l_{ij},l_j,\\\vh_j^{\star}\big)\end{gathered} &
\begin{gathered}\mathbf C_{1,i}\\
{}=\textstyle\sum\nolimits_{j\in\mathcal N(i)}\\
h_w\bigl(l_i,l_{ij},l_j,\\
\vh_j^{\star}\bigr),\\
\mixmsg_i=\mathbf C_{1,i}\\
{\scriptstyle(\boxplus:\mathrm{id})}\end{gathered} &
\begin{gathered}\vh_i^\star=\mixmsg_i,\\
g_w(\vh_i^\star,l_i)\\
{\scriptstyle(\text{external readout})}\end{gathered}
\\[6pt]

\hline
\rowcolor{spatrowB}
IGNN~\cite{gu2020implicit} &
V &
\{1\} &
\mathbf A^\top &
\mathbf X^{\star}\mathbf W &
\begin{gathered}\mixmsg^{(\ell)}=\mathbf P_1\Psi_1\\
{\scriptstyle(\boxplus:\mathrm{id};}\\
{\scriptstyle\text{fixed-point slice})}\end{gathered} &
\begin{gathered}\phi\!\big(\mixmsg^{(\ell)}\\+b_\Omega(\mathbf X_{\mathrm{in}})\big)\end{gathered}
\\[6pt]

\hline
\rowcolor{spatrowB}
EIGNN~\cite{liu2021eignn} &
V &
\{1\} &
\gamma\,\hat{\mathbf A}\ {\scriptstyle(\text{fixed})} &
\mathbf X^{\star}g(\mathbf F)^\top &
\begin{gathered}\mixmsg^{(\ell)}=\mathbf P_1\Psi_1\\
{\scriptstyle(\boxplus:\mathrm{id};}\\
{\scriptstyle\text{fixed-point slice})}\end{gathered} &
\begin{gathered}\mathbf X^{\star}{:}\\
\mathbf X^{\star}{=}\mixmsg^{(\ell)}{+}\mathbf X\\
{\scriptstyle(\text{Sylvester eqm.})}\end{gathered}
\\[6pt]

\hline
\rowcolor{spatrowB}
GIND~\cite{chen2022optimization} &
V &
\{1\} &
\begin{gathered}\left[\mathbf P_1\right]_{ij}\\
{}=\dfrac{\mathbf A_{ij}}{\sqrt{2\tilde d_i}}\\
{\scriptstyle(\text{weighted one-hop support})}\end{gathered} &
\begin{gathered}
\boldsymbol\Delta_{ji}\\
{}=\frac{\vu_j^\star}{\sqrt{2\tilde d_j}}\\
{}-\frac{\vu_i^\star}{\sqrt{2\tilde d_i}},\\
\Psi_{ij}\\
{}=\tanh(\boldsymbol\Delta_{ji}\mathbf K^\top)\mathbf K
\end{gathered} &
\begin{gathered}\mathbf C_{1,i}\\
{}=\textstyle\sum_{j\in\mathcal N(i)}\\
[\mathbf P_1]_{ij}\Psi_{ij},\\
\mixmsg_i=\mathbf C_{1,i}\\
{\scriptstyle(\boxplus:\mathrm{id};}\\
{\scriptstyle\text{fixed-point slice})}\end{gathered} &
\begin{gathered}\vu_i^\star\\
{}=b_\Omega(\mathbf X_{\rm in})_i+\mixmsg_i,\\
\mathbf H^\star\\
{}=\mathbf U^\star-b_\Omega(\mathbf X_{\rm in}),\\
g_\Theta(\mathbf X+\mathbf H^\star)\\
{\scriptstyle(\text{external readout})}
\end{gathered}
\\[6pt]

\hline
\rowcolor{spatrowB}
JKNet~\cite{xu2018representation}$^\dagger$ &
V &
\{1\} &
\begin{gathered}\mathbf P_{\rm base}^{(\ell)}\\{\scriptstyle(\text{GCN/SAGE/GAT layer})}\end{gathered} &
\Psi_{\rm base}^{(\ell)}(\mathbf H^{(\ell)}) &
\begin{gathered}\mixmsg^{(\ell)}\\
{}=\mathbf P_{\rm base}^{(\ell)}\\
\Psi_{\rm base}^{(\ell)}\\
{\scriptstyle(\boxplus:\mathrm{id})}\end{gathered} &
\begin{gathered}\phi_{\rm base}^{(\ell)}\\
(\mathcal B_{\rm base}^{(\ell)},\mixmsg^{(\ell)});\\
\rho_{\rm JK}(\{\mathbf H^{(r)}\}_{r=1}^{L})
\\{\scriptstyle(\text{external readout})}\end{gathered}
\\[6pt]

\hline
\rowcolor{spatrowB}
PNA~\cite{corso2020principal} &
V &
\mathcal A_{\rm agg}\!\times\!\mathcal S &
\begin{gathered}\mathbf P_{a,s}=\Diag(s(\mathbf d))\mathbf A\\
{\scriptstyle(a\in\mathcal A_{\rm agg},\ s\in\mathcal S)}\end{gathered} &
\begin{gathered}\Psi_{ij}^{(\ell)}=\Psi_\ell\big(\vh_i^{(\ell)},\\
\vh_j^{(\ell)},\ve_{ij}\big)\end{gathered} &
\begin{gathered}\mathbf C_{a,s,i}\\
{}=a\!\bigl(\\
\left\{[\mathbf P_{a,s}]_{ij}\Psi_{ij}\right\}_{j\in\mathcal N(i)}\\
\bigr),\\
\mixmsg_i\\
{}=\big\Vert_{(a,s)\in\mathcal K}\mathbf C_{a,s,i}\\
{\scriptstyle(\boxplus:\Vert)}
\end{gathered} &
\phi_\ell(\mathbf H^{(\ell)},\mixmsg^{(\ell)})
\\[6pt]

\hline
\rowcolor{spatrowB}
Molecular FP~\cite{duvenaud2015convolutional} &
V &
\{1\} &
\mathbf A &
\mathbf H^{(\ell)} &
\begin{gathered}\mixmsg^{(\ell)}=\mathbf P_1\Psi_1\\{\scriptstyle(\boxplus:\mathrm{id})}\end{gathered} &
\begin{gathered}\sigma\!\big((\mathbf H^{(\ell)}+\mixmsg^{(\ell)})\\\mathbf W_{d_i}^{(\ell)}\big)\end{gathered}
\\[6pt]

\hline
\rowcolor{spatrowB}
GGNN~\cite{li2016gated} &
V &
\mathcal R_{\mathrm{rel}} &
\mathbf A_r &
\mathbf H^{(\ell)}\mathbf W_r &
\begin{gathered}\mixmsg^{(\ell)}\\
{}=\textstyle\sum\nolimits_{r\in\mathcal R_{\mathrm{rel}}}\\
\mathbf C_r^{(\ell)}\\
{\scriptstyle(\boxplus:\sum)}\end{gathered} &
\begin{gathered}\GRU\big(\mathbf H^{(\ell)},\\\mixmsg^{(\ell)}\big)\end{gathered}
\\[6pt]

\hline
\rowcolor{spatrowB}
GN block~\cite{battaglia2018relational}$^\ddagger$ &
V &
\{1\} &
\mathbf A &
\begin{gathered}\phi_\ell^{e}\big(\ve_{ij}^{(\ell)},\vh_i^{(\ell)},\\\vh_j^{(\ell)},u\big)\end{gathered} &
\begin{gathered}\mathbf C_{1,i}\\
{}=\Agg_{e\to v}\bigl(\\
\{\phi_\ell^{e}\}_{j\in\mathcal N(i)}\bigr),\\
\mixmsg_i=\mathbf C_{1,i}\\
{\scriptstyle(\boxplus:\mathrm{id})}\end{gathered} &
\begin{gathered}\phi_\ell^{v}\big(\mixmsg_i^{(\ell)},\\\vh_i^{(\ell)},u\big)\end{gathered}
\\[6pt]

\hline
\rowcolor{spatrowB}
ResGatedGCN~\cite{bresson2017residual} &
V &
\{1\} &
\mathbf A &
\begin{gathered}\sigma\big(\vh_i^{(\ell)}\mathbf W_a^{(\ell)}\\{+}\vh_j^{(\ell)}\mathbf W_b^{(\ell)}\big)\\\odot \vh_j^{(\ell)}\mathbf W_n^{(\ell)}\end{gathered} &
\begin{gathered}\mathbf C_{1,i}\\
{}=\textstyle\sum\nolimits_{j\in\mathcal N(i)}\\
\sigma\bigl(\vh_i^{(\ell)}\mathbf W_a^{(\ell)}\\
{}+\vh_j^{(\ell)}\mathbf W_b^{(\ell)}\bigr)\\
\odot \vh_j^{(\ell)}\mathbf W_n^{(\ell)},\\
\mixmsg_i=\mathbf C_{1,i}\\
{\scriptstyle(\boxplus:\mathrm{id})}\end{gathered} &
\begin{gathered}\mathbf H^{(\ell)}{+}\\\ReLU\big(\mathbf H^{(\ell)}\mathbf W_s^{(\ell)}\\{+}\mixmsg^{(\ell)}\big)\end{gathered}
\\[6pt]

\hline
\rowcolor{spatrowB}
DeeperGCN / GENConv~\cite{li2020deepergcn} &
V &
\{1\} &
\mathbf A &
\begin{gathered}\mathbf Z^{(\ell)}\\
{}=\ReLU\bigl(\\
\mathrm{Norm}(\mathbf H^{(\ell)})\bigr),\\
\Psi_{ij}\\
{}=\ReLU\bigl(\\
\vz_j^{(\ell)}{+}\mathbf 1\{\ve_{ij}\}\\
\ve_{ij}^{(\ell)}\bigr){+}\varepsilon\end{gathered} &
\begin{gathered}\mathbf C_{1,i}\\
{}=\Agg_{\beta/p}\bigl(\\
\{\Psi_\ell\}_{j\in\mathcal N(i)}\bigr),\\
\mixmsg_i=\mathbf C_{1,i}\\
{\scriptstyle(\boxplus:\mathrm{id})}\end{gathered} &
\begin{gathered}\mathbf H^{(\ell+1)}\\
{}=\mathbf H^{(\ell)}+\\
\MLP\!\big(\mathbf Z^{(\ell)}+\\
\MsgNorm_s(\mathbf Z^{(\ell)},\\
\mixmsg^{(\ell)})\big)
\end{gathered}
\\[6pt]

\hline
\rowcolor{spatrowB}
AMP~\cite{Errica2025AMP} &
V &
\{1\} &
\begin{gathered}\mathbf A\\{\scriptstyle(\text{fixed support})}\end{gathered} &
\begin{gathered}\mathbf F(j,\ell{-}1)\odot\\
\psi_\ell(\vh_j^{(\ell-1)},a_{ji})\\
{\scriptstyle(\mathbf F(j,\ell{-}1)\in[0,1]^d)}\end{gathered} &
\begin{gathered}\mathbf C_{1,i}\\
{}=\Agg\bigl(\\
\{\Psi_{ij}^{(\ell)}:\ j\in\mathcal N(i)\}\bigr),\\
\mixmsg_i=\mathbf C_{1,i}\\
{\scriptstyle(\boxplus:\mathrm{id})}\end{gathered} &
\begin{gathered}\phi_\ell\bigl(\\
\vh_i^{(\ell-1)},\\
\mixmsg_i^{(\ell)}\bigr)\\
{\scriptstyle(L\sim q(L),}\\
{\scriptstyle\rho_\ell\text{ external})}\end{gathered}
\\[6pt]

\hline
\rowcolor{spatrowB}
$\omega$GNN (channelwise)~\cite{Eliasof2023omegaGNN} &
V &
\{1\} &
\begin{gathered}\mathbf P_1=\mathbf S^{(\ell)}\\
{\scriptstyle(\mathbf S^{(\ell)}:\ \text{base GNN}}\\
{\scriptstyle\text{operator})}\end{gathered} &
\begin{gathered}\Psi_1=\mathbf H^{(\ell)}\boldsymbol\Omega^{(\ell)}\\
{\scriptstyle(\boldsymbol\Omega^{(\ell)}}\\
{\scriptstyle{}=\Diag(\boldsymbol\omega^{(\ell)}))}\end{gathered} &
\begin{gathered}\mixmsg^{(\ell)}=\mathbf P_1\Psi_1\\{\scriptstyle(\boxplus:\mathrm{id})}\end{gathered} &
\begin{gathered}\mathcal B_\omega(\mathbf H)\\
{}=\mathbf H(\mathbf I_d-\boldsymbol\Omega^{(\ell)}),\\
\sigma\!\big((\mathcal B_\omega(\mathbf H^{(\ell)})\\
{}+\mixmsg^{(\ell)})\mathbf K^{(\ell)}\big)
\end{gathered}
\\[6pt]

\hline
\rowcolor{spatrowB}
GSN-v (GIN inst.)~\cite{bouritsas2023improving} &
V &
\{1\} &
\mathbf A &
[\mathbf H^{(\ell)}\,\Vert\,\mathbf c_S] &
\begin{gathered}\mixmsg^{(\ell)}=\mathbf P_1\Psi_1\\{\scriptstyle(\boxplus:\mathrm{id})}\end{gathered} &
\begin{gathered}\MLP^{(\ell)}\big(\\
(1+\epsilon_\ell)[\mathbf H^{(\ell)}\Vert\mathbf c_S]\\
{}+\mixmsg^{(\ell)}\big)\end{gathered}
\\[6pt]

\hline
\rowcolor{spatrowB}
GraphSNN~\cite{wijesinghe2022graphsnn} &
V &
\{1\} &
\begin{gathered}\big[(\widetilde\omega_{ij}{+}1)A_{ij}\big]_{ij}\\{\scriptstyle(\widetilde\omega_{ij}:\ \text{row-normalized}}\\{\scriptstyle\text{overlap coefficient})}\end{gathered} &
\mathbf H^{(\ell)} &
\begin{gathered}\mixmsg^{(\ell)}=\mathbf P_1\Psi_1\\{\scriptstyle(\boxplus:\mathrm{id})}\end{gathered} &
\begin{gathered}\MLP^{(\ell)}\!\bigl(\\
\gamma^{(\ell)}\!\left[1+\textstyle\sum_j\widetilde\omega_{ij}\right]\\
\mathbf H_i^{(\ell)}\\
{}+\mixmsg_i^{(\ell)}\bigr)\end{gathered}
\\[6pt]

\rowcolor{spatbarC}
\multicolumn{7}{@{}p{16.38cm}@{}}{%
\textcolor{white}{\textbf{1c\ \ Adaptive-support and capsule-indexed extensions}\quad
\emph{state-dependent support or capsule-indexed messages}}}
\\[6pt]

\hline
\rowcolor{spatrowC}
CoGNN~\cite{Finkelshtein2024CoGNN} &
V &
\{1\} &
\begin{gathered}\Diag(\mathbf 1\{a\!\in\!\{L,S\}\})\,\mathbf A\\{\boldsymbol{\cdot}}\Diag(\mathbf 1\{a\!\in\!\{B,S\}\})\end{gathered} &
\mathbf H^{(\ell)} &
\begin{gathered}\mixmsg^{(\ell)}=\mathbf P_1\Psi_1\\{\scriptstyle(\boxplus:\mathrm{id};\ \text{masked-MP slice})}\end{gathered} &
\eta^{(\ell)}(\mathbf H^{(\ell)},\mixmsg^{(\ell)})
\\[6pt]

\hline
\rowcolor{spatrowC}
NCGNN~\cite{Yang2024NCGNN} &
V &
\{1,\dots,C\} &
\begin{gathered}\mathbf P_s=\bar{\mathbf A}=f_\xi(\widetilde{\mathbf A})\\
{\scriptstyle(\text{sparsified polynomial}}\\
{\scriptstyle\text{filter})}\end{gathered} &
\begin{gathered}\Psi_s(j)=\vp_j^{(s)}\\
{}=\textstyle\sum_{k=1}^{K}c_{j,k,s}^{(\ell)}\,\\
\vh_j^{(k)}\mathbf W_{k,s}^{(\ell)}\end{gathered} &
\begin{gathered}\mixmsg_i^{(\ell)}\\
{}=\big\Vert_{s=1}^{C}\mathbf C_{i,s},\\
\mathbf C_{i,s}\\
{}=\textstyle\sum_j[\mathbf P_s]_{ij}\Psi_s(j)\\
{\scriptstyle(\boxplus:\Vert;\ \text{routing slice})}
\end{gathered} &
\begin{gathered}\big\Vert_{s=1}^{C}\\
\squash\!\left(\mathbf C_{i,s}+\mathbf b_s\right)\end{gathered}
\\[6pt]

\hline
\rowcolor{spatrowC}
Flow2GNN~\cite{Huang2024Flow2GNN} &
V &
\{0,1,2\} &
\begin{gathered}g(\mathbf A),\;g(\mathbf A\!\odot\!G^{(\ell)}),\;\\g(\mathbf A\!\odot\!(1{-}G^{(\ell)}))\\
{\scriptstyle(G^{(\ell)}\text{: complementary}}\\
{\scriptstyle\text{Bernoulli mask};}\\
{\scriptstyle\;g(\cdot)\text{: plug-in MP op})}\end{gathered} &
\mathbf H^{(\ell)}\mathbf W_k^{(\ell)} &
\begin{gathered}\mixmsg_i^{(\ell)}=\big\Vert_{k=0}^{2}\mathbf C_{i,k},\\
\mathbf C_k=\mathbf P_k\Psi_k\\
{\scriptstyle(\boxplus:\Vert)}
\end{gathered} &
\begin{gathered}\textstyle\sum_{k=0}^{2}p_{i,k}^{(\ell)}\\
{}\odot\sigma(\mathbf C_{i,k})\\
{\scriptstyle(p_{i,k}^{(\ell)}\text{: cosine-sim}}\\
{\scriptstyle\text{softmax computed}}\\
{\scriptstyle\text{node-locally from }\mathbf H^{(0)}}\\
{\scriptstyle\text{and}}\\
{\scriptstyle\{\sigma(\mathbf C_{i,r})\}_r)}
\end{gathered}
\\[6pt]

\rowcolor{spatbarD}
\multicolumn{7}{@{}p{16.38cm}@{}}{%
\textcolor{white}{\textbf{1d\ \ Fixed operator banks}\quad
\emph{non-state-dependent propagation through directional fields}}}
\\[6pt]

\hline
\rowcolor{spatrowD}
DGN~\cite{Beaini2021DGN} &
V &
\begin{gathered}\mathcal K=\mathcal K_0\!\times\\
\{-1,0,1\},\\
\mathcal K_0=\{0\}\cup\\
\{(q,\mathrm{av}),\\
(q,\mathrm{dx}):\\
q=1,\ldots,Q\}
\end{gathered} &
\begin{gathered}\mathbf P_{k,a}\\
{}=\Diag(S_a(\mathbf d))\mathbf P_k,\\
\{\mathbf P_k\}_{k\in\mathcal K_0}\\
{}=\{\mathbf D^{-1}\mathbf A,\\
B_{\rm av}(\nabla u_q),\\
B_{\rm dx}(\nabla u_q)\}_{q=1}^{Q}\\[3pt]
\end{gathered} &
\mathbf H^{(\ell)} &
\begin{gathered}\mixmsg^{(\ell)}\\
{}=\big\Vert_{(k,a)\in\mathcal K}\\
\mathbf P_{k,a}\mathbf H^{(\ell)}\\
{\scriptstyle(\boxplus:\Vert)}
\end{gathered} &
\begin{gathered}U^{(\ell)}\!\bigl(\\
\vh_i^{(\ell)},\\
T_{\rm dx}(\mixmsg_i^{(\ell)})\bigr)\\
{\scriptstyle(U\text{ may omit }\vh_i}\\
{\scriptstyle\text{in the simple form})}\end{gathered}
\\[6pt]

\hline
\rowcolor{spatrowD}
Dir-GNN~\cite{rossi2024edge} &
V &
\{\leftarrow,\rightarrow\} &
\begin{gathered}\operatorname{supp}(\mathbf P_{\leftarrow})\\
{}=\operatorname{supp}(\mathbf A^\top),\\
\operatorname{supp}(\mathbf P_{\rightarrow})\\
{}=\operatorname{supp}(\mathbf A)\\
{\scriptstyle(\text{base-aggregator scalar}}\\
{\scriptstyle\text{weights})}\end{gathered} &
\begin{gathered}\Psi_k(i,j)=\\
(\vh_j^{(\ell)},\vh_i^{(\ell)})\end{gathered} &
\begin{gathered}\mathbf C_{i,k}^{(\ell)}\\
{}=\Agg_k^{(\ell)}\!\bigl(\\
\bigl\{\!\bigl\{\\
[\mathbf P_k]_{ij}\Psi_k(i,j)\\
\bigr\}\!\bigr\}_{\substack{j:\\{}[\mathbf P_k]_{ij}\ne0}}\\
\bigr),\\
\mixmsg_i^{(\ell)}\\
{}=\mathbf C_{i,\leftarrow}^{(\ell)}\\
\Vert\mathbf C_{i,\rightarrow}^{(\ell)}\\
{\scriptstyle(\boxplus:\Vert)}
\end{gathered} &
\begin{gathered}\mathrm{COM}^{(\ell)}\!\bigl(\\
\vh_i^{(\ell)},\\
\mathbf C_{i,\leftarrow}^{(\ell)},\\
\mathbf C_{i,\rightarrow}^{(\ell)}\bigr)
\end{gathered}
\\[6pt]

\end{longtable}
\endgroup

\colorlet{attbarA}{repAttention}
\colorlet{attbarB}{repAttention!80}
\colorlet{attbarC}{repAttention!62}
\colorlet{attrowA}{repAttention!9}
\colorlet{attrowB}{repAttention!9}
\colorlet{attrowC}{repAttention!9}

\providecommand{\MLP}{\operatorname{MLP}}
\providecommand{\GRU}{\operatorname{GRU}}
\providecommand{\Diag}{\operatorname{diag}}
\providecommand{\LReLU}{\operatorname{LReLU}}
\providecommand{\sm}{\operatorname{softmax}}
\providecommand{\Conv}{\operatorname{Conv1D}}
\providecommand{\FC}{\operatorname{FC}}
\providecommand{\FF}{\operatorname{FF}}
\providecommand{\FFN}{\operatorname{FFN}}
\providecommand{\LN}{\operatorname{LN}}

\newcommand{\prop}[1]{\scriptscriptstyle #1}

\makeatletter
\@ifundefined{NC@find@Y}{\newcolumntype{Y}[1]{>{\raggedright\arraybackslash}p{#1}}}{}
\makeatother

\begingroup
\scriptsize
\renewcommand{\arraystretch}{1.5}
\arrayrulecolor{black!45}
\setlength{\arrayrulewidth}{0.5pt}
\setlength{\tabcolsep}{3.67pt}
\setlength{\LTpre}{6pt}
\setlength{\LTpost}{6pt}
\setlength{\LTcapwidth}{\textwidth}
\begin{longtable}{@{}>{\raggedright\arraybackslash}p{2.22cm}>{\centering\arraybackslash$}p{1.04cm}<{$}>{\centering\arraybackslash$}p{1.48cm}<{$}>{\raggedright\arraybackslash$}p{3.17cm}<{$}>{\raggedright\arraybackslash$}p{2.17cm}<{$}>{\raggedright\arraybackslash$}p{2.37cm}<{$}>{\raggedright\arraybackslash$}p{2.37cm}<{$}@{}}
\caption{\scriptsize\textbf{Family 2. Attention-Based GNNs.}
Unified-component decompositions of attention-based message passing. Scalar attention weights
belong to $\mathbf P_k$, edge-conditioned values to $\Psi_k$, and node-wise gates or residuals
to the update. Pairwise aggregation forms each channel contribution before head or branch
mixing. DAGNN$^\dagger$ reports one update in its sequential topological scan.}
\label{tab:family2_attention_slots}\\[6pt]

\toprule
\rowcolor{attbarA}
\textcolor{white}{\textbf{Method}} &
\multicolumn{1}{>{\centering\arraybackslash}p{1.04cm}}{\textcolor{white}{\(\boldsymbol{\mathcal X}\)}} &
\multicolumn{1}{>{\centering\arraybackslash}p{1.48cm}}{\textcolor{white}{\(\boldsymbol{\mathcal K}\)}} &
\multicolumn{1}{>{\centering\arraybackslash}p{3.17cm}}{\textcolor{white}{\makecell[c]{\textbf{Propagation}\\\(\boldsymbol{\mathbf P_k^{(\ell)}}\)}}} &
\multicolumn{1}{>{\centering\arraybackslash}p{2.17cm}}{\textcolor{white}{\makecell[c]{\textbf{Message}\\\(\boldsymbol{\Psi_k^{(\ell)}}\)}}} &
\multicolumn{1}{>{\centering\arraybackslash}p{2.37cm}}{\textcolor{white}{\makecell[c]{\textbf{Mixing}\\\(\boldsymbol{\boxplus^{(\ell)}}\)}}} &
\multicolumn{1}{>{\centering\arraybackslash}p{2.37cm}}{\textcolor{white}{\makecell[c]{\textbf{Ego/Update}\\\(\boldsymbol{(\mathcal B_\ell,\phi_\ell)}\)}}}\\[6pt]
\midrule
\endfirsthead

\multicolumn{7}{c}{\tablename\ \thetable\ -- continued}\\[6pt]
\toprule
\rowcolor{attbarA}
\textcolor{white}{\textbf{Method}} &
\multicolumn{1}{>{\centering\arraybackslash}p{1.04cm}}{\textcolor{white}{\(\boldsymbol{\mathcal X}\)}} &
\multicolumn{1}{>{\centering\arraybackslash}p{1.48cm}}{\textcolor{white}{\(\boldsymbol{\mathcal K}\)}} &
\multicolumn{1}{>{\centering\arraybackslash}p{3.17cm}}{\textcolor{white}{\makecell[c]{\textbf{Propagation}\\\(\boldsymbol{\mathbf P_k^{(\ell)}}\)}}} &
\multicolumn{1}{>{\centering\arraybackslash}p{2.17cm}}{\textcolor{white}{\makecell[c]{\textbf{Message}\\\(\boldsymbol{\Psi_k^{(\ell)}}\)}}} &
\multicolumn{1}{>{\centering\arraybackslash}p{2.37cm}}{\textcolor{white}{\makecell[c]{\textbf{Mixing}\\\(\boldsymbol{\boxplus^{(\ell)}}\)}}} &
\multicolumn{1}{>{\centering\arraybackslash}p{2.37cm}}{\textcolor{white}{\makecell[c]{\textbf{Ego/Update}\\\(\boldsymbol{(\mathcal B_\ell,\phi_\ell)}\)}}}\\[6pt]
\midrule
\endhead

\midrule
\multicolumn{7}{r}{\textit{continued on next page}}\\[6pt]
\endfoot

\bottomrule
\endlastfoot

\rowcolor{attbarA}
\multicolumn{7}{@{}p{16.38cm}@{}}{\textcolor{white}{\textbf{2a\ \ Query--key attention}\quad
\emph{static and dynamic attention operators}}}\\[6pt]

\hline
\rowcolor{attrowA}
GAT~\cite{velickovic2018gat} &
V &
\{1,\dots,M\} &
{\prop{\begin{gathered}\sm_{j\in\widetilde{\mathcal N}(i)}\!\bigl(\\
\LReLU\!\bigl(\\
[\vh_i\mathbf W_m\Vert\vh_j\mathbf W_m]\\
\mathbf a_m\bigr)\bigr)
\end{gathered}}} &
\mathbf H^{(\ell)}\mathbf W_m^{(\ell)} &
\begin{gathered}\mathbf C_m=\mathbf P_m\Psi_m,\\
\mixmsg^{(\ell)}=\big\Vert_m\mathbf C_m\\
{\scriptstyle(\boxplus:\Vert,}\\
{\scriptstyle\mathrm{avg.\ at\ final\ layer})}\end{gathered} &
\begin{gathered}\mathcal B_\ell=0,\\ \sigma(\mixmsg^{(\ell)})\end{gathered}\\[6pt]

\hline
\rowcolor{attrowA}
GATv2~\cite{brody2022attentive} &
V &
\{1,\dots,M\} &
{\prop{\begin{gathered}\sm_{j\in\widetilde{\mathcal N}(i)}\!\bigl(\\
\LReLU\!\bigl(\\
\vh_i\mathbf W_{q,m}\\
{}+\vh_j\mathbf W_{k,m}\bigr)\\
\mathbf a_m\bigr)
\end{gathered}}} &
\mathbf H^{(\ell)}\mathbf W_{k,m}^{(\ell)} &
\begin{gathered}\mathbf C_m=\mathbf P_m\Psi_m,\\
\mixmsg^{(\ell)}=\big\Vert_m\mathbf C_m\\
{\scriptstyle(\boxplus:\Vert,}\\
{\scriptstyle\mathrm{avg.\ at\ final\ layer})}\end{gathered} &
\begin{gathered}\mathcal B_\ell=0,\\ \sigma(\mixmsg^{(\ell)})\end{gathered}\\[6pt]

\hline
\rowcolor{attrowA}
GAT-POS~\cite{ma2021gatpos} &
V &
\{1,\dots,M\} &
{\prop{\begin{gathered}\sm_{j\in\widetilde{\mathcal N}(i)}\!\bigl(\\
\LReLU\!\bigl(\\
[\vh_i\mathbf W_m{+}p_i\mathbf U_m\\
{}\Vert\vh_j\mathbf W_m{+}p_j\mathbf U_m]\\
\mathbf a_m\bigr)\bigr)
\end{gathered}}} &
\mathbf H^{(\ell)}\mathbf W_m^{(\ell)}\ {\scriptstyle(\text{no }p)} &
\begin{gathered}\mathbf C_m=\mathbf P_m\Psi_m,\\
\mixmsg^{(\ell)}=\big\Vert_m\mathbf C_m\\
{\scriptstyle(\boxplus:\Vert,}\\
{\scriptstyle\mathrm{avg.\ at\ final\ layer})}\end{gathered} &
\begin{gathered}\mathcal B_\ell=0,\\ \sigma(\mixmsg^{(\ell)})\end{gathered}\\[6pt]

\hline
\rowcolor{attrowA}
AGNN~\cite{thekumparampil2018attention} &
V &
\{1\} &
{\prop{\sm_{j\in\widetilde{\mathcal N}(i)}(\beta_\ell\cos(\vh_i,\vh_j))}} &
\mathbf H^{(\ell)} &
\begin{gathered}\mixmsg^{(\ell)}=\mathbf P_1\Psi_1\\{\scriptstyle(\boxplus:\mathrm{id})}\end{gathered} &
\begin{gathered}\mathcal B_\ell=0,\\ \phi_\ell(\mixmsg)=\mixmsg\end{gathered}\\[6pt]

\hline
\rowcolor{attrowA}
GaAN~\cite{zhang2018gaan} &
V &
\{1,\dots,M\} &
{\prop{\begin{gathered}\sm_{j\in\mathcal N(i)}\langle \vh_i{\scriptstyle\mathbf W_{xa}^{(m)}},\\\vh_j{\scriptstyle\mathbf W_{za}^{(m)}}\rangle\end{gathered}}} &
\FC_{\theta_v^{(m)}}^{h}(\vh_j) &
\begin{gathered}\mathbf C_{m,i}=\mathbf P_m\Psi_m,\\
\mixmsg_i^{(\ell)}=\big\Vert_m g_i^{(m)}\mathbf C_{m,i}\\
{\scriptstyle(\boxplus:\Vert\ \text{with head gates})}\end{gathered} &
\begin{gathered}{\scriptstyle\mathcal B_{\ell,i}=\vh_i^{(\ell)},}\\
{\scriptstyle\mathbf q_i=\vh_i\Vert}\\
{\scriptstyle\max_{j\in\mathcal N(i)}\FC_m(\vh_j)}\\
{\scriptstyle{}\Vert\operatorname{mean}_{j\in\mathcal N(i)}\vh_j,}\\
{\scriptstyle\mathbf g_i=\sigma(\FC(\mathbf q_i)),}\\
\FC\bigl(\\
\mathcal B_{\ell,i}\Vert\mixmsg_i^{(\ell)}\bigr)\\[3pt]
\end{gathered}\\[6pt]

\hline
\rowcolor{attrowA}
CPA (f-Scaled)~\cite{zhang2020cardinality} &
V &
\{1\} &
{\prop{\begin{gathered}P_{ij}=|\widetilde{\mathcal N}(i)|\,\alpha_{ij},\\
\alpha_{ij}=\sm_{j\in\widetilde{\mathcal N}(i)}(e_{ij})\end{gathered}}} &
\mathbf H^{(\ell)} &
\begin{gathered}\mathbf C_{1,i}\\
{}=\textstyle\sum\nolimits_{j\in\tilde{\mathcal N}(i)}\\
\mathbf P_{1,ij}\Psi_{1,j},\\
\mixmsg_i=\mathbf C_{1,i}\\
{\scriptstyle(\boxplus:\mathrm{id})}\end{gathered} &
f^{(\ell)}\!\big(\mixmsg_i^{(\ell)}\big)\\[6pt]

\hline
\rowcolor{attbarB}
\multicolumn{7}{@{}p{16.38cm}@{}}{\textcolor{white}{\textbf{2b\ \ Edge-conditioned and generated weights}\quad
\emph{edge features, gating, hypernetworks, diffusion attention}}}\\[6pt]

\hline
\rowcolor{attrowB}
ECC~\cite{simonovsky2017dynamic} &
V &
\{1\} &
{\prop{\begin{gathered}P_{ij}=\mathbf{1}_{\{j\in\widetilde{\mathcal N}(i)\}}\\
/|\widetilde{\mathcal N}(i)|\end{gathered}}} &
\begin{gathered}\vh_j\,F^{(\ell)}(L(j,i))^\top\\
\end{gathered} &
\begin{gathered}\mixmsg^{(\ell)}=\mathbf P_1\Psi_1\\{\scriptstyle(\boxplus:\mathrm{id})}\end{gathered} &
\begin{gathered}\mathcal B_\ell=0,\\ \mixmsg^{(\ell)}+b_\ell\end{gathered}\\[6pt]

\hline
\rowcolor{attrowB}
DMP (edge-wise)~\cite{yang2021diverse} &
V &
\{1\} &
[\mathbf P_1]_{vu}=\mathbf 1_{\{u\in\mathcal N(v)\cup\{v\}\}} &
\begin{gathered}\mathbf 
\tanh([\vh_v\Vert \vh_u]\mathbf W_c),\\
\Psi_1(\vh_v,\vh_u)\\
\odot \vh_u\end{gathered} &
\begin{gathered}\mathbf C_{1,v}\\
{}=\textstyle\sum\nolimits_{u\in\mathcal N(v)\cup\{v\}}\\
[\mathbf P_1]_{vu}\Psi_1(\vh_v,\vh_u),\\
\mixmsg_v=\mathbf C_{1,v}\\
{\scriptstyle(\boxplus:\mathrm{id})}\end{gathered} &
\begin{gathered}\mathcal B_\ell=0,\\
\sigma\big(\mixmsg_v^{(\ell)}\mathbf W^{(\ell)}\big)
\end{gathered}\\[6pt]

\hline
\rowcolor{attrowB}
MAGNA~\cite{wang2021magna} &
V &
\begin{gathered}\{(m,k):\\1{\leq}m{\leq}M,\\0{\leq}k{\leq}K\}\end{gathered} &
{\prop{\begin{gathered}c_k\big(\sm(\mathbf S_m^{(\ell)})\big)^k\\
{\scriptstyle(\text{finite unrolled})},\\
c_k{=}\alpha(1{-}\alpha)^k\\
(k{<}K),\\ c_K{=}(1{-}\alpha)^K
\end{gathered}}} &
\LN(\mathbf H^{(\ell)}) &
\begin{gathered}\mathbf z_m\\
{}=\textstyle\sum_{k=0}^{K}\mathbf P_{m,k}\Psi_{m,k},\\
\mixmsg^{(\ell)}\\
{}=(\Vert_m\mathbf z_m)\mathbf W_o\\
{\scriptstyle(\boxplus:\Vert_m\ \text{then}}\\
{\scriptstyle\text{output projection})}
\end{gathered} &
\begin{gathered}\mathbf R\\
{}=\mathbf H^{(\ell)}+\mixmsg^{(\ell)},\\
\mathbf H^{(\ell+1)}\\
{}=\mathbf R+\FFN(\LN(\mathbf R)),\\
{\scriptstyle\FFN(\mathbf Z)}\\
{\scriptstyle{}=\mathrm{ReLU}(\mathbf Z\mathbf W_1)\mathbf W_2}
\end{gathered}\\[6pt]

\hline
\rowcolor{attrowB}
VR-GNN~\cite{Shi2024VRGNN} &
V &
\{1\} &
{\prop{\begin{gathered}P_{ij}=\theta\,\beta_{ij},\\
\beta_{ij}=\sm_{j\in\mathcal N(i)}\!\bigl(\\
\langle\vh_i,\,\vh_j\mathbf W^{(\ell)}\\
{}+\mathbf z_{ji}^{(\ell)}\rangle\bigr)\end{gathered}}} &
\begin{gathered}\vh_j\mathbf W^{(\ell)}\\
{}+\mathbf z_{ji}^{(\ell)}\\
{\scriptstyle(\mathbf z_{ji}^{(\ell+1)}\text{ is parallel}}
\end{gathered} &
\begin{gathered}\mixmsg^{(\ell)}=\mathbf P_1\Psi_1\\{\scriptstyle(\boxplus:\mathrm{id},\ \text{node slice})}\end{gathered} &
\begin{gathered}\mathcal B_{\ell,i}\\
{}=(1-\theta)\vh_i^{(0)},\\
\phi_\ell(\mathcal B_{\ell,i},\mixmsg_i)\\
{}=\mixmsg_i+\mathcal B_{\ell,i},\\
{\scriptstyle\vh_i^{(0)}=\mathrm{ReLU}(\MLP(\vx_i)),}\\
{\scriptstyle\mathbf z_{ji}^{(\ell+1)}}\\
{\scriptstyle{}=\mathbf z_{ji}^{(\ell)}\mathbf W_{\rm rel}^{(\ell)}}
\end{gathered}\\[6pt]

\hline
\rowcolor{attrowB}
PMP-GAT~\cite{He2024PMPGNN} &
V &
\{1\} &
{\prop{\begin{gathered}P_{ij}=\\
E_{ij}\Big/\!\textstyle\sum_{k\in\mathcal N(i)}E_{ik},\\
E_{ij}=A_{ij}\exp\!\bigl(\\
C_{ij}-\beta S_{ij}\bigr),\\
{\scriptstyle C:\ \text{feature+position}}\\
{\scriptstyle\text{correlation},}\\
{\scriptstyle S:\ \text{feature+position}}\\
{\scriptstyle\text{distance}}
\end{gathered}}} &
\hspace{4pt}\mathbf H^{(\ell)}\mathbf W_h^{(\ell)} &
\begin{gathered}\mixmsg^{(\ell)}=\mathbf P_1\Psi_1\\{\scriptstyle(\boxplus:\mathrm{id})}\end{gathered} &
\begin{gathered}\mathcal B_{\ell,i}\\
{}=\tfrac{\theta}{|\mathcal N(i)|}\Psi_{1,i},\\
\mathcal B_{\ell,i}{+}\mixmsg_i^{(\ell)}
\end{gathered}\\[6pt]

\hline
\rowcolor{attrowB}
GDAMNs (stable E-step)~\cite{Chen2023GDAMNs} &
V &
\{1\} &
{\prop{\begin{gathered}\mathbf A^{\rm stable}\\
{}=\bigl(\mathbf A\odot\\
\sigma(\hat{\mathbf Y}Q\hat{\mathbf Y}^{\top})\bigr)\\
\odot\sm_j^{\mathcal N(i)}\!\bigl(\\
-\cos(\mathbf g_i,\mathbf g_j)\bigr),\\
\mathbf g_i=\mathbf s\odot\\
\mathrm{ReLU}(\mathbf x_i\mathbf W_{\rm proj})
\end{gathered}}} &
\mathbf H^{(\ell)} &
\begin{gathered}\mathbf C_{1,i}\\
{}=\textstyle\sum\nolimits_{j\in\mathcal N(i)}\\
\mathbf P_{1,ij}\Psi_{1,j},\\
\mixmsg_i=\mathbf C_{1,i}\\
{\scriptstyle(\boxplus:\mathrm{id})}\end{gathered} &
\begin{gathered}\phi_\theta^Q\!\bigl(\\
\mixmsg^{(\ell)},\mathbf H^{(\ell)}\bigr)
\end{gathered}\\[6pt]

\hline
\rowcolor{attbarC}
\multicolumn{7}{@{}p{16.38cm}@{}}{\textcolor{white}{\textbf{2c\ \ Adaptive routing and receptive-field-adaptive attention}}}\\[6pt]

\hline
\rowcolor{attrowC}
DAGNN$^\dagger$~\cite{thost2021directed} &
V &
\{1\} &
{\prop{\begin{gathered}P_{vu}^{(\ell)}=\mathbf{1}_{\{u\in\mathcal P(v)\}}\\
\sm_{u\in\mathcal P(v)}\!\bigl(\\
\vh_v^{(\ell-1)}\vw_1^{(\ell)}\\
{}+\vh_u^{(\ell)}\vw_2^{(\ell)}\bigr)
\end{gathered}}} &
\begin{gathered}\mathbf H^{(\ell)}\\
\end{gathered} &
\begin{gathered}\mixmsg_v^{(\ell)}=\mathbf P_1\Psi_1\\
{\scriptstyle(\boxplus:\mathrm{id},}\\
{\scriptstyle\text{topological-scan slice})}\end{gathered} &
\begin{gathered}\vh_v^{(\ell)}=\GRU\big(
\underbrace{\vh_v^{(\ell-1)}}_{\rm input},\\
\underbrace{\mixmsg_v^{(\ell)}}_{\rm state}\big)
\end{gathered}\\[6pt]

\hline
\rowcolor{attrowC}
Deformable GCN~\cite{park2022deformable} &
V &
\begin{gathered}\{(q,r):\\0{\leq}q{\leq}L{+}1,\\1{\leq}r{\leq}R\}\end{gathered} &
{\prop{\begin{gathered}\mathbf{1}_{\{u\in\mathcal N_q(v)\}}\\
{}\cdot\sm_{u\in\mathcal N_q(v)}\!\bigl(\\
\mathbf r_{uv}^{(q)\top}\bigl(\\
\widetilde{\boldsymbol\phi}_r+\\
\Delta_r(\mathbf e_v^{(q)})\bigr)\bigr),\\
{\scriptstyle q{\leq}L:\ \text{fixed smoothed-feature}}\\
{\scriptstyle\text{kNN},\ q{=}L{+}1:\ G}
\end{gathered}}} &
\hspace{4pt}\mathbf H^{(\ell)}\mathbf W_r &
\begin{gathered}\mathbf z_v^{(q)}\\
{}=\textstyle\sum_{r=1}^{R}\big[\mathbf P_{q,r}\Psi_r\big]_{v},\\
\mathbf y_v^{(q)}\\
{}=\mathbf z_v^{(q)}/\|\mathbf z_v^{(q)}\|_2,\\
\mixmsg_v\\
{}=\textstyle\sum_{q=0}^{L+1}s_v^{(q)}\mathbf y_v^{(q)},\\
s_v^{(q)}\\
{}=\sm_q(\mathbf z^\top\mathbf y_v^{(q)})\\
{\scriptstyle(\boxplus:\sum_q s_v^{(q)}(\cdot))}\\[3pt]
\end{gathered} &
\begin{gathered}\mathcal B_\ell=0,\\ \phi_\ell(\mixmsg)=\mixmsg\end{gathered}\\[6pt]

\hline
\rowcolor{attrowC}
DeGTA~\cite{Wang2024DeGTA} &
V &
\{\mathrm{loc},\mathrm{glob}\} &
{\prop{\begin{gathered}P^{\rm loc}_{ij}=\sm_{j\in\mathcal N(i)}\!\big(\\
\alpha p_{ij}{+}\beta s_{ij}{+}\gamma a_{ij}\big),\\
P^{\rm glob}_{ij}=\sm_{j\in\mathcal K_i}\!\big(\\
\alpha U^s_{ij}{+}\beta U^p_{ij}{+}\gamma\widehat U^a_{ij}\big),\\
{\scriptstyle\mathcal K_i:\ \text{hard-sampled}}\\
{\scriptstyle\text{non-neighbor set}}
\end{gathered}}} &
\begin{gathered}\mathbf H_a^{(\ell)}\\
{}=E_a(\mathbf H^{(\ell-1)})\end{gathered} &
\begin{gathered}\mathbf C_{\rm loc}\\
{}=\mathbf P_{\rm loc}\mathbf H_a^{(\ell)},\\
\mathbf C_{\rm glob}\\
{}=\mathbf P_{\rm glob}\mathbf H_a^{(\ell)},\\
\mixmsg^{(\ell)}\\
{}=\mathbf C_{\rm loc}\mathbf W_l^{(\ell)}\\
{}+\mathbf C_{\rm glob}\mathbf W_g^{(\ell)}\\
{\scriptstyle(\boxplus:\text{learned local--global}}\\
{\scriptstyle\text{sum, fixed-sample slice})}
\end{gathered} &
\begin{gathered}\mathcal B_\ell=0,\\
\phi_\ell(\mixmsg^{(\ell)})\\
{}=\mixmsg^{(\ell)}\\
{\scriptstyle(\text{full hard-sampling}}\\
{\scriptstyle\text{model is boundary})}
\end{gathered}\\[6pt]

\hline
\rowcolor{attrowC}
GeniePath~\cite{liu2019geniepath} &
V &
\{1\} &
{\prop{\begin{gathered}\sm_{j\in\widetilde{\mathcal N}(i)}\!\bigl(\\
\tanh\bigl(\\
\vh_i\mathbf W_s{+}\vh_j\mathbf W_d\\
\bigr)\mathbf v\bigr)
\end{gathered}}} &
\mathbf H^{(\ell)} &
\begin{gathered}\mixmsg^{(\ell)}=\mathbf P_1\Psi_1\\{\scriptstyle(\boxplus:\mathrm{id})}\end{gathered} &
\begin{gathered}{\scriptstyle\mathcal B_\ell=\mathbf C_i^{(\ell)},}\\
\mathbf u_i=\tanh(\mixmsg_i\mathbf W),\\
{\scriptstyle\mathbf i_i=\sigma(\mathbf u_i\mathbf W_i),}\\
{\scriptstyle\mathbf f_i=\sigma(\mathbf u_i\mathbf W_f),}\\
{\scriptstyle\mathbf o_i=\sigma(\mathbf u_i\mathbf W_o),}\\
{\scriptstyle\widetilde{\mathbf C}_i}\\
{\scriptstyle{}=\tanh(\mathbf u_i\mathbf W_c),}\\
\mathbf C_i^+\\
{}=\mathbf f_i\odot\mathbf C_i\\
{}+\mathbf i_i\odot\widetilde{\mathbf C}_i,\\
\vh_i^+\\
{}=\mathbf o_i\odot\tanh(\mathbf C_i^+)\\[3pt]
\end{gathered}\\[6pt]

\hline
\rowcolor{attbarC}
\multicolumn{7}{@{}p{16.38cm}@{}}{\textcolor{white}{\textbf{2d\ \ Outside the declared single-stage channel template}\quad
\emph{neighborhood mixers and sequence routing listed for coverage}}}\\[6pt]

\hline
\rowcolor{attrowC}
\multicolumn{7}{@{}p{16.51cm}@{}}{\textbf{GraphHyperConv}~\cite{Lell2024HyperAggregation}. \emph{Neighborhood-mixer boundary.}
For \(Y=\FF(\mathbf H^{(\ell)})\), HyperAggregation generates
\(W_{\rm tar}=(\sigma(Y_{\widetilde{\mathcal N}(i)}W_A))W_B\) and applies
\(\mathrm{HA}(Y_{\widetilde{\mathcal N}(i)})
 =((\sigma(Y_{\widetilde{\mathcal N}(i)}^\top W_{\rm tar}))W_{\rm tar}^\top)^\top\).
GHC then uses root or mean selection followed by \(\FF\), optionally concatenating the root.
This whole-neighborhood generated mixer is not one endpoint-local pairwise-message channel.}\\[6pt]

\hline
\rowcolor{attrowC}
\multicolumn{7}{@{}p{16.51cm}@{}}{\textbf{NLGNN}~\cite{liu2021non}. \emph{Sequence routing.} Local embedding \(z_v\) is hard-sorted by attention score \(a_v=\mathrm{Attend}(c,z_v)\). The convolution acts on the sorted, attention-weighted sequence \((a_1z_1,\ldots,a_nz_n)\), and prediction uses \([z_v\Vert\hat z_v]\). This hard reindexing followed by sequence convolution is a multi-stage operator, not a direct linear-value or invariantly aggregated pairwise-message channel. Score sorting alone does not imply failure of relabeling equivariance. Tied scores require an equivariant tie rule.}\\[6pt]

\hline
\rowcolor{attrowC}
\multicolumn{7}{@{}p{16.51cm}@{}}{\textbf{GPNN}~\cite{yang2022graph}. \emph{Sequence routing.} A pointer network selects an ordered subsequence from a sampled multi-hop neighborhood. Convolution and pooling produce a non-local vector \(z_v\), and prediction uses \([x_vW\Vert\widehat x_v\Vert z_v]\). Its BFS enumeration, truncation, and LSTM pointer are generally sequence-order-sensitive. The pointer, convolution, and pooling pipeline is not one declared channel contribution.}\\[6pt]

\end{longtable}
\endgroup

\colorlet{specbarA}{repSpectral}
\colorlet{specbarB}{repSpectral!80}
\colorlet{specbarC}{repSpectral!62}
\colorlet{specrowA}{repSpectral!9}
\colorlet{specrowB}{repSpectral!9}
\colorlet{specrowC}{repSpectral!9}

\providecommand{\MLP}{\operatorname{MLP}}
\providecommand{\GRU}{\operatorname{GRU}}
\providecommand{\Diag}{\operatorname{diag}}
\providecommand{\Conv}{\operatorname{Conv1D}}
\providecommand{\sm}{\operatorname{softmax}}

\makeatletter
\@ifundefined{NC@find@L}{\newcolumntype{L}[1]{>{\raggedright\arraybackslash}p{#1}}}{}
\@ifundefined{NC@find@C}{\newcolumntype{C}[1]{>{\centering\arraybackslash}p{#1}}}{}
\makeatother

\begingroup
\scriptsize
\renewcommand{\arraystretch}{1.5}
\arrayrulecolor{black!45}
\setlength{\arrayrulewidth}{0.5pt}
\setlength{\tabcolsep}{3.67pt}
\setlength{\LTpre}{6pt}
\setlength{\LTpost}{6pt}
\setlength{\LTcapwidth}{\textwidth}

\begin{longtable}{
@{}
L{2.22cm}
C{1.04cm}
C{1.48cm}
L{2.67cm}
L{2.37cm}
L{2.67cm}
L{2.37cm}
@{}}

\caption{\scriptsize\textbf{Family 3. Spectral, Diffusion, and Compatibility-Aware Graph Operators.}
Unified-component decompositions of spectral, diffusion, and compatibility-aware operators.
Spectral and channel coefficients are absorbed into $\mathbf P_k$, while restart and residual
terms remain in the update. Pairwise aggregation precedes cross-channel mixing. Rows marked
$\dagger$ report the source-supported slice of signal-dependent, recurrent, selected, or
composite operators without imposing a false fixed factorization.}
\label{tab:family3_spectral_slots}\\[6pt]

\toprule
\rowcolor{specbarA}
\textcolor{white}{\textbf{Method}} &
\multicolumn{1}{C{1.04cm}}{\textcolor{white}{$\boldsymbol{\mathcal X}$}} &
\multicolumn{1}{C{1.48cm}}{\textcolor{white}{$\boldsymbol{\mathcal K}$}} &
\multicolumn{1}{C{2.67cm}}{\textcolor{white}{\makecell[c]{\textbf{Propagation}\\\(\boldsymbol{\mathbf P_k^{(\ell)}}\)}}} &
\multicolumn{1}{C{2.37cm}}{\textcolor{white}{\makecell[c]{\textbf{Message}\\\(\boldsymbol{\Psi_k^{(\ell)}}\)}}} &
\multicolumn{1}{C{2.67cm}}{\textcolor{white}{\makecell[c]{\textbf{Mixing}\\\(\boldsymbol{\boxplus^{(\ell)}}\)}}} &
\multicolumn{1}{C{2.37cm}}{\textcolor{white}{\makecell[c]{\textbf{Ego/Update}\\\(\boldsymbol{(\mathcal B_\ell,\phi_\ell)}\)}}}
\\[6pt]
\midrule
\endfirsthead

\multicolumn{7}{c}{\tablename\ \thetable\ -- continued} \\[6pt]
\toprule
\rowcolor{specbarA}
\textcolor{white}{\textbf{Method}} &
\multicolumn{1}{C{1.04cm}}{\textcolor{white}{$\boldsymbol{\mathcal X}$}} &
\multicolumn{1}{C{1.48cm}}{\textcolor{white}{$\boldsymbol{\mathcal K}$}} &
\multicolumn{1}{C{2.67cm}}{\textcolor{white}{\makecell[c]{\textbf{Propagation}\\\(\boldsymbol{\mathbf P_k^{(\ell)}}\)}}} &
\multicolumn{1}{C{2.37cm}}{\textcolor{white}{\makecell[c]{\textbf{Message}\\\(\boldsymbol{\Psi_k^{(\ell)}}\)}}} &
\multicolumn{1}{C{2.67cm}}{\textcolor{white}{\makecell[c]{\textbf{Mixing}\\\(\boldsymbol{\boxplus^{(\ell)}}\)}}} &
\multicolumn{1}{C{2.37cm}}{\textcolor{white}{\makecell[c]{\textbf{Ego/Update}\\\(\boldsymbol{(\mathcal B_\ell,\phi_\ell)}\)}}}
\\[6pt]
\midrule
\endhead

\midrule
\multicolumn{7}{r}{\textit{continued on next page}} \\[6pt]
\endfoot

\bottomrule
\endlastfoot

\rowcolor{specbarA}
\multicolumn{7}{@{}p{16.38cm}@{}}{%
\textcolor{white}{\textbf{3a\ \ Explicit spectral and polynomial filters}\quad
\emph{fixed/learned filter banks, linear in the signal}}}
\\[6pt]

\hline
\rowcolor{specrowA}
SpectralCNN~\cite{bruna2014spectral} &
\(V\) &
\(\{(p,q)\}\) &
\(\mathbf U\Diag(\boldsymbol\theta_{pq})\mathbf U^\top\) {\tiny(input/output channel pair)} &
\(\mathbf H^{(\ell)}_{:,p}\mathbf e_q^\top\) &
\(\textstyle\sum\nolimits_{p,q}\mathbf P_{pq}\Psi_{pq}\) &
\(\sigma(\mixmsg^{(\ell)})\)
\\[6pt]

\hline
\rowcolor{specrowA}
ChebNet~\cite{defferrard2016convolutional} &
\(V\) &
\(\{0,\dots,K{-}1\}\) &
\(T_k(\tilde{\mathbf L}_{\mathrm{ch}})\) {\tiny($[-1,1]$-rescaled \(\mathbf L_{\mathrm{sym}}\), distinct from \(\tilde{\mathbf L}\))} &
\(\mathbf H^{(\ell)}\boldsymbol\Theta_k\) &
\(\textstyle\sum\nolimits_k \mathbf P_k \Psi_k\) &
\(\sigma(\mixmsg^{(\ell)})\)
\\[6pt]

\hline
\rowcolor{specrowA}
CayleyNet~\cite{levie2019cayleynets} &
\(V\) &
\(\begin{gathered}\{0\}\cup\\
\{(j,p,q):\\1{\le}j{\le}r\}\end{gathered}\) &
\(\begin{gathered}
\mathbf P_0{=}c_0\mathbf I,\\
\mathbf P_{j,p,q}{=}c_{j,pq}\mathbf C_h^j,\\
\mathbf C_h\\
{}=(h\boldsymbol\Delta{-}i\mathbf I)\\
\mathbin{\cdot}(h\boldsymbol\Delta{+}i\mathbf I)^{-1}
\end{gathered}\) &
\(\begin{gathered}
\Psi_0{=}\mathbf H^{(\ell)}\boldsymbol\Theta_0,\\
\Psi_{j,p,q}\\
{}=\mathbf H^{(\ell)}_{:,p}\mathbf e_q^\top
\end{gathered}\)
{\tiny(complex coefficients folded into \(\mathbf P\))} &
\(\begin{gathered}
\mathbf P_0\Psi_0\\
{}+2\mathrm{Re}\!\textstyle\sum\nolimits_{j,p,q}\\
\mathbf P_{j,p,q}\Psi_{j,p,q}
\end{gathered}\) &
\(\sigma(\mixmsg^{(\ell)})\)
\\[6pt]

\hline
\rowcolor{specrowA}
GWNN~\cite{xu2019graph} &
\(V\) &
\(\{(p,q)\}\) &
\(\boldsymbol\psi_s\,\Diag(\mathbf f_{pq})\,\boldsymbol\psi_s^{-1}\) {\tiny(wavelet; input/output channel pair)} &
\(\mathbf H^{(\ell)}_{:,p}\mathbf e_q^\top\) &
\(\textstyle\sum\nolimits_{p,q}\mathbf P_{pq} \Psi_{pq}\) &
\(\sigma(\mixmsg^{(\ell)})\)
\\
\end{longtable}
\endgroup

\colorlet{tfbarA}{repTransformer}
\colorlet{tfbarB}{repTransformer!80}
\colorlet{tfbarC}{repTransformer!62}
\colorlet{tfrowA}{repTransformer!9}
\colorlet{tfrowB}{repTransformer!9}
\colorlet{tfrowC}{repTransformer!9}

\providecommand{\MLP}{\operatorname{MLP}}
\providecommand{\MHA}{\operatorname{MHA}}
\providecommand{\MSA}{\operatorname{MSA}}
\providecommand{\GCN}{\operatorname{GCN}}
\providecommand{\GN}{\operatorname{GN}}
\providecommand{\FFN}{\operatorname{FFN}}
\providecommand{\SGC}{\operatorname{SGC}}
\providecommand{\LN}{\operatorname{LN}}
\providecommand{\sm}{\operatorname{softmax}}
\providecommand{\rownorm}{\operatorname{rownorm}}
\providecommand{\LReLU}{\operatorname{LReLU}}

\providecommand{\prop}[1]{\scriptstyle #1}
\providecommand{\propML}[1]{\begin{array}[t]{@{}l@{}}#1\end{array}}
\newcommand{\tfMethodCell}[1]{\multicolumn{1}{>{\cellcolor{tfrowA}\raggedright\arraybackslash}m{1.70cm}}{#1}}
\newcommand{\tfXCell}[1]{\multicolumn{1}{>{\centering\arraybackslash$}m{1.35cm}<{$}}{#1}}
\newcommand{\tfKCell}[1]{\multicolumn{1}{>{\centering\arraybackslash$}m{1.65cm}<{$}}{#1}}
\newcommand{\tfPCell}[1]{\multicolumn{1}{>{\centering\arraybackslash$}m{3.10cm}<{$}}{#1}}
\newcommand{\tfPsiCell}[1]{\multicolumn{1}{>{\centering\arraybackslash$}m{2.15cm}<{$}}{#1}}
\newcommand{\tfMixCell}[1]{\multicolumn{1}{>{\centering\arraybackslash$}m{2.45cm}<{$}}{#1}}
\newcommand{\tfUpdateCell}[1]{\multicolumn{1}{>{\centering\arraybackslash$}m{3.00cm}<{$}}{#1}}

\makeatletter
\@ifundefined{NC@find@Y}{\newcolumntype{Y}[1]{>{\raggedright\arraybackslash}p{#1}}}{}
\makeatother

\begingroup
\scriptsize
\renewcommand{\arraystretch}{1.78}
\setlength{\extrarowheight}{1.2pt}
\arrayrulecolor{black!45}
\setlength{\arrayrulewidth}{0.5pt}
\setlength{\tabcolsep}{2pt}
\setlength{\LTpre}{6pt}
\setlength{\LTpost}{6pt}
\setlength{\LTcapwidth}{\textwidth}

\begin{longtable}{
@{}
>{\cellcolor{tfrowA}\raggedright\arraybackslash}p{1.70cm}
@{\color{tfrowA}\vrule width4pt}
>{\centering\arraybackslash$}p{1.35cm}<{$}
@{\color{tfrowA}\vrule width4pt}
>{\centering\arraybackslash$}p{1.65cm}<{$}
@{\color{tfrowA}\vrule width4pt}
>{\raggedright\arraybackslash$}p{3.10cm}<{$}
@{\color{tfrowA}\vrule width4pt}
>{\raggedright\arraybackslash$}p{2.15cm}<{$}
@{\color{tfrowA}\vrule width4pt}
>{\raggedright\arraybackslash$}p{2.45cm}<{$}
@{\color{tfrowA}\vrule width4pt}
>{\raggedright\arraybackslash$}p{3.00cm}<{$}
@{}}
\caption{\scriptsize\textbf{Family 4. Graph Transformers.}
Unified-component decompositions of graph Transformer layers. Attention defines $\mathbf P_k$;
operator-level structural biases enter its score, input encodings enter $\mathcal X$, and
$\boxplus$ combines heads or branches. Pair-dependent values retain their explicit neighbor
contraction, and Block~4d records learned mixing over typed token axes. Flags denote gates
$^{\otimes}$, softmax-free operators $^{\circ}$, dynamic boundaries $^{\natural}$,
hierarchical folds $^{\triangledown}$, sequential stages $^{\dagger}$, staged composites
$^{\star}$, and local--global token concatenation $^{\ddagger}$.}
\label{tab:family-transformer}\\[6pt]

\toprule
\rowcolor{tfbarA}
\multicolumn{1}{>{\raggedright\arraybackslash}p{1.70cm}}{\textcolor{white}{\textbf{Method}}} &
\multicolumn{1}{C{1.35cm}}{\textcolor{white}{$\boldsymbol{\mathcal X}$}} &
\multicolumn{1}{C{1.65cm}}{\textcolor{white}{$\boldsymbol{\mathcal K}$}} &
\multicolumn{1}{C{3.10cm}}{\textcolor{white}{\makecell[c]{\textbf{Propagation}\\\(\boldsymbol{\mathbf P_k^{(\ell)}}\)}}} &
\multicolumn{1}{C{2.15cm}}{\textcolor{white}{\makecell[c]{\textbf{Message}\\\(\boldsymbol{\Psi_k^{(\ell)}}\)}}} &
\multicolumn{1}{C{2.45cm}}{\textcolor{white}{\makecell[c]{\textbf{Mixing}\\\(\boldsymbol{\boxplus^{(\ell)}}\)}}} &
\multicolumn{1}{C{3.00cm}}{\textcolor{white}{\makecell[c]{\textbf{Ego/Update}\\\(\boldsymbol{(\mathcal B_\ell,\phi_\ell)}\)}}}\\[6pt]
\midrule
\endfirsthead

\multicolumn{7}{c}{\tablename\ \thetable\ -- continued}\\[6pt]
\toprule
\rowcolor{tfbarA}
\multicolumn{1}{>{\raggedright\arraybackslash}p{1.70cm}}{\textcolor{white}{\textbf{Method}}} &
\multicolumn{1}{C{1.35cm}}{\textcolor{white}{$\boldsymbol{\mathcal X}$}} &
\multicolumn{1}{C{1.65cm}}{\textcolor{white}{$\boldsymbol{\mathcal K}$}} &
\multicolumn{1}{C{3.10cm}}{\textcolor{white}{\makecell[c]{\textbf{Propagation}\\\(\boldsymbol{\mathbf P_k^{(\ell)}}\)}}} &
\multicolumn{1}{C{2.15cm}}{\textcolor{white}{\makecell[c]{\textbf{Message}\\\(\boldsymbol{\Psi_k^{(\ell)}}\)}}} &
\multicolumn{1}{C{2.45cm}}{\textcolor{white}{\makecell[c]{\textbf{Mixing}\\\(\boldsymbol{\boxplus^{(\ell)}}\)}}} &
\multicolumn{1}{C{3.00cm}}{\textcolor{white}{\makecell[c]{\textbf{Ego/Update}\\\(\boldsymbol{(\mathcal B_\ell,\phi_\ell)}\)}}}\\[6pt]
\endhead

\multicolumn{7}{r}{\textit{continued on next page}}\\[6pt]
\endfoot

\bottomrule
\endlastfoot

\rowcolor{tfbarA}
\multicolumn{7}{@{}l@{}}{\parbox{16.38cm}{\color{white}\textbf{4a\ \ Dense global attention with structural bias}\quad
\emph{full or subgraph all-pairs softmax; bias inside the operator}\\[-1pt]
\emph{Graph Trans.~\cite{dwivedi2021generalization} is the one exception, restricting
$\mathbf P_k$ to $\mathcal N(i)$ (genuinely local, not all-pairs),}\\[-1pt]
\emph{grouped here for its Transformer-block lineage rather than its support}}}\\[6pt]

\hline
\rowcolor{tfrowA}
\multicolumn{1}{>{\cellcolor{tfrowA}\raggedright\arraybackslash}m{1.70cm}}{\makecell[l]{Graph-BERT\\[-1pt]
{\scriptsize\citeauthor{zhang2020graphbert}}\\[-1pt]
{\scriptsize(\citeyear{zhang2020graphbert})}}} &
\multicolumn{1}{>{\centering\arraybackslash$}m{1.35cm}<{$}}{\begin{gathered}
g_i=(V_i,\emptyset)\\[1pt]
V_i=\{v_i\}\\[-1pt]
{}\cup\Gamma(v_i)\\[-1pt]
{}+\mathrm{WL}\\[-1pt]
{}+\mathrm{PPR}\mathrm{rank}\\[-1pt]
{}+\mathrm{hop}
\end{gathered}} &
\multicolumn{1}{>{\centering\arraybackslash$}m{1.65cm}<{$}}{\{1\}} &
\multicolumn{1}{>{\centering\arraybackslash$}m{3.10cm}<{$}}{\begin{gathered}
\sm(\mathbf Q\mathbf K^\top/\sqrt d)\\[-1pt]
\text{on linkless context }V_i
\end{gathered}} &
\multicolumn{1}{>{\centering\arraybackslash$}m{2.15cm}<{$}}{\mathbf H_i^{(\ell)}\mathbf W^V} &
\multicolumn{1}{>{\centering\arraybackslash$}m{2.45cm}<{$}}{\mathbf P\Psi} &
\multicolumn{1}{>{\centering\arraybackslash$}m{3.00cm}<{$}}{\begin{gathered}
\mixmsg_i+\mathrm{G\text{-}Res}(\mathbf H_i,\mathbf X_i);\\[2pt]
{\scriptstyle z_i=\mathrm{Fusion}(H_i^D)\text{ external}}
\end{gathered}}\\
\end{longtable}

\endgroup

\colorlet{hetbarA}{repHetero}
\colorlet{hetbarB}{repHetero!80}
\colorlet{hetbarC}{repHetero!62}
\colorlet{hetrowA}{repHetero!9}
\colorlet{hetrowB}{repHetero!9}
\colorlet{hetrowC}{repHetero!9}

\makeatletter
\@ifundefined{color@darkred}{\definecolor{darkred}{HTML}{8B0000}}{}
\makeatother

\providecommand{\MLP}{\operatorname{MLP}}
\providecommand{\GNN}{\operatorname{GNN}}
\providecommand{\POOL}{\operatorname{POOL}}
\providecommand{\AGG}{\operatorname{AGG}}

\makeatletter
\@ifundefined{NC@find@L}{\newcolumntype{L}[1]{>{\raggedright\arraybackslash}p{#1}}}{}
\@ifundefined{NC@find@C}{\newcolumntype{C}[1]{>{\centering\arraybackslash}p{#1}}}{}
\makeatother

\begingroup
\scriptsize
\renewcommand{\arraystretch}{1.72}
\setlength{\extrarowheight}{1pt}
\arrayrulecolor{black!45}
\setlength{\arrayrulewidth}{0.5pt}
\setlength{\tabcolsep}{2pt}
\setlength{\LTpre}{6pt}
\setlength{\LTpost}{6pt}
\setlength{\LTcapwidth}{\textwidth}

\begin{longtable}{
@{}
L{2.00cm}
C{1.15cm}
C{1.65cm}
C{2.85cm}
C{2.50cm}
C{2.75cm}
C{2.50cm}
@{}}

\caption{\scriptsize\textbf{Family 5. Heterogeneous and Multi-Relational GNNs.}
Source-grounded decompositions of heterogeneous and multi-relational layers. Relation types,
meta-paths, and typed heads index channels; scalar supports and attention scores belong to
$\mathbf P_k$, while typed projections and composed values belong to $\Psi_k$. Mixing combines
completed relation or head contributions, and nonlinear selection is shown directly rather
than forced into a fixed matrix product. Boundary and architecture-search rows report only
the source-supported layer components.}
\label{tab:family-heterogeneous}\\[6pt]

\toprule
\rowcolor{hetbarA}
\textcolor{white}{\textbf{Method}} &
\multicolumn{1}{C{1.15cm}}{\textcolor{white}{$\boldsymbol{\mathcal X}$}} &
\multicolumn{1}{C{1.65cm}}{\textcolor{white}{$\boldsymbol{\mathcal K}$}} &
\multicolumn{1}{C{2.85cm}}{\textcolor{white}{\makecell[c]{\textbf{Propagation}\\\(\boldsymbol{\mathbf P_k^{(\ell)}}\)}}} &
\multicolumn{1}{C{2.50cm}}{\textcolor{white}{\makecell[c]{\textbf{Message}\\\(\boldsymbol{\Psi_k^{(\ell)}}\)}}} &
\multicolumn{1}{C{2.75cm}}{\textcolor{white}{\makecell[c]{\textbf{Mixing}\\\(\boldsymbol{\boxplus^{(\ell)}}\)}}} &
\multicolumn{1}{C{2.50cm}}{\textcolor{white}{\makecell[c]{\textbf{Ego/Update}\\\(\boldsymbol{(\mathcal B_\ell,\phi_\ell)}\)}}}
\\[6pt]
\midrule
\endfirsthead

\multicolumn{7}{c}{\tablename\ \thetable\ -- continued} \\[6pt]
\toprule
\rowcolor{hetbarA}
\textcolor{white}{\textbf{Method}} &
\multicolumn{1}{C{1.15cm}}{\textcolor{white}{$\boldsymbol{\mathcal X}$}} &
\multicolumn{1}{C{1.65cm}}{\textcolor{white}{$\boldsymbol{\mathcal K}$}} &
\multicolumn{1}{C{2.85cm}}{\textcolor{white}{\makecell[c]{\textbf{Propagation}\\\(\boldsymbol{\mathbf P_k^{(\ell)}}\)}}} &
\multicolumn{1}{C{2.50cm}}{\textcolor{white}{\makecell[c]{\textbf{Message}\\\(\boldsymbol{\Psi_k^{(\ell)}}\)}}} &
\multicolumn{1}{C{2.75cm}}{\textcolor{white}{\makecell[c]{\textbf{Mixing}\\\(\boldsymbol{\boxplus^{(\ell)}}\)}}} &
\multicolumn{1}{C{2.50cm}}{\textcolor{white}{\makecell[c]{\textbf{Ego/Update}\\\(\boldsymbol{(\mathcal B_\ell,\phi_\ell)}\)}}}
\\[6pt]
\midrule
\endhead

\midrule
\multicolumn{7}{r}{\textit{continued on next page}} \\[6pt]
\endfoot

\bottomrule
\endlastfoot

\rowcolor{hetbarA}
\multicolumn{7}{@{}l}{%
\textcolor{white}{\textbf{5a\ \ Per-relation operators and weights}\quad
\emph{explicit per-relation supports, weights, and relation/KG scoring}}}
\\[6pt]

\hline
\rowcolor{hetrowA}
R-GCN~\cite{schlichtkrull2018modeling} &
\(V\) &
\(\mathcal R\) &
\(\begin{gathered}\tfrac{1}{c_{i,r}}\mathbf A_r\\{\scriptstyle\text{(per-rel.\ norm.)}}\end{gathered}\) &
\(\mathbf H^{(\ell)}\mathbf W_r\) &
\(\textstyle\sum\nolimits_{k\in\mathcal K} \mathbf P_k \Psi_k\) &
\(\begin{gathered}\sigma\!\big(\mathbf H^{(\ell)}\mathbf W_0+\\\mixmsg^{(\ell)}\big)\end{gathered}\)
\\[6pt]

\hline
\rowcolor{hetrowA}
RGAT~\cite{busbridge2019relational} &
\(V\) &
\(\begin{gathered}\{(r,m):\\r\in\mathcal R,\\1{\le}m{\le}M\}\end{gathered}\) &
\(\begin{gathered}\alpha_{ij}^{(r,m)}(\mathbf H)\,\mathbf A_r\\{\scriptstyle\text{WIR: softmax over }j\text{ per }r;}\\{\scriptstyle\text{AR: softmax over }(j,r)}\end{gathered}\) &
\(\mathbf H^{(\ell)}\mathbf W_{r,m}\) &
\(\begin{aligned}\mathbf C_m&=\textstyle\sum_{r\in\mathcal R}\\&\quad\mathbf P_{r,m}\Psi_{r,m},\\\mixmsg&=\big\Vert_{m=1}^{M}\mathbf C_m\end{aligned}\) &
\(\begin{gathered}\mathbf H'=\big\Vert_{m=1}^{M}\sigma(\mathbf C_m)\\{\scriptstyle(M=1:\ \text{base rule})}\end{gathered}\)
\\[6pt]

\hline
\rowcolor{hetrowA}
CompGCN~\cite{vashishth2020compgcn} &
\(V\sqcup\mathcal R\) &
\(\begin{gathered}\mathcal R\cup\mathcal R^{-1}\\{}\cup\{\top\}\end{gathered}\) &
\(\begin{gathered}
\mathbf A_r\\
{\scriptstyle\text{direct/inverse/}}\\
{\scriptstyle\text{self support;}}\\
{\scriptstyle\text{unnormalized}}
\end{gathered}\) &
\(\begin{gathered}\xi(\vh_j,\vr_r)\\{}\mathbf W_{\lambda(r)}\\{\scriptstyle\text{(entity--rel.\ comp.;}}\\{\scriptstyle\lambda(r)\in\{O,I,S\}\text{)}}\end{gathered}\) &
\(\textstyle\sum\nolimits_{k\in\mathcal K} \mathbf P_k \Psi_k\) &
\(\begin{gathered}\vh_v^{(k+1)}=f(\mixmsg_v^{(k)});\\\vr_r^{(k+1)}=\vr_r^{(k)}\mathbf W_{\mathrm{rel}}^{(k)}\end{gathered}\)
\\
\end{longtable}
\endgroup

\colorlet{barA}{repHigher}
\colorlet{barB}{repHigher!80}
\colorlet{barC}{repHigher!62}
\colorlet{rowA}{repHigher!16}
\colorlet{rowB}{repHigher!16}
\colorlet{rowC}{repHigher!16}

\providecommand{\MLP}{\operatorname{MLP}}
\providecommand{\GNN}{\operatorname{GNN}}
\providecommand{\POOL}{\operatorname{POOL}}
\providecommand{\AGG}{\operatorname{AGG}}
\providecommand{\AGGR}{\operatorname{AGGR}}
\providecommand{\MSG}{\operatorname{MSG}}

\makeatletter
\@ifundefined{NC@find@L}{\newcolumntype{L}[1]{>{\raggedright\arraybackslash}p{#1}}}{}
\@ifundefined{NC@find@C}{\newcolumntype{C}[1]{>{\centering\arraybackslash}p{#1}}}{}
\@ifundefined{NC@find@T}{\newcolumntype{T}[1]{>{\centering\arraybackslash}p{#1}}}{}
\makeatother

\begingroup
\scriptsize
\let\tiny\scriptsize
\renewcommand{\arraystretch}{1.72}
\setlength{\extrarowheight}{1pt}
\arrayrulecolor{black!45}
\setlength{\arrayrulewidth}{0.5pt}
\setlength{\tabcolsep}{2pt}
\setlength{\LTpre}{6pt}
\setlength{\LTpost}{6pt}
\setlength{\LTcapwidth}{\textwidth}

\begin{longtable}{
@{}
L{1.85cm}
C{1.00cm}
C{1.30cm}
L{2.90cm}
L{2.80cm}
L{2.85cm}
L{2.70cm}
@{}}

\caption{\scriptsize\textbf{Family 6. Higher-Order and Subgraph GNNs.}
Source-grounded decompositions in the graded extension, with $\mathcal X$ lifted to tuples,
subsets, rooted subgraphs, bags, or cells. Within-object aggregation forms a channel
contribution before cross-channel mixing. Rows marked $^{\S}$ use transient lifted routing
while persistent state remains on nodes; $^{\ddagger}$ marks schematic multi-stage subgraph
constructions whose pooling, sampling, and readout remain external.}
\label{tab:family-higher-order}\\[6pt]

\toprule
\rowcolor{barA}
\textcolor{white}{\textbf{Method}} &
\multicolumn{1}{C{1.00cm}}{\textcolor{white}{$\boldsymbol{\mathcal X}$}} &
\multicolumn{1}{C{1.30cm}}{\textcolor{white}{$\boldsymbol{\mathcal K}$}} &
\multicolumn{1}{T{2.90cm}}{\textcolor{white}{\makecell[c]{\textbf{Propagation}\\$\boldsymbol{\mathbf P_k^{(\ell)}}$}}} &
\multicolumn{1}{T{2.80cm}}{\textcolor{white}{\makecell[c]{\textbf{Message}\\$\boldsymbol{\Psi_k^{(\ell)}}$}}} &
\multicolumn{1}{T{2.85cm}}{\textcolor{white}{\makecell[c]{\textbf{Mixing}\\$\boldsymbol{\boxplus^{(\ell)}}$}}} &
\multicolumn{1}{T{2.70cm}}{\textcolor{white}{\makecell[c]{\textbf{Ego/Update}\\$\boldsymbol{(\mathcal B_\ell,\phi_\ell)}$}}}
\\[6pt]
\midrule
\endfirsthead

\multicolumn{7}{c}{\tablename\ \thetable\ -- continued} \\[6pt]
\toprule
\rowcolor{barA}
\textcolor{white}{\textbf{Method}} &
\multicolumn{1}{C{1.00cm}}{\textcolor{white}{$\boldsymbol{\mathcal X}$}} &
\multicolumn{1}{C{1.30cm}}{\textcolor{white}{$\boldsymbol{\mathcal K}$}} &
\multicolumn{1}{T{2.90cm}}{\textcolor{white}{\makecell[c]{\textbf{Propagation}\\$\boldsymbol{\mathbf P_k^{(\ell)}}$}}} &
\multicolumn{1}{T{2.80cm}}{\textcolor{white}{\makecell[c]{\textbf{Message}\\$\boldsymbol{\Psi_k^{(\ell)}}$}}} &
\multicolumn{1}{T{2.85cm}}{\textcolor{white}{\makecell[c]{\textbf{Mixing}\\$\boldsymbol{\boxplus^{(\ell)}}$}}} &
\multicolumn{1}{T{2.70cm}}{\textcolor{white}{\makecell[c]{\textbf{Ego/Update}\\$\boldsymbol{(\mathcal B_\ell,\phi_\ell)}$}}}
\\[6pt]
\midrule
\endhead

\midrule
\multicolumn{7}{r}{\textit{continued on next page}} \\[6pt]
\endfoot

\bottomrule
\endlastfoot

\rowcolor{barA}
\multicolumn{7}{@{}p{\textwidth}@{}}{%
\textcolor{white}{\textbf{6a\ \ Tuple-based ($k$-order / $k$-WL) models}\quad
\emph{states over $k$-tuples; equivariant linear maps and FWL}}}
\\[6pt]

\hline
\rowcolor{rowA}
IEGN~\cite{maron2019invariant} &
\(V^k\!\to V^l\) &
\([n]^{k+l}/\!\sim\) &
\(\mathbf B_\pi\) {\tiny(equality-pattern equivariant basis; $|\Pi_{k,l}|=b(k{+}l)$)} &
\(\mathbf H^{(\ell)}\mathbf W_\pi^{(\ell)}\) &
\(\begin{gathered}\textstyle\sum_\pi\mathbf B_\pi\Psi_\pi+\\\textstyle\sum_\lambda\mathbf C_\lambda\mathbf b_\lambda\end{gathered}\) {\tiny(including equivariant bias basis)} &
\(\sigma(\mixmsg^{(\ell)})\)
\\[6pt]

\hline
\rowcolor{rowA}
$k$-GNN~\cite{morris2019weisfeiler} &
\([V]^k\) &
\(\{1\}\) {\tiny(or optional loc/glob split)} &
\(\mathbf P\) {\tiny(support $N_L(s)\cup N_G(s)$; local variant uses $N_L$ only)} &
\(\mathbf H^{(\ell)}\mathbf W_2^{(\ell)}\) &
\(\mathbf P\Psi\) {\tiny(optional loc/glob split sums channel contributions)} &
\(\sigma\!\big(\mathbf H_r^{(\ell)}\mathbf W_1+\mixmsg^{(\ell)}\big)\)
\\[6pt]

\hline
\rowcolor{rowA}
PPGN~\cite{maron2019provably} &
\(V^2\) &
\(\{c=1{:}b\}\) {\tiny(product channels; shown as $\times$)} &
\(\MLP_1(\mathbf H^{(\ell)})\) {\tiny(state-dep.\ operator; matrix prod.\ $=$ 2-FWL)} &
\(\MLP_2(\mathbf H^{(\ell)})\) &
\(\mathbf P_\times \Psi_\times\) &
\(\big[\MLP_3(\mathbf H^{(\ell)})\,\Vert\,\mixmsg^{(\ell)}\big]\) {\tiny($\MLP_3$ gives $B_\ell$; concat in $\phi_\ell$)}
\\[6pt]

\hline
\rowcolor{rowA}
Ring-GNN~\cite{chen2019ring}\ {\tiny(partial slice)} &
\(V^2\) &
\(\{1,\times\}\) &
\(\begin{gathered}\{L_\alpha(\cdot),\\L_\beta(\cdot)\}\end{gathered}\) {\tiny(equiv.\ linear maps; biases included)} &
\(\begin{gathered}\{1,\\L_\gamma(\cdot)\}\end{gathered}\) &
\(\begin{gathered}\mathbf C_1\Vert\mathbf C_\times\\{\scriptstyle\mathbf C_1=L_\alpha(A),}\\{\scriptstyle\mathbf C_\times=L_\beta(A)L_\gamma(A)}\end{gathered}\) &
\(\begin{gathered}k_1\sigma(\mathbf C_1)\\{}+k_2\sigma(\mathbf C_\times)\\{\scriptstyle\text{branchwise }\sigma\text{ and}}\\{\scriptstyle\text{post-}\sigma\text{ scalars}}\end{gathered}\)
\\[6pt]

\hline
\rowcolor{rowA}
$\delta$-$k$-LGNN~\cite{morris2020sparse} &
\(V^k\) &
\(\{j=1{:}k\}\) &
\(\mathbf P_j^\delta\) {\tiny(selects $\phi_j(\mathbf v,w)$ with $w\in N(v_j)$)} &
\(\Psi_j(\mathbf H)=\mathbf H\) {\tiny(value $\vh_{\phi_j(\mathbf v,w)}^{(\ell)}$)} &
\(\mathrm{id}\) {\tiny(replacement multisets feed joint tuple aggregation)} &
\(\begin{gathered}f_{\rm mrg}^{\mathbf W_1}\!\big(\\\vh_{\mathbf v}^{(\ell)},\\f_{\rm agg}^{\mathbf W_2}\!\big(\\\{\!\{\Psi_1\}\!\},\ldots,\\\{\!\{\Psi_k\}\!\}\big)\big)\end{gathered}\)
\\[6pt]

\hline
\rowcolor{rowA}
SpeqNets~\cite{morris2022speqnets} &
\(V_s^k\) &
\(\{1\}\) &
\(\mathbf P_j^{(k,s)}\) {\tiny(local replacements retained only in $V_s^k$)} &
\(\begin{gathered}(\Psi_1,\ldots,\Psi_k),\\ \Psi_j=\{\!\{\vh_{\phi_j(\mathbf v,w)}^{(\ell)}\}\!\}_{w\in N(v_j)}\end{gathered}\) {\tiny(with $\phi_j(\mathbf v,w)\in V_s^k$)} &
\(\mathrm{id}\) {\tiny(replacement multisets feed joint tuple aggregation)} &
\(\begin{gathered}f_{\rm mrg}^{\mathbf W_1}\!\big(\\\vh_{\mathbf v}^{(\ell)},\\f_{\rm agg}^{\mathbf W_2}\!(\\\Psi_1,\ldots,\Psi_k)\big)\end{gathered}\)
\\[6pt]

\rowcolor{barB}
\multicolumn{7}{@{}p{\textwidth}@{}}{%
\textcolor{white}{\textbf{6b\ \ Subgraph-based models}\quad
\emph{rooted subgraphs, bags of subgraphs, position/identity-aware supports}}}
\\[6pt]

\hline
\rowcolor{rowB}
NGNN~\cite{zhang2021nested}$^{\ddagger}$ &
\(\mathcal S\) &
\(\{1\}\) {\tiny($G_w^h$ indexes copied rooted subgraphs)} &
\(\mathbf A_{\mathcal S_i}\) {\tiny(rooted subgr.)} &
\(M_t(\vh_{v,G_w^h}^t,\vh_{u,G_w^h}^t,e_{vu})\) &
\(\begin{gathered}
\mathbf C_{1,v}^{t+1}=\\
\sum_{u\in N(v\mid G_w^h)}\\
\mathbf P_{vu}\Psi_{vu},\\
\mathbf m_{v,G_w^h}^{t+1}=\\
\mathbf C_{1,v}^{t+1}\\
{\scriptstyle(\boxplus:\mathrm{id})}
\end{gathered}\) &
\(\begin{gathered}\vh_{v,G_w^h}^{t+1}=U_t\!\big(\\\vh_{v,G_w^h}^t,\\\mathbf m_{v,G_w^h}^{t+1}\big),\\[-1pt]{\scriptstyle\text{after }T:\ R_0\text{ subgraph pooling,}}\\{\scriptstyle\text{then }R_1\text{ graph pooling}}\\{\scriptstyle\text{(only the inner update}}\\{\scriptstyle\text{is in the layer)}}\end{gathered}\)
\\[6pt]

\hline
\rowcolor{rowB}
GNN-AK~\cite{zhao2022from}$^{\ddagger}$ &
\(\mathcal S\) &
\(\mathcal K_{\rm base}\) &
\(P_k\) {\tiny(base-GNN support inside each $\operatorname{Sub}^{(\ell)}[v]$)} &
\(\Psi_k\) {\tiny(base-GNN message; $\operatorname{Emb}$ is the inner output)} &
\(\boxplus_{\rm base}\) {\tiny(inner slice only; $\POOL_{\GNN}$ and FUSE are outer operations)} &
\(\phi_{\rm base}\) {\tiny(after inner pass: FUSE($h_{\rm ctr},h_{\rm sub}$))}
\\[6pt]

\hline
\rowcolor{rowB}
ESAN~\cite{bevilacqua2022equivariant} &
\(\mathfrak S\) &
\(\begin{gathered}
\{\mathrm{self},\\
\mathrm{share}\}
\end{gathered}\) &
\(\{P_{\rm self},P_{\rm share}\}\) {\tiny(base-encoder supports; share broadcasts bag aggregate)} &
\(\begin{gathered}
L_1(\mathbf A_i,\\
\mathbf X_i),\\
L_2(\AGG_j\mathbf A_j,\\
\AGG_j\mathbf X_j)
\end{gathered}\) &
\(C_{\rm self}+C_{\rm share}\) {\tiny(DSS fusion; DS sets $L_2=0$)} &
\(\mixmsg\) {\tiny($R_{\rm subgraphs}$ and $E_{\rm sets}$ are external readout)}
\\[6pt]

\hline
\rowcolor{rowB}
Subgraph\-ormer~\cite{barshalom2024subgraphormer} &
\(V\!\times\!V\) &
\(\begin{gathered}
\{G,\\
GS,\\
\mathrm{pt}\}
\end{gathered}\) &
\(\begin{gathered}\mathbf P_G=\alpha_G(\mathbf Q_G,\mathbf K_G\mid\mathbf A_G),\\\mathbf P_{GS}=\alpha_{GS}(\mathbf Q_{GS},\mathbf K_{GS}\mid\mathbf A_{GS}),\\\mathbf P_{\rm pt}=\mathbf A_{\rm pt}\end{gathered}\) {\tiny($\mathbf A_G=\mathbf I\otimes\mathbf A$, $\mathbf A_{GS}=\mathbf A\otimes\mathbf I$)} &
\(\begin{gathered}\Psi_G=\mathbf V_G,\\\Psi_{GS}=\mathbf V_{GS},\\\Psi_{\rm pt}=\mathbf X\end{gathered}\) &
\(\begin{gathered}\mathbf C_G=\mathbf P_G\Psi_G,\\\mathbf C_{GS}=\mathbf P_{GS}\Psi_{GS},\\\mathbf C_{\rm pt}=\MLP_{\rm pt}\!\big(\\(1{+}\epsilon)\mathbf X\\{}+\mathbf A_{\rm pt}\mathbf X\big),\\\mixmsg=\mathbf C_G\\{}\Vert\mathbf C_{GS}\Vert\mathbf C_{\rm pt}\end{gathered}\) &
\(\MLP(\mixmsg)\) {\tiny(pooling and PE are external/input)}
\\[6pt]

\hline
\rowcolor{rowB}
HyMN (sampling/input wrapper)~\cite{Southern2025HyMN}\ {\tiny(boundary wrapper)} &
\(--\) &
\(--\) &
\(--\) &
\(--\) &
\(--\) &
\(\begin{gathered}\mathrm{CSE}_{v,k}=(\mathbf A^k)_{vv}/k!,\\M=\operatorname{select\text{-}top}\!\big(\\\sum_k\mathrm{CSE}_{:,k},T\big),\\y_G=f\!\big(\mathbf A,\\\mathbf X\Vert\mathrm{CSE},M\big)\\{\scriptstyle\text{(wrapper only; no}}\\{\scriptstyle\text{standalone layer)}}\end{gathered}\)
\\[6pt]

\hline
\rowcolor{rowB}
ID-GNN~\cite{you2021identity} &
\(\mathcal S\) &
\(\{\mathrm{id},\overline{\mathrm{id}}\}\) &
\(\mathbf A_{G_v^{(K)}}\) {\tiny(ego-net adjacency unchanged; root identity selects the message type)} &
\(\MSG_{\indicator[s=v]}^{(k)}(\vh_s^{(k-1)})\) &
\(\AGG_{s\in N(u)}\mathbf P_k\Psi_k\) {\tiny(source leaves AGG generic)} &
\(\AGG^{(k)}(\mixmsg_{u\mid v}^{(k)},\vh_{u\mid v}^{(k-1)})\)
\\[6pt]

\hline
\rowcolor{rowB}
I\textsuperscript{2}-GNN~\cite{huang2023i2gnn}$^{\ddagger}$ &
\(\begin{gathered}
\{(i,j,k):\\
i\in V,\\
j\in N(i),\\
k\in V_i\}
\end{gathered}\) &
\(\{1\}\) &
\(\mathbf A_{G_{i,j}}\) {\tiny($K$-hop ego-net; distinct root/branch identifiers)} &
\(\begin{gathered}
M_t\!\big(\vh_{i,j,k}^t,\\
\vh_{i,j,l}^t,\\
e_{kl}\big)
\end{gathered}\) &
\(\begin{gathered}
\mathbf C_{1,i,j,k}^{t}=\\
\sum_{l\in N_i(k)}\mathbf P_{kl}\Psi_{kl},\\
\mathbf m_{i,j,k}^{t}=\\
\mathbf C_{1,i,j,k}^{t}\\
{\scriptstyle(\boxplus:\mathrm{id})}
\end{gathered}\) &
\(\begin{gathered}\vh_{i,j,k}^{t+1}\\{}=U_t\!\big(\vh_{i,j,k}^t,\\\mathbf m_{i,j,k}^{t}\big),\\[-1pt]{\scriptstyle\text{after }T:\ R_{\rm edge}}\\{\scriptstyle{}\to R_{\rm node}\to R_{\rm graph}}\\{\scriptstyle\text{(the three readouts follow}}\\{\scriptstyle\text{the inner updates)}}\end{gathered}\)
\\[6pt]

\hline
\rowcolor{rowB}
OSAN~\cite{qian2022ordered} &
\(V\!\times\!\mathcal G^k\) &
\(\{1\}\) &
\(\begin{gathered}
P_\square\\
{\scriptstyle\text{block support for fixed }g;}\\
{\scriptstyle\square=N_G(v)\text{ or }V(G)}
\end{gathered}\) &
\(\vh_{u,g}^{(\ell)}\) &
\(\begin{gathered}
C_{v,g}^{(\ell)}=\\
\AGG\!\big(\{\!\{\vh_{u,g}^{(\ell)}:\\
u\in\square(v)\}\!\}\big),\\
\mixmsg_{v,g}^{(\ell)}=C_{v,g}^{(\ell)}\\
{\scriptstyle(\boxplus:\mathrm{id})}
\end{gathered}\) &
\(\begin{gathered}
\vh_{v,g}^{(\ell+1)}=\\
\operatorname{UPD}^{(\ell+1)}\!\big(\\
\vh_{v,g}^{(\ell)},\\
\mixmsg_{v,g}^{(\ell)}\big)\\
{\scriptstyle\text{SAGG and READOUT}}\\
{\scriptstyle\text{post-stack}}
\end{gathered}\)
\\[6pt]

\hline
\rowcolor{rowB}
DropGNN~\cite{Papp2021DropGNN}\ {\tiny(boundary wrapper)} &
\(V\!\times\![r]\) &
\(\mathcal K_{\rm base}\) &
\(\begin{gathered}
P_{k,s}\\
{\scriptstyle\text{base-GNN support}}\\
{\scriptstyle\text{on }G\setminus\mathcal D_s}
\end{gathered}\) &
\(\Psi_{k,\theta}\) {\tiny(base-GNN message map)} &
\(\boxplus_{\rm base}\) &
\(\phi_\theta\) {\tiny(base layer only; after $d$ layers, run aggregation then READOUT external)}
\\[6pt]

\hline
\rowcolor{rowB}
\makecell[l]{\(k\)-\\
Reconstruction\\
GNN\\[-1pt]
{\tiny\citeauthor{cotta2021reconstruction}}\\[-1pt]
{\tiny(\citeyear{cotta2021reconstruction})}} &
\(\mathfrak S\) {\tiny(boundary)} &
\(\{S\in[V]^k\}\) {\tiny(readout index, not layer channel)} &
\(\mathbf A_{G[S]}\) {\tiny(input to inner base GNN, not one global $\mathbf P_k$)} &
\(\begin{gathered}
h_{W_2}^{\GNN}(G[S])\\
{\scriptstyle\text{output of inner GNN}}\\
{\scriptstyle\text{on each induced subgraph}}
\end{gathered}\) &
\(\begin{gathered}
\rho_{W_1}\!\Bigg(\\
\sum_{S\in S(k)}\\
\phi_{W_2}\!\Big(\\
h_{W_3}^{\GNN}(G[S])\\
\Big)\Bigg)\\
{\scriptstyle\text{Deep-Sets graph readout;}}\\
{\scriptstyle\text{external}}
\end{gathered}\) &
\(\begin{gathered}\text{boundary:}\\\text{enumerate/sample}\\\text{subgraphs}\\{}\to\text{inner GNN}\\{}\to\text{multiset readout}\end{gathered}\)
\\[6pt]

\hline
\rowcolor{rowB}
LGAN\textsuperscript{\S}~\cite{Du2026LGAN} &
\(V\) &
\(\{t,n\}\) &
\(\begin{gathered}\mathcal E_t(t)=\{\{t,p\}:\\p\in N(t)\},\\\mathcal E_n(t)=\{\{p,q\}\in E:\\p,q\in N(t)\}\\{\scriptstyle\text{(pair-to-target supports;}}\\{\scriptstyle\text{transient line graph of }G_t\text{)}}\end{gathered}\) &
\(\begin{gathered}\vh_{\{u,v\}}^{(\ell-1)}=\\\operatorname{COMB}_{\rm pair}\!\big(\\\vh_u^{(\ell-1)},\\\vh_v^{(\ell-1)}\big)\end{gathered}\) &
\(\begin{gathered}\Vert\\{\scriptstyle\text{after }\AGGR_t(\{\!\{\vh_e:e\in\mathcal E_t(t)\}\!\})}\\{\scriptstyle\text{and }\AGGR_n(\{\!\{\vh_e:e\in\mathcal E_n(t)\}\!\})}\end{gathered}\) &
\(\begin{gathered}\phi(\mixmsg_t)\\{\scriptstyle\text{(LGAN-res uses}}\\{\scriptstyle\phi'(\mathbf W\vh_t^{(\ell-1)}+\mixmsg_t)\text{;}}\\{\scriptstyle\text{routing-boundary case)}}\end{gathered}\)
\\[6pt]

\hline
\rowcolor{rowB}
SA-GAT~\cite{Gao2021SAGAT} &
\(\begin{gathered}V(G^k)\\{\scriptstyle(k\text{-})}\\{\scriptstyle\text{substr.}}\\{\scriptstyle\text{graph}}\end{gathered}\) &
\(\begin{gathered}\{\mathrm{sa},\\\mathrm{ia}\}\end{gathered}\) &
\(\begin{gathered}\mathrm{sa}:a\,\alpha(C_k,T_k),\\ \mathrm{ia}:(1{-}a)\,\bar\beta(T_k,S_k)\end{gathered}\) {\tiny(scalar channel weights folded into $P$)} &
\(\begin{gathered}\Psi_{\rm sa}(T_k)=\mathbf u(T_k)\mathbf W_1,\\\Psi_{\rm ia}(T_k,S_k)=\mathbf u(T_k)\odot\mathbf u(S_k)\end{gathered}\) &
\(C_{\rm sa}+C_{\rm ia}\) &
\(\sigma(\mixmsg)\)
\\[6pt]

\rowcolor{barC}
\multicolumn{7}{@{}p{\textwidth}@{}}{%
\textcolor{white}{\textbf{6c\ \ Simplicial, cell, hypergraph, and pair domains}\quad
\emph{lifted support via (co)boundary maps, hyperedge incidence, or node-pair coupling}}}
\\[6pt]

\hline
\rowcolor{rowC}
MPSN~\cite{bodnar2021simplicial} &
\(\bigsqcup_p X_p\) &
\(\begin{gathered}\{\mathrm B,\mathrm C,\\\downarrow,\uparrow\}\end{gathered}\) &
\(\begin{gathered}\{B(\sigma),C(\sigma),\\N_\downarrow(\sigma),\\N_\uparrow(\sigma)\}\\{\scriptstyle\text{(boundary, co-boundary,}}\\{\scriptstyle\text{lower, upper supports)}}\end{gathered}\) &
\(\begin{gathered}M_B(\vh_\sigma,\vh_\tau),\\M_C(\vh_\sigma,\vh_\tau),\\M_\downarrow\!\big(\vh_\sigma,\vh_\tau,\\\vh_{\sigma\cap\tau}\big),\\M_\uparrow\!\big(\vh_\sigma,\vh_\tau,\\\vh_{\sigma\cup\tau}\big)\end{gathered}\) &
\(\begin{gathered}\big[m_B\Vert m_C\\{}\Vert m_\downarrow\\{}\Vert m_\uparrow\big]\end{gathered}\) &
\(\begin{gathered}U\!\big(\vh_\sigma^{(\ell)},\\\mixmsg_\sigma^{(\ell)}\big)\end{gathered}\)
\\[8pt]

\hline
\rowcolor{rowC}
CW Networks~\cite{Bodnar2021CWNetworks} &
\(\mathcal X_p\) &
\(\{\mathrm{B},\uparrow\}\) &
\(\{B(\sigma),N_\uparrow(\sigma),C(\sigma,\tau)\}\) {\tiny(boundary/upper supports)} &
\(\begin{gathered}
\Psi_{\rm B}(\sigma,\tau)=\\
\vh_\tau,\\
\Psi_\uparrow(\sigma,\tau,\delta)=\\
\MLP_M\!\big(\\
\vh_\tau\Vert\vh_\delta\big)\\
{\scriptstyle\delta\text{ is a shared coface}}
\end{gathered}\) &
\(\mixmsg=[\mathbf m_{\rm B}(\sigma)\Vert\mathbf m_\uparrow(\sigma)]\) {\tiny(concat., not sum)} &
\(\begin{gathered}\MLP_U\!\big[\\\MLP_{\rm B}\!\big((1{+}\epsilon_{\rm B})\vh_\sigma\\{}+\mathbf m_{\rm B}\big)\ \Vert\\\MLP_\uparrow\!\big((1{+}\epsilon_\uparrow)\vh_\sigma\\{}+\mathbf m_\uparrow\big)\big]\end{gathered}\)
\\[6pt]

\hline
\rowcolor{rowC}
FSpec\-GNN~\cite{Wang2026FSpecGNN} &
\(V^2\) &
\(\{g\}\) &
\(\begin{gathered}g(\mathbf L\otimes\mathbf I,\\\mathbf I\otimes\mathbf L)\end{gathered}\) {\tiny(learnable bivariate full-spectrum filter on $V\times V$)} &
\(\mathbf H_{V^2}^{(\ell)}\mathbf W^{(\ell)}\) {\tiny(pair features from $\phi(\mathbf X_u,\mathbf X_v,\mathbf E_{uv})$)} &
\(\mathbf P_g\Psi_g\quad{\scriptstyle(\boxplus:\mathrm{id})}\) &
\(\sigma(\mixmsg^{(\ell)})\) {\tiny(polynomial/low-rank implementations avoid explicit $n^2\!\times n^2$ operator)}
\\[6pt]

\hline
\rowcolor{rowC}
HGNN\textsuperscript{\S}~\cite{feng2019hypergraph} &
\(V\) &
\(\{1\}\) &
\(\begin{gathered}\mathbf D_v^{-1/2}\mathbf H_{\rm inc}\mathbf W_E\mathbf D_e^{-1}\\{}\mathbf H_{\rm inc}^{\top}\mathbf D_v^{-1/2}\end{gathered}\) {\tiny(node-hyperedge-node incidence operator)} &
\(\mathbf H^{(\ell)}\boldsymbol\Theta\) &
\(\mathbf P_1\Psi_1\quad{\scriptstyle(\boxplus:\mathrm{id})}\) &
\(\sigma(\mixmsg^{(\ell)})\)
\\[6pt]

\hline
\rowcolor{rowC}
HyperGCN\textsuperscript{\S}~\cite{yadati2019hypergcn} &
\(V\) &
\(\{1\}\) &
\(\overline{\mathbf A}_{\rm med}^{(\ell)}\) {\tiny(after hard mediator construction: each hyperedge selects $(i_e,j_e)=\arg\max_{i,j\in e}\|h_i^{(\ell)}\Theta^{(\ell)}-h_j^{(\ell)}\Theta^{(\ell)}\|_2$; self-loops included; boundary support step)} &
\(\mathbf H^{(\ell)}\boldsymbol\Theta^{(\ell)}\) &
\(\mathbf P_1\Psi_1\quad{\scriptstyle(\boxplus:\mathrm{id})}\) &
\(\sigma(\mixmsg^{(\ell)})\) {\tiny(covered slice after $\overline{\mathbf A}_{\rm med}^{(\ell)}$ is fixed; FastHyperGCN fixes it from input features)}
\\[6pt]

\hline
\rowcolor{rowC}
AllSet~\cite{chien2022allset} &
\(V\cup E\) &
\(\begin{gathered}V\!\to\!E\\ \text{then }E\!\to\!V\end{gathered}\) {\tiny(schedule, not simultaneous channels)} &
\(\mathbf Q^\top\ \text{then}\ \mathbf Q\) {\tiny(incidence support for the two half-steps)} &
\(\begin{gathered}
\Psi_{V\to E}:\\
\{\mathbf X_u^{(t)}:\ u\in e\},\\
\Psi_{E\to V}:\\
\{\mathbf Z_e^{(t+1)}:\ v\in e\}\\
{\scriptstyle\text{AllDeepSets: row MLP before sum;}}\\
{\scriptstyle\text{Transformer: set attention}}
\end{gathered}\) &
\(\mathrm{id}\) {\tiny(per half-step; sequential, not channel mixing)} &
\(\begin{gathered}\mathbf Z_e^{(t+1)}=f_{V\to E}\!\big(\\\mathcal V_{e,X}^{(t)};\mathbf Z_e^{(t)}\big),\\\mathbf X_v^{(t+1)}=f_{E\to V}\!\big(\\\mathcal E_{v,Z}^{(t+1)};\mathbf X_v^{(t)}\big)\\{\scriptstyle\text{(boundary two-stage update)}}\end{gathered}\)
\\[6pt]

\hline
\rowcolor{rowC}
ED-HNN\textsuperscript{\S}~\cite{Wang2023EDHNN} &
\(V\) &
\(\{1\}\) {\tiny(final node-update slice; $V\!\to\!E$ then $E\!\to\!V$ is a schedule)} &
\(\mathbf Q\) {\tiny(E-to-V incidence after transient $\mathbf m_e=\sum_{u\in e}\widehat\phi(\vh_u^{(t)})$ is built)} &
\(\Psi(e,v)=\widehat\rho\!\left(\vh_v^{(t)},\sum_{u\in e}\widehat\phi(\vh_u^{(t)})\right)\) &
\(\begin{gathered}\mathbf C_{1,v}^{(t)}=\sum_{e\ni v}\Psi(e,v),\\\mixmsg_v^{(t)}=\mathbf C_{1,v}^{(t)}\quad{\scriptstyle(\boxplus:\mathrm{id})}\end{gathered}\) {\tiny(boundary: $\Psi$ contains prior V-to-E aggregation)} &
\(\widehat\varphi(\vh_v^{(t)},\mixmsg_v^{(t)},\vx_v,d_v)\) {\tiny(approximates GD; ADMM only under an extra assumption)}
\\[6pt]

\end{longtable}
\endgroup

\colorlet{geobarA}{repGeometric}
\colorlet{geobarB}{repGeometric!80}
\colorlet{geobarC}{repGeometric!62}
\colorlet{georowA}{repGeometric!9}
\colorlet{georowB}{repGeometric!9}
\colorlet{georowC}{repGeometric!9}

\providecommand{\MLP}{\operatorname{MLP}}
\providecommand{\GNN}{\operatorname{GNN}}
\providecommand{\GVP}{\operatorname{GVP}}
\providecommand{\LN}{\operatorname{LN}}
\providecommand{\geomethod}[3]{%
\begin{minipage}[c]{1.82cm}\raggedright
#1\\[-1pt]
{\tiny\citeauthor{#2}}\\[-1pt]
{\tiny(\citeyear{#2})}\\[-1pt]
{\tiny #3}
\end{minipage}}

\makeatletter
\@ifundefined{NC@find@L}{\newcolumntype{L}[1]{>{\raggedright\arraybackslash}p{#1}}}{}
\@ifundefined{NC@find@C}{\newcolumntype{C}[1]{>{\centering\arraybackslash}p{#1}}}{}
\@ifundefined{color@darkred}{\definecolor{darkred}{HTML}{8B0000}}{}
\makeatother

\begingroup
\scriptsize
\renewcommand{\arraystretch}{1.5}
\arrayrulecolor{black!45}
\setlength{\arrayrulewidth}{0.5pt}
\setlength{\tabcolsep}{2pt}
\setlength{\LTpre}{6pt}
\setlength{\LTpost}{6pt}
\setlength{\LTcapwidth}{\textwidth}

\begin{longtable}{
@{}
L{1.95cm}
C{1.60cm}
C{1.15cm}
L{2.70cm}
L{2.55cm}
L{2.30cm}
L{2.80cm}
@{}}

\caption{\scriptsize\textbf{Family 7. Geometric and Equivariant GNNs.}
Source-grounded decompositions of coordinate-, distance-, angle-, vector-, tensor-, and
algebra-valued geometric layers. Geometric context enters $\mathcal X$; typed channel
contributions are fused by tuples, direct sums, or concatenation rather than forced scalar
sums. The stated symmetry group is exactly the one established by the source. Rows marked
$^{\natural}$ preserve that source-level group claim, while $^{\ddagger}$ reports only the
supported slice of discrete-orientation or sequential geometric constructions.}
\label{tab:family-geometric}\\[6pt]

\toprule
\rowcolor{geobarA}
\textcolor{white}{\textbf{Method}} &
\multicolumn{1}{C{1.60cm}}{\textcolor{white}{$\boldsymbol{\mathcal X}$}} &
\multicolumn{1}{C{1.15cm}}{\textcolor{white}{$\boldsymbol{\mathcal K}$}} &
\multicolumn{1}{C{2.70cm}}{\textcolor{white}{\makecell[c]{\textbf{Propagation}\\$\boldsymbol{\mathbf P_k^{(\ell)}}$}}} &
\multicolumn{1}{C{2.55cm}}{\textcolor{white}{\makecell[c]{\textbf{Message}\\$\boldsymbol{\Psi_k^{(\ell)}}$}}} &
\multicolumn{1}{C{2.30cm}}{\textcolor{white}{\makecell[c]{\textbf{Mixing}\\$\boldsymbol{\boxplus^{(\ell)}}$}}} &
\multicolumn{1}{C{2.80cm}}{\textcolor{white}{\makecell[c]{\textbf{Ego/Update}\\$\boldsymbol{(\mathcal B_\ell,\phi_\ell)}$}}}
\\[6pt]
\midrule
\endfirsthead

\multicolumn{7}{c}{\tablename\ \thetable\ -- continued} \\[6pt]
\toprule
\rowcolor{geobarA}
\textcolor{white}{\textbf{Method}} &
\multicolumn{1}{C{1.60cm}}{\textcolor{white}{$\boldsymbol{\mathcal X}$}} &
\multicolumn{1}{C{1.15cm}}{\textcolor{white}{$\boldsymbol{\mathcal K}$}} &
\multicolumn{1}{C{2.70cm}}{\textcolor{white}{\makecell[c]{\textbf{Propagation}\\$\boldsymbol{\mathbf P_k^{(\ell)}}$}}} &
\multicolumn{1}{C{2.55cm}}{\textcolor{white}{\makecell[c]{\textbf{Message}\\$\boldsymbol{\Psi_k^{(\ell)}}$}}} &
\multicolumn{1}{C{2.30cm}}{\textcolor{white}{\makecell[c]{\textbf{Mixing}\\$\boldsymbol{\boxplus^{(\ell)}}$}}} &
\multicolumn{1}{C{2.80cm}}{\textcolor{white}{\makecell[c]{\textbf{Ego/Update}\\$\boldsymbol{(\mathcal B_\ell,\phi_\ell)}$}}}
\\[6pt]
\midrule
\endhead

\midrule
\multicolumn{7}{r}{\textit{continued on next page}} \\[6pt]
\endfoot

\bottomrule
\endlastfoot

\rowcolor{geobarA}
\multicolumn{7}{@{}p{16.02cm}@{}}{%
\textcolor{white}{\makecell[l]{\textbf{7a\ \ Invariant (distance / angle / torsion) messages}\\[2pt]
\emph{$G$-invariant scalars; coordinates enter $\mathcal X$ only through invariants}}}}
\\[8pt]

\hline
\rowcolor{georowA}
\begin{minipage}[c]{1.82cm}\raggedright
MoNet\\[-1pt]
{\tiny\citeauthor{monti2017geometric}}\\[-1pt]
{\tiny(\citeyear{monti2017geometric})}
\end{minipage} &
$\begin{gathered}V;\\u_{ij}\\{\scriptstyle(\text{pseudo-coord.})}\end{gathered}$ &
$\{1,\dots,J\}$ &
$\begin{gathered}{}[\mathbf P_m]_{ij}=\\\indicator[j\in\mathcal N(i)]\\{}\cdot w_m(u_{ij})\\{\scriptstyle(\text{Gaussian})}\\{\scriptstyle(\text{pseudo-coordinate})}\\{\scriptstyle(\text{kernel})}\end{gathered}$ &
$\vh_j^{(\ell)}\mathbf W_m^{(\ell)}$ &
$\textstyle\sum_m \mathbf C_m$ &
$\begin{gathered}\mathcal B_\ell=0,\\\phi_\ell=\sigma(\mixmsg^{(\ell)})\end{gathered}$ \\[8pt]

\hline
\rowcolor{georowA}
\geomethod{SchNet}{schutt2017schnet}{$E(3)$} &
$\begin{gathered}V;\\r_{ij}\end{gathered}$ &
$\{1\}$ &
$\begin{gathered}{}[\mathbf P_1]_{ij}=1\\{\scriptstyle(\text{atom-pair support})}\end{gathered}$ &
$\begin{gathered}(\vh_j^{(\ell)}\mathbf W_{\rm in})\\{}\odot W_\theta\!\big(\\\mathrm{RBF}(r_{ij})\big)\\{\scriptstyle(\text{cont.-filter msg})}\end{gathered}$ &
$\begin{gathered}\mathrm{id}\\C_{1,i}=\\\sum_jP_{ij}\Psi_{ij}\end{gathered}$ &
$\begin{gathered}\mathcal B_\ell=\vh_i^{(\ell)},\\\phi_\ell(b,\mixmsg)=\\b+\MLP(\mixmsg)\end{gathered}$ \\
\end{longtable}

\endgroup

\end{document}